\pdfoutput=1

\documentclass[a4paper,12pt,times,numbered,print,index]{PhDThesisPSnPDF}

\input{Preamble/preamble}

\title{Long-term Traffic Scene Prediction via Polynomial Representations in Autonomous Driving}

\renewcommand{\submissiontext}{\textbf{Dissertation} \\ zur Erlangung des Grades eines Doktors \\ der
Naturwissenschaften (Dr. rer. nat.) 
\vspace{1.5cm}

am Fachbereich Mathematik und Informatik \\ der Freien Universität Berlin}

\author{vorgelegt von \\ \textbf{Yue Yao}}

\makeatletter
\newcommand{\gutachteronelabel}{Erstgutachter/in:}
\newcommand{\gutachtertwolabel}{Zweitgutachter/in:}
\newcommand{\disputationdatelabel}{Tag der Disputation:}

\newcommand{\gutachterone}{Prof. Dr. Daniel G{ö}hring}
\newcommand{\gutachtertwo}{Prof. Dr. Matthias Schreier}
\newcommand{\disputationdate}{15.07.2026} 
\makeatother

\ifdefineAbstract
\fi

\ifdefineChapter
\fi

\begin{document}
\frontmatter

\maketitle

\clearpage
\thispagestyle{plain} 
\begingroup
\setlength{\parindent}{0pt}
\setlength{\parskip}{6pt}
{\raggedright
\textbf{\gutachteronelabel} \ \gutachterone\par
\textbf{\gutachtertwolabel} \ \gutachtertwo\par
\vspace{1em}
\textbf{\disputationdatelabel} \ \disputationdate\par
}
\endgroup
\clearpage


\begin{declaration}

Name: Yao\\
Vorname: Yue
\vspace{1.5em}

Ich erkläre gegenüber der Freien Universität Berlin, dass ich die vorliegende Dissertation
selbstständig und ohne Benutzung anderer als der angegebenen Quellen und Hilfsmittel angefertigt
habe. Die vorliegende Arbeit ist frei von Plagiaten. Alle Ausführungen, die wörtlich oder inhaltlich
aus anderen Schriften entnommen sind, habe ich als solche kenntlich gemacht. Diese Dissertation
wurde in gleicher oder ähnlicher Form noch in keinem früheren Promotionsverfahren eingereicht.

Zur sprachlichen Überarbeitung (insbesondere zur Korrektur von Grammatik
und Formulierungen) habe ich das KI-gestützte Textverarbeitungssystem
ChatGPT (Modell: GPT-5.2 instant, OpenAI) verwendet. Die Nutzung beschränkte
sich ausschließlich auf sprachliche Korrekturen; inhaltliche, konzeptionelle
oder wissenschaftliche Leistungen wurden nicht durch das System erbracht.

Mit einer Prüfung meiner Arbeit durch ein Plagiatsprüfungsprogramm erkläre ich mich
einverstanden.

\vspace{1.5em}

Datum:  \: \: \: \:  \: \:  \: \: Unterschrift: 

\end{declaration}

\begin{ieeecopyright}

\noindent In reference to IEEE copyrighted material which is used with permission in this thesis, the IEEE does not endorse any of Freie Universität Berlin's products or services. Internal or personal use of this material is permitted. If interested in reprinting/republishing IEEE copyrighted material for advertising or promotional purposes or for creating new collective works for resale or redistribution, please go to \url{https://www.ieee.org/publications/rights/rights-link.html} to learn how to obtain a License from RightsLink.

\end{ieeecopyright}


\begin{acknowledgements}      
This doctoral thesis was carried out as part of my research at the AI Lab of Aumovio SE in collaboration with the Freie Universität Berlin, and I would like to express my sincere gratitude to all those who made this work possible and accompanied me on this journey.

First and foremost, I would like to thank Prof. Dr. Daniel Göhring for his supervision of this dissertation and for providing the opportunity to pursue my doctoral studies at the Freie Universität Berlin. I am sincerely grateful for his feedback and constructive suggestions, which contributed to shaping and improving this work. I am also grateful to Prof. Dr. Matthias Schreier for being the second examiner of this thesis and for his valuable time and feedback. Furthermore, I would like to thank Dr. Andreas Philipp for his many insightful discussions and helpful advice throughout the research process.

Special thanks are also directed to all my colleagues at Aumovio SE and the Freie Universität Berlin for the great working atmosphere, fruitful discussions, and the enjoyable social events. My deepest gratitude goes to Dr. Jörg Reichardt, my company supervisor at Aumovio, whose continuous guidance, scientific advice, and constructive criticism were instrumental to the success of this work. His mentorship profoundly shaped my approach to research design, critical thinking, and scientific writing. I am deeply indebted for his support during every stage of this dissertation, from early concept discussions to paper rebuttals and revisions. 

I would also like to thank my colleagues Mohamed-Khalil Bouzidi, Jonas Neuhöfer, and Christian Schlauch for their collaboration, inspiring technical conversations, and the pleasant working atmosphere we shared in the lab. My sincere thanks further go to Manas Kumar, Georgii Mikriukov, and Moussa Kassem Sbeyti for many enriching discussions and valuable advice. I am also grateful to Dr. Andreas Weinlich for his excellent management and organization of the lab, which created an ideal environment for research and exchange.

I also wish to especially acknowledge Shengchao Yan from the University of Freiburg for his collaboration and contributions to our joint research, which greatly enriched and strengthened the foundations of this work.

Finally, and most importantly, I would like to express my deepest gratitude to my parents, Jiankang Yao and Yongjun Zhong, and my wife, Jing Sun, for their unwavering love, understanding, and encouragement. Their patience and belief in me have been my greatest source of strength throughout this journey.

\end{acknowledgements}

\begin{abstract}
This thesis addresses fundamental challenges in traffic scene prediction for autonomous driving by introducing robust and computationally efficient models based on polynomial representations. While conventional sequence-based representations often struggle with noise and generalization, this work demonstrates that polynomial representations offer significant advantages in computational efficiency, generalization, and prediction plausibility.

Through theoretical analysis and empirical validation, this thesis demonstrates that moderate-degree polynomials capture real-world motion dynamics with high fidelity without constraining predictive performance. Building on this foundation, a prediction model representing both trajectories and map geometry with polynomial representations achieves near state-of-the-art accuracy on standard benchmarks while substantially improving generalization under distribution shift. Extending this concept, a diffusion-based generative framework enables multi-agent scene generation, producing traffic continuations that are more plausible and kinematically consistent than those generated by conventional baselines.

Evaluations on the Argoverse 2 and Waymo Open datasets confirm that polynomial representations reduce computational cost, enhance cross-dataset generalization, and yield smoother trajectories and higher behavioral plausibility. The findings reveal that standard in-distribution evaluation and regression-based metrics may fail to reflect true model generalization and prediction plausibility.

By providing theoretical justification and empirical validation, this dissertation establishes polynomial trajectory representations as an efficient, expressive, and generalizable foundation for traffic scene prediction in safety critical autonomous driving.
\end{abstract}

\begin{kurzfassung}
Diese Dissertation behandelt grundlegende Herausforderungen der Verkehrsszenenvorhersage für das autonome Fahren, indem robuste und recheneffiziente Modelle auf der Basis polynomieller Repräsentationen entwickelt werden. Während herkömmliche sequenzbasierte Darstellungen häufig mit Rauschen und mangelnder Generalisierungsfähigkeit zu kämpfen haben, zeigt diese Arbeit, dass polynomielle Repräsentationen deutliche Vorteile hinsichtlich Rechenaufwand, Generalisierung und Vorhersageplausibilität bieten.

Durch theoretische Analysen und empirische Validierungen wird gezeigt, dass Polynome reale Bewegungsdynamiken mit hoher Genauigkeit erfassen, ohne die Vorhersageleistung einzuschränken. Aufbauend auf dieser Grundlage wird ein Vorhersagemodell vorgestellt, das sowohl Trajektorien als auch Kartengeometrien mittels polynomieller Repräsentationen beschreibt. Dieses Modell erreicht nahezu den Stand der Technik in Bezug auf Genauigkeit, verbessert jedoch die Generalisierungsfähigkeit unter Verteilungsverschiebungen erheblich.

Darüber hinaus ermöglicht ein darauf aufbauendes, diffusionsbasiertes generatives Framework die gemeinsame Vorhersage und Generierung mehrerer interagierender Verkehrsteilnehmer. Dadurch entstehen Verkehrsszenen, die plausibler und kinematisch konsistenter sind als jene konventioneller Vergleichsmodelle.

Auswertungen auf den Datensätzen Argoverse 2 und Waymo Open bestätigen, dass polynomielle Repräsentationen den Rechenaufwand reduzieren, die Generalisierung über Datensätze hinweg verbessern und zu glatteren Trajektorien mit höherer Verhaltensplausibilität führen. Die Ergebnisse verdeutlichen zudem, dass gängige In-Distribution-Evaluierungen und regressionsbasierte Metriken die tatsächliche Generalisierungsfähigkeit und Vorhersageplausibilität eines Modells häufig nicht adäquat widerspiegeln.

Durch die Kombination theoretischer Begründungen und empirischer Validierung etabliert diese Dissertation polynomielle Trajektorienrepräsentationen als effiziente, ausdrucksstarke und generalisierbare Grundlage für die Verkehrsszenenvorhersage in sicherheitskritischen autonomen Fahrsystemen.
\end{kurzfassung}


\tableofcontents

\listoffigures

\listoftables


\printnomenclature

\nomenclature[a-a0]{$A$}{number of agents}
\nomenclature[a-a1]{$\mathcal{A}$}{set of agents}
\nomenclature[a-c1]{$\mathbf{c}(\tau)$}{polynomial modeled states parameterized by $\tau$}
\nomenclature[a-c0]{$\boldsymbol{C}$}{conditioning information (tokens)}
\nomenclature[a-d]{$\mathcal{D}$}{set of data}
\nomenclature[a-d]{$D$}{hidden dimension}
\nomenclature[a-d1]{$d$}{spatial dimension}
\nomenclature[a-d]{$D_{KL}$}{Kullback-Leibler divergence}
\nomenclature[a-d]{$D^{\text{key}}$}{dimension of key in attention mechanism}

\nomenclature[a-i]{$\mathbf{I}_d$}{$d \times d$ identity matrix}
\nomenclature[a-k]{$K$}{number of prediction modes}
\nomenclature[a-m]{$M$}{number of map elements}
\nomenclature[a-m]{$\mathcal{M}$}{set of map elements}
\nomenclature[a-n]{$N$}{polynomial degree}
\nomenclature[a-q]{$\boldsymbol{Q}$}{query matrix in attention mechanism}
\nomenclature[a-k2]{$\boldsymbol{K}$}{key matrix in attention mechanism}
\nomenclature[a-v]{$\boldsymbol{V}$}{value matrix in attention mechanism}
\nomenclature[a-s]{$S$}{total number of diffusion steps}
\nomenclature[a-t]{$T$}{total number of timesteps}
\nomenclature[a-x0]{$\mathbf{x}$}{state vector}
\nomenclature[a-x1]{$\mathbf{x}_t$}{state vector at timestep $t$}
\nomenclature[a-x2]{$\mathbf{x}_s$}{noised data vector at diffusion step $s$}
\nomenclature[a-w1]{$\mathbf{w}_n$}{$n$-th degree polynomial parameter}
\nomenclature[a-w1]{$\mathbf{W}_n$}{$n$-th degree polynomial parameters of a set of elements}
\nomenclature[a-z1]{$\mathbf{z}$}{measurement vector}
\nomenclature[a-z4]{$\mathbf{\hat{z}}$}{predicted measurement vector}

\nomenclature[a-t]{$t'$}{timestamp in second at timestep $t$}
\nomenclature[a-T]{$T'$}{total time in second}
\nomenclature[a-T]{$\boldsymbol{T}$}{token feature}
\nomenclature[a-r]{$\mathbb{R}$}{space of real numbers}
\nomenclature[a-r]{$\mathbb{R}_{\geq0}$}{space of non-negative real numbers}
\nomenclature[a-z]{$\mathbb{Z}$}{space of integer numbers}
\nomenclature[a-r]{$r$}{radial distance in polar coordinate}
\nomenclature[a-r]{$\boldsymbol{R}^{\mathrm{rot}}$}{rotation matrix}

\nomenclature[a-f]{$\mathbf{F}$}{state transition matrix in Kalman filter}
\nomenclature[a-H]{$\mathbf{H}$}{observation matrix in Kalman filter}
\nomenclature[a-q]{$\mathbf{Q}_t$}{process noise covariance in Kalman filter at timestep $t$}
\nomenclature[a-r]{$\mathbf{R}_t$}{observation noise covariance in Kalman filter at timestep $t$}
\nomenclature[a-p]{$\mathbf{P}$}{state covariance matrix}
\nomenclature[a-k]{$\mathbf{K}_t$}{Kalman gain at timestep $t$}
\nomenclature[a-x2]{$\hat{\mathbf{x}}$}{predicted (estimated) state vector}
\nomenclature[a-x2]{$\hat{\mathbf{x}}^s$}{smoothed state vector from Rauch-Tung-Striebel smoother}
\nomenclature[a-p]{$\mathbf{P}^s$}{smoothed state covariance matrix}
\nomenclature[a-c]{$\mathbf{C}_t$}{Rauch-Tung-Striebel smoothing gain}

\nomenclature[a-n]{$\hat N$}{optimal polynomial degree}
\nomenclature[a-p1]{$p_x$}{x-position}
\nomenclature[a-p1]{$p_y$}{y-position}
\nomenclature[a-v1]{$v_x$}{x-velocity}
\nomenclature[a-v1]{$v_{y}$}{y-velocity}
\nomenclature[a-a1]{$a_x$}{x-acceleration}
\nomenclature[a-a1]{$a_y$}{y-acceleration}
\nomenclature[g-alpha]{$\alpha_s$}{noise schedule parameter at diffusion step $s$}
\nomenclature[g-alphabar]{$\bar{\alpha}_s$}{cumulative noise schedule parameter}
\nomenclature[g-beta]{$\beta_s$}{noise parameter at diffusion step $s$}
\nomenclature[g-delta]{$\Delta$}{control point vector difference}
\nomenclature[g-epsilon1]{$\boldsymbol{\epsilon}$}{noise (residual) vector}
\nomenclature[g-epsilon1]{$\hat{\boldsymbol{\epsilon}}$}{predicted noise (residual) vector}
\nomenclature[g-theta]{$\boldsymbol{\theta}$}{model parameters}
\nomenclature[g-ell]{$\ell$}{loss objective}
\nomenclature[g-tau]{$\tau$}{normalized input variable $\in [0,1]$}
\nomenclature[g-phi1]{$\phi(\cdot)$}{basis function}
\nomenclature[g-pi]{$\pi$}{predicted probability}
\nomenclature[g-sigma]{$\boldsymbol{\Sigma}$}{covariance matrix}

\nomenclature[g-omega]{$\boldsymbol{\omega}$}{vector of polynomial parameters (control points)}
\nomenclature[g-theta]{$\nabla_{\boldsymbol{\theta}}$}{gradient with respect to ${\boldsymbol{\theta}}$}

\nomenclature[g-mu]{${\boldsymbol{\mu}}$}{mean of multivariate normal distribution}

\nomenclature[g-phi2]{$\dot\phi(\cdot)$}{first derivative of basis function}
\nomenclature[g-phi3]{$\ddot\phi(\cdot)$}{second derivative of basis function}
\nomenclature[g-phi]{$\boldsymbol{\Phi}$}{full basis matrix including spatial dimensions}
\nomenclature[g-phi]{$\boldsymbol{\Phi}_{\text{B}}$}{basis matrix for a single spatial dimension}
\nomenclature[g-sigma]{$\sigma$}{standard deviation of Gaussian distribution}
\nomenclature[g-psi]{$\psi$}{angle in polar coordinate}
\nomenclature[g-Sigma]{$\boldsymbol{\Sigma}_{o}$}{observation noise covariance matrix}
\nomenclature[g-Sigma]{$\boldsymbol{\Sigma}_{\boldsymbol{\omega}}$}{prior covariance matrix of polynomial parameters}
\nomenclature[g-epsilon2]{$\boldsymbol{\epsilon}_{p}$}{process noise vector in Kalman filter}
\nomenclature[g-epsilon2]{$\boldsymbol{\epsilon}_{o}$}{observation noise vector in Kalman filter}
\nomenclature[g-gamma]{$\gamma$}{heading}
\nomenclature[g-gamma]{$\gamma^{-}$}{heading difference}

\nomenclature[z-lidar]{LiDAR}{Light Detection and Ranging}
\nomenclature[z-gps]{GPS}{Global Positioning System}
\nomenclature[z-sota]{SotA}{State-of-the-Art}
\nomenclature[z-id]{ID}{In-Distribution}
\nomenclature[z-ood]{OoD}{Out-of-Distribution}
\nomenclature[z-ego]{Ego}{the autonomous vehicle}
\nomenclature[z-a1]{A1}{Argoverse 1 dataset}
\nomenclature[z-a2]{A2}{Argoverse 2 dataset}
\nomenclature[z-wo]{WO}{Waymo Open dataset}
\nomenclature[z-hd]{HD}{High-Definition}
\nomenclature[z-cv]{CV}{Computer Vision}
\nomenclature[z-bev]{BEV}{Bird's-Eye View}
\nomenclature[z-cnn]{CNN}{Convolutional Neural Network}
\nomenclature[z-elbo]{ELBO}{Evidence Lower Bound}
\nomenclature[z-rnn]{RNN}{Recurrent Neural Network}
\nomenclature[z-lstm]{LSTM}{Long Short-Term Memory}
\nomenclature[z-vae]{VAE}{Variational Autoencoder}
\nomenclature[z-gan]{GAN}{Generative Adversarial Network}
\nomenclature[z-ddpm]{DDPM}{Denoising Diffusion Probabilistic Model}
\nomenclature[z-ddim]{DDIM}{Denoising Diffusion Implicit Model}
\nomenclature[z-nlp]{NLP}{Natural Language Processing}
\nomenclature[z-gpt]{GPT}{Generative Pre-trained Transformer}
\nomenclature[z-mlp]{MLP}{Multi-Layer Perceptron}
\nomenclature[z-minade]{minADE}{minimum Average Displacement Error}
\nomenclature[z-minfde]{minFDE}{minimum Final Displacement Error}
\nomenclature[z-minade]{minSADE}{minimum Scene Average Displacement Error}
\nomenclature[z-minfde]{minSFDE}{minimum Scene Final Displacement Error}
\nomenclature[z-afe]{AFE}{Average Fit Error}

\nomenclature[z-rts]{RTS}{Rauch-Tung-Striebel (smoother)}
\nomenclature[z-aic]{AIC}{Akaike Information Criterion}
\nomenclature[z-bic]{BIC}{Bayesian Information Criterion}

\nomenclature[z-mae]{MAE}{Masked Autoencoder}
\nomenclature[z-tls]{TLS}{Total Least Squares}
\nomenclature[z-ols]{OLS}{Ordinary Least Squares}

\nomenclature[z-ep0]{EP}{Everything Polynomial (proposed)}
\nomenclature[z-ep1-f]{EP-F}{Everything Polynomial with heterogeneous augmentation (proposed)}
\nomenclature[z-ep1-q]{EP-Q}{Everything Polynomial with homogeneous augmentation (proposed)}
\nomenclature[z-ep1-diffuser]{EP-Diffuser}{Everything Polynomial Diffuser (proposed)}

\nomenclature[z-fmae0]{FMAE}{Forecast-MAE (baseline)}
\nomenclature[z-fmae1-ma]{FMAE-MA}{Forecast-MAE-multiagent (baseline)}
\nomenclature[z-qcnet]{QCNet}{Query-Centric Trajectory Prediction Network (baseline)}
\nomenclature[z-opttrajdiff]{OptTrajDiff}{Optimized Trajectory Diffusion (baseline)}

\nomenclature[z-tw]{TW}{Time Window}
\nomenclature[z-pi]{PI}{Positional Information}
\nomenclature[z-AT]{AT}{Agent Type}
\nomenclature[z-AT]{MT}{Map Element Type}

\nomenclature[s-i]{$i$}{agent index}
\nomenclature[s-t]{$t$}{timestep index}
\nomenclature[s-t1]{$t_0$}{reference (current) timestep index}
\nomenclature[s-l]{$l$}{sample point index}
\nomenclature[s-n]{$n$}{polynomial parameter index}
\nomenclature[s-k]{$k$}{prediction mode index}
\nomenclature[s-s]{$s$}{diffusion step index}

\nomenclature[x-n]{$||\cdot||$}{Norm (magnitude) of a vector}
\nomenclature[x-g]{$\mathcal{N}(\cdot, \cdot)$}{multivariate normal distribution}
\nomenclature[x-e]{$\mathbb{E}(a)$}{the expectation of a random variable $a$}
\nomenclature[x-e1]{$\mathbb{E}_b(a)$}{the expectation of a random variable $a$ with respect to $b$}
\nomenclature[x-diag]{$\text{diag}(\boldsymbol{a})$}{diagonal matrix with $\boldsymbol{a}$ as diagonal entities}
\nomenclature[x-concat]{$\text{concat}(\cdot,\cdot)$}{concatenation operation}
\nomenclature[x-kronecker]{$\otimes$}{Kronecker product}
\nomenclature[x-magnitude]{$\lVert\cdot\rVert_2$}{magnitude of a vector}

\newcommand{\TOA}{A2$\rightarrow$A2}
\newcommand{\TA}{A2*$\rightarrow$WO*}
\newcommand{\TWO}{WO*$\rightarrow$A2*}

\mainmatter


\chapter{Introduction}  

\section{Human Intuition in Traffic Scene Prediction} 
In daily traffic, humans continuously predict the movements of other road users. Whether as drivers, cyclists, or pedestrians, we instinctively anticipate how others will move based on past experience, contextual cues, and social interactions. For example, a driver approaching an intersection expects another vehicle to decelerate at a red light, just as a pedestrian crossing the street assumes an oncoming car will yield. This remarkable human ability to anticipate traffic behavior is grounded in several key characteristics, including smooth motion, generalization across environments, and the capacity to consider multiple possible futures.

\paragraph{Smooth and Feasible Motion.}
A fundamental property of human movement is smoothness. People tend to minimize abrupt changes in velocity, acceleration, and jerk, making motion comfortable and dynamically feasible \cite{hayati_jerk_2020}. Most drivers avoid sudden accelerations or sharp turns unless responding to an emergency \cite{bae_self_2020, macadam_understanding_2003}, and pedestrians and cyclists follow continuous, predictable motion patterns that adhere to physical and biomechanical constraints \cite{famiglietti_bicycle_2020}.

This intrinsic smoothness is key to how humans anticipate the actions of others. When predicting the movement of a vehicle, pedestrians assume that it will maintain a gradual change in speed and direction, rather than making erratic, unnatural maneuvers. Similarly, when merging onto a highway, drivers expect surrounding vehicles to follow well-behaved trajectories, allowing for safe and coordinated interactions. The preference for predictable, dynamically feasible motion is an essential reason why humans are so effective at anticipating and reacting to other traffic participants' behaviors in traffic.

\paragraph{Generalization in Prediction.}
Beyond producing smooth movements, human behavior is also shaped by a shared understanding of traffic dynamics, shaped through social interactions and constrained by traffic rules. This shared framework creates a set of generalized motion patterns that enable humans to generalize motion predictions remarkably well across diverse environments. Even in unfamiliar settings, humans can make inferences about how others will behave. For instance, a driver unfamiliar with a city may still predict how other vehicles will merge onto a highway or navigate roundabouts by drawing on learned traffic dynamics from previous experiences. This ability to transfer knowledge between different contexts allows humans to adapt to unseen scenarios with ease, ensuring safe and efficient movement even in highly variable traffic environments.

\paragraph{Anticipating Multiple Futures.}
The shared understanding of motion dynamics and generalized traffic patterns forms the basis of our prior knowledge of traffic. This enables the most sophisticated aspect of human prediction: the ability to anticipate multiple possible futures rather than committing to a single deterministic outcome. 

As the time horizon increases, the movements of traffic participants reveal multiple plausible futures, influenced by road topology and interactions between traffic participants. Humans do not assume a single trajectory for each traffic participant; instead, they consider a range of plausible behaviors based on their prior experience.
For instance, when approaching an intersection, a driver evaluates several possible actions of nearby vehicles based on their prior experience: (\lowerromannumeral{1}) the car in front might slow down to yield, (\lowerromannumeral{2}) it might proceed aggressively, or (\lowerromannumeral{3}) it might turn.

By keeping multiple possible outcomes in mind, humans remain adaptable and react accordingly to sudden changes, ensuring safe operation in highly dynamic and uncertain environments.

Together, these human capabilities highlight the sophistication of intuitive traffic prediction and set a high bar for computational approaches. The next section discusses the key challenges in replicating such human-like predictive capability in autonomous driving systems.

\section{Challenges in Computational Traffic Scene Prediction}

\begin{figure}[thb]
\centering
\includegraphics[width=\textwidth]{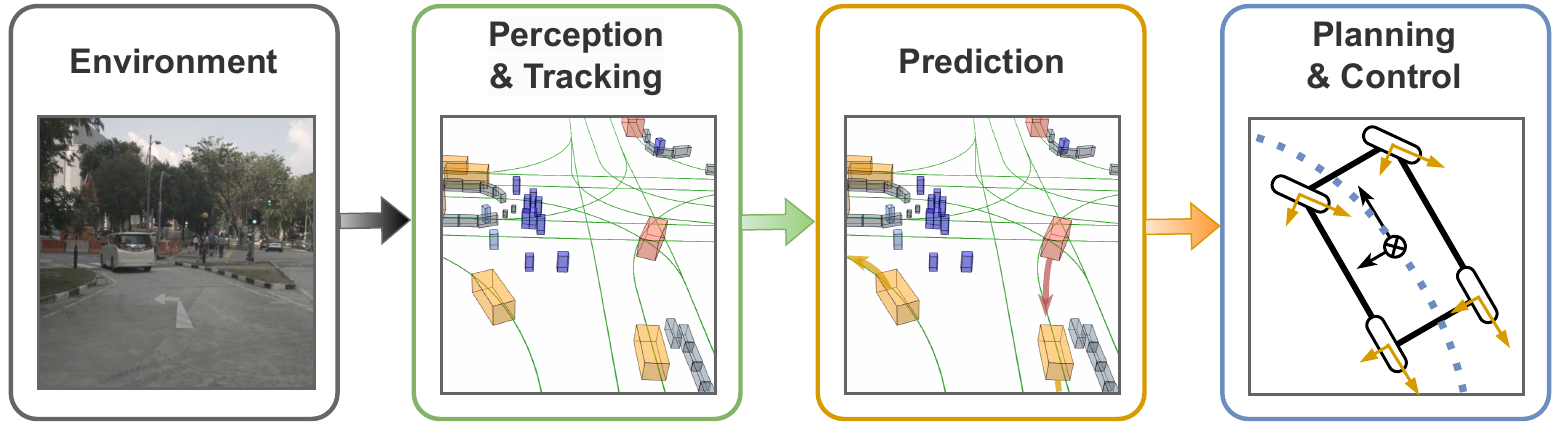}
\caption{A typical modular architecture of an autonomous driving function stack}
\label{fig: AV_pipeline}
\end{figure}

Prediction is an essential component of autonomous driving systems. As shown in Figure \ref{fig: AV_pipeline}, the prediction algorithm typically receives an abstract environment model from the perception module, which includes information about other traffic participants and the map. It then predicts the future motions of surrounding \emph{agents}\footnote{In the remainder of this dissertation, the term \emph{agents} refers to all traffic participants, including the ego vehicle, although the ego vehicle’s trajectory is determined by the planning module rather than predicted.} near the \emph{ego vehicle} (the autonomous vehicle itself), by considering the interactions between agents and map elements. These predicted motions serve as constraints in the planning module to output safe and feasible trajectories for the autonomous vehicle.

In parallel to the modular architecture shown in Figure \ref{fig: AV_pipeline}, recent research trains a single network end-to-end from sensor input to a planned trajectory \cite{hu_planning_2023, jiang_vad_2023, sun_sparsedrive_2025}. Competitive end-to-end systems typically retain motion prediction as an explicit intermediate task rather than discarding it. Since this work studies a trajectory and map representation rather than a stack layout, the findings are not confined to the modular case.

A further simplification in Figure \ref{fig: AV_pipeline} is the unidirectional interaction between prediction and planning. Since the ego vehicle is itself a traffic participant, its planned trajectory alters the likely futures of the surrounding agents. Prediction and planning are therefore coupled, and treating them as a strict cascade is known to yield overly conservative behavior \cite{trautman_unfreezing_2010}. Joint formulations of prediction and planning have accordingly received increasing attention \cite{hagedorn_integration_2024}. This work nevertheless addresses the formulation in which prediction is not conditioned on the ego's plan, in line with the benchmarks used in Chapters \ref{cpt: improving generalization} and \ref{cpt: generative model}. The extension toward joint prediction and planning is discussed in Section \ref{sec 6.2}.

While humans predict traffic behavior intuitively, replicating this capability in computational models remains a significant challenge. The research community initially employed rule-based approaches to encode human driving behaviors through predefined heuristics and explicit modeling \cite{kesting_general_2007, kesting_enhanced_2010, philipp_perception_2021}. However, the field has undergone a paradigm shift toward learning-based methods that can capture the complex patterns and implicit rules governing traffic dynamics \cite{alahi_social_2016, liang_learning_2020, zhou_hivt_2022}. Currently, state-of-the-art (SotA) performance in traffic scene prediction is dominated by learning-based models that leverage large datasets to identify subtle interaction patterns that would be highly challenging to manually encode. Despite the success of learning-based approaches, significant challenges remain in developing models with human-like predictive capabilities:

\paragraph{Data Representation.}
Modern autonomous vehicle platforms \cite{caesar_nuscenes_2020, wilson_argoverse2_2021, ettinger_waymo_2021} and traffic monitoring systems \cite{zhan_interaction_2019, krajewski_highd_2018} fuse and track agent state information using one or multiple sensors -- such as cameras, LiDAR, radar, and GPS -- operating at fixed sampling frequencies. Agents' trajectories are typically represented as time sequences of measured states at discrete timestamps. While this representation offers high flexibility and expressiveness, it can also include measurement noise, potentially leading to physically implausible movements in the recorded data. Furthermore, this measurement noise can propagate into prediction systems, resulting in unstable predictions.

Finding a trajectory representation that balances expressiveness with physical feasibility -- similar to humans’ natural understanding of motion dynamics -- remains an open challenge.

\paragraph{Generalization Across Environments.}
Unlike humans, who can readily apply prior traffic knowledge across diverse scenarios, computational models often optimize their performance for recorded data, which may fail to generalize to new environments with unfamiliar road layouts, traffic densities, and cultural driving styles \cite{bahari_vehicle_2022, feng_unitraj_2024}. Additionally, artifacts introduced during data collection and pre-processing can vary with different sensor configurations, software stacks, and environmental factors \cite{yao_empirical_2023}, leading to degraded model performance in new scenarios.

This limitation is particularly critical for autonomous driving systems, which must operate reliably across diverse and previously unseen environments.

\paragraph{Multi-Modality.}
Traffic is a complex phenomenon where multiple agents
interact in a shared space and influence each other’s behavior.
The motion of agents over long time horizons is governed by an inherently \emph{multi-modal} probability distribution, as multiple plausible futures exist depending on road topology and agent interactions. While modeling multi-modality for a single agent already presents a significant challenge for prediction algorithms, modeling multiple interacting agents simultaneously to predict coherent traffic scene continuations presents an even greater challenge.

Reliably handling numerous interacting agents and predicting about multiple possible future evolutions remains among the central challenges for autonomous driving prediction systems.

\paragraph{Computational Efficiency.}
An additional practical challenge is computational efficiency. For real-time applications in autonomous vehicles, predictions must be generated within strict time constraints while maintaining accuracy and robustness. However, as the complexity of traffic scenes increases -- with more agents and more intricate road layouts -- meeting this requirement becomes increasingly challenging for recent approaches, as highlighted in \cite[p.~107]{philipp_perception_2021}\cite{salzmann_trajectron_2020}.

In summary, despite significant progress in learning-based motion prediction, challenges remain in achieving computationally efficient, generalizable, and physically consistent models capable of handling complex multi-agent interactions. These challenges motivate the research presented in this thesis.

\newcommand{\oodfootnote}{Out-of-Distribution refers to test samples that come from a different statistical distribution than the training data, representing the model's ability to generalize to new environments and conditions not seen during training.}

\section{Research Contributions}
This thesis aims to address these limitations by developing an efficient and robust prediction methodology. This approach should be capable of predicting multiple future evolutions of traffic scenes while considering the interactions between traffic participants and map elements. Additionally, the approach should be robust against various input artifacts, such as changes in sensor setup, pre-processing, and map representation, as well as unseen scenarios.




\noindent To accomplish this, the thesis introduces a principled framework built upon three interconnected pillars: trajectory modeling, cross-dataset evaluation, and multi-agent traffic scene prediction. The research contributions progress systematically from foundational analysis of trajectory representations to the practical implementation of scalable multi-agent prediction systems.

The key contributions of this work are:

\begin{itemize}[leftmargin=*]

\item  \textbf{Contribution 1}: A principled justification of polynomial trajectory representations in prediction systems using an empirical Bayes\footnote{Empirical Bayes is a statistical approach in which the parameters of the prior distribution (hyper-parameters) are estimated directly from the data, rather than being specified purely from prior knowledge.} analysis on real-world trajectory data.



\item \textbf{Contribution 2}: An efficient multi-modal prediction model focused on individual agent motion. 
This model achieves near SotA In-Distribution (ID) performance and superior Out-of-Distribution (OoD)\footnote{\oodfootnote} robustness through polynomial representation of trajectories and map features.

\item \textbf{Contribution 3}: An extension of Contribution 2 using a diffusion-based model for multi-agent traffic scene prediction
, demonstrating improved prediction plausibility and enhanced generalization in OoD testing.
\end{itemize}

Together, these contributions substantiate the main research thesis of this dissertation:
{Polynomial representations provide a compact, generalizable, and physically consistent foundation for traffic scene prediction, enabling improved model efficiency, robustness, and prediction plausibility under distribution shift while maintaining competitive predictive performance.}






\section{Structure of the Following Chapters}
This work is structured into the following chapters:
\begin{itemize}[leftmargin=*]
\item Chapter 2 provides an overview of the state-of-the-art (SotA) in the field of traffic scene prediction, including public motion datasets and the associated motion prediction competitions along with their evaluation frameworks. It also reviews recent advancements in model development, focusing on data representations and novel model architectures such as attention mechanisms in Transformer models \cite{vaswani_attention_2017} and generative models.

\item Chapter \ref{cpt: Empirical Bayes Analysis} analyzes the fundamental questions involved in employing polynomial representations in prediction systems. It addresses whether polynomial representations are suitable for real-world trajectories by presenting an in-depth empirical analysis of the trade-off between model complexity and fit error in modeling agent trajectories. This chapter builds upon a paper by the author, published at the \textit{IEEE Intelligent Transportation Systems Conference (ITSC)} 2023 \cite{yao_empirical_2023}.

\item Chapter \ref{cpt: improving generalization} presents an OoD testing protocol that standardizes data formats and prediction tasks across two large-scale motion datasets. Additionally, a novel model is proposed for individual agent motion prediction, leveraging the polynomial representations analyzed in Chapter \ref{cpt: Empirical Bayes Analysis}. This model is benchmarked with two SotA models. Specifically, the model's performance is evaluated not only on the ID samples from the dataset used for training but also on the OoD samples from a different dataset. This chapter builds upon a paper by the author, published at the \textit{IEEE/RSJ International Conference on Intelligent Robots and Systems (IROS)} 2024 \cite{yao_improving_2024}.

\item Chapter \ref{cpt: generative model} extends the prediction model proposed in Chapter \ref{cpt: improving generalization} to address the multi-agent joint prediction task using a diffusion-based approach. Beyond the traditional metrics focusing solely on prediction accuracy, the evaluation also considers the perspective of prediction plausibility and diversity. Furthermore, the model generalization capability is evaluated using the OoD testing protocol proposed in Chapter \ref{cpt: improving generalization}. This chapter builds upon a paper by the author, published in the \textit{IEEE Robotics and Automation Letters (RA-L)} 2025 \cite{yao_ep_2025}.

\item Chapter 6 summarizes the work and provides an outlook on potential future extensions of the different approaches presented in this dissertation.

\end{itemize}


\chapter{Background and Related Work}
\section{Advancements in Motion Datasets}
\label{sec: advancements in motion datasets}
This section provides an overview of advancements in motion datasets by briefly introducing several state-of-the-art (SotA) motion datasets, which are studied and employed in the following chapters. Along with the datasets, the associated prediction competitions and evaluation frameworks are also introduced for a broader understanding of the current traffic scene prediction landscape and how performance metrics shape modeling approaches in this domain.

\subsection{Motion Datasets}
A variety of public motion datasets have been recently introduced to advance research in traffic scene prediction and autonomous driving. These datasets provide large-scale, real-world traffic scenes and have become essential benchmarks for evaluating prediction models. 

Motion datasets contain an extensive collection of diverse traffic scenes along with maps, strategically partitioned into three distinct and disjoint splits -- train, validation (val), and test. The scenes are captured from a moving sensor
platform, such as an ego vehicle \cite{chang_argoverse_2019, wilson_argoverse2_2021, ettinger_waymo_2021, caesar_nuscenes_2020}, or a drone-based system \cite{krajewski_highd_2018, zhan_interaction_2019}, during measurement campaigns with a fixed sampling frequency. The recorded agent states are tracked and processed -- such as by labeling agent types -- before being organized as sequences at discrete timesteps. Each dataset specifies a reference time point that divides the traffic scene into two segments: a history (past observations) and a future (the target to be predicted). Together, these segments form the basis for prediction tasks. To enable objective comparison and evaluation of model generalization, the future trajectories in the test split are held-out and not disclosed to model developers.

To give a clearer impression of the data format, the following outlines a generalized and adapted version of the scene structure used in the Argoverse 2 dataset \cite{wilson_argoverse2_2021}:
\begin{itemize}
  \item \textbf{Scene meta information:}
  \begin{itemize}
    \item \texttt{scene\_id}: Unique identifier for the scene.
    \item \texttt{map\_id}: Identifier for the associated map.
    \item \texttt{timestamps\_ns}: One-dimensional array of timestamps.
  \end{itemize}

  \item \textbf{Agent tracks}: An \emph{unordered} list of tracked agents. Each track contains:
  \begin{itemize}
    \item \texttt{track\_id}: Unique identifier for the agent.
    \item \texttt{agent\_type}: Type of agent (e.g., vehicle, pedestrian).
    \item \texttt{agent\_states}: Temporally ordered list of agent states. Each state includes:
    \begin{itemize}
      \item \texttt{observed}: Whether the agent was observed at this timestep.
      \item \texttt{position}: 2D coordinates of the agent.
      \item \texttt{heading}: Heading of the agent.
      \item \texttt{velocity}: 2D velocity of the agent.
    \end{itemize}
  \end{itemize}
\end{itemize}
While the example is based on Argoverse 2, it has been slightly simplified for readability and illustrative purposes, and does not reflect the exact schema used in all datasets. Nonetheless, the overall structure -- comprising scene metadata, agent tracks, and temporally ordered states -- is broadly representative of most large-scale motion datasets.

Most motion datasets also provide high-definition (HD) maps to supply the necessary road context for prediction. These maps typically contain various map elements, such as lane segments and crosswalks, represented as sequences of positional points. To describe road topology, the maps typically provide the relationships between lane segments, including successors and predecessors. 

In addition to geometric information, datasets often include categorical attributes, such as lane type (e.g., vehicle lane or bus lane). Some datasets further enrich this context with information about traffic light \cite{ettinger_waymo_2021} or traffic sign \cite{caesar_nuscenes_2020}, enhancing the environmental understanding required for accurate prediction.

Similar to the scene data, the following presents a generalized and adapted version of the map structure used in the Argoverse 2 dataset, simplified for clarity and illustrative purposes:
\begin{itemize}
  \item \textbf{Map meta information:}
  \begin{itemize}
    \item \texttt{map\_id}: Unique identifier associated with the map.
  \end{itemize}

  \item \textbf{Lane segments}: A list of all lane segments in the map. Each lane segment includes:
  \begin{itemize}
    \item \texttt{lane\_id}: Unique identifier for the lane.
    \item \texttt{lane\_type}: Type of lane (e.g., vehicle lane, bus lane).
    \item \texttt{predecessors}: List of predecessor lane segment identifiers.
    \item \texttt{successors}: List of successor lane segment identifiers.
    \item \texttt{lane\_boundary}: Geometric boundary of the lane, consisting of:
    \begin{itemize}
      \item \texttt{waypoints}: A list of 2D points representing the lane geometry.
    \end{itemize}
  \end{itemize}

  \item \textbf{Crosswalks}: A list of all crosswalks in the map. Each crosswalk includes:
  \begin{itemize}
    \item \texttt{crosswalk\_id}: Unique identifier for the crosswalk.
    \item \texttt{edge}: Geometric boundary of the crosswalk, consisting of:
    \begin{itemize}
      \item \texttt{waypoints}: A list of 2D points representing the crosswalk geometry.
    \end{itemize}
  \end{itemize}
\end{itemize}



\begin{table}[!ht]
\vspace{-0.0em}
\caption[Summary of motion datasets]{Summary of motion datasets}
\centering
\begin{tabularx}{\textwidth}{c c >{\centering\arraybackslash}X >{\centering\arraybackslash}X >{\centering\arraybackslash}X}
\Xhline{3\arrayrulewidth}
\multicolumn{2}{c}{Dataset} & Argoverse 1  \cite{chang_argoverse_2019} & Argoverse 2 \cite{wilson_argoverse2_2021} & Waymo \cite{ettinger_waymo_2021} \\
\Xhline{3\arrayrulewidth}
\multicolumn{2}{l}{\#scenes} & \SI{324}{k} & \SI{250}{k}  & \SI{576}{k}\\
\hline
\multicolumn{2}{l}{time length [s]} & \multirow{2}{*}{2 / 3 / 5} & \multirow{2}{*}{5 / 6 / 11} & \multirow{2}{*}{1.1 / 8 / 9.1} \\
\multicolumn{2}{l}{(history / future / total)} & & \\
\hline
\multicolumn{2}{l}{\#cities} & 2 & 6 & 6 \\
\hline
\multicolumn{2}{l}{sampling rate [Hz]} & 10&10&10\\
\hline
& position & 2D & 2D & 3D\\
\cline{2-5}
agent&velocity & - & 2D & 2D \\
\cline{2-5}
features&heading & - & \checkmark & \checkmark\\
\cline{2-5}
&agent size & - & - & \checkmark \\
\hline
& lane & \checkmark & \checkmark & \checkmark \\
\cline{2-5}
map&crosswalk & - & \checkmark & \checkmark \\
\cline{2-5}
features&traffic light & - & - & \checkmark \\
\cline{2-5}
&coordinates & 2D & 3D & 3D \\
\hline
prediction & marginal & \checkmark & \checkmark & \checkmark \\
\cline{2-5}
tasks&joint & - & \checkmark & \checkmark \\
\Xhline{3\arrayrulewidth}
\label{tab: dataset summary (chapter 2)}
\vspace{-1em}
\end{tabularx}
\end{table}

A summary of several popular large-scale public motion datasets is provided in Table \ref{tab: dataset summary (chapter 2)}. In addition to those listed, several other notable public datasets are worth mentioning. For example, Interaction \cite{zhan_interaction_2019} offers high-quality measurements captured using a drone platform and emphasizes highly interactive agents. Shifts \cite{malinin_shifts_2021} focuses on distribution shifts in recorded data, while nuPlan \cite{caesar_nuplan_2021} is specifically designed for planning algorithms. Furthermore, the nuScenes dataset \cite{caesar_nuscenes_2020} stands out for its comprehensive collection of raw sensor data, including synchronized camera, LiDAR, and radar streams, making it particularly valuable for research on perception, sensor fusion, and end-to-end autonomous driving systems.

Among the many available motion datasets, this work will focus on three in the following chapters:
\begin{itemize}[leftmargin=*]
\item \textbf{Argoverse 1 (2019)} \cite{chang_argoverse_2019}: A relatively lightweight motion dataset comprising approximately \SI{324}{k} scenes, each with 2 seconds of history and 3 seconds of future trajectory data.
\item \textbf{Argoverse 2 (2021)} \cite{wilson_argoverse2_2021}: The successor to Argoverse 1, featuring longer recordings (11 seconds in total, with 5 seconds of history and 6 seconds of future). This dataset includes about \SI{250}{k} scenes and provides richer agent types and map features than Argoverse 1.
\item \textbf{Waymo Open (2021)} \cite{ettinger_waymo_2021}: A large-scale and diverse motion dataset containing around \SI{576}{k} scenes. Each scene spans 9.1 seconds, including 1.1 seconds of history and 8 seconds of future trajectory. Compared to Argoverse 2, it provides more detailed agent features (e.g., agent size) and richer map features, such as traffic light information.
\end{itemize}

All three share a sampling rate of \SI{10}{\hertz}, which makes the fitting results comparable across datasets in Chapter \ref{cpt: Empirical Bayes Analysis} and enables the cross-dataset evaluation in Chapter \ref{cpt: improving generalization}. Argoverse 2 and Waymo Open additionally define a joint prediction task, as required in Chapter \ref{cpt: generative model}. Other datasets are not considered: nuScenes \cite{caesar_nuscenes_2020} provides annotated agent states at a lower sampling rate, the Interaction dataset \cite{zhan_interaction_2019} is considerably smaller in scale, and nuPlan \cite{caesar_nuplan_2021} targets planning rather than prediction. Shifts \cite{malinin_shifts_2021} addresses distribution shift within a single data collection pipeline, whereas the out-of-distribution protocol introduced in Chapter \ref{cpt: improving generalization} deliberately evaluates across two independently collected benchmarks.

\subsection{Prediction Competitions}
\label{sec 2.1.2: prediction competitions}
\begin{figure}[thb]
\centering
\includegraphics[width=\textwidth]{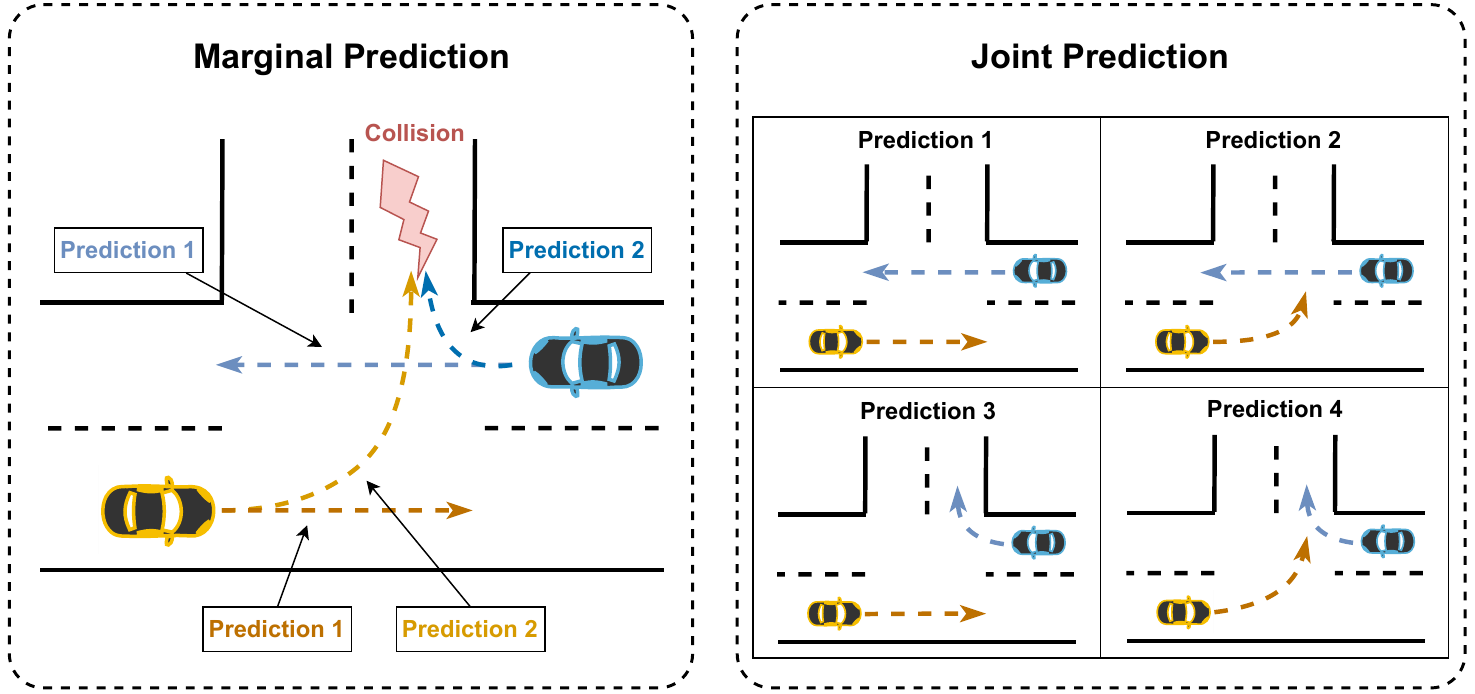}
\caption{Marginal and joint prediction}
\label{fig: marginal and joint prediction}
\end{figure}

Motion datasets providers often host corresponding prediction competitions to advance model development in this domain. In each recorded scene, among all agents $\mathcal{A}$, one or multiple \emph{target agents} $\mathcal{A}^{\text{ta}} \subseteq \mathcal{A}$ -- whose predictions are scored -- are designated by the dataset creator. Participating models are typically tasked with forecasting multiple plausible future trajectories -- along with their associated probabilities -- for each target agent, to capture the inherent uncertainty and multiple possibilities in future agent behavior. The traffic prediction task in competitions can be categorized into two main types:

\begin{itemize}[leftmargin=*]
\item \textbf{Marginal Prediction}: Forecasting individual agent trajectories without ensuring they collectively form consistent scenes. The interactions between agents during the prediction horizon are not explicitly modeled and can result in unrealistic predicted traffic scenes, such as collisions, as shown in the left panel of Figure \ref{fig: marginal and joint prediction}. 

\item \textbf{Joint Prediction}: Modeling multiple interacting agents simultaneously to predict coherent traffic scene continuations as shown in the right panel of Figure \ref{fig: marginal and joint prediction}. Despite its greater task complexity and computational demands, joint prediction requires models to capture the inter-dependencies between agents, where each agent's behavior influences and is influenced by others in the shared environment.
\end{itemize}

Table \ref{tab: dataset summary (chapter 2)} summarizes the prediction tasks in different datasets. Although marginal prediction simplifies the problem by considering the prediction of each agent independently, it is more popular and has been widely adopted in competitions due to its computational efficiency and easier evaluation framework. However, recent research has increasingly recognized the limitations of this approach, leading to a growing focus on joint prediction. This shift is reflected in newer competitions with joint prediction tasks, such as those in Argoverse 2 and Waymo Open, which emphasize modeling agent interactions and scene consistency.

\subsection{Evaluation Metrics}
\label{sec: evaluation metrics}
The development of evaluation frameworks has evolved alongside motion datasets and associated competitions. Early competitions primarily focused on marginal prediction tasks, which led to the design of metrics that assess the prediction accuracy of individual agents, rather than evaluating the overall consistency of predicted traffic scenes.

Before introducing the available evaluation metrics in detail, the following notations are defined for clarity:
\begin{itemize}[leftmargin=*]
\item The observed states of the $i$-th agent are denoted as $\mathbf{z}_i = [\mathbf{z}_{i,1}, \mathbf{z}_{i,2}, ..., \mathbf{z}_{i,T}]$, where $\mathbf{z}_{i,t}$ denotes the observed state of the $i$-th agent at timestep $t$, and $T$ denotes the total number of timesteps. Depending on the dataset, the observed state $\mathbf{z}_{i,t}$ can include position, heading, and velocity information, as presented in Table \ref{tab: dataset summary (chapter 2)}. In this work, unless otherwise specified, $\mathbf{z}_{i,t}$ represents only the 2D-position coordinates.

\item Let $t=t_0$ be the reference timestep defined by the dataset provider, which segments the scene into history and future. The historical and future positions of the $i$-agent are thus denoted as:
\begin{itemize}[]
\item $\mathbf{z}_i^{\mathrm{hist}} = [\mathbf{z}_{i,1}, \mathbf{z}_{i,2}, ..., \mathbf{z}_{i, t_0}]$.
\item $\mathbf{z}_i^{\mathrm{fut}} = [\mathbf{z}_{i, t_0+1}, \mathbf{z}_{i, t_0+2}, ..., \mathbf{z}_{i, T}]$.
\end{itemize}

\item Since the prediction model outputs multiple future trajectories for each agent, the $k$-th predicted trajectory with positions for the $i$-th agent is denoted as $\hat{\mathbf{z}}_{i,k} = [\hat{\mathbf{z}}_{i, t_0+1, k}, \hat{\mathbf{z}}_{i, t_0+2, k}, ..., \hat{\mathbf{z}}_{i, T, k}]$, where $\hat{\mathbf{z}}_{i,t, k}$ is the predicted position at timestep $t$  for the $k$-th prediction. 

\item In marginal prediction, the model is required to output the probability $\pi_{i, k}$ for each individual predicted trajectory of each agent. In contrast, in joint prediction, the model outputs a single probability $\pi_{k}$ for each predicted traffic scene as a whole.
\end{itemize}

\begin{figure}[thb]
\centering
\includegraphics[width=\textwidth]{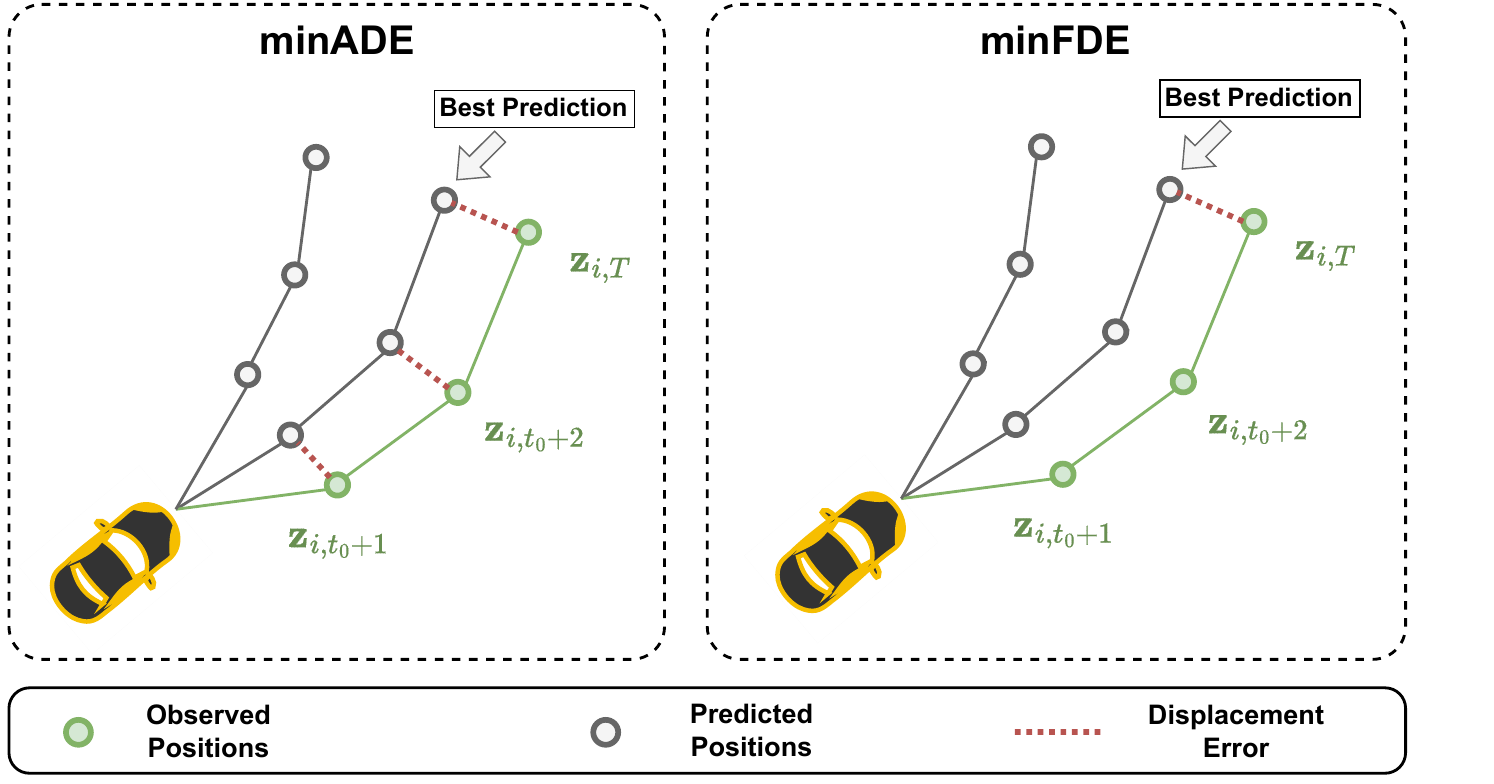}
\caption[Minimum average and final displacement error]{Minimum average displacement error (minADE) and minimum final displacement error (minFDE).}
\label{fig: minADE and minFDE}
\end{figure}

To assess prediction performance, a commonly used approach is to measure the Euclidean distance between the predicted positions $\hat{\mathbf{z}}_{i,t,k}$ and the observed future positions of agents\footnote{
Throughout this dissertation, the term ``observed'' refers to recorded measurements provided by the dataset. These measurements
may contain noise and do not represent noise-free ground-truth motion.
} $\mathbf{z}_{i,t}$. In the remainder of this dissertation, this Euclidean distance is referred to as the \emph{displacement error}. This approach exhibits multiple variants in \emph{marginal prediction}:

\begin{itemize}[leftmargin=*]
\item \textbf{minimum Average Displacement Error ($\text{minADE}_K$)} calculates the Euclidean distance between the observed future trajectory and the best of $K$ predicted trajectories as an average of all future timesteps. This metric is expressed as:

\begin{equation}
\text{minADE}_K = \frac{1}{A (T - t_0)} \sum_{i=1}^{A} \left( \min_{k \in \{1, ..., K\}}  \sum_{t = t_0 + 1}^{T} \left\| \hat{\mathbf{z}}_{i,t,k} - \mathbf{z}_{i,t} \right\|_2 \right)
\end{equation}

where $A$ denotes the number of considered agents in a scene. 

In practice, the number of prediction modes $K$ is commonly set to $K=1$ and $K=6$ to evaluate single-best and multi-modal prediction performance \cite{chang_argoverse_2019, wilson_argoverse2_2021, ettinger_waymo_2021}. When a model outputs more than $K$ predictions, the $K$ predictions with the highest predicted probabilities are selected for evaluation.


\item \textbf{minimum Final Displacement Error ($\text{minFDE}_K$)} focuses solely on the
displacement error at the final timestep $T$ from $K$ best predicted trajectories, emphasizing long-term performance. It is expressed as:

\begin{equation}
\text{minFDE}_K = \frac{1}{A} \sum_{i=1}^{A} \left( \min_{k \in \{1, ..., K\}} \left\| \hat{\mathbf{z}}_{i,T,k} - \mathbf{z}_{i,T} \right\|_2 \right)
\end{equation}

\item \textbf{Miss Rate (MR)} calculates how frequent the prediction ``miss'' the observed trajectory. It defines a miss as the state when none of the individual $K$ predictions for an agent are within a given distance threshold of the observed trajectory at a given timestep $t$. The distance threshold can either be a fixed radius (2 meters), as used in Argoverse 2 \cite{wilson_argoverse2_2021}, or split into longitudinal and lateral components that scale with the agent’s velocity, as in Waymo Open \cite{ettinger_waymo_2021}.
\end{itemize}

These metrics proposed for marginal prediction primarily reward predictions that are positionally close to the observed future trajectory of each agent. Recently, as joint prediction has gained more attention in the research community, new metrics have been introduced to evaluate performance in \emph{joint prediction} tasks specifically: 

\begin{itemize}[leftmargin=*]
\item \textbf{minimum Scene Average Displacement Error ($\text{minSADE}_K$)} differs from $\text{minADE}_K$ primarily in how the best prediction is determined. While $\text{minADE}_K$ selects the best trajectory for each agent independently, the $\text{minSADE}_K$ considers the best prediction to be the one that minimizes the overall displacement error across all agents in a scene. This metric is expressed as:

\begin{equation}
\text{minSADE}_K = \min_{k \in \{1, ..., K\}} \left( \frac{1}{A (T - t_0)} \sum_{i=1}^{A} \sum_{t = t_0 + 1}^{T} \left\| \hat{\mathbf{z}}_{i,t,k} - \mathbf{z}_{i,t} \right\|_2 \right)
\end{equation}

where $A$ denotes the number of considered agents in a scene.

\item \textbf{minimum Scene Final Displacement Error ($\text{minSFDE}_K$)} is similar to $\text{minSADE}_K$ with the focus on the the final timestep. It is expressed as:

\begin{equation}
\text{minSFDE}_K = \min_{k \in \{1, ..., K\}} \left( 
\frac{1}{A} \sum_{i=1}^{A}  \left\| \hat{\mathbf{z}}_{i,T,k} - \mathbf{z}_{i,T} \right\|_2 \right)
\end{equation}

\item \textbf{Drivable Area Compliance (DAC)} measures the percentage of an agent's predicted positions that fall outside the drivable area defined by the map.

\item \textbf{Collision Rate (CR)} quantifies the proportion of predicted collisions between agents over time within a traffic scene.

\end{itemize}

As seen above, the metrics for joint prediction (minSADE and minSFDE) are natural extensions of their marginal counterparts (minADE and minFDE), adapted to evaluate multi-agent scenes by considering the best prediction set across all agents. Beyond trajectory accuracy, joint prediction introduces additional metrics such as DAC and CR to assess scene-level consistency -- capturing whether predicted agent interactions respect collision constraints and temporal coordination. These scene consistency metrics reflect a fundamental shift in evaluation philosophy: from measuring individual trajectory quality to assessing the plausibility of predicted multi-agent futures as coherent, interactive scenarios.

In summary, motion datasets -- together with their associated competition protocols and evaluation metrics -- play a defining role in shaping the design and development of prediction models. Their structure, coverage, and scoring procedures implicitly guide what models learn to prioritize, influencing both architectural choices and training objectives. Understanding these characteristics and their limitations is therefore essential for developing prediction systems that are generalizable and truly reflective of real-world driving behavior.






\section{Advancements in Model Development}
\label{sec: 2.2 advancements in model}
Early approaches to motion prediction propagate agent states with a kinematic
motion model, or first estimate a
discrete maneuver -- using formalisms such as hidden Markov models \cite{berndt_continuous_2008} and (dynamic)
Bayesian networks~\cite{gindele_probabilistic_2010} -- and
predict a trajectory conditioned on it \cite{lefevre_survey_2014}. Such models
are interpretable and computationally inexpensive. However, kinematic models extrapolate low-level dynamics and are therefore reliable only over short horizons, while maneuver-based models are restricted to a manually specified set of maneuvers whose parameters are difficult to tune in dense urban scenes. In both cases, interactions between agents are captured only to a limited extent.

The availability of large-scale datasets (cf. Section \ref{sec: advancements in motion datasets})
has shifted the field towards learning motion behavior, including agent
interactions, directly from data. Recent surveys provide a detailed overview of this development \cite{huang_survey_2022, karle_scenario_2022}. 

The remainder of this section therefore focuses on learning-based approaches, covering data representations, model architectures, and generative models.

\subsection{Data Representation}
\label{sec: 2_2_1_data_representation}
The choice of data representation plays a critical role in the performance and efficiency of traffic scene prediction models. Different representations encode motion and map information in distinct ways, affecting how well a model can generalize, interpret interactions, and generate feasible predictions. Broadly, data representations can be categorized into rasterization-based, sequence-based (vectorization), and parametric representations.

\begin{figure}[thb]
\centering
\includegraphics[width=\textwidth]{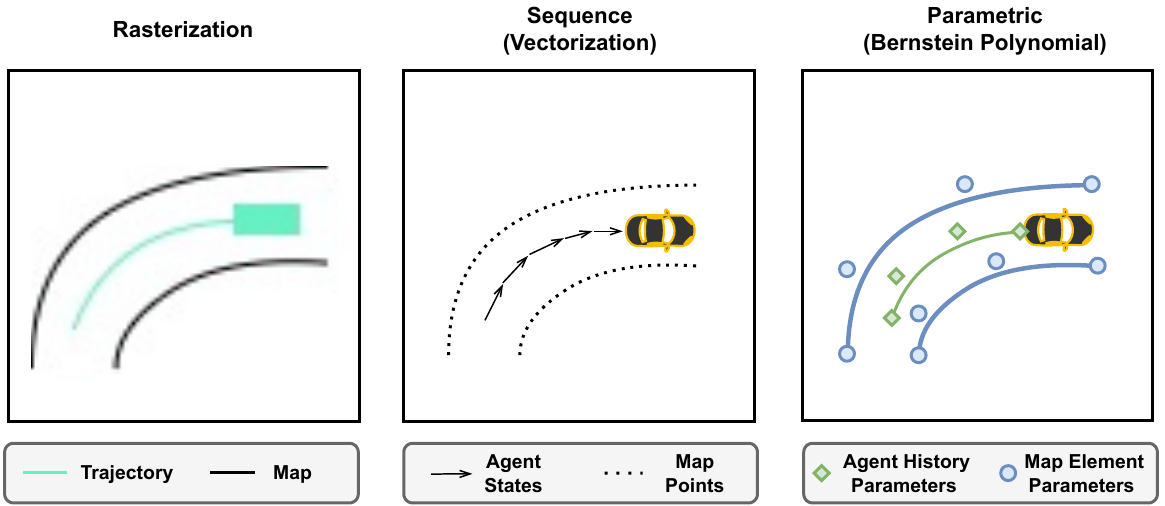}
\caption{Data representations for agent history trajectory and map}{Data representations for agent history trajectory and map.}
\label{fig: data representation}
\end{figure}

\subsection*{Rasterization-Based Representations}
Rasterization-based representations treat the prediction task as a computer vision (CV) problem by rendering the surrounding environment within a specific range of the ego vehicle into ``bird's-eye view'' (BEV) image grids, as shown in the left panel of Figure \ref{fig: data representation}. These grids typically include rich contextual information, such as road layouts and agent motion. Several notable works, including CoverNet \cite{phan_covernet_2020} and TrajectoryCNN \cite{liu_trajectorycnn_2020}, adopt this approach and apply Convolutional Neural Networks (CNNs) \cite{krizhevsky_imagenet_2012} to extract spatial and temporal features from these rasterized inputs.

While rasterization offers a comprehensive way to integrate heterogeneous data sources, it also comes with several limitations:

\begin{itemize}[leftmargin=*]
\item High computational cost and discretizations: Storing and processing image-like inputs is resource-intensive. Moreover, the data dimensionality scales quadratically with both the grid resolution and the spatial extent around the ego vehicle. This typically forces the use of lower-resolution grids with limited range, leading to high discretization artifacts.
\item Viewpoint sensitivity: Rasterized views are usually centered on the ego vehicle and aligned with its heading, causing the traffic scene to rotate accordingly. While this ego-centric projection ensures a consistent viewpoint for the ego vehicle, it introduces unnecessary rotational variance for predicting surrounding agents' motion -- a challenge that standard CNNs struggle to model efficiently \cite{worrall_harmonic_2017, marcos_rotation_2017}.
\item Sparse and redundant input space: A significant portion of the grid often contains empty or irrelevant areas, introducing uninformative inputs and increasing the learning difficulty for the model.
\end{itemize}


\subsection*{Sequence-Based Representations}
Recent SotA models \cite{liang_learning_2020, zhou_query_2023, cheng_forecast_2023, seff_motionlm_2023, jiang_motiondiffuser_2023, wang_optimizing_2025} employ the sequence-based representation, also known as \emph{vectorization} \cite{gao_vectornet_2020}, for agents' trajectories and map elements, as shown in the middle panel in Figure \ref{fig: data representation}. Unlike rasterization, which renders scenes into image grids, the sequence-based approach aligns directly with the structure of motion datasets. It represents agent motion through measured sequences of position, heading, velocity, and acceleration at discrete time steps, while road geometry is captured as sampled positional points.

Compared to rasterization, this format is significantly more efficient and leverages the structured nature of the data to effectively capture temporal dynamics and contextual information. However, despite its advantages, the sequence-based representation remains discretized and introduces several limitations:

\begin{itemize}[leftmargin=*]
\item Temporal inconsistency: While flexible, sequence-based data can also reflect measurement noise and jitter, leading to physically implausible trajectory segments. These inconsistencies may propagate through the model, leading to predicted trajectories that violate physical constraints.

\item Data dimensionality: The dimensionality of sequence-based representations scales with the temporal resolution, spatial sampling density, and the length of agent trajectories or map elements. As a result, longer or more detailed sequences lead to increased computational cost and memory usage, making real-time inference more challenging -- especially in dense or complex scenes.

\item Cross-dataset compatibility: Sequence-based representations are tightly coupled to the temporal resolution of agent trajectories and the spatial sampling of map features, which, in turn, depend on the specific perception setup and map format of a dataset. As a result, transferring models across datasets with differing formats -- such as \SI{2}{Hz} sampling in nuScenes \cite{caesar_nuscenes_2020} versus \SI{10}{Hz} in Waymo Open \cite{ettinger_waymo_2021} -- often requires resampling or interpolation, introducing additional complexity and potential sources of error.
\end{itemize}

\subsection*{Parametric Representations}
Unlike discrete representations such as rasterized grids or temporal sequences, parametric representations are continuous and describe trajectories and map elements as combinations of basis functions. Depending on how these basis functions are constructed, parametric representations can be broadly categorized into two approaches.

A common approach employs \emph{analytic} basis functions \cite{huang_uncertainty_2019, buhet_plop_2020, su_temporally_2021}, where the functions are explicitly formulated based on known variables -- such as time or spatial coordinates. Alternatively, \emph{data-driven} methods derive basis functions directly from a given dataset, such as through Principal Component Analysis (PCA) \cite{jiang_motiondiffuser_2023}. While effective in capturing patterns in the data, these bases are inherently data-dependent, as they are tailored to the empirical data distribution. Consequently, they may inherit dataset-specific biases, potentially limiting generalization to unseen traffic scenes. Since this contradicts our goal of developing a robust and transferable prediction system, we therefore do \emph{not} focus on data-driven bases in this dissertation.

Representative examples of the analytic approach are \emph{polynomial} representations. In particular, positions along agent trajectories or map elements (e.g., lane centerlines) can be represented as linear combinations of polynomial basis functions. Formally, this can be written as:
\begin{equation*}
    \mathbf{c}(\tau) = \sum_{n=0}^N  \phi_n(\tau)\mathbf{w}_n
\end{equation*}
where \( \mathbf{c}(\tau) \in \mathbb{R}^2 \) is the sampled point along the curve parameterized by the variable \( \tau \), expressed in the $x$- and $y$-coordinates; \( \mathbf{w}_n \in \mathbb{R}^2 \) are polynomial parameters; \( \phi_n(\tau) \) are basis functions (e.g., monomial or Bernstein bases). The parameter $N$ denotes the model complexity, corresponding to the polynomial degree.

Another representative analytic approach is the use of splines (e.g., B-splines), i.e., piecewise polynomials. Splines offer greater expressiveness at a given degree through their piecewise definition with local support, which also improves numerical stability at higher degrees. This flexibility, however, requires selecting the number and placement of knots, introducing a model-selection task beyond the choice of degree. This work therefore adopts a single-segment polynomial representation, leaving the polynomial degree as the sole model-selection parameter. The additional flexibility of splines is left to future work.

Parametric representations significantly reduce data dimensionality by encoding an entire trajectory or map element using only a few parameters. These parameters are physically meaningful -- for example, the parameters of Bernstein polynomials are known as \emph{control points}, which define the convex hull of the curve, as visualized in the right panel in Figure \ref{fig: data representation}. Moreover, parametric representations inherently satisfy continuity of trajectories in position, velocity, and acceleration \cite{su_temporally_2021}, while also regularizing measurement noise through priors (cf. Chapter \ref{cpt: Empirical Bayes Analysis}) -- both of which are common issues in discretized formats.

Despite these advantages, parametric representations are still underexplored in the field, largely due to the following limitations:

\begin{itemize}[leftmargin=*]
\item Additional pre-processing: Most motion datasets provide trajectory and map data in the sequence-based format. To adopt parametric representations, additional steps -- such as curve fitting or trajectory smoothing -- are required to convert sequence data into parametric form.
\item Parametric model selection ambiguity: Parametric representations are approximations of real-world motion. Several key questions remain under-researched, including the trade-off between model complexity and approximation error, and how to determine the optimal model complexity (e.g., polynomial degree) for capturing diverse trajectory behaviors. As a result, recent studies typically select the polynomial degree for trajectories empirically, without a principled investigation \cite{huang_uncertainty_2019, buhet_plop_2020, su_temporally_2021}.
\end{itemize}

As summarized in Table \ref{tab: data representation summary (chapter 2)}, each representation method offers distinct trade-offs: rasterization enables unified spatial encoding but suffers from high memory and resolution issues; sequence-based formats align well with dataset structures but are prone to measurement noise and cross-dataset inconsistencies; parametric representations offer compact, physically consistent modeling at the cost of preprocessing and approximation challenges.

\begin{table}[thb]
\caption{Comparison of data representations}
\centering
\begin{tabularx}{\textwidth}{>{\raggedright\arraybackslash}p{3.0cm} |>{\centering\arraybackslash}X |> {\centering\arraybackslash}X |>{\centering\arraybackslash}X}
\toprule
Criteria & Rasterization & Sequence & Parametric \\
\hline
 & \fullstar\emptystar\emptystar  & \fullstar\fullstar\emptystar & \fullstar\fullstar\fullstar \\
Computational Efficiency& Scales quadratically with grid resolution and spatial extent. &  Scales with resolution and sequence length. & Requires only a small set of parameters.\\
\hline
 & \fullstar\fullstar\emptystar & \fullstar\fullstar\fullstar & \fullstar\fullstar\emptystar \\
Expressiveness& Limited by grid resolution. & Preserves exact measurements, including noise. & Approximates motion and map. \\
\hline
& \fullstar\emptystar\emptystar & \fullstar\fullstar\emptystar & \fullstar\fullstar\fullstar\\
Temporal Consistency and Continuity & Temporal consistency not guaranteed. & Sensitive to noise; continuity not enforced. & Inherently ensures continuity in agent kinematics \\
&&&\\
\hline
& \fullstar\emptystar\emptystar & \fullstar\fullstar\fullstar & \fullstar\fullstar\emptystar \\
Pre-processing Simplicity  & Requires rendering scene data into the grid format. & Uses structured data formats directly from the dataset. & Requires curve fitting or smoothing to derive parameters. \\

\bottomrule
\end{tabularx}
\label{tab: data representation summary (chapter 2)}
\end{table}

Although parametric representations are less commonly employed in recent state-of-the-art models and several open questions remain, their benefits in ensuring motion continuity and potential for real-time prediction systems provide strong motivation to explore their use in prediction tasks. Given these advantages and the identified challenges -- particularly the parametric model selection ambiguity -- this dissertation focuses specifically on polynomial representations, particularly Bernstein polynomials, to systematically address the optimal model complexity selection and quantify representation fidelity. Therefore, Chapter \ref{cpt: Empirical Bayes Analysis} directly addresses these questions through an empirical Bayes analysis of polynomial trajectory representations.

\subsection{Environment Encoding and Modeling}
\label{sec 2.2.2: Environment Encoding and Modeling}
Following the discussion on data representation, the next step in the modeling pipeline is to encode the environment -- including agent dynamics and map context -- into a format suitable for learning. The quality of this encoding directly affects the model’s ability to reason about agent interactions and the structure of the surrounding map. Since agent motion history and the map geometry are among the most informative features for traffic scene prediction \cite{makansi_you_2021}, two neural architectures have been widely adopted for encoding this information: Convolutional Neural Networks (CNNs) \cite{krizhevsky_imagenet_2012} and attention mechanisms from the Transformer architecture \cite{vaswani_attention_2017}.

\subsubsection{Convolutional Neural Network (CNNs)}
\begin{figure}[thb]
\centering
\includegraphics[width=\textwidth]{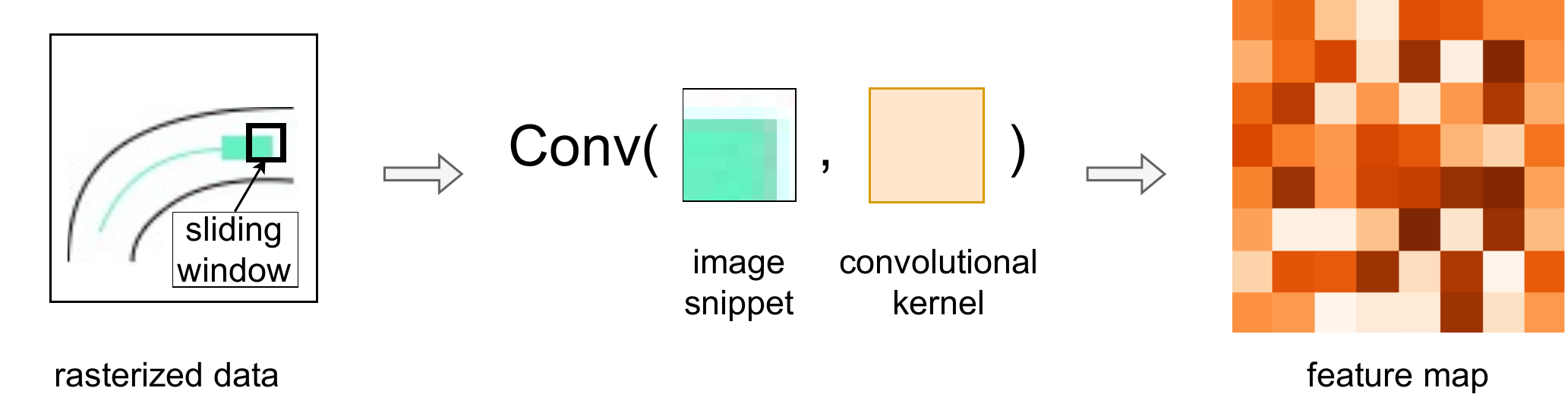}
\caption{Feature extraction in convolutional layers}
\label{fig: CNN}
\end{figure}

CNNs gained widespread popularity after AlexNet's \cite{krizhevsky_imagenet_2012} breakthrough performance in 2012. They have been increasingly employed in trajectory prediction studies since around 2020 \cite{chai_multipath_2019, phan_covernet_2020, liu_trajectorycnn_2020} for encoding environmental information. They are commonly associated with rasterization-based representations, where the traffic scene -- including map context and agent motion history -- is rendered as image grids. These models typically consist of multiple convolutional layers, each applying a set of kernels (filters) to capture local patterns within the grid. By sliding these kernels over the image, the full input is transformed into feature maps, a process illustrated in Figure \ref{fig: CNN}. Through multiple convolutional layers organized in hierarchical architectures, CNNs can efficiently extract spatial features across different levels of abstraction, enabling the model to interpret complex spatial relationships within the traffic scene.

Despite their strong performance on computer vision tasks, CNNs exhibit notable limitations in traffic scene prediction tasks. They rely on local convolution operations, resulting in fixed receptive fields that limit their ability to capture long-range dependencies and subtle interactions between distant agents -- a critical requirement for understanding complex traffic dynamics. This limitation is further amplified by the constrained spatial extent of rasterized representations, as distant but potentially relevant scene elements may fall entirely outside the rendered image. Moreover, the absence of explicit agent kinematics and the viewpoint sensitivity introduced by ego-centric rasterization (cf. Section \ref{sec: 2_2_1_data_representation}) further increases the learning complexity for CNN-based architectures.

To overcome these limitations, recent research has increasingly turned to attention mechanisms, which offer greater flexibility and global reasoning capabilities.

\subsection*{Attention Mechanism}
The attention mechanism \cite{bahdanau_neural_2015}, first introduced in 2015, was further refined and popularized as the central component of the Transformer architecture \cite{vaswani_attention_2017}, which revolutionized sequence modeling tasks in natural language processing (NLP). Compared to its predecessors -- such as recurrent neural networks (RNNs) and Long Short-Term Memory networks (LSTMs) \cite{hochreiter_long_1997} -- the attention mechanism excels at capturing long-range dependencies within data sequences and enables parallel processing, significantly improving both modeling capacity and computational efficiency. 

Recent studies have successfully employed the attention mechanism in the traffic scene prediction task, achieving SotA performance in multiple prediction competitions \cite{zhou_hivt_2022, zhou_query_2023, shi_mtr++_2023}. The attention mechanism offers two major advantages in this domain: First, in contrast to CNNs, which rely on fixed receptive fields and image-based inputs, the attention mechanism in SotA models are typically coupled with sequence-based representations and is more efficient at encoding complex interactions between multiple agents and map features over a wide spatial range. Second, attention mechanism enables each element to attend to all others, regardless of their position in the sequence. This permutation invariance\footnote{A function or model is permutation invariant if its output does not change when the input elements are reordered.} is particularly valuable for modeling interactions between agents and map elements, which typically lack a predefined sequential order.



The following provides a brief introduction to the attention mechanism for traffic scene prediction tasks.

\begin{figure}[thb]
\centering
\includegraphics[width=\textwidth]{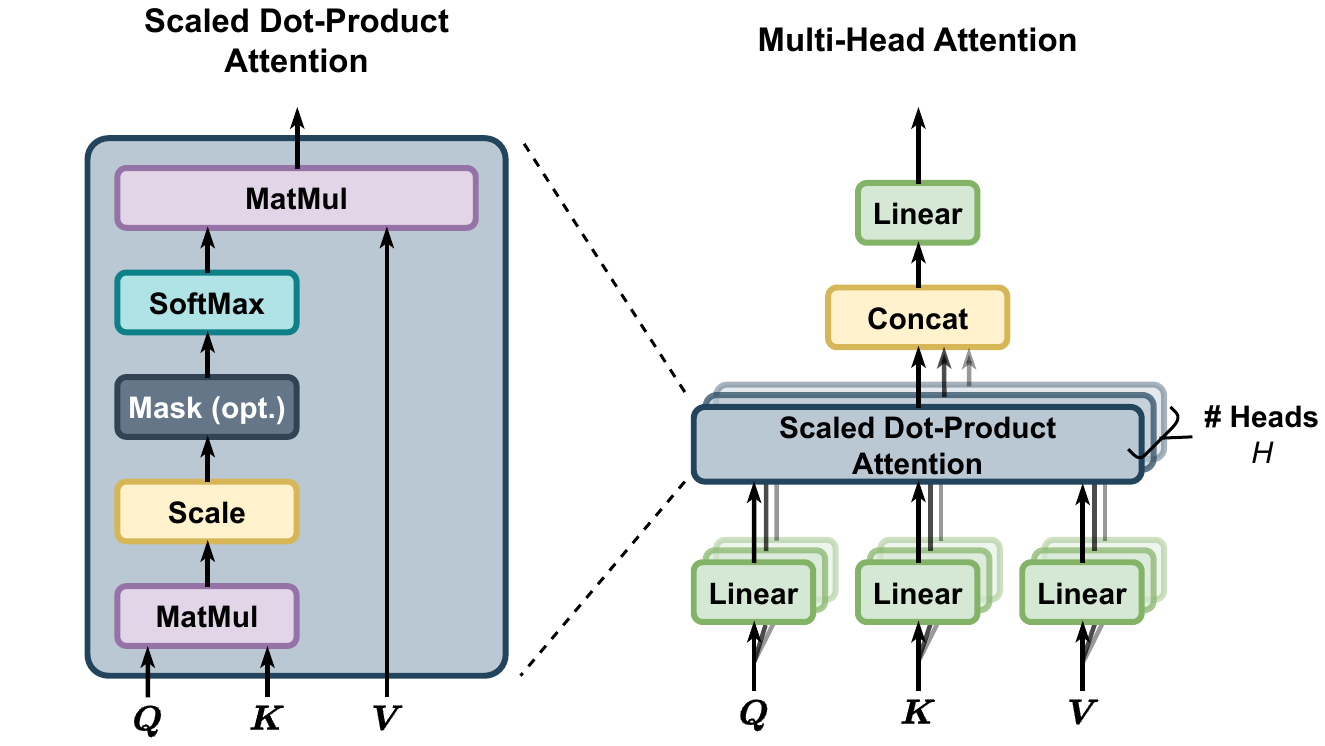}
\caption[Scaled dot-product attention and multi-head attention]{Scaled dot-product attention and multi-head attention \cite{vaswani_attention_2017}}
\label{fig: attention}
\end{figure}

In the context of traffic scene prediction, the input to the attention mechanism consists of \emph{tokens}, which are learned embeddings that can represent various elements of the traffic scene. Depending on the specific model implementation, a token might encode an agent's state at a specific timestep, an entire agent trajectory, or a map element such as a lane segment. 

The input tokens are typically segmented in query $\boldsymbol{Q}$, key $\boldsymbol{K}$, and value $\boldsymbol{V}$. Intuitively:
\begin{itemize}[leftmargin=*]
\item A query represents what a given token is ``looking for''.
\item A key describes the content of each token to be attended to.
\item A value contains the information that will be passed forward if the key is attended to.
\end{itemize}
The attention weights are computed by assessing the similarity between each query and key using a scaled dot-product operation:
\begin{equation}
\label{eqn: attention}
    \text{Attention}(\boldsymbol{Q}, \boldsymbol{K}, \boldsymbol{V}) = \text{softmax}\left(\frac{\boldsymbol{Q}\boldsymbol{K}^{\top}}{\sqrt{D^{\text{key}}}}\right)\boldsymbol{V}
\end{equation}
here $D^{\text{key}}$ is the dimensionality of the key vectors used for scaling.

Depending on how the query, key, and value tokens are constructed, the attention mechanism can operate in different modes.
\begin{itemize}[leftmargin=*]
 \item \textbf{Self-attention}: The queries, keys, and values are derived from the \emph{same} input sequence. This allows each token to attend to all other tokens in the sequence. In traffic scene prediction, self-attention is commonly used to model interactions between agents, relationships among map elements, or temporal dependencies within an agent's own trajectory.
 \item \textbf{Cross-attention}: The queries are derived from a different input source than the keys and values. This is typically used to model agent-map interactions in SotA models \cite{nayakanti_wayformer_2022, zhou_query_2023, cheng_forecast_2023}, where agent features serve as queries and map elements provide the keys and values. This enables the model to condition its predictions on the static context -- such as how road geometry influences an agent's motion.
\end{itemize}
These attention mechanisms can be layered and combined to build a hierarchical understanding of the scene, enabling the model to reason jointly over agent dynamics and structured map information. Attention masks are often applied during this process to restrict the attention scope -- for example, by spatial distance or token type -- ensuring that the model focuses on semantically relevant interactions.

While a single attention mechanism allows each token to attend to others in the input, it may be limited in the types of relationships it can capture. Multi-head attention addresses this by applying multiple attention operations in parallel, as visualized in the right panel in Figure \ref{fig: attention}. This allows the model to attend to different subspaces of the input features and capture diverse interaction patterns simultaneously.



\begin{figure}[thb]
\centering
\includegraphics[width=\textwidth]{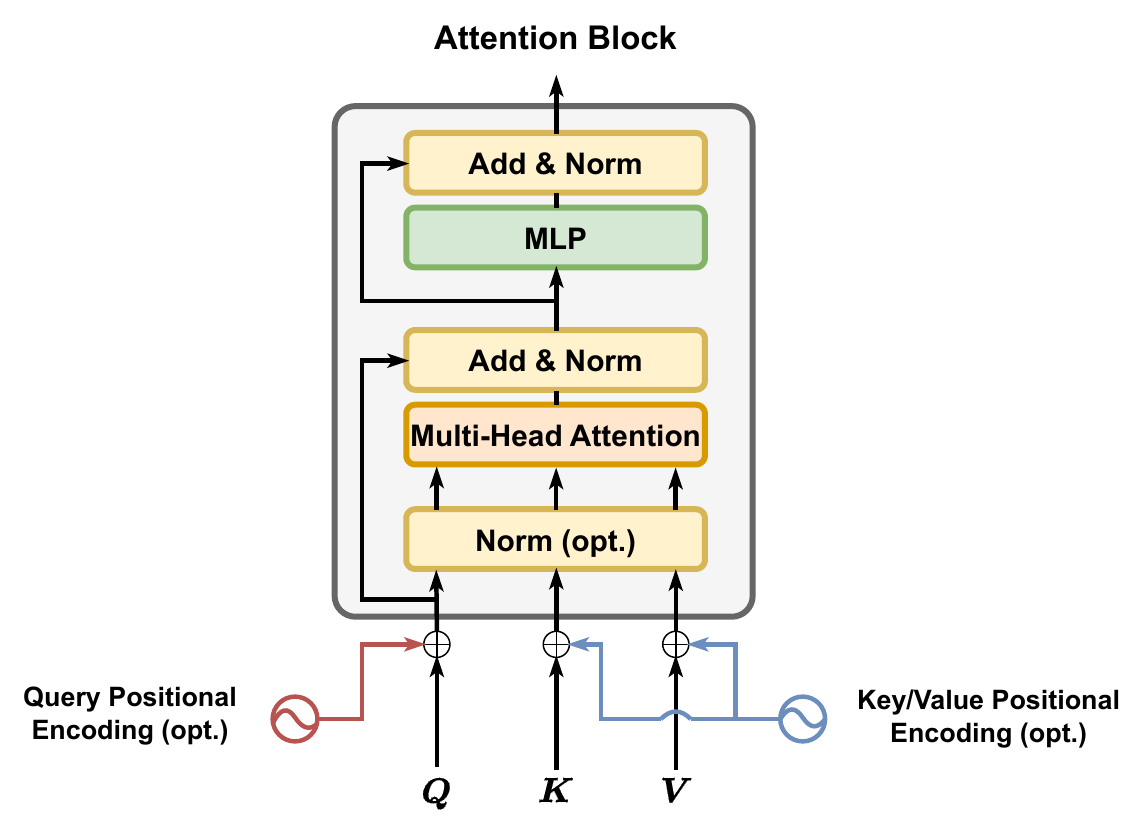}
\caption[Architecture of attention block]{Architecture of attention block \cite{vaswani_attention_2017}}
\label{fig: attention block}
\end{figure}

The multi-head attention, together with multi-layer perceptrons (MLP), residual connections \cite{he_deep_2016}, and layer normalization \cite{ba_layer_2016} forms the foundation of the Transformer architecture \cite{vaswani_attention_2017}. In traffic scene prediction, researchers often adapt these core components into modular \emph{attention blocks} \cite{zhou_query_2023, shi_mtr++_2023}, conceptually analogous to the encoder module of the Transformer architecture \cite{vaswani_attention_2017}, as shown in Figure \ref{fig: attention block}. These blocks can be flexibly stacked to build deeper models and support both self-attention and cross-attention operations. Since the attention is inherently permutation-invariant, modeling ordered inputs -- such as time series of measured agent states -- requires positional encoding to inject information about the relative or absolute positions of tokens in a sequence.

While the attention mechanism offers strong modeling capacity, it also has notable limitations. In particular, the computational complexity of self-attention scales quadratically with the number of input tokens, resulting in high computational and memory costs -- especially in dense traffic scenes involving many interacting agents and map elements. Moreover, high-capacity attention mechanisms are data-hungry and may struggle to generalize when training data is limited or lacks scene diversity \cite{touvron_training_2021, khan_transformers_2022}.

Another practical challenge arises from the varying number of agents and map elements across scenes, which requires additional pre-processing to batch the inputs into fixed sizes during training. Consequently, complex scenes often need to be truncated, limiting the number of agents and map tokens that can be processed. In contrast, simpler scenes require padding with non-existent entities, leading to computational inefficiencies.


Despite these limitations, the high flexibility and scalability have made attention-based architectures a dominant choice in recent traffic scene prediction models. Notably, their application has expanded beyond deterministic prediction: recent works have integrated attention mechanisms into generative frameworks, such as auto-regressive models and diffusion-based models, which aim not merely to learn a direct mapping from input to output, but to model the underlying distribution of future trajectories. This shift reflects a growing emphasis on capturing the uncertainty and multi-modality inherent in traffic scene prediction. As a result, attention-based models are increasingly used not only to encode interaction dynamics, but also to generate diverse and plausible future behaviors.

In this work, attention-based architectures serve as the backbone for both benchmark models and proposed methods presented in Chapter \ref{cpt: improving generalization} and \ref{cpt: generative model}.

\subsection{Generative Models}
\label{sec: generative model introduction}
Beyond improvements in data representation and architectural design, recent research has also reconsidered the underlying modeling paradigm for traffic scene prediction.

Since motion datasets provide one observed future for each traffic scene, earlier approaches predominantly framed trajectory prediction as a \emph{regression} task \cite{shi_mtr++_2023, zhou_query_2023, cheng_forecast_2023} -- learning a direct mapping from past observations to one or several likely future outcomes, often by minimizing a loss with respect to a single observed future trajectory for each agent.  While most of these models allow for multi-modal outputs to fulfill requirements from motion prediction competitions, they typically do not model the full distribution of possible futures and often collapse to predict the average or most likely trajectories \cite{gupta_social_2018, bahari_vehicle_2022}.

To better capture the uncertainty of traffic scene continuation, recent research has shifted toward \emph{generative models}, which aim to learn the full distribution over possible future continuations conditioned on the observed past and scene context. Generative models enable sampling from the modeled distribution -- enabling the generation of diverse, plausible future scenes that are critical for robust decision-making and planning in autonomous systems. 

Earlier generative approaches typically address the \emph{marginal prediction} task and have included methods based on variational autoencoders (VAEs) \cite{lee_desire_2017,salzmann_trajectron_2020} and generative adversarial networks (GANs) \cite{gupta_social_2018, sadeghian_sophie_2019, roy_vehicle_2019}, both of which introduced stochasticity into prediction by modeling a distribution over agent future trajectories. VAEs learn a latent space that represents possible future behaviors, regularized by a prior distribution; sampling from this space and conditioning on past motion and context yields diverse trajectory predictions. GANs take a different approach, generating realistic trajectories by training a generator to fool a discriminator. Despite their success in producing diverse outputs, these methods often suffer from practical limitations: VAEs can experience posterior collapse\footnote{Posterior collapse refers to a phenomenon in VAEs where the decoder ignores the latent variable and relies entirely on the conditioning input, effectively making the latent space uninformative.}  \cite{lucas_understanding_2019}, and GANs can be unstable to train and prone to mode collapse\footnote{Mode collapse occurs when the model produces limited diversity in outputs, collapsing to a few modes of the target distribution and ignoring others.} \cite{salimans_improved_2016}.

Inspired by the success of \emph{auto-regressive models} in natural language processing and \emph{diffusion models} \cite{ho_denoising_2020, song_denoising_2021} in image and video generation, recent work in trajectory prediction has begun to explore these two generative paradigms. In parallel with the rising emphasis on the \emph{joint prediction} task, as outlined in Section~\ref{sec 2.1.2: prediction competitions}, these approaches offer more expressive and flexible frameworks for modeling complex, multi-agent interactions and scene dynamics.

\subsection*{Autoregressive Model}
Auto-regressive models have achieved remarkable success in natural language processing (e.g., GPT \cite{brown_language_2020}) and have recently gained traction in traffic scene prediction tasks \cite{seff_motionlm_2023, wu_smart_2024, zhou_behaviorgpt_2024}. These models frame an agent’s future trajectory as a sequence of tokens and reformulate the prediction task as iterative next-token prediction, where each step is conditioned on all previously generated tokens and the observed scene context. 

Recent studies have adopted the GPT-style next-token prediction by framing the token generation as a classification task. In this setting, the model selects the next token from a predefined set of motion primitives \cite{seff_motionlm_2023, wu_smart_2024, zhou_behaviorgpt_2024, jia_amp_2024}. This sequential formulation allows auto-regressive models to capture complex temporal dependencies and inter-agent interactions, enabling fine-grained temporal coherence and flexible, multi-modal trajectory generation through iterative sampling.

Despite their expressiveness, autoregressive models suffer from two key limitations. First, they are inherently sequential, meaning that predictions cannot be parallelized during inference, making them less efficient in real-time settings. For instance, MotionLM \cite{seff_motionlm_2023} first generates low-density agent states at \SI{0.5}{\second} intervals and then interpolates intermediate states to mitigate this bottleneck. Second, they are prone to error accumulation, where small mistakes in early steps can compound and lead to unrealistic or inconsistent long-term predictions \cite{zhang_closed_2025}. These issues have motivated the exploration of alternative generative frameworks, such as diffusion models, which decouple temporal coherence from inference order and offer improved stability and diversity in sampling.


\subsection*{Diffusion Model}
\label{sec: diffusion model}
Diffusion probabilistic models were first introduced by Sohl-Dickstein et al. \cite{sohl_deep_2015}, laying the theoretical foundation for a class of generative models based on iterative denoising. This framework was later extended and popularized through the development of the Denoising Diffusion Probabilistic Model (DDPM) \cite{ho_denoising_2020}, which offers a scalable and effective training and sampling procedure. Diffusion models have recently emerged as a powerful class of generative models, achieving state-of-the-art performance in domains such as image and video synthesis \cite{rombach_high_2022, hu_gaia_2023}, and more recently, in traffic scene generation and prediction tasks \cite{jiang_motiondiffuser_2023, wang_optimizing_2025}.

\begin{figure}[thb]
\centering
\includegraphics[width=\textwidth]{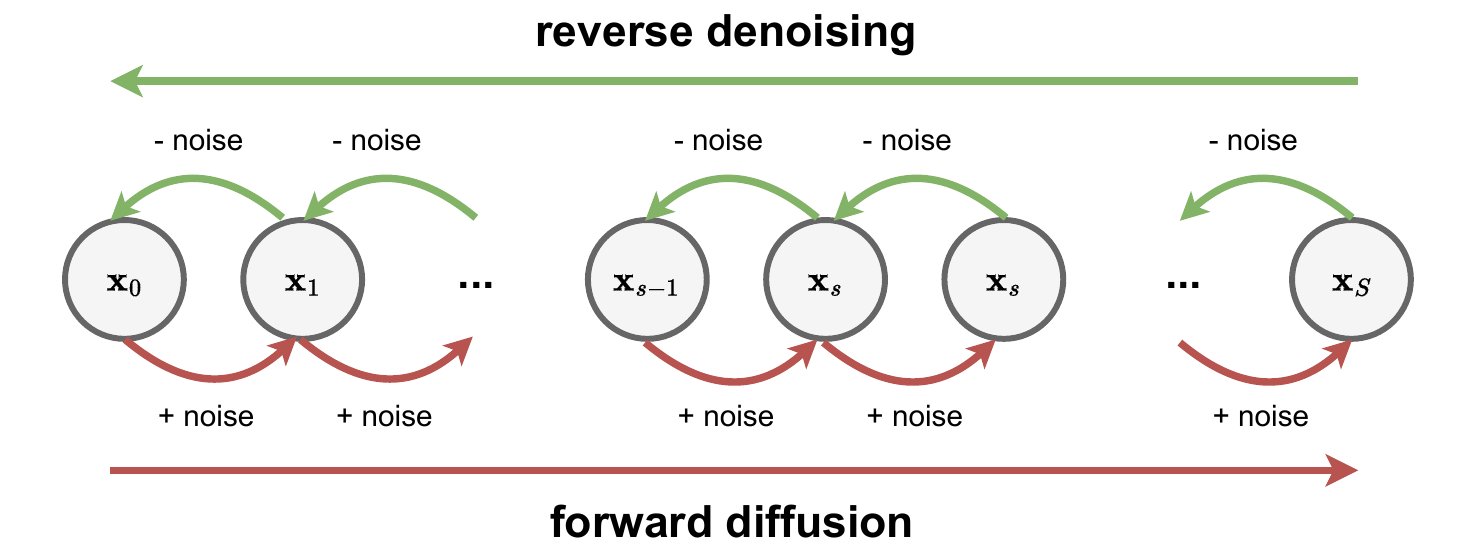}
\caption{Forward diffusion and reverse denoising}
\label{fig: autoregressive}
\end{figure}

The core idea behind diffusion models is to learn the data distribution \( p(\mathbf{x}_0) \) through a two-stage process. In the first stage, called the the \emph{forward diffusion} process, a clean data sample \( \mathbf{x}_0 \sim p(\mathbf{x}_0) \) is gradually perturbed over a series of \( S \) steps into pure Gaussian noise \( \mathbf{x}_S \). Here, \( \mathbf{x}_0 \) denotes the original clean data, and \( S \) is the total number of diffusion steps. The second stage, known as the \emph{reverse denoising} process, involves training a neural network to invert this noising procedure, reconstructing the original data from the noisy sample.

To guide the generation process, conditioning information $\boldsymbol{C}$ is often incorporated during denoising. This conditioning can influence the output in desired ways (e.g., by enforcing class labels, text prompts, or other modalities). There are multiple strategies to integrate conditioning information, such as as applying attention between the noisy sample $\mathbf{x}_s$ and $\boldsymbol{C}$, or by adding, multiplying or concatenating $\boldsymbol{C}$ to $\mathbf{x}_s$, depending on the model design.

In the context of traffic scene prediction, the data sample \( \mathbf{x}_0 \) typically represents either the full trajectory of an agent~\cite{jiang_motiondiffuser_2023}, or individual states of the agent's motion~\cite{wang_optimizing_2025, mao_leapfrog_2023, choi_dice_2024}, depending on the specific implementation. The fully noised data sample \( \mathbf{x}_S \) is expected to be statistically independent of the original trajectory, effectively containing no information about \( \mathbf{x}_0 \). The conditioning information $\boldsymbol{C}$ generally encodes relevant contextual data required to generate plausible future trajectories, such as the agent’s history and the surrounding map geometry.

We now describe the forward diffusion and reverse denoising processes, as well as the training and sampling approaches.

\paragraph{Forward Diffusion Process:}
The forward diffusion process is a Markov chain that incrementally adds Gaussian noise to the clean data sample $\mathbf{x}_0$ over $S$ diffusion steps, transforming it into a sample that approximates an isotropic Gaussian distribution. This process is expressed as:
\begin{equation}
\begin{aligned}
    q(\mathbf{x}_{1:S} \mid \mathbf{x}_0) &:= \prod_{s=1}^{S} q(\mathbf{x}_s \mid \mathbf{x}_{s-1})
\end{aligned}
\end{equation}
where each transition $q(\mathbf{x}_s | \mathbf{x}_{s-1})$ is a Gaussian distribution:
\begin{equation}
\begin{aligned}
    q(\mathbf{x}_s \mid \mathbf{x}_{s-1}) &:= \mathcal{N}(\mathbf{x}_s; \sqrt{\alpha_s} \, \mathbf{x}_{s-1}, (1-\alpha_s) \mathbf{I}).
\end{aligned}
\end{equation}
Here, $\alpha_s \in (0,1)$ is a variance-preserving noise schedule that controls the amount of noise added at each step.

Since the forward process consists of Gaussian transitions in a Markov chain, we can sample $\mathbf{x}_s$ at an arbitrary step $s$ from $\mathbf{x}_0$ in closed form by marginalizing out the intermediate steps. With the notation $\bar{\alpha}_s:= \prod_{k=1}^{s}\alpha_k$, this gives:
\begin{equation}
    q(\mathbf{x}_s \mid \mathbf{x}_{0}) =  \mathcal{N}(\mathbf{x}_s; \sqrt{\bar{\alpha}_s} \, \mathbf{x}_{0}, (1-\bar{\alpha}_s) \mathbf{I})
\end{equation}
Consequently, $\mathbf{x}_s$ can be expressed as a linear combination of the original data $\mathbf{x}_0$ and Gaussian noise $\boldsymbol{\epsilon}$:
\begin{equation}
\label{eqn: forward diffusion}
\begin{aligned}
\mathbf{x}_s = \sqrt{\bar{\alpha}_s} \, \mathbf{x}_0 + \sqrt{1 - \bar{\alpha}_s} \, \boldsymbol{\epsilon}, \quad \boldsymbol{\epsilon} \sim \mathcal{N}(\mathbf{0}, \mathbf{I})
\end{aligned}
\end{equation}
This formulation (Equation~\ref{eqn: forward diffusion}) allows for direct sampling of $\mathbf{x}_s$ at an arbitrary step $s$ without iteratively applying all previous noise steps, which is especially advantageous during training for computational efficiency.

\paragraph{Reverse Denoising Process:}
The reverse denoising process is also modeled as a Gaussian Markov chain, beginning from the prior distribution:
\begin{equation}
    p(\mathbf{x}_S):=\mathcal{N}(\mathbf{x}_S; \mathbf{0}, \mathbf{I})
\end{equation}
The complete reverse denoising process is defined as:
\begin{equation}
p_\theta(\mathbf{x}_{0:S}) = p(\mathbf{x}_S) \prod_{s=1}^{S} p_\theta(\mathbf{x}_{s-1} \mid \mathbf{x}_s).
\end{equation}
Each reverse transition is parameterized as a Gaussian distribution:
\begin{equation}
\label{eqn: reverse transition}
p_\theta(\mathbf{x}_{s-1} \mid \mathbf{x}_s) := \mathcal{N}(\mathbf{x}_{s-1}; \boldsymbol{\mu}_\theta(\mathbf{x}_s), \mathbf{\Sigma}_\theta(\mathbf{x}_s)).
\end{equation}
Here, the mean $\boldsymbol{\mu}_\theta $ and covariance $\boldsymbol{\Sigma}_\theta$ are predicted by a neural network with learnable parameters $\theta$ based on the noisy input \( \mathbf{x}_s \).  In practice, the model is also conditioned on the diffusion step $s$ and conditioning information $\boldsymbol{C}$, which we omit from the notation in this chapter for clarity.

\paragraph{Learning Objective:}
Training diffusion model aims to minimize the negative log-likelihood $-\log p_\theta(\mathbf{x}_0)$, bounded above by the Evidence Lower Bound (ELBO):
\begin{equation}
\mathbb{E}[-\log p_\theta(\mathbf{x}_0)] 
\leq 
\underbrace{
\mathbb{E}_q \left[ -\log \frac{p_\theta(\mathbf{x}_{0:S})}{q(\mathbf{x}_{1:S} \mid \mathbf{x}_0)} \right]
}_{\text{\scriptsize ELBO}} 
= 
\mathbb{E}_q \left[ -\log p(\mathbf{x}_S) - \sum_{s \geq 1} \log \frac{p_\theta(\mathbf{x}_{s-1} \mid \mathbf{x}_s)}{q(\mathbf{x}_s \mid \mathbf{x}_{s-1})} \right]
=: \ell.
\end{equation}
Based on the derivation of Ho et al.~\cite{ho_denoising_2020}, this can be rewritten as a sum of Kullback-Leibler (KL) divergences \cite{kullback_information_1951} between the distributions of the forward and reverse steps:
\begin{equation}
\label{eqn: ELBO}
\begin{aligned}
\text{ELBO} = \mathbb{E}_q \Big[ \,
& + \underbrace{D_{\mathrm{KL}}\big(q(\mathbf{x}_S \mid \mathbf{x}_0) \,\|\, p(\mathbf{x}_S)\big)}_{\ell_S
}\\
& + \sum_{s > 1} 
\underbrace{D_{\mathrm{KL}}\big(q(\mathbf{x}_{s-1} \mid \mathbf{x}_s, \mathbf{x}_0) \,\|\, p_\theta(\mathbf{x}_{s-1} \mid \mathbf{x}_s)\big)}_{\ell_{s-1}} \\
& - \underbrace{\log p_\theta(\mathbf{x}_0 \mid \mathbf{x}_1)}_{\ell_0} 
\, \Big]
\end{aligned}
\end{equation}
where $D_{KL}$ refers to the KL divergence. This breaks the loss into three components:
\begin{itemize}
\item The KL divergence term $\ell_S$ measures how well the forward process $q(\mathbf{x}_S \mid \mathbf{x}_0)$ transforms the clean data $\mathbf{x}_0$ into a sample $\mathbf{x}_S$ that matches the prior distribution $p(\mathbf{x}_S)$. Since the forward process is entirely determined by the predefined noise schedule $\alpha_s$, this term does not involve any learnable parameters.

\item The KL divergence term $\ell_{s-1}$ measures the discrepancy between the Bayesian posterior of the forward transition $q(\mathbf{x}_{s-1}\mid \mathbf{x}_s, \mathbf{x}_0)$ and the learned reverse transition $p_\theta(\mathbf{x}_{s-1} \mid \mathbf{x}_s)$. Using Bayes’ theorem, it can be shown that:
\begin{equation}
\begin{aligned}
q(\mathbf{x}_{s-1} \mid \mathbf{x}_s, \mathbf{x}_0) 
&= \frac{q(\mathbf{x}_s \mid \mathbf{x}_{s-1}, \mathbf{x}_0) q(\mathbf{x}_{s-1} \mid \mathbf{x}_0)}{q(\mathbf{x}_s \mid \mathbf{x}_0)} \\
&= \frac{q(\mathbf{x}_s \mid \mathbf{x}_{s-1}) q(\mathbf{x}_{s-1} \mid \mathbf{x}_0)}{q(\mathbf{x}_s \mid \mathbf{x}_0)} \\
&= \mathcal{N}\left(\mathbf{x}_{s-1}; \tilde{\boldsymbol{\mu}}_s(\mathbf{x}_s, \mathbf{x}_0), \tilde{\sigma}_s^2 \mathbf{I} \right) \\
\end{aligned}
\end{equation}
where $\quad \tilde{\boldsymbol{\mu}}_s(\mathbf{x}_s, \mathbf{x}_0) := \frac{\sqrt{\alpha_s}(1 - \bar{\alpha}_{s-1})}{1 - \bar{\alpha}_s} \, \mathbf{x}_s + \frac{(1-\alpha_s)\sqrt{\bar{\alpha}_{s-1}} }{1 - \bar{\alpha}_s}\mathbf{x}_0 
\quad \text{and} \quad \tilde{\sigma}_s^2 := \frac{(1-\alpha_s)(1 - \bar{\alpha}_{s-1})}{1 - \bar{\alpha}_s}$. The first equality follows from Bayes’ rule, the second from the Markovian property of the forward diffusion process, and the third is a rearrangement of Gaussian density functions. Since both $q(\mathbf{x}_{s-1} \mid \mathbf{x}_s, \mathbf{x}_0)$ and $p_\theta(\mathbf{x}_{s-1} \mid \mathbf{x}_s)$ are Gaussian, the KL divergence in $\ell_{s-1}$ admits a closed-form expression. To further simplify the optimization, it is common practice to fix the covariance of the reverse transition $p_\theta(\mathbf{x}_{s-1} \mid \mathbf{x}_s)$ to match that of the forward posterior $q(\mathbf{x}_{s-1} \mid \mathbf{x}_s, \mathbf{x}_0)$: $\mathbf{\Sigma}_\theta(\mathbf{x}_s) = \tilde{\sigma}_s^2\mathbf{I}$. Under this assumption, minimizing the KL divergence $\ell_{s-1}$ reduces to minimizing the squared difference between the means of the two Gaussians: 
\begin{equation}
\begin{aligned}
\|\tilde{\boldsymbol{\mu}}_s(\mathbf{x}_s, \mathbf{x}_0) - \boldsymbol{\mu}_{\theta}(\mathbf{x}_s)\|^2_2
\end{aligned}
\end{equation}

\item The term $\ell_0$ denotes the reconstruction loss, measuring how well the model (associated with $p_\theta$) can reconstruct the original clean data $\mathbf{x}_0$ from the latent variable $\mathbf{x}_1$, corresponding to the final step in the reverse denoising process. This can be simplified to:
\begin{equation}
\begin{aligned}
\log p_\theta(\mathbf{x}_0 \mid \mathbf{x}_1) 
&= \log \mathcal{N}(\mathbf{x}_0 \mid \boldsymbol{\mu}_\theta(\mathbf{x}_1), \tilde{\sigma}_1^2 \mathbf{I}) \\
&= \log \left( \frac{1}{(\sqrt{2\pi \tilde{\sigma}_1^2})^d} \exp\left\{ -\frac{\|\mathbf{x}_0 - \boldsymbol{\mu}_\theta(\mathbf{x}_1)\|^2_2}{2\tilde{\sigma}_1^2} \right\} \right) \\
&= -\frac{\|\mathbf{x}_0 - \boldsymbol{\mu}_\theta(\mathbf{x}_1)\|^2_2}{2\tilde{\sigma}_1^2} 
- \underbrace{\frac{d}{2} \log \left( 2\pi \tilde{\sigma}_1^2 \right)}_{\text{constant}} 
\end{aligned}
\end{equation}
where $d$ denotes the dimension of $\mathbf{x}_0$.
\end{itemize}
Therefore, by ignoring the non-learnable part $\ell_S$ and the constant term in $\ell_0$ and $\ell_{s-1}$, the training aims to minimize:
\begin{equation}
\label{eqn: intermediate loss}
    \frac{1}{2\tilde{\sigma}_s^2} 
\left\| 
\underbrace{\tilde{\boldsymbol{\mu}}_s(\mathbf{x}_s, \mathbf{x}_0)}_{\text{known}} 
- 
\underbrace{\boldsymbol{\mu}_\theta(\mathbf{x}_s)}_{\text{learnable}} 
\right\|^2_2
\end{equation}
The term $\boldsymbol{\mu}_{\theta}(\mathbf{x}_s)$ can be constructed in the same way as $\tilde{\boldsymbol{\mu}}_s(\mathbf{x}_s, \mathbf{x}_0)$ by introducing the learnable $\hat{\mathbf{x}}_\theta(\mathbf{x}_s)$ as:
\begin{equation}
\label{eqn: decomposite mu_theta}
\underbrace{\boldsymbol{\mu}_\theta(\mathbf{x}_s)}_{\text{learnable}} 
:=\frac{\sqrt{\alpha_s}(1 - \bar{\alpha}_{s-1}) }{1 - \bar{\alpha}_s} \, \mathbf{x}_s 
+ \frac{(1 - \alpha_s) \sqrt{\bar{\alpha}_{s-1}}}{1 - \bar{\alpha}_s} \, 
\underbrace{\hat{\mathbf{x}}_\theta(\mathbf{x}_s)}_{\text{learnable}}.
\end{equation}
Substituting Equation \ref{eqn: intermediate loss} and Equation \ref{eqn: decomposite mu_theta} into Equation \ref{eqn: ELBO}, the ELBO for DDPM is simplfied to:
\begin{equation}
\text{ELBO}_\theta(\mathbf{x}) = - \sum_{s=1}^{S} 
\frac{1}{2\tilde{\sigma}_s^2} 
\cdot \frac{(1 - \alpha_s)^2 \bar{\alpha}_{s-1}}{(1 - \bar{\alpha}_s)^2}
\, \mathbb{E}_{q(\mathbf{x}_s \mid \mathbf{x}_0)} 
\left[ \left\| \hat{\mathbf{x}}_\theta(\mathbf{x}_s) - \mathbf{x}_0 \right\|_2^2 \right]
\end{equation}
This indicates that the model is required to predict the clean data $\mathbf{x}_0$ given the noisy data $\mathbf{x}_s$ at an arbitrary diffusion step $s$. Ho et al. \cite{ho_denoising_2020} show that predicting the added noise $\boldsymbol{\epsilon}_{\theta}(\mathbf{x}_s)$, corresponding to the noise term $\boldsymbol{\epsilon}$ in Equation \ref{eqn: forward diffusion}, rather than the clean data $\hat{\mathbf{x}}_\theta(\mathbf{x}_s)$ of each forward step leads to a simplified loss function:
\begin{equation}
    \ell_{\text{simple}} := \mathbb{E}_{s, \mathbf{x}_0, \boldsymbol{\epsilon}} \left[ \left\| \boldsymbol{\epsilon} - \boldsymbol{\epsilon}_\theta(\mathbf{x}_s) \right\|_2^2 \right].
\end{equation}

\paragraph{Training and Sampling:}
The training process for diffusion-based traffic scene prediction is outlined in Algorithm~\ref{alg: ddpm-training}. During training, the model learns to predict the noise added to clean data at a randomly sampled diffusion step. At each iteration, a clean trajectory \( \mathbf{x}_0 \) is corrupted into a noisy version \( \mathbf{x}_s \) using the forward diffusion process. The model \( \boldsymbol{\epsilon}_\theta \) is then trained to recover the original noise \( \boldsymbol{\epsilon} \) using a simple mean squared error (MSE) loss. For completeness, the noise prediction is conditioned on both the diffusion step \( s \) and the conditioning information \( \boldsymbol{C} \), such as the agent’s motion history and the surrounding map.

 \begin{algorithm}
\caption{Training Procedure for Diffusion-Based Traffic Scene Prediction}
\label{alg: ddpm-training}
\SetAlgoLined
\SetKwInOut{Input}{Input}
\SetKwInOut{KwResult}{Output}
\Input{Training dataset $\mathcal{D}$ containing clean future trajectories $\mathbf{x}_0$;\\
Total number of diffusion steps $S$;\\
Conditioning information $\boldsymbol{C}$ (e.g., map, agent history);}
\KwResult{Trained model parameters $\boldsymbol{\theta}$;}
\ForEach{training iteration}
{
    Sample a clean trajectory $\mathbf{x}_0 \sim \mathcal{D}$\;
    Sample a diffusion timestep $s \sim \text{Uniform}(1, S)$\;
    Sample noise $\boldsymbol{\epsilon} \sim \mathcal{N}(\mathbf{0}, \mathbf{I})$\;
    Compute noisy trajectory: $\mathbf{x}_s = \sqrt{\bar{\alpha}_s} \mathbf{x}_0 + \sqrt{1 - \bar{\alpha}_s} \boldsymbol{\epsilon}$\;
    Predict noise: $\hat{\boldsymbol{\epsilon}} = \boldsymbol{\epsilon}_\theta(\mathbf{x}_s, s, \boldsymbol{C})$\;
    Compute loss: $\ell = \|\boldsymbol{\epsilon} - \hat{\boldsymbol{\epsilon}}\|^2_2$\;
    Update model parameters $\boldsymbol{\theta}$ via gradient descent using $\nabla_{\boldsymbol{\theta}} \ell$\;
}
\Return{$\boldsymbol{\theta}$}
\end{algorithm}

Once the diffusion model is trained, it can be used to generate diverse trajectory samples. There are various sampling procedures available, depending on the design choices and desired trade-offs between speed and quality. 

\begin{algorithm}[tbh]
\caption{DDPM Sampling for Diffusion-Based Traffic Scene Prediction}
\label{alg: ddpm-sampling}
\SetAlgoLined
\SetKwInOut{Input}{Input}
\SetKwInOut{Output}{Output}

\Input{Trained model $\boldsymbol{\epsilon}_\theta$;\\
Total number of diffusion steps $S$;\\
Conditioning information $\boldsymbol{C}$ (e.g., map, agent history)}
\Output{Generated data sample $\hat{\mathbf{x}}_0$}
\textbf{Initialize:} Sample $\mathbf{x}_S \sim \mathcal{N}(\mathbf{0}, \mathbf{I})$\;
\For{$s = S, S-1, \dots, 1$}{
    Predict noise: $\hat{\boldsymbol{\epsilon}} = \boldsymbol{\epsilon}_\theta(\mathbf{x}_s, s, \boldsymbol{C})$\;
    
    Compute predicted mean:
    \[
    \boldsymbol{\mu}_\theta(\mathbf{x}_s, s, \boldsymbol{C}) = \frac{1}{\sqrt{\alpha_s}} \left( \mathbf{x}_s - \frac{1-\alpha_s}{\sqrt{1-\bar{\alpha}_s}} \hat{\boldsymbol{\epsilon}} \right)
    \]
    \eIf{$s > 1$}{
        Sample noise $\boldsymbol{\epsilon}^{*} \sim \mathcal{N}(\mathbf{0}, \mathbf{I})$\;
        
        Update:
        \[
        \mathbf{x}_{s-1} = \boldsymbol{\mu}_\theta(\mathbf{x}_s, s, \boldsymbol{C}) + \tilde{\sigma}_s \boldsymbol{\epsilon}^{*}
        \]
    }{
        Set:
        \[
        \hat{\mathbf{x}}_0 = \boldsymbol{\mu}_\theta(\mathbf{x}_1, 1, \boldsymbol{C})
        \]
    }
}
\Return{$\hat{\mathbf{x}}_0$}
\end{algorithm}
Algorithm~\ref{alg: ddpm-sampling} outlines the DDPM sampling procedure \cite{ho_denoising_2020} used in diffusion-based traffic scene prediction models. The process begins from a Gaussian prior \( \mathbf{x}_S \sim \mathcal{N}(\mathbf{0}, \mathbf{I}) \), and proceeds iteratively through all \( S \) reverse diffusion steps to generate a data sample \( \hat{\mathbf{x}}_0 \). At each step \( s \), the model \( \boldsymbol{\epsilon}_\theta \) predicts the noise component in \( \mathbf{x}_s \), conditioned on the timestep \( s \) and the condition information \( \boldsymbol{C} \) (e.g., map and agent history). The predicted noise is then used to compute the mean of the reverse Gaussian distribution. 

To preserve stochasticity and allow for diverse outputs, Gaussian noise \( \boldsymbol{\epsilon}^{*} \sim \mathcal{N}(\mathbf{0}, \mathbf{I}) \) is injected at each step \( s > 1 \). This allows the model to produce diverse trajectory samples during generation. The final denoised output \( \hat{\mathbf{x}}_0 \) is returned after completing all \( S \) steps.

An obvious drawback of DDPM sampling is its inefficiency: the number of reverse denoising steps required during sampling is equal to the number of forward diffusion steps, often in the hundreds or even thousands. This leads to significant computational overhead, especially in real-time or resource-constrained applications.

To address this limitation, Denoising Diffusion Implicit Models (DDIM)~\cite{song_denoising_2021} was proposed as a non-Markovian alternative that enables deterministic and faster sampling. DDIM reinterprets the reverse process as a deterministic transformation conditioned on the same learned noise prediction model \( \boldsymbol{\epsilon}_\theta \), allowing the number of sampling steps to be reduced without retraining the model. By using a carefully designed update rule, DDIM can generate high-quality samples in significantly fewer steps than DDPM, while maintaining compatibility with the original training objective.
Algorithm~\ref{alg: ddim-sampling} outlines the deterministic DDIM sampling procedure. Unlike DDPM, which iterates over all \( S \) denoising steps, DDIM accelerates sampling by using a stride $\delta_s$, so that only every \( \delta_s \)-th step is used for denoising. In practice \cite{song_denoising_2021}, the stride is typically chosen to divide evenly into the total diffusion steps $S$, resulting in a significantly reduced number of denoising steps \( \frac{S}{\delta_s} \ll S \). For example, the original DDIM paper uses \( S = 1000 \) and \( \delta_s = 100 \), reducing sampling to just 10 steps.

Starting from pure Gaussian noise \( \mathbf{x}_{S} \sim \mathcal{N}(\mathbf{0}, \mathbf{I}) \), the model iteratively denoises by predicting the noise \( \hat{\boldsymbol{\epsilon}} \) at the current noisy state \( \mathbf{x}_{s} \), which is used to estimate the clean sample \( \hat{\mathbf{x}}_0 \). The next sample \( \mathbf{x}_{s-\delta_s} \) is computed as a weighted combination of \( \hat{\mathbf{x}}_0 \) and \( \hat{\boldsymbol{\epsilon}} \), without injecting any additional randomness. This deterministic approach enables efficient sampling while maintaining high sample quality.
\begin{algorithm}
\caption{Deterministic DDIM Sampling for Diffusion-Based Traffic Scene Prediction}
\label{alg: ddim-sampling}
\SetAlgoLined
\SetKwInOut{Input}{Input}
\SetKwInOut{Output}{Output}

\Input{Trained model $\boldsymbol{\epsilon}_\theta$;\\
Denoising stride $\delta_s$;\\
Conditioning information $\boldsymbol{C}$ (e.g., map, agent history);\\}
\Output{Generated data sample $\hat{\mathbf{x}}_0$;}
\textbf{Initialize:} Sample $\mathbf{x}_{S} \sim \mathcal{N}(\mathbf{0}, \mathbf{I})$\;
\For{$s = S, S-\delta_s, S-2\delta_s, \ldots, \delta_s$}{
    
    Predict noise: $\hat{\boldsymbol{\epsilon}} = \boldsymbol{\epsilon}_\theta(\mathbf{x}_s, s, \boldsymbol{C})$\;
    
    Compute predicted clean sample:
    \[
    \hat{\mathbf{x}}_0 = \frac{1}{\sqrt{\bar{\alpha}_s}} \left( \mathbf{x}_s - \sqrt{1 - \bar{\alpha}_s} \, \hat{\boldsymbol{\epsilon}} \right)
    \]

    \eIf{$s > \delta_s$}{
    Compute DDIM update:
    \[
    \mathbf{x}_{s - \delta_s} = \sqrt{\bar{\alpha}_{s - \delta_s}} \, \hat{\mathbf{x}}_0 + \sqrt{1 - \bar{\alpha}_{s - \delta_s}} \hat{\boldsymbol{\epsilon}} 
    \]
    }
    {\Return{$\hat{\mathbf{x}}_0$}}
}
\end{algorithm}

Despite their computational overhead, diffusion models have emerged as a powerful framework for traffic scene generation, offering several advantages over earlier generative approaches. Unlike VAEs and GANs, diffusion models provide stable training dynamics and can generate diverse, high-quality samples without mode collapse. The iterative denoising process naturally captures the multi-modal nature of multi-agent interactions, making diffusion models particularly well-suited for joint prediction tasks.

Relative to the classical maneuver- and kinematic-based methods (cf. Section \ref{sec: 2.2 advancements in model}), diffusion models learn scene-level interactions directly from data and generate an unrestricted set of outcomes, at the expense of the interpretability and low computational cost of those earlier approaches.

However, several challenges remain in applying diffusion models to traffic scene prediction. The sequential nature of the denoising process can be computationally expensive for real-time applications, though techniques like DDIM sampling help mitigate this issue. Diffusion models also provide no explicit likelihood over the generated scenes, so the relative probability of the sampled outcomes is not directly available. Additionally, the choice of data representation -- whether sequence-based or parametric -- significantly impacts both the quality of generated trajectories and the model's ability to enforce physical constraints.

The integration of parametric representations, such as polynomials, with generative frameworks presents an opportunity to combine the expressiveness of diffusion models with the physical interpretability and computational efficiency of parametric representations. This motivates the investigation of parametric trajectory representations presented in Chapter \ref{cpt: Empirical Bayes Analysis}, which establishes the theoretical foundation for the generative models developed later in this work.
\chapter{Empirical Bayes Analysis of Agent Trajectory Models}
\label{cpt: Empirical Bayes Analysis}

This chapter builds upon the author's conference paper published at the \textit{IEEE International Conference on Intelligent Transportation Systems (ITSC)} 2023 \cite{yao_empirical_2023}\footnote{\copyright~2023 IEEE. Reprinted, with permission, from Y. Yao, D. Goehring, and J. Reichardt, ``An Empirical Bayes Analysis of Object Trajectory Representation Models,'' in \textit{Proc. IEEE Int. Conf. Intelligent Transportation Systems (ITSC)}, pp.~902--909, 2023.}, co-authored with Prof. Dr. Daniel Goehring (advisor) and Dr. Joerg Reichardt. The work has been significantly extended and refined for this chapter.

The author was responsible for the conceptual development of the empirical Bayes framework, the complete implementation, and all experiments and evaluations presented in the original paper and in this chapter. Prof. Dr. Daniel Goehring and Dr. Joerg Reichardt contributed through scientific discussions and supervision.

The corresponding implementation is publicly available at: \url{https://github.com/aumovio/empirical-bayes-analysis-of-object-trajectory}.

\section{Motivation and Problem Description}
Efficient and accurate trajectory representation is a foundational requirement for prediction and planning systems. When choosing a trajectory representation, developers navigate an inherent trade-off between model complexity and representation fidelity. Sequence-based representations, which store states at discrete timesteps, provide maximum flexibility but lack inherent physical constraints and scale poorly with increasing trajectory duration. In contrast, parametric representations encode entire trajectories as compact sets of parameters, enforcing motion continuity and significantly reducing computational demands. However, as approximations of real-world motion, they introduce representation error -- i.e., bias -- that has not been systematically quantified across different agent types and trajectory durations.  Additionally, although parametric representations, specifically polynomial representations, are increasingly applied in recent prediction systems and achieve competitive performance \cite{buhet_plop_2020, su_temporally_2021}, the choice of model complexity (i.e., polynomial degree) is typically made empirically without systematic justification. 

In this dissertation, we are motivated to reliably integrate \emph{polynomial} representations into prediction systems. Before that, three principal questions must be addressed and validated in this chapter:
\begin{itemize}[leftmargin=*]
 \item How accurately can polynomial representations capture real-world trajectories?
 \item Does the bias introduced by polynomial representations fundamentally limit prediction performance?
 \item What level of model complexity (i.e., polynomial degree) best fits different agent types and trajectory durations?
\end{itemize}

\begin{figure}[thb]
\centering
\includegraphics[width=\textwidth]{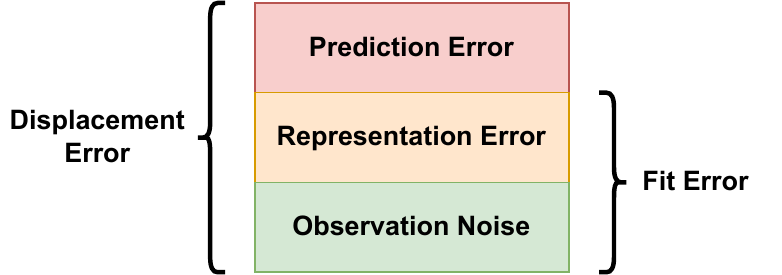}
\caption{Components of displacement error}
\label{fig: components of displacement error}
\end{figure}

To systematically address these questions, it is first essential to understand the composition of the displacement error in prediction systems. As outlined in Section \ref{sec: evaluation metrics}, displacement errors are typically the spatial deviations between predicted trajectories and the \emph{noisy observations} provided in the dataset. Therefore, it can be decomposed into three distinct components, as shown in Figure \ref{fig: components of displacement error}:
\begin{itemize}[leftmargin=*]
 \item \textbf{Observation noise}: The measurement error intrinsic to the dataset, representing the deviation between the (unobserved) ground-truth trajectory and the recorded observations.
 \item \textbf{Representation error}: The deviation introduced by the choice of trajectory representation model, which constrains the space of possible trajectories.
  \item \textbf{Prediction error}: The deviation caused by imperfections in the prediction system itself, including missing or incomplete input information, such as agents' intentions.
\end{itemize}
In an ideal setting, deviations between predicted and ground-truth trajectories would arise solely from prediction error and representation error. However, motion datasets contain only noisy observations, without direct access to the ground-truth trajectories. As a result, observation noise cannot be separated from the other two components. We therefore define the sum of observation noise and representation error as the \emph{fit error} -- an upper bound on the representation error. Under a perfect prediction algorithm with zero prediction error, this fit error represents the minimum achievable displacement error for the given trajectory representation model. Quantifying this fit error relative to displacement error is essential for assessing whether polynomial representations introduce substantial bias that might limit overall prediction performance.

To address these questions, this chapter conducts an extensive empirical Bayes analysis of polynomial trajectory representations across three widely-used public motion datasets: Argoverse 1 (A1) \cite{chang_argoverse_2019}, Argoverse 2 (A2) \cite{wilson_argoverse2_2021}, and Waymo Open (WO) \cite{ettinger_waymo_2021}. We focus specifically on agent trajectory representations rather than map elements, as trajectories exhibit measurement noise and temporal dynamics that require systematic analysis, while map elements in these datasets are typically constructed with high precision and minimal noise \cite{wilson_argoverse2_2021, ettinger_waymo_2021}. The results of this analysis provide theoretical justification for the adoption of polynomial representations in the prediction models developed in subsequent chapters.

The remainder of this chapter is organized as follows. Section \ref{sec: 3.2 parametric representation} formulates our polynomial trajectory representation and outlines its theoretical advantages. Section \ref{sec 3.3: Bayesian regression} introduces Bayesian regression for trajectory fitting and discusses its reliance on prior and observation noise parameters, which are typically unavailable. To address this,  Section \ref{sec 3.4: Empirical Bayes Method} presents an empirical Bayes framework for estimating prior distributions over model parameters and observation noise directly from real-world datasets. Section \ref{sec 3.5: outlier filtering} describes our outlier filtering approach. Section \ref{sec 3.6: Experiments} reports experiments analyzing the trade-off between model complexity and data fit quality, using information-theoretic criteria and the average fit error. This analysis is conducted across three large-scale public datasets and multiple agent types, evaluating trajectories of varying trajectory durations to identify optimal model complexities. We also compare the resulting fit errors with displacement errors reported by state-of-the-art prediction approaches, demonstrating that polynomial representations offer high fidelity with moderate complexity -- supporting their suitability for traffic scene prediction models. Section \ref{sec 3.7: Conclusion and Discussion} concludes with a discussion of the results, their limitations, and their implications for the following chapters.

\section{Polynomial Trajectory Representation}
\label{sec: 3.2 parametric representation}

Agent trajectories can be modeled using various polynomial bases, including monomial, Bernstein (commonly known as Bézier curves), and Chebyshev bases. While different polynomial bases of the same degree offer equivalent expressive power -- as they can be linearly transformed into one another -- this work adopts \emph{Bernstein} polynomials due to their advantageous properties in coordinate transformations. Bernstein polynomials define trajectories through a set of \emph{control points},  to which common geometric operations -- such as rotation, translation, and scaling -- can be applied directly. This makes spatial transformations both intuitive and computationally efficient.

We denote $t'$ and $T'$ as the continuous-time (in seconds) counterparts of the discrete timestep $t$ and the total number of timesteps $T$, respectively. This distinction is necessary because the time intervals between successive discrete steps are not always uniform, due to variations in the actual sampling rate. 

Given the polynomial degree $N$, the position of an agent $\mathbf{c}(\tau_t)\in \mathbb{R}^d$, where $d$ is the spatial dimension, at the rescaled time $\tau_t=\frac{t'}{T'}\in [0,1]$  can be expressed as a linear combination of $N+1$ basis functions of time $\phi_n(\tau_t) : \mathbb{R}\rightarrow\mathbb{R}$ and corresponding parameters $\mathbf{w}_n\in \mathbb{R}^d$, where $n=0, 1, \ldots, N$:
\begin{equation}
\label{eqn: linear_combination}
\mathbf{c}(\tau_t) = \sum_{n=0}^N\phi_n(\tau_t)\mathbf{w}_n.
\end{equation}
The Bernstein basis functions of degree \( N \) are defined as:
\begin{equation}
\label{eqn: bernstein_basis}
\phi_n(\tau_t) = \binom{N}{n}(1 - \tau_t)^{N - n} (\tau_t)^n, \quad \text{for } n = 0, \ldots, N \text{ and } \tau_t \in [0,1],
\end{equation}
where \( \binom{N}{n} \) is the binomial coefficient. These Bernstein basis functions are non-negative and sum to one for all \( \tau_t \in [0,1] \), i.e., \( \sum_{n=0}^{N} \phi_n(\tau_t) = 1 \), a property known as the partition of unity. This ensures that the resulting trajectory \( \mathbf{c}(\tau_t) \) lies within the convex hull of its control points \( \mathbf{w}_n \).

In this dissertation, we focus on 2D agent centroid positions provided in each dataset (i.e., \( d = 2 \)), as this information is consistently available across datasets. To express Equation \ref{eqn: linear_combination} more compactly, we define the parameter vector $\boldsymbol{\omega}\in \mathbb{R}^{(N+1)d}$ as $\boldsymbol{\omega}= [\mathbf{w}_0; \mathbf{w}_1;\ldots;\mathbf{w}_N]$ and a vector of basis functions $\boldsymbol{\phi}(\tau_t)\in \mathbb{R}^{N+1}$ by $\boldsymbol{\phi}(\tau_t)=[\phi_0(\tau_t); \phi_1(\tau_t);\ldots;\phi_N(\tau_t)]$. With this notation, Equation \ref{eqn: linear_combination} can be rewritten as:
\begin{equation}
\label{eqn: trajectory point}
\mathbf{c}(\tau_t)=(\boldsymbol{\phi}^{\top}(\tau_t)\otimes\mathbf{I}_d)\boldsymbol{\omega}
\end{equation}
where $\mathbf{I}_d$ is the $d\times d$ identity matrix and the Kronecker product $\otimes$ distributes the basis functions over the $d$ spatial dimensions.

\begin{figure}[thb]
\centering
\includegraphics[width=\textwidth]{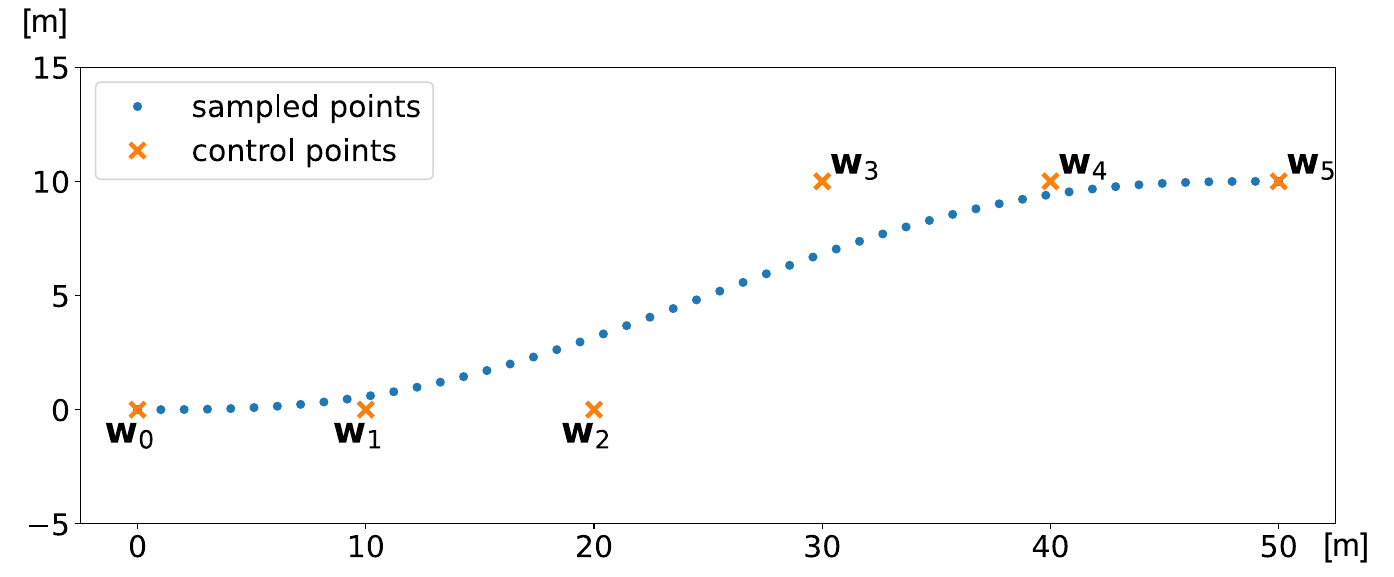}
\caption[An example of Bernstein polynomial]{An example of a 5-degree polynomial with 50 uniformly sampled points.}
\label{fig: Bernstein polynomial example}
\end{figure}

Now assume that the complete trajectory consists of $T$ timesteps $\boldsymbol{\tau} = [\tau_{1}, \ldots,\tau_{T}]$. These points can be stacked into a vector $\mathbf{c}\in\mathbb{R}^{Td}$, defined as $\mathbf{c}=[\mathbf{c}(\tau_{1});\ldots;\mathbf{c}(\tau_{T})]$. By constructing the matrix $\boldsymbol{\Phi}= \boldsymbol{\Phi}_\text{B} \otimes\mathbf{I}_d \in \mathbb{R}^{(N+1)d\times Td}$ with $\boldsymbol{\Phi}_\text{B}=[\boldsymbol{\phi}(\tau_{1}), \ldots,\boldsymbol{\phi}(\tau_{T})]$, the full trajectory can be compactly expressed as:
\begin{equation}
\label{eqn: trajectory representation}
\mathbf{c}=\boldsymbol{\Phi}^{\top}\boldsymbol{\omega}
\end{equation}
Equation \ref{eqn: trajectory representation} forms the foundation for the trajectory parameterization used throughout this dissertation, enabling efficient learning and prediction in downstream tasks. An illustrative example is shown in Figure~\ref{fig: Bernstein polynomial example}, where the sampled trajectory points $\mathbf{c}(\tau_1), \ldots, \mathbf{c}(\tau_T)$ are drawn as dots and the control points $\mathbf{w}_0, \ldots, \mathbf{w}_N$ as crosses.

Although Equation \ref{eqn: trajectory representation} represents only trajectory position, derivatives such as velocity and acceleration can be obtained by applying the derivative operators on the basis functions.

Following \cite{reichardt_trajectories_2022}, we employ a linear derivative operator $\textbf{D}_\text{B} \in \mathbb{Z}^{(N+1)\times (N+1)}$ for Bernstein basis functions, defined as:
\begin{equation}
    \textbf{D}_\text{B} = \textbf{M}^{-1} \textbf{D}_\text{M} \textbf{M}.
\end{equation}
where $\textbf{M}\in \mathbb{Z}^{(N+1)\times (N+1)}$ is the linear transformation matrix between monomial and Bernstein basis functions, with entries:
\begin{equation}
M_{pq}=
\begin{cases}
\binom{N}{q-1} \binom{N - q + 1}{p - q} (-1)^{((p+q)\bmod 2)} & \text{if } q \leq p \\
0 & \text{otherwise.}
\end{cases}
\end{equation}
The matrix $\textbf{D}_\text{M}\in \mathbb{Z}^{(N+1)\times (N+1)}$ is the linear derivative operator for monomial basis functions, defined as:
\begin{equation}
    \textbf{D}_\text{M} =
\begin{bmatrix}
0 & 1 & 0 & 0 & \cdots \\
0 & 0 & 2 & 0 & \cdots \\
0 & 0 & 0 & 3 & \cdots \\
\vdots & \vdots & \vdots & \vdots & \ddots
\end{bmatrix}.
\end{equation}
The velocity and acceleration can be then expressed as: 
\begin{equation}
\label{eqn: polynomial derivative}
\begin{aligned}
    {\mathbf{\dot{c}}}(\tau_t)&=({\boldsymbol{\dot{\phi}}}^{\top}(\tau_t)\otimes\mathbf{I}_d)\boldsymbol{\omega},\\
    {\mathbf{\ddot{c}}}(\tau_t)&=({\boldsymbol{\ddot\phi}}^{\top}(\tau_t)\otimes\mathbf{I}_d)\boldsymbol{\omega},
\end{aligned}
\end{equation}
where:
\begin{equation}
\label{eqn: basis function derivative}
\begin{aligned}
    \boldsymbol{\dot{\phi}}^{\top}(\tau_{t}) &= \boldsymbol{\phi}^{\top}(\tau_{t}) \textbf{D}_\text{B},\\
    \boldsymbol{\ddot{\phi}}^{\top}(\tau_{t}) &= \boldsymbol{\dot{\phi}}^{\top}(\tau_{t}) \textbf{D}_\text{B} \\ &= \boldsymbol{\phi}^{\top}(\tau_{t}) \textbf{D}_\text{B} \textbf{D}_\text{B}.
\end{aligned}
\end{equation}
This formulation enables the computation of kinematic properties directly from the polynomial representation, maintaining consistency with Equation \ref{eqn: trajectory point}.

\section{Bayesian Regression for Modeling Trajectory Data}
\label{sec 3.3: Bayesian regression}
After presenting the formulation of Bernstein polynomial representations for agent trajectories, the next step is to evaluate how well Bernstein polynomials approximate real-world motion. To perform this evaluation at scale, we analyze data from three large-scale motion datasets: Argoverse 1 (A1) \cite{chang_argoverse_2019}, Argoverse 2 (A2) \cite{wilson_argoverse2_2021}, and Waymo Open (WO) \cite{ettinger_waymo_2021}. 



It is important to note again that these datasets do \emph{not} provide ground truth trajectories, but rather noisy estimates of agent positions and kinematics. Consequently, applying standard linear regression, such as Ordinary Least Squares (OLS), would result in fitting both the ground-truth trajectory and the measurement noise, potentially distorting the analysis. To explicitly account for uncertainty during model fitting, we employ \emph{Bayesian regression}, which integrates both a prior over the trajectory polynomial parameters and the observation noise. 

\begin{figure}[thb]
\centering
\includegraphics[width=0.7\textwidth]{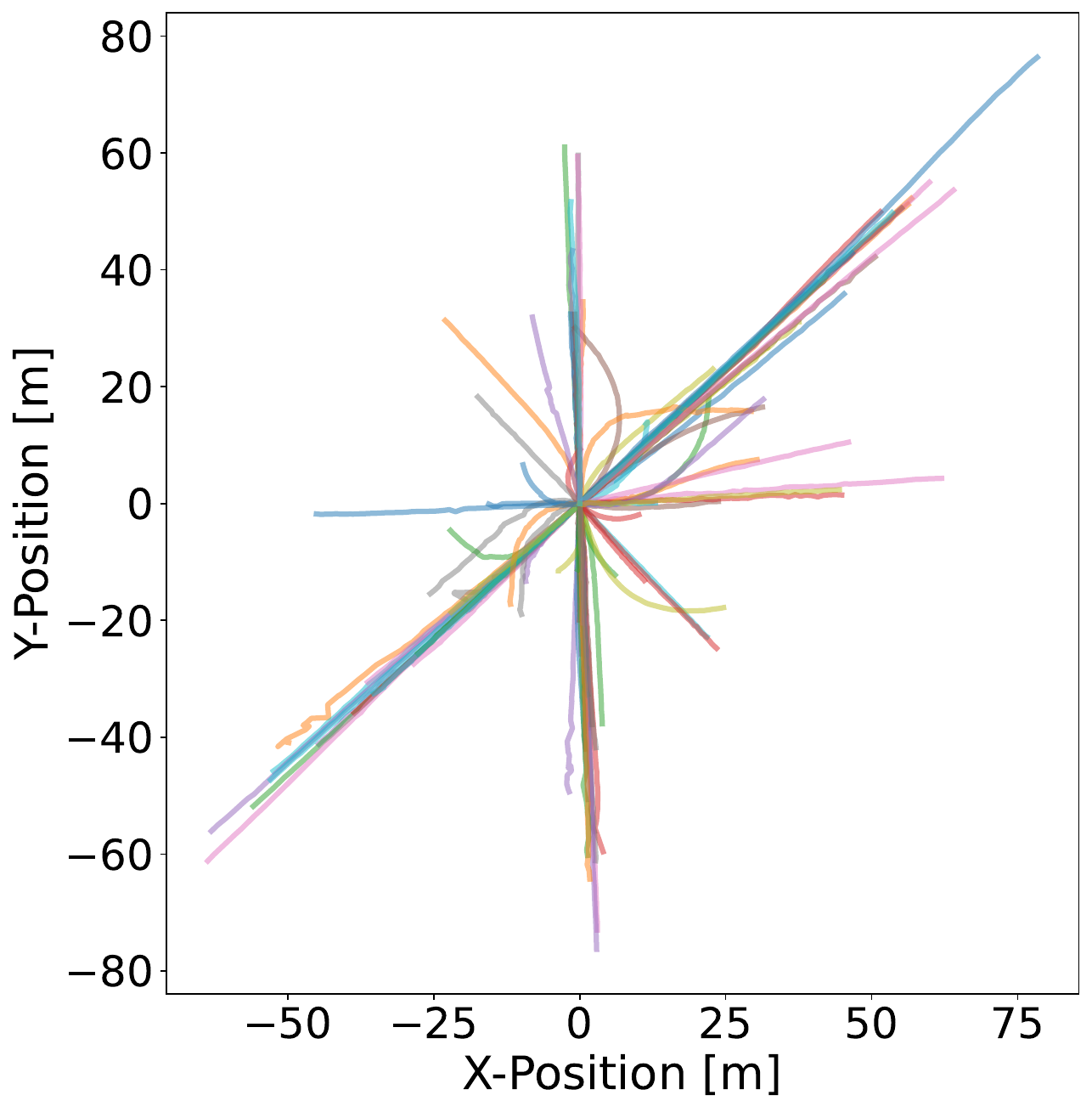}
\caption[Visualization of translated trajectories]{Visualization of 100 translated trajectories from the Argoverse 1 dataset \cite{chang_argoverse_2019}, all aligned to start at the origin \((0, 0)\).}
\label{fig: 100 translated trajectories}
\end{figure}


We first simplify our formulation of Equation \ref{eqn: trajectory representation}. An agent's initial position in global coordinates does not reflect its motion pattern and is not of interest for trajectory fitting. To eliminate its influence, all trajectories are translated to start at the origin position \( (0, 0) \), as visualized in Figure \ref{fig: 100 translated trajectories}. 

This translation allows us to fix the first control point $\mathbf{w}_0 = \mathbf{0}$, corresponding to the start position, and exclude it from the fitting process. Consequently, the parameter vector is reduced to $\boldsymbol{\omega} = [\mathbf{w}_1; \mathbf{w}_2; \ldots; \mathbf{w}_N] \in \mathbb{R}^{Nd}$, and the corresponding basis matrix becomes $\boldsymbol{\Phi} \in \mathbb{R}^{Nd \times Td}$. This dimensionality reduction not only simplifies the fitting process but also ensures that the analysis focuses on the dynamics of trajectories rather than their start positions. We adopt this simplified formulation throughout the remainder of this chapter.

To enable Bayesian regression, we make the following assumptions regarding the prior distribution and observation noise:
\begin{itemize}[leftmargin=*]
\item \textbf{Prior}: We assume a Gaussian prior over the trajectory parameter vector $\boldsymbol{\omega} \in \mathbb{R}^{Nd}$, given by:
\begin{equation}
\label{eqn: omega prior}
    p(\boldsymbol{\omega}) = \mathcal{N}(\boldsymbol{\omega}|\boldmath{0}, \boldsymbol{\Sigma}_{\boldsymbol{\omega}})
\end{equation}
where $\boldsymbol{\Sigma}_{\boldsymbol{\omega}} \in \mathbb{R}^{Nd \times Nd}$ is the prior covariance matrix. This distribution reflects how trajectory parameters are statistically distributed across the motion datasets. The distribution is assumed to be zero-mean, based on the assumption that agent motion in the \emph{global} coordinate frame is approximately symmetric -- e.g., agents can move from left to right or right to left with similar likelihood. The covariance $\boldsymbol{\Sigma}_{\boldsymbol{\omega}}$ captures variability and correlation across polynomial parameters and spatial dimensions.
\item \textbf{Observation Noise}: We denote the observed trajectory of $i$-th agent as $\mathbf{z}_i$, and its observed position at timestep $t$ as $\mathbf{z}_{i, t}$, to distinguish them from the fitted trajectory model. We assume additive, zero-mean Gaussian observation noise $\boldsymbol{\epsilon}$ for each observed position: 
\begin{equation}
\label{eqn: observation noise}
\begin{aligned}
\mathbf{z}_{i,t} = \mathbf{c}_{i}(\tau_t) + \boldsymbol{\epsilon}, \:\:
\boldsymbol{\epsilon} \sim \mathcal{N}(\mathbf{0}, \boldsymbol{\Sigma}_{o, i, t}) 
\end{aligned}
\end{equation}
where $\boldsymbol{\Sigma}_{o,i,t} \in \mathbb{R}^{d\times d}$ is the covariance matrix for the $t$-th timestep of agent $i$. Assuming temporally independent noise along a single trajectory, we construct the observation noise covariance for the complete trajectory $\mathbf{c}_i$ as a block diagonal matrix $\boldsymbol{\Sigma}_{o, i}\in \mathbb{R}^{Td \times Td}$, expressed as:
\begin{equation}
\label{eqn: noise block}
\boldsymbol{\Sigma}_{o,i} = 
\begin{bmatrix}
\boldsymbol{\Sigma}_{o,i,1} & \mathbf{0} & \cdots & \mathbf{0} \\
\mathbf{0} & \boldsymbol{\Sigma}_{o,i,2} & \cdots & \mathbf{0} \\
\vdots & \vdots & \ddots & \vdots \\
\mathbf{0} & \mathbf{0} & \cdots & \boldsymbol{\Sigma}_{o,i,T}
\end{bmatrix}
\end{equation}
where the $t$-th diagonal block corresponds to $\boldsymbol{\Sigma}_{o,i,t}$.
\end{itemize}

Given these assumptions, the posterior estimate of model parameters for a single trajectory $\mathbf{c}_i$ is given in closed form \cite[pp. 232--234]{murphy_machine_2012}:
\begin{equation}
\label{eqn:posterior_mean}
\begin{aligned}
\boldsymbol{\Sigma}^{\mathrm{post}}_{\boldsymbol{\omega}, i} = & (\boldsymbol{\Sigma}_{\boldsymbol{\omega}}^{-1} + \boldsymbol{\Phi}  \boldsymbol{\Sigma}_{o, i}^{-1} \boldsymbol{\Phi}^{\top})^{-1}\\
\boldsymbol{\omega}^{\mathrm{post}}_i = & \boldsymbol{\Sigma}^\mathrm{post}_{\boldsymbol{\omega}, i}\boldsymbol{\Phi} \boldsymbol{\Sigma}_{o,i}^{-1} \mathbf{z}_i
\end{aligned}
\end{equation}
where $\boldsymbol{\Phi} \in \mathbb{R}^{Nd \times Td}$ is the matrix encoding the basis functions at each time step, and $\boldsymbol{\omega}^{\mathrm{post}}_i$ and $\boldsymbol{\Sigma}^{\mathrm{post}}_{\boldsymbol{\omega}, i}$ are the posterior mean and covariance of the trajectory parameters, respectively.

Using these posterior estimates, we can then compute the \emph{average fit error} (AFE) across trajectories as:
\begin{equation}
\mathrm{AFE} = \frac{1}{AT}\sum_i^A\sum_t^T ||(\boldsymbol{\phi}^{\top}(\tau_{t}) \otimes \mathbf{I}_d)\boldsymbol{\omega}_i^{\mathrm{post}} - \mathbf{z}_{i,t}||_2
\end{equation}
where $A$ denotes the number of considered agents, and $\boldsymbol{\phi}(\tau_t)$ is is the vector of basis functions evaluated at time $\tau_t$. 

By comparing the AFE with the average displacement error reported by SotA models, we can assess whether using polynomial representations limits overall prediction performance. However, the above estimation of fit error requires specifying the observation covariance $\boldsymbol{\Sigma}_{o, i}$, the prior covariance $\boldsymbol{\Sigma}_\omega$, and the model complexity $N$ -- none of which is given a priori. This motivates us to employ the \emph{empirical Bayes} method to estimate all three quantities.

\section{Empirical Bayes Method}
\label{sec 3.4: Empirical Bayes Method}
The empirical Bayes method \cite{efron_large_2012} enables the estimation of prior distributions over model parameters if many independent samples of the same phenomenon are observed, such as the agent trajectories in our datasets. The core idea is to express the likelihood of \emph{all} observed agent trajectories $\mathbf{Z} = \{\mathbf{z}_1, \cdots, \mathbf{z}_A \}$ as a function of the prior parameters alone by marginalizing out the trajectory parameters for each trajectory. The optimal prior parameters are obtained by maximizing the resulting type-\upperRomannumeral{2} likelihood (also known as the marginal likelihood). Based on our assumptions of Equation \ref{eqn: omega prior} and \ref{eqn: observation noise}, the type-\upperRomannumeral{2} likelihood is expressed as: 
\begin{equation}
\begin{aligned}
p(\mathbf{Z} \mid \boldsymbol{\Sigma}_{o,i}, \boldsymbol{\Sigma}_{\boldsymbol{\omega}}) 
&= \prod_{i=1}^A \int p(\mathbf{z}_i \mid \boldsymbol{\omega}_i, \boldsymbol{\Sigma}_{o,i}) \, p(\boldsymbol{\omega}_i \mid \boldsymbol{\Sigma}_{\boldsymbol{\omega}}) \, d\boldsymbol{\omega}_i \\
&= \prod_{i=1}^A \int \underbrace{\mathcal{N}(\mathbf{z}_i \mid \boldsymbol{\Phi}_{i}^{\top} \boldsymbol{\omega}_i, \boldsymbol{\Sigma}_{o,i})}_{\shortstack{\text{\scriptsize likelihood} \\ \text{ \scriptsize (polynomial with noise)}}} \, \underbrace{\mathcal{N}(\boldsymbol{\omega}_i \mid \mathbf{0}, \boldsymbol{\Sigma}_{\boldsymbol{\omega}})}_{\text{\scriptsize prior}} \, d\boldsymbol{\omega}_i
\end{aligned}
\label{eqn: type2_likelihood}
\end{equation}
Since both the likelihood and the prior are Gaussian distributions, this integral has a closed-form solution. Following \cite[pp. 172--176]{murphy_machine_2012}:
\begin{equation}
\begin{aligned}
p(\mathbf{Z} \mid \boldsymbol{\Sigma}_{o,i}, \boldsymbol{\Sigma}_{\boldsymbol{\omega}}) 
= &\prod_{i=1}^A \mathcal{N}(\mathbf{z}_i|\boldsymbol{0}, \boldsymbol{\Sigma}_{o,i} + \boldsymbol{\Phi}^{\top}_{i} \boldsymbol{\Sigma}_{\boldsymbol{\omega}} \boldsymbol{\Phi}_{i})\\ 
\end{aligned}
\label{eqn: type2_likelihood_simplified}
\end{equation}
Ideally, the only prior parameter to estimate would be the prior covariance $\boldsymbol{\Sigma}_{\boldsymbol{\omega}}$. The components of observation noise covariance matrices $\boldsymbol{\Sigma}_{o,i}$ would be derived from the ego vehicle's sensor setup and provided directly with the dataset. Unfortunately, none of the datasets offer ground-truth observation noise statistics. As a result, we estimate these quantities from data using the empirical Bayes approach.


In the following sections, we present our approach to parameterize the prior covariance $\boldsymbol{\Sigma}_{\boldsymbol{\omega}}$, the observation noise covariance $\boldsymbol{\Sigma}_{o,i}$, the optimizing objective, and the selection of optimal polynomial degree $\hat{N}$.

\subsection{Modeling the Prior Covariance}
\label{sec 3.4.1: Modeling the Prior Covariance}
As a valid covariance matrix, the prior covariance $\boldsymbol{\Sigma}_{\boldsymbol{\omega}}$ must be symmetric and positive semi-definite. To ensure these properties and enable efficient optimization, we parameterize it via its Cholesky decomposition:
\begin{equation}
\label{eqn: prior cov decomp} 
\begin{aligned}
\boldsymbol{\Sigma}_{\boldsymbol{\omega}} = \boldsymbol{L}_{\boldsymbol{\omega}}\boldsymbol{L}_{\boldsymbol{\omega}}^{\top}
\end{aligned}
\end{equation}
where $\boldsymbol{L}_{\boldsymbol{\omega}} \in \mathbb{R}^{Nd \times Nd}$ is the lower triangular matrix with positive diagonal entries.

Equation \ref{eqn: prior cov decomp} guarantees that $\boldsymbol{\Sigma}_{\boldsymbol{\omega}}$ is positive definite and avoids the need for constrained optimization. The  Cholesky factor $\boldsymbol{L}_{\boldsymbol{\omega}}$ contains $\frac{1}{2}Nd(Nd+1)$ independent parameters, which fully define the structure of the prior covariance. 


\subsection{Modeling the Observation Noise Covariance}
\label{sec 3.4.2: observation noise modeling}
As defined in Equation \ref{eqn: noise block}, the observation noise covariance $\boldsymbol{\Sigma}_{o,i}$ is constructed as a block-diagonal matrix, where each diagonal block corresponds to a per-timestep covariance $\boldsymbol{\Sigma}_{o,i,t}$. Estimating each $\boldsymbol{\Sigma}_{o,i,t}$
independently for every agent and timestep is infeasible. Instead, we introduce a structured parameterization in the form: 
\begin{equation}
\boldsymbol{\Sigma}_{o,i}=\boldsymbol{\Sigma}_{o,i}(\boldsymbol{\theta})
\end{equation}
where $\boldsymbol{\theta}$ denotes a shared set of parameters to be optimized.

Additionally, ego and non-ego trajectories originate from fundamentally different sensing pipelines. Ego vehicle trajectories are recorded using direct localization methods such as Global Navigation Satellite System (GNSS) and Inertial Measurement Unit (IMU) systems, providing relatively high-precision measurements. In contrast, non-ego agent trajectories are obtained through perception-based tracking using the ego vehicle's sensors (LiDAR, cameras, radar) \cite{wilson_argoverse2_2021, ettinger_waymo_2021}. Therefore, we adopt distinct parameterizations for their respective observation noise models: $\boldsymbol{\theta}=[\boldsymbol{\theta}^{\text{ego}}, \boldsymbol{\theta}^{\text{ne}}]$.

\subsubsection{Observation Noise Covariance for Ego Trajectories}
The observation noise covariance for ego trajectories is directly modeled in \emph{global} coordinates. We assume the observation noise of ego position -- the localization error -- at each timestep follows the same zero-mean Gaussian distribution. The per-timestep covariance is constructed as:
\begin{equation}
\label{eqn: ego_cov}
\begin{aligned}
\boldsymbol{\Sigma}_{o,i,t}^{\mathrm{ego, global}} & = \boldsymbol{\Sigma}_{o}^{\mathrm{ego, global}} = 
\begin{bmatrix}
\sigma^2_{\mathrm{x}} & \sigma_{\mathrm{xy}} \\
\sigma_{\mathrm{xy}} & \sigma^2_{\mathrm{y}} \\
\end{bmatrix} 
\end{aligned}
\end{equation}
where we assume isotropic variance $\sigma^2_{\mathrm{x}} = \sigma^2_{\mathrm{y}} = \sigma^2_{\mathrm{diag}}$ and off-diagonal covariance $\sigma_{\mathrm{xy}} = \sigma_{\mathrm{cov}}$. The first superscript indicates the \emph{ego} trajectory, while the second denotes the \emph{global} coordinate frame. Although the subscript $i$ is not strictly necessary for ego trajectories (since there is only one ego vehicle), we retain it for consistency with the notation used for non-ego agent trajectories. 

Assuming that observation noise is identically distributed and temporally independent, the full observation noise covariance for an ego trajectory of timesteps $T$ is constructed as a block-diagonal matrix:  
\begin{equation}
\label{eqn: ego noise block}
\boldsymbol{\Sigma}_{o,i}(\boldsymbol{\theta}^{\mathrm{ego}}) = 
\begin{bmatrix}
\boldsymbol{\Sigma}_{o}^{\mathrm{ego, global}} & \mathbf{0} & \cdots & \mathbf{0} \\
\mathbf{0} & \boldsymbol{\Sigma}_{o}^{\mathrm{ego, global}} & \cdots & \mathbf{0} \\
\vdots & \vdots & \ddots & \vdots \\
\mathbf{0} & \mathbf{0} & \cdots & \boldsymbol{\Sigma}_{o}^{\mathrm{ego, global}}
\end{bmatrix} = \mathbf{I}_T \otimes \boldsymbol{\Sigma}_o^{\mathrm{ego, global}}
\end{equation}
Under this formulation, the observation noise model for ego trajectories is governed by just two scalar parameters: 
\begin{equation}
\label{eqn: ego_ob_parameters}
\begin{aligned}
\boldsymbol{\theta}^{\mathrm{ego}}=[\sigma_{\mathrm{diag}}, \sigma_{\mathrm{cov}}]. 
\end{aligned}
\end{equation}

\subsubsection{Observation Noise Covariance for Non-Ego Trajectories}
Non-ego agent trajectories are obtained via the ego vehicle’s perception and tracking system. In the datasets considered, all ego vehicles are equipped with high-resolution LiDAR sensors, and the reported positions of surrounding agents are primarily influenced by LiDAR-based detections. 

\begin{figure}[thb]
\centering
\includegraphics[width=0.5\textwidth]{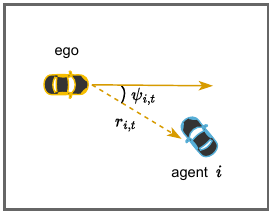}
\caption{Measurement in the polar coordinate}
\label{fig: lidar model}
\end{figure}

Since LiDAR measurements are originally acquired in polar coordinates, we model the observation uncertainty for non-ego agent positions in the same coordinate system. Let $\psi_{i,t}$ denote the measured angle, and $r_{i,t}$ be the radial distance from the ego vehicle to agent $i$ at timestep $t$, as visualized in Figure \ref{fig: lidar model}. We assume a constant angular uncertainty $\sigma^2_{\psi}$ and a distance-dependent radial uncertainty $\sigma^2_{r, i,t}$. The resulting observation noise covariance matrix for the $t$-th measurement of agent $i$ in polar coordinates is given by:
\begin{equation}
\label{eqn_cov_polar}
\begin{aligned}
\boldsymbol{\Sigma}_{o, i,t}^{\mathrm{ne, polar}} &= 
\begin{bmatrix}
\sigma^2_{\psi} & 0 \\
0 &  \sigma^2_{r, i,t}\\
\end{bmatrix}
\end{aligned}
\end{equation}
To model radial uncertainty, we acknowledge that while LiDAR sensors can estimate distances with high accuracy and low variance for individual point measurements -- even at long ranges \cite{lambert_performance_2020} -- the reported agent positions represent centroids reconstructed from multiple LiDAR returns and fused with other perception sensors \cite{chang_argoverse_2019, wilson_argoverse2_2021, ettinger_waymo_2021}. The accuracy of this reconstruction process can degrade significantly with distance \cite{jiang_perception_2025} due to factors such as reduced sensor resolution, lower signal-to-noise ratios, and increased tracking uncertainty. Therefore, the radial uncertainty $\sigma_{r,i,t}^2$ is modeled as a quadratic function of the measured distance $r_{i,t}$ between ego and agent $i$ at the $t$-$\mathrm{th}$ timestep, using three non-negative parameters $[b_0, b_1, b_2] \in \mathbb{R}^3_{\geq0}$:
\begin{equation}
\label{eqn: cov_r}
\sigma^2_{r,i,t} = b_{0}+ b_{1} r_{i,t} + b_{2}r_{i,t}^2
\end{equation}
This functional form captures the empirically observed degradation in perception performance by allowing for non-linear growth in measurement uncertainty with distance.

\begin{figure}[thb]
\centering
\includegraphics[width=\textwidth]{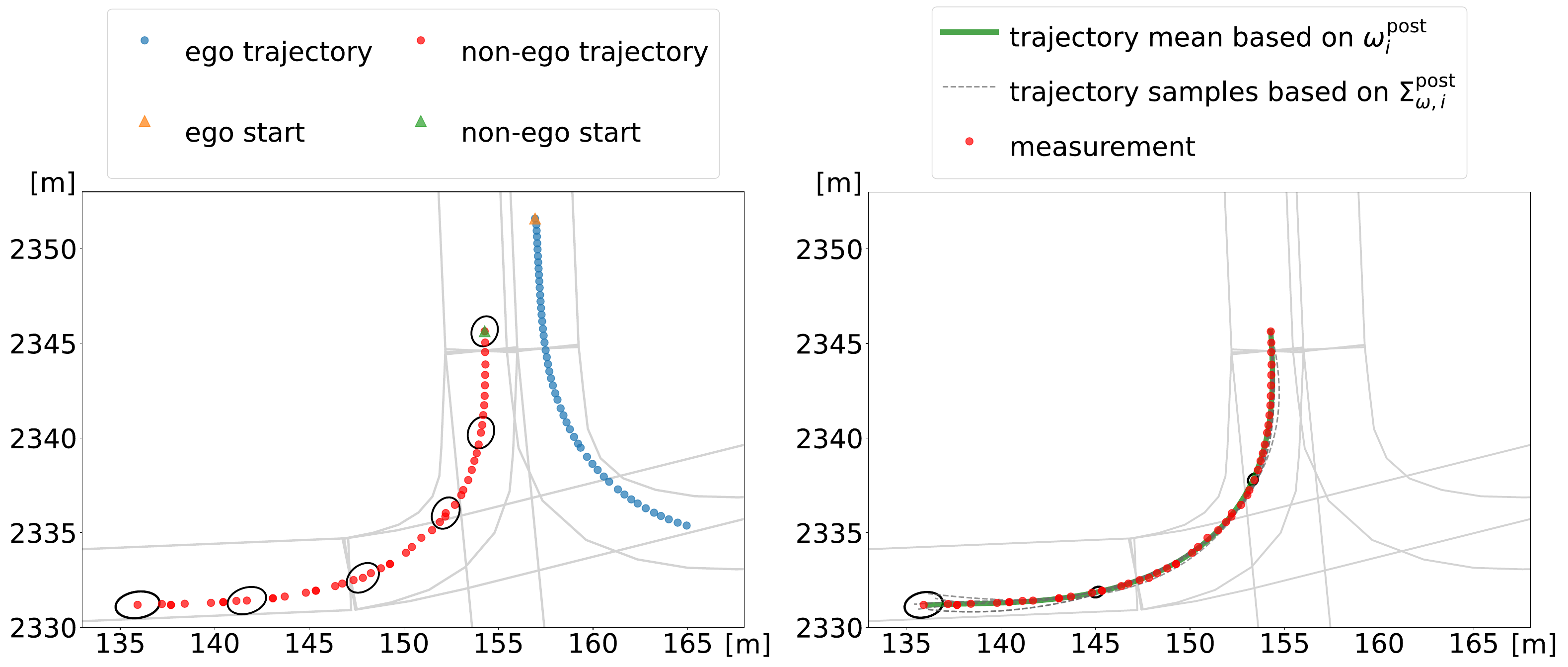}
\caption[Observation noise and posterior estimates]{\textbf{Left:} A typical scene from Argoverse 1 Motion dataset \cite{chang_argoverse_2019}. Data is gathered by a moving sensor platform (ego vehicle) and subsequently transformed into a fixed world coordinate frame. As the distance and angle between the sensor and non-ego agent change during recording, the observation covariance of non-ego agent locations stretches and rotates over time. We show all sample points and a few $95\%$ confidence ellipses for agent position, enlarged by a factor of $4$ for better visibility. \textbf{Right:} The same agent trajectory as in the \textbf{left} panel but fitted with a $5$-degree polynomial trajectory representation estimated via Equation \ref{eqn:posterior_mean}. The resulting posterior covariances for agent positions are enlarged by a factor of $8$ for better visibility.}
\label{fig: observation noise and posterior}
\end{figure}

To express the observation noise model in global coordinates, we first transform this polar-coordinate covariance into the ego-vehicle's local Cartesian frame using the relative agent-ego geometry. Following \cite[p. 77]{kampchen_feature_2007}, the transformed observation covariance in the $ego$ frame is expressed as:
\begin{equation}
\label{eqn: agt cov in ego}
\begin{aligned}
\boldsymbol{\Sigma}_{o, i,t}^{\mathrm{ne, ego}}=\begin{bmatrix}
\sigma^2_{\mathrm{lon, lon}} & \sigma_{\mathrm{lon,lat}} \\
\sigma_{\mathrm{lon,lat}} & \sigma^2_{\mathrm{lat, lat}} \\
\end{bmatrix}
\end{aligned}
\end{equation}
where the variances in the ego's longitudinal and lateral directions, as well as their covariance, are defined as:
\begin{equation}
\begin{aligned}
\sigma^2_{\mathrm{lon, lon}} &= \sigma^2_{r,i,t} \cos^2(\psi_{i,t}) + \sigma^2_{\psi} r_{i,t}^2 \sin^2(\psi_{i,t}) \\
\sigma^2_{\mathrm{lat,lat}} &= \sigma^2_{r,i,t} \sin^2(\psi_{i,t}) + \sigma^2_{\psi} r_{i,t}^2 \cos^2(\psi_{i,t}) \\
\sigma_{\mathrm{lon,lat}} &= (\sigma^2_{r,i,t} - \sigma^2_{\psi} r_{i,t}^2) \sin(\psi_{i,t}) \cos(\psi_{i,t})
\end{aligned}
\end{equation}
While the observation covariance of non-ego agents has been constructed from a perception-centric perspective, other factors also influence perception quality -- such as dataset-specific post-processing artifacts \cite{chang_argoverse_2019, wilson_argoverse2_2021, ettinger_waymo_2021}. To account for these residual sources of noise, we incorporate an additive isotropic term $\sigma_{c}^2$ along the diagonal, representing uncertainty arising from all remaining, unmodeled factors. The final observation covariance in the global coordinate frame is obtained by rotating the adjusted ego-frame covariance matrix:
\begin{equation}
\label{eqn_cov_ego}
\begin{aligned}
\boldsymbol{\Sigma}_{o, i,t}^{\mathrm{ne, global}} = \boldsymbol{R}_{t}^{\mathrm{rot}}(\boldsymbol{\Sigma}_{o, i,t}^{\mathrm{ne, ego}} +  \sigma_{c}^2 \mathbf{I}_d)(\boldsymbol{R}^{\mathrm{rot}}_{t})^{\top}
\end{aligned}
\end{equation}
where $\boldsymbol{R}^{\mathrm{rot}}_{t} \in \mathbb{R}^{2 \times 2}$ denotes the rotation matrix defined by the ego vehicle's heading $\gamma^{\text{ego, global}}_t$ in the global coordinate at timestep $t$ as:
\begin{equation}
\begin{aligned}
\boldsymbol{R}^{\mathrm{rot}}_{t}=\begin{bmatrix}
\cos(\gamma^{\text{ego,global}}_t) & -\sin(\gamma^{\text{ego,global}}_t) \\
\sin(\gamma^{\text{ego,global}}_t) & \cos(\gamma^{\text{ego,global}}_t) \\
\end{bmatrix}
\end{aligned}
\end{equation}
The observation covariance $\boldsymbol{\Sigma}_{o,i}(\boldsymbol{\theta}^{\mathrm{ne}})$ for one complete non-ego trajectory consists of an $Td\times Td$ block diagonal matrix, where the $t$-th block corresponds to $\boldsymbol{\Sigma}_{o, i,t}^{\mathrm{ne, global}}$. In total, we have only five parameters to estimate: 
\begin{equation}
\boldsymbol{\theta}^{\mathrm{ne}}=[\sigma_\psi, b_0, b_1, b_2, \sigma_c]
\end{equation}
An example of the modeled observation noise is visualized in the left panel of Figure \ref{fig: observation noise and posterior}, with covariance marked as black ellipses.

\subsection{Optimizing Objective}
With the prior covariance $\boldsymbol{\Sigma}_{\boldsymbol{\omega}}$ and observation noise covariance $\boldsymbol{\Sigma}_{o,i}(\boldsymbol{\theta})$ modeled, the type-\upperRomannumeral{2} likelihood (cf. Equation \ref{eqn: type2_likelihood_simplified}) becomes:
\begin{equation}
\label{eqn: optimizing target}
\begin{aligned}
p(\mathbf{Z} \mid \boldsymbol{\Sigma}_{o,i}(\boldsymbol{\theta}), \boldsymbol{\Sigma}_{\boldsymbol{\omega}}) 
= &\prod_{i=1}^A \mathcal{N}(\mathbf{z}_i|\boldsymbol{0}, \boldsymbol{\Sigma}_{o,i}(\boldsymbol{\theta}) + \boldsymbol{\Phi}^{\top}_{i} \boldsymbol{\Sigma}_{\boldsymbol{\omega}} \boldsymbol{\Phi}_{i})
\end{aligned}
\end{equation}
The optimizing objective is to maximize the log-likelihood with respect to prior and observation noise parameters, $\boldsymbol{\Sigma}_{\boldsymbol{\omega}}$ and $\boldsymbol{\theta}$ respectively. We perform this optimization using the gradient descent method, following \cite{dillon_tensorflow_2017}:
\begin{equation}
\label{eqn: prior_optimization}
\hat{\boldsymbol{\theta}}, \hat{\boldsymbol{\Sigma}}_{\boldsymbol{\omega}} =  \argmax_{\boldsymbol{\theta}, \boldsymbol{\Sigma}_{\boldsymbol{\omega}}} \log(p(\mathbf{Z} | \boldsymbol{\Sigma}_{o,i}(\boldsymbol{\theta}), \boldsymbol{\Sigma}_{\boldsymbol{\omega}}))
\end{equation}
Given a polynomial degree $N$, the posterior estimates can be obtained by substituting the optimized parameters $\hat{\boldsymbol{\Sigma}}_{\boldsymbol{\omega}}$ and $\boldsymbol{\Sigma}_{o, i}(\hat{\boldsymbol{\theta}})$ into Equation \ref{eqn:posterior_mean}. The right panel of Figure \ref{fig: observation noise and posterior} visualizes an example of the estimated posterior.

\subsection{Estimating Optimal Polynomial Degree}
The expressiveness of the Bernstein polynomial representation increases with the polynomial degree $N$, leading to reduced fit error and improved fit quality. However, this benefit comes with the risk of overfitting, as the estimated parameters may begin to capture noise rather than the underlying ground-truth motions. It is therefore essential to balance fitting fidelity with model complexity.

To quantify this trade-off, we employ the Akaike Information Criterion (AIC) \cite{akaike_AIC_1973} and the Bayesian Information Criterion (BIC) \cite{schwarz_BIC_1978}, which score models based on their log-likelihood penalized by polynomial degree. For our setting, AIC and BIC are defined as:
\begin{equation}
\begin{aligned}
\label{eqn:AIC}
\mathrm{AIC} &= \frac{\log(p(\mathbf{Z}| \boldsymbol{\Sigma}_{o,i}(\boldsymbol{\theta}),\boldsymbol{\Sigma}_{\boldsymbol{\omega}}))}{A}  - \mathrm{dof}(\boldsymbol{\theta}, \boldsymbol{\Sigma}_{\boldsymbol{\omega}})\\
\mathrm{BIC} & = \frac{\log(p(\mathbf{Z} | \boldsymbol{\Sigma}_{o,i}(\boldsymbol{\theta}), \boldsymbol{\Sigma}_{\boldsymbol{\omega}}))}{A} - \frac{\mathrm{dof}(\boldsymbol{\theta}, \boldsymbol{\Sigma}_{\boldsymbol{\omega}})}{2} \log(T)
\end{aligned}
\end{equation}
Here, $T$ denotes the trajectory timesteps, and $\mathrm{dof}(\boldsymbol{\theta},\boldsymbol{\Sigma}_{\boldsymbol{\omega}})$ represents the degrees of freedom (i.e., the number of parameters), computed as:
\begin{equation}
\begin{aligned}
\mathrm{dof}(\boldsymbol{\theta}, \boldsymbol{\Sigma}_{\boldsymbol{\omega}}) &= \mathrm{dof}(\boldsymbol{\theta}) + \mathrm{dof}(\boldsymbol{\Sigma}_{\boldsymbol{\omega}}) \\ &= \mathrm{dof}(\boldsymbol{\theta}) + \frac{1}{2} Nd(Nd + 1)
\end{aligned}
\end{equation}
Here, $\mathrm{dof}(\boldsymbol{\Sigma}_{\boldsymbol{\omega}}) = \frac{1}{2} Nd(Nd + 1)$ represents the number of parameters in the Cholesky factor $\boldsymbol{L}_{\boldsymbol{\omega}}$ (cf. Section \ref{sec 3.4.1: Modeling the Prior Covariance}). For ego and non-ego agents, we have $\mathrm{dof}(\boldsymbol{\theta}^{\mathrm{ego}})=2$ and $\mathrm{dof}(\boldsymbol{\theta}^{\mathrm{ne}})=5$, respectively. The polynomial degree $N$ that maximizes AIC or BIC is considered the optimal polynomial degree $\hat{N}$, as it optimally balances goodness of fit and generalization. In general, BIC imposes a stronger penalty on model complexity than AIC and therefore tends to favor simpler models.

\section{Outlier Filtering}
\label{sec 3.5: outlier filtering}
Since both the prior covariance and observation noise covariance are assumed to follow Gaussian distributions, outlier trajectories can significantly distort empirical Bayes estimates and should be filtered before analysis. Several factors may cause a trajectory to be classified as an outlier, including timing inconsistencies, static non-informative agents, and artifacts introduced by tracking systems. In the following, we describe the procedures used to detect and remove such outlier trajectories from the dataset.

\subsection{Timing Inconsistency and Static Trajectory}
Recorded trajectories can exhibit deviations from the nominal sampling rate (\SI{10}{Hz} in the studied datasets). To account for this, we include all trajectory samples that fall within the time window \([T' - 0.5,\, T' + 0.5)\) seconds when analyzing trajectories with a target duration $T'$. This approach allows for minor deviations from the nominal sampling rate.

Static trajectories are trivial to fit and typically yield near-zero approximation error, resulting in artificially low optimal polynomial degrees. To avoid bias, we discard all trajectories whose spatial length is less than or equal to $0.5$ meters.

\subsection{Tracking Issues}
\label{sec 3.5.2: tracking issues}
\begin{figure}[thb]
\centering
\includegraphics[width=0.9\textwidth]{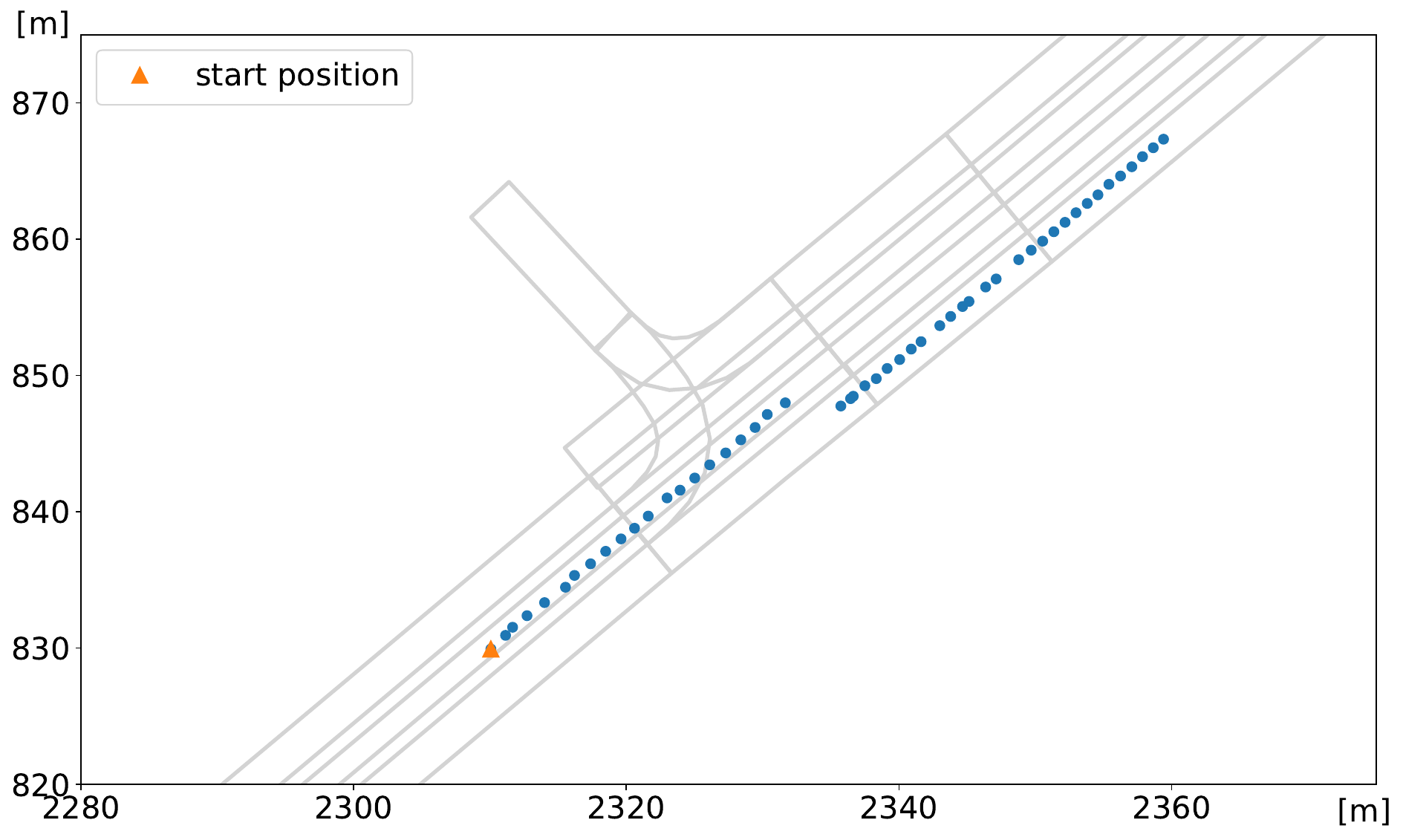}
\caption[Example of a tracking outlier]{Tracking outlier in A1 \cite{chang_argoverse_2019} caused by misaligned track association, resulting in physically implausible agent motion.}
\label{fig: tracking outlier}
\end{figure}

We observe tracking-related outliers, i.e., agents are reported at physically implausible locations between successive frames. These are typically caused by missing data due to occlusion or sensor range limitations, or by tracking artifacts such as misaligned track associations, as visualized in Figure \ref{fig: tracking outlier}.

To automatically detect such anomalies, we employ Rauch-Tung-Striebel (RTS) \cite{rauch_maximum_1965} smoother, a fixed-interval smoothing algorithm based on the Kalman filter \cite{kalman_new_1960}. Compared to the Kalman filter with a single forward pass \cite{kalman_new_1960}, the RTS smoother employs a two-pass process, which leverages both past and future observations, yielding trajectories that are more consistent with expected kinematic behavior and better suited for anomaly detection.

In the following, we describe our approach for using the RTS smoother to identify outlier trajectories.

\subsubsection{State Transition Model}
We model agent motion using a double integrator in 2D, where the state vector is defined as:
\begin{equation}
    \mathbf{x}_{i, t} = 
\begin{bmatrix}
p_x \\ p_y \\ v_x \\ v_y \\ a_x \\ a_y
\end{bmatrix}_{i, t}
\end{equation}
This state represents the position $(p_x, p_y)$, velocity $(v_x, v_y)$, and acceleration $(a_x, a_y)$ of the $i$-th agent at timestep $t$. For clarity, we omit the agent index $i$ in the remainder of this section.

Assuming a time interval $\delta_t$ between successive timesteps $t$ and $t+1$, the agent's motion follows a linear dynamical model with constant acceleration. The state transition is expressed as:
\begin{equation}
    \label{eqn: state transition model}
    \mathbf{x}_{t+1} = \mathbf{F}_t \mathbf{x}_t + \boldsymbol{\epsilon}_{p,t}, \quad \boldsymbol{\epsilon}_{p,t} \sim \mathcal{N}(\mathbf{0}, \mathbf{Q}_t)
\end{equation}
where $\mathbf{F}_t$ is the state transition matrix:
\begin{equation}
    \mathbf{F}_t = \begin{bmatrix}
1 &  \delta_t & \frac{1}{2}\delta_t^2  \\
0 & 1 & \delta_t \\
0 & 0& 1  \\
\end{bmatrix} \otimes \textbf{I}_2
\end{equation}
The Kronecker product $\otimes$ with the \(2 \times 2\) identity matrix $\mathbf{I}_2$ extends the 1D motion model to two dimensions, resulting in a $6 \times 6$ transition matrix for the full state vector \((p_x, p_y, v_x, v_y, a_x, a_y)\).

The process noise $\boldsymbol{\epsilon}_{p,t}$ with covariance matrix $\mathbf{Q}_t$ accounts for model uncertainty and unmodeled dynamics. $\mathbf{Q}_t$  is constructed following~\cite{reece_introduction_2010}:
\begin{equation}
   \mathbf{Q}_t = q\begin{bmatrix}
\frac{\delta_t^5 }{20}& \frac{\delta_t^4 }{8} & \frac{\delta_t^3 }{6}  \\
\frac{\delta_t^4 }{8}  & \frac{\delta_t^3 }{3} &\frac{\delta_t^2 }{2} \\
\frac{\delta_t^3 }{6} & \frac{\delta_t^2 }{2}& \delta_t  \\
\end{bmatrix} \otimes \textbf{I}_2
\end{equation}
where $q \geq 0$ is the intensity of the process noise. Since the smoother is
used only to detect physically implausible outliers rather than to reconstruct
true kinematic states, its behavior depends on the relative weighting of process
and observation noise rather than on the absolute scale of $q$. $q$ is therefore
fixed to $q = 1$, and the weighting is set through the observation-noise
covariance $\mathbf{R}_t$ described below.

\subsubsection{Observation Model}
The observation model relates the true state to the observed measurements as follows:
\begin{equation}
    \label{eqn: observation model}
    \mathbf{z}_t = \mathbf{H} \mathbf{x}_t + \boldsymbol{\epsilon}_{o,t}, \quad \boldsymbol{\epsilon}_{o,t} \sim \mathcal{N}(\mathbf{0}, \mathbf{R}_t)
\end{equation}
Here, $\mathbf{z}_t$ is the observation vector, $\mathbf{H}$ is the observation matrix, and $\boldsymbol{\epsilon}_{o,t}$ is zero-mean Gaussian observation noise with covariance $\mathbf{R}_t$. 

The structure of $\mathbf{H}$ depends on the dataset:
\begin{itemize}
    \item \textbf{A1 dataset:} The observation vector $\mathbf{z}_t$ only contains position measurements. The corresponding observation matrix is:
    \begin{equation}
        \mathbf{H}_{\text{A1}} = 
        \begin{bmatrix}
            1 & 0 & 0 & 0 & 0 & 0 \\
            0 & 1 & 0 & 0 & 0 & 0
        \end{bmatrix}
    \end{equation}

    \item \textbf{A2 and WO datasets:} The observation vector $\mathbf{z}_t$ contains both position and velocity measurements. The corresponding observation matrix is:
    \begin{equation}
        \mathbf{H}_{\text{A2/WO}} = 
        \begin{bmatrix}
            1 & 0 & 0 & 0 & 0 & 0 \\
            0 & 1 & 0 & 0 & 0 & 0 \\
            0 & 0 & 1 & 0 & 0 & 0 \\
            0 & 0 & 0 & 1 & 0 & 0
        \end{bmatrix}
    \end{equation}
\end{itemize}

The observation noise covariance matrix $\mathbf{R}_t$ is a tunable parameter and can be chosen based on the observation noise $\boldsymbol{\Sigma}_{o,i,t}$ estimated in the empirical Bayes analysis. However, the purpose of the RTS smoother in our framework is not to optimally reconstruct the true kinematic states, but rather to detect physically implausible outliers during preprocessing. For this purpose, we assume $\mathbf{R}_t$ to be a diagonal, time-invariant matrix while retaining the subscript $t$ for notational consistency. Its values are selected empirically based on datasets, and the specific settings used for each dataset are provided in Table~\ref{tab: rts-params}.

\begin{table}[tbh]
\caption[Observation noise covariance of Rauch-Tung-Striebel smoother]{Observation noise covariance $\mathbf{R}_t$ of RTS smoother}
\centering
\begin{tabularx}{\textwidth}{c >{\centering\arraybackslash}X  >{\centering\arraybackslash}X >{\centering\arraybackslash}X}
\Xhline{3\arrayrulewidth}
Datasets & A1 & A2 & WO\\
\Xhline{3\arrayrulewidth}
ego &$\operatorname{diag}\scriptstyle(0.1, 0.1)$& $\operatorname{diag}\scriptstyle(0.005,0.005, 0.1,0.1)$ & $\operatorname{diag}\scriptstyle(0.005, 0.005, 0.1, 0.1)$\\
\hline
non-ego &$\operatorname{diag}\scriptstyle(1, 1)$& $\operatorname{diag}\scriptstyle(0.01, 0.01, 1, 1)$ & $\operatorname{diag}\scriptstyle(0.01, 0.01, 1, 1)$\\
\Xhline{3\arrayrulewidth}
\label{tab: rts-params}
\end{tabularx}
\end{table}

\subsubsection{Forward Pass}
The forward pass of the RTS smoother is identical to the standard Kalman filter implementation. The Kalman filter recursively estimates the state using a predict-update cycle, expressed as:

    \begin{enumerate}
        \item \textbf{Prediction Step:}
        \begin{equation}\label{eq:kalman_prediction}
        \begin{aligned}
        \hat{\mathbf{x}}_{t+1 | t} &= \mathbf{F}_t \hat{\mathbf{x}}_{t} \\
        \mathbf{P}_{t+1|t} &= \mathbf{F}_t \mathbf{P}_{t} \mathbf{F}_t^\top + \mathbf{Q}_t
        \end{aligned}
        \end{equation}
        where:
        \begin{itemize}
            \item $\hat{\mathbf{x}}_{t+1|t}$ is the predicted state at next timestep $t+1$,
            \item $\mathbf{P}_{t+1|t}$ is the predicted covariance,
            \item $\hat{\mathbf{x}}_{t}$ and $\mathbf{P}_{t}$ are the filtered state and covariance from the $t$-th timestep.
        \end{itemize}
        We initialize the state estimate $\hat{\mathbf{x}}_1$ using the available observed components at the first timestep. Specifically:
        \begin{itemize}
            \item In the \textbf{A1 dataset}, only the initial position is observed, so we set the position components accordingly and initialize the velocity and acceleration components to zero.
            \item In the \textbf{A2 and WO datasets}, both position and velocity are observed at the first timestep; these components are used to initialize the corresponding entries in $\hat{\mathbf{x}}_1$, while acceleration is initialized to zero.
        \end{itemize}
        
        The initial covariance matrix $\mathbf{P}_1$ is assumed to be diagonal. For observed components, the diagonal entries are set to the corresponding values from the observation noise covariance in Table \ref{tab: rts-params}. For unobserved components, we use fixed values: velocity variance is set to 10 in A1 (where velocity is unobserved), and acceleration variance is set to 10 in all datasets.
        \item \textbf{Update Step:}
        \begin{equation}\label{eq:kalman_update}
        \begin{aligned}
        \mathbf{K}_{t+1} &= \mathbf{P}_{t+1|t} \mathbf{H}^\top \left( \mathbf{H} \mathbf{P}_{t+1|t} \mathbf{H}^\top + \mathbf{R}_{t+1} \right)^{-1} \\
        \hat{\mathbf{x}}_{t+1} &= \hat{\mathbf{x}}_{t+1|t} + \mathbf{K}_{t+1} \left( \mathbf{z}_{t+1} - \mathbf{H} \hat{\mathbf{x}}_{t+1|t} \right) \\
        \mathbf{P}_{t+1} &= \left( \mathbf{I} - \mathbf{K}_{t+1} \mathbf{H} \right) \mathbf{P}_{t+1|t}
        \end{aligned}
        \end{equation}
        where:
        \begin{itemize}
            \item $\mathbf{K}_{t+1}$ is the Kalman gain,
            \item $\hat{\mathbf{x}}_{t+1}$ and $\mathbf{P}_{t+1}$ are the filtered state and covariance at timestep $t+1$.
        \end{itemize}
    \end{enumerate}

\subsubsection{Backward Pass}
The backward pass distinguishes the RTS smoother from the standard Kalman filter by incorporating future information to refine past estimates. Starting from the final timestep, we refine the filtered estimates using future information:
    \begin{equation}\label{eq:rts_smooth}
    \begin{aligned}
    \mathbf{C}_t &= \mathbf{P}_t \mathbf{F}_{t+1}^\top \left( \mathbf{P}_{t+1|t} \right)^{-1} \\
    \hat{\mathbf{x}}_t^s &= \hat{\mathbf{x}}_t + \mathbf{C}_t \left( \hat{\mathbf{x}}_{t+1}^s - \hat{\mathbf{x}}_{t+1|t} \right) \\
    \mathbf{P}_t^s &= \mathbf{P}_t + \mathbf{C}_t \left( \mathbf{P}_{t+1}^s - \mathbf{P}_{t+1|t} \right) \mathbf{C}_t^\top
    \end{aligned}
    \end{equation}
    where:
    \begin{itemize}
        \item $\mathbf{C}_t$ is the RTS smoothing gain,
         \item $\hat{\mathbf{x}}_t^s$ and $\mathbf{P}_t^s$ are the smoothed state and covariance at timestep $t$,
        \item $\hat{\mathbf{x}}_t$ and $\mathbf{P}_t$ are the filtered estimates from the forward pass,
        \item $\hat{\mathbf{x}}_{t+1}^s$ and $\mathbf{P}_{t+1}^s$ are the smoothed estimates at timestep $t+1$,
        \item $\hat{\mathbf{x}}_{t+1|t}$ and $\mathbf{P}_{t+1|t}$ are the predicted state and covariance from the forward pass.
    \end{itemize}
The initial values for the smoothed state and covariance are set as $\hat{\mathbf{x}}_{T}^s = \hat{\mathbf{x}}_{T}$ and  $\mathbf{P}_{T}^s = \mathbf{P}_T$, respectively.

Once the states are estimated using the RTS smoother, outlier detection is applied based on the smoothed state estimates. Additionally, in the A1 dataset -- where heading information is not provided -- the estimated velocity direction is used as a proxy for the agent's heading.

\subsubsection{Outlier Filtering Criteria}
Based on the estimates of agent states from the RTS smoother, a trajectory is classified as an outlier and discarded if:
\begin{itemize}[leftmargin=*]
\item The RTS-smoothed position deviates from the observed position by more than 2 meters, or
\item The estimated longitudinal acceleration (or deceleration) exceeds physically plausible limits:
\begin{itemize}
    \item For vehicles \cite{bokare_acceleration_2017}: [\( -10\,\mathrm{m/s}^2 \),\( +6\,\mathrm{m/s}^2 \)] 
    \item For cyclists \cite{famiglietti_bicycle_2020}: [\(  -4\,\mathrm{m/s}^2, +2\,\mathrm{m/s}^2 \)] 
    \item For pedestrians \cite{zkebala_pedestrian_2012}: [\( -3\,\mathrm{m/s}^2 \),\( +2\,\mathrm{m/s}^2 \)] 
\end{itemize}
\end{itemize}

These thresholds are designed to strike a balance between filtering out physically implausible trajectories and retaining as much data as possible. For vehicles, we allow a relatively wider acceleration range to preserve more samples. For cyclists and pedestrians -- where limited empirical research exists to quantify typical acceleration bounds -- we select the thresholds based on observed motion patterns in the dataset. This empirical tuning ensures that filtering remains conservative while still eliminating the most egregious outliers.

The results of outlier filtering are reported in Section \ref{sec: outlier filtering results}.






\section{Experiments}
\label{sec 3.6: Experiments}
\subsection{Experimental Setup}
We perform the empirical Bayes analysis on recorded trajectories from the training split of each dataset. Detailed counts of trajectories by agent type are provided in Table~\ref{tab: dataset characteristics cpt3}. 

Separate analyses are performed for the ego vehicle and for each type of non-ego agent -- including vehicles (excluding ego), cyclists, and pedestrians. For these non-ego categories, we select the annotated \emph{target agents} to analyze their trajectories across all datasets. Since A1 does not annotate agent types other than vehicles, it is excluded from the analysis of cyclist and pedestrian trajectories.

\begin{table}[!th]
\vspace{-0.0em}
\caption{Characteristics of the datasets under study}
\centering
\begin{tabularx}{\textwidth}{c c >{\centering\arraybackslash}X >{\centering\arraybackslash}X >{\centering\arraybackslash}X}
\Xhline{3\arrayrulewidth}
\multicolumn{2}{c}{Dataset (training split)} & A1  \cite{chang_argoverse_2019} & A2 \cite{wilson_argoverse2_2021} & WO \cite{ettinger_waymo_2021}\\
\Xhline{3\arrayrulewidth}
\multicolumn{2}{l}{\#scenarios, \#ego trajectories} & \num{205942} & \num{199908} & \num{487002}\\
\hline
\multicolumn{2}{l}{\#agent (vehicle) trajectories}  & \num{205942} & \num{175602} & \num{1829161} \\
\hline
\multicolumn{2}{l}{\#agent (cyclist) trajectories}  & - & \num{2995} & \num{62785} \\
\hline
\multicolumn{2}{l}{\#agent (pedestrian) trajectories}  & - & \num{14417} & \num{231508} \\
\hline
\multicolumn{2}{l}{maximal time horizon [s]} & 5 & 11 & 9\\
\hline
\multicolumn{2}{l}{\#cities} & 2 & 6 & 6\\
\hline
\multicolumn{2}{l}{sampling rate} & \SI{10}{Hz}& \SI{10}{Hz} & \SI{10}{Hz}\\
\hline
\multirow{4}{*}{\thead{trajectory \\ information}} & position & 2D & 2D & 3D\\
\cline{2-5}
&velocity & - & 2D & 2D\\
\cline{2-5}
&heading & - & \checkmark & \checkmark\\
\cline{2-5}
&timestamp & \checkmark & \checkmark & \checkmark\\
\Xhline{3\arrayrulewidth}
\label{tab: dataset characteristics cpt3}
\vspace{-1em}
\end{tabularx}
\end{table}

Our analysis spans a range of trajectory durations $T' = 1\,\text{s}, 2\,\text{s}, \ldots, 9\,\text{s}$. For cases where $T'$ is shorter than the maximum available trajectory duration, we extract trajectory segments differently depending on the dataset:
\begin{itemize}[leftmargin=*]
    \item \textbf{A1:} We use a sliding window of size $T'$ with a stride of 1 second. This is due to timing inconsistencies in A1 (as shown in the left panel of Figure \ref{fig: issues in A1 and A2}), where measurements are not consistently recorded with \SI{10}{Hz}, resulting in fewer trajectories with valid trajectory durations.
    \item \textbf{A2:} We randomly sample a single segment of duration $T'$ from each trajectory.
    \item \textbf{WO:} Following the same sampling strategy as A2, we randomly extract one $T'$-second segment per trajectory. Since WO annotates multiple target agents per scene, it yields a significantly larger number of vehicle trajectories. To reduce computational overhead, we randomly subsample 300,000 non-outlier vehicle trajectories from the WO training split, following the outlier criteria described in Section \ref{sec 3.5: outlier filtering}.
\end{itemize}
The optimization is gradient-based, and the hyper-parameters used for training in the individual dataset are listed in Table \ref{tab: hyper parameters of EB}. 
\begin{table}[tbh]
\caption{Hyper-parameters of the empirical Bayes method}
\centering
\begin{tabularx}{\textwidth}{c  >{\centering\arraybackslash}X>{\centering\arraybackslash}X >{\centering\arraybackslash}X >{\centering\arraybackslash}X}
\Xhline{3\arrayrulewidth}
\multirow{2}{*}{dataset} &\multicolumn{4}{|c}{ego / non-ego} \\
\cline{2-5}
& \multicolumn{1}{|c}{optimizer} & learning rate & epochs & batch size \\
\Xhline{3\arrayrulewidth}
A1 & \multicolumn{1}{|c}{Adam \cite{kingma_adam_2014}}& 1e-3 / 5e-3& 300 / 200 & 1024 \\
\cline{1-5}
A2 &  \multicolumn{1}{|c}{Adam \cite{kingma_adam_2014}} & 1e-3 & 200 / 150  &1024\\
\cline{1-5}
WO &  \multicolumn{1}{|c}{Adam \cite{kingma_adam_2014}} & 1e-3 / 5e-3 & 150 / 100 & 1024\\
\Xhline{3\arrayrulewidth}
\label{tab: hyper parameters of EB}
\end{tabularx}
\end{table}

In addition to the overall Average Fit Error (AFE), we also report its projection onto the longitudinal and lateral directions of motion, as these components are of particular interest for driving applications.
Unless stated otherwise, AFE denotes the Euclidean distance rather than its longitudinal and lateral components.

\subsection{Outlier Filtering Results}
\label{sec: outlier filtering results}
Table \ref{tab: outlier} summarizes the percentage of trajectories discarded due to outlier detection for a trajectory duration of $T'=5\,\text{s}$ -- a setting widely used in Chapter \ref{cpt: improving generalization} and \ref{cpt: generative model}.

\begin{table}[!b]
\caption[Percentage of outliers in datasets]{Percentage of outliers for $T'=5$s trajectory. For the WO dataset, the percentage of outliers -- including pre-flagged invalid data -- is shown in parentheses.}
\centering
\begin{tabularx}{\textwidth}{c c !{\vrule width 1.5pt} >{\centering\arraybackslash}X |  >{\centering\arraybackslash}X | >{\centering\arraybackslash}X}
\Xhline{3\arrayrulewidth}
\multicolumn{2}{c}{Datasets}&\multicolumn{1}{c}{A1} & \multicolumn{1}{c}{A2} & \multicolumn{1}{c}{WO}\\
\Xhline{3\arrayrulewidth}
\multirow{5}{*}{ego} & time &22.81& 0 & 0.02\\
\cline{2-5}
&static &23.95&20.66& 25.41\\
\cline{2-5}
&out of view &0&0& 0\\
\cline{2-5}
&RTS &0&1.18& 0\\
\cline{2-5}
&total &42.95&21.84& 25.42\\
\hline
\hline
\multirow{5}{*}{\thead{agent \\ vehicle}} & time &22.81& 0 & 0.01 \; (0.02)\\
\cline{2-5}
&static &0&4.95& 1.35 \; (1.70)\\
\cline{2-5}
&out of view &0&0& 0 \; \; (19.40)\\
\cline{2-5}
&RTS &6.81&0.86& 0.05 (19.45)\\
\cline{2-5}
&total &28.11&5.81& 1.41 (20.83)\\
\hline
\hline
\multirow{5}{*}{\thead{agent \\ cyclist}} & time &-& 0 & 0.01 \; (0.02)\\
\cline{2-5}
&static &-&1.04& 1.38 \; (1.43)\\
\cline{2-5}
&out of view &-&0&  0 \; \; (24.12) \\
\cline{2-5}
&RTS &-&3.91& 0.62 (24.58)\\
\cline{2-5}
&total &-&4.94& 2.01 (25.62) \\
\hline
\hline
\multirow{5}{*}{\thead{agent \\ pedestrian}} & time &-& 0 & 0.01 \; (0.02)\\
\cline{2-5}
&static &-&0.19& 1.43 \; (1.32)\\
\cline{2-5}
&out of view &-&0& 0 \; \;  (32.20)\\
\cline{2-5}
&RTS &-&0.76&  0.07 (32.25)\\
\cline{2-5}
&total &-&0.93& 1.51 (33.12) \\
\Xhline{3\arrayrulewidth}
\label{tab: outlier}
\end{tabularx}
\end{table}

\begin{figure}[thb]
\centering
\includegraphics[width=\textwidth]{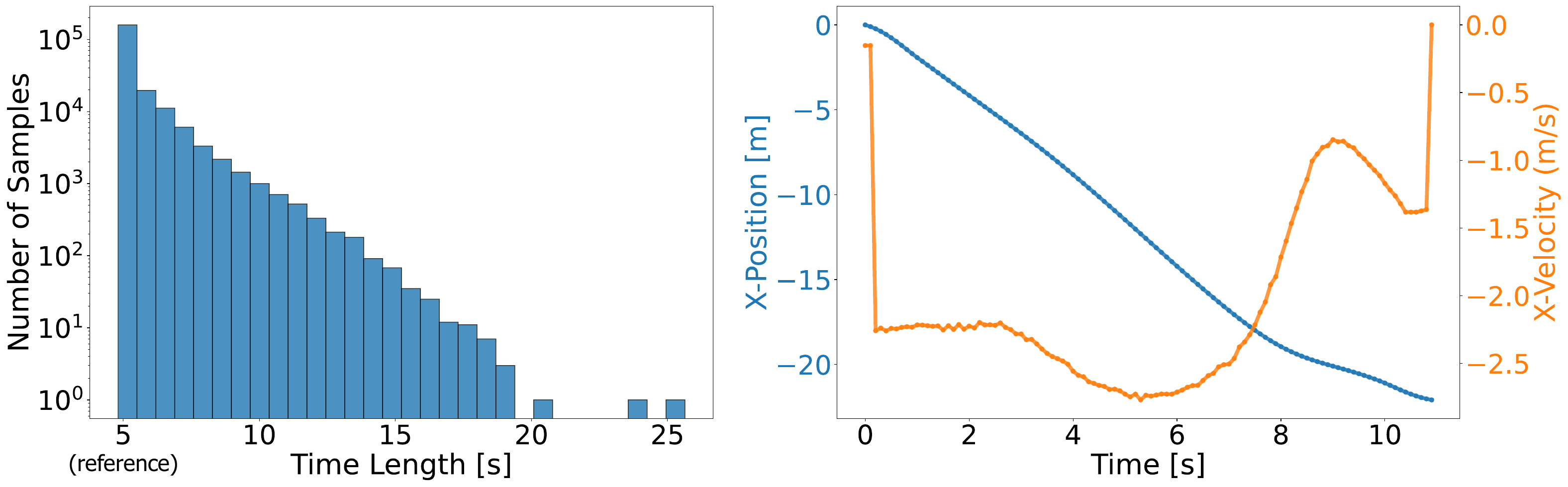}
\caption[Data issues in Argoverse 1 and Argoverse 2]{\textbf{Left:} Timing inconsistencies in the Argoverse 1 dataset, where a number of trajectories deviate from the nominal 5-second duration. \textbf{Right}: Position and velocity along the x-axis of an 11-second ego trajectory from the Argoverse 2 dataset, illustrating unstable velocity estimation at the beginning and end of the trajectory.}
\label{fig: issues in A1 and A2}
\end{figure}

Each dataset exhibits different issues in recordings. In A1, $22.81\%$ of trajectories deviate from the nominal sampling rate -- for instance, trajectories labeled as 5 seconds in duration can range from \SI{4.81}{s} to \SI{25.64}{s}, as visualized in the left panel of Figure \ref{fig: issues in A1 and A2}. These deviations are attributable to irregular timestamp spacing or inconsistencies in the effective sampling frequency, which nominally corresponds to \SI{10}{Hz}. A2 is the only dataset in which the ego vehicle occasionally exhibits physically implausible trajectories, primarily due to unstable velocity estimates at the beginning or end of the recorded segments, as visualized in the right panel of Figure \ref{fig: issues in A1 and A2}. 

The WO dataset pre-flags invalid observations for target agents, typically caused by occlusion or sensor range limitations (reported as ``out of view'' in Table~\ref{tab: outlier}). The RTS smoother also detects these pre-flagged observations as outliers. In Table~\ref{tab: outlier}, we report the percentage of outliers including these pre-flagged points in parentheses, while the main percentages exclude them.

Despite these pre-flagged invalid observations, WO demonstrates the highest overall tracking quality for non-ego agents, as reflected by the lowest rate of physically infeasible trajectories across all agent categories detected by the RTS smoother.


\subsection{Estimation of Observation Noise}
\label{sec: 3.6.3: observation noise estimation}
Estimated observation noise can vary significantly with trajectory durations and polynomial degrees. To mitigate this, we report the representative results for $T'=5\,\mathrm{s}$ in Table \ref{tab: observation noise}, using the optimal polynomial degree $N=\hat{N}$ that maximizes AIC (cf. Section \ref{sec: optimal polynomial degree}). The values of \( \sigma_r \) are reported at selected distances \( r = 10\,\mathrm{m},\, 20\,\mathrm{m},\, 40\,\mathrm{m} \) to provide better intuition.

Although observation noise is modeled differently for ego and non-ego agents -- making direct magnitude comparisons difficult -- it is evident that the $\sigma_c$ component of vehicle trajectory noise alone exceeds the total ego noise $\sigma_{diag}$ (e.g., \SI{0.044}{m} vs. \SI{0.012}{m} in the A2 dataset). This indicates that ego trajectories exhibit substantially lower observation noise than non-ego (vehicle) trajectories in all datasets, as expected.

\begin{table}[!tb]
\centering
\caption[Estimated observation noise]{Estimated observation noise for $T'=5$s at $\hat{N}$ maximizing AIC}
\vspace{-0.5em}
\begin{tabularx}{\textwidth}{c c !{\vrule width 1.5pt} >{\centering\arraybackslash}X | >{\centering\arraybackslash}X | >{\centering\arraybackslash}X}
\Xhline{3\arrayrulewidth}
\multicolumn{2}{c}{Datasets}&\multicolumn{1}{c}{A1} & \multicolumn{1}{c}{A2} & \multicolumn{1}{c}{WO}\\
\Xhline{3\arrayrulewidth}
\multirow{3}{*}{$\boldsymbol{\theta}^{\mathrm{ego}}$} & $\hat{N}$ & 5 & 6 & 6\\
\cline{2-5} 
& $\sigma_{diag}$ [m] &0.024& 0.012 & 0.008\\
\cline{2-5} 
&$\sigma_{cov}$ $[m^2]$ &2e-4&3e-6& -1e-7\\
\hline
\hline
\multirow{6}{*}{\thead{$\boldsymbol{\theta}^{\mathrm{ne}}$ \\ vehicle}} 
 & $\hat{N}$ & 3 & 5 & 5\\
\cline{2-5}
& $\sigma_{\alpha}$ [rad] &1e-3& 6e-4& 3e-4\\
\cline{2-5}
& $\sigma_{c}$ [m] &0.161 & 0.044& 0.017\\
\cline{2-5}
& \multirow{3}{*}{\thead{$\sigma_{r}$ [m] \\  $r=10\,\mathrm{m}, 20\,\mathrm{m}, 40\,\mathrm{m}$}} &0.128&0.055& 0.019\\
&&0.176&0.062& 0.027\\
& & 0.246&0.085& 0.042\\
\hline
\hline
\multirow{5}{*}{\thead{$\boldsymbol{\theta}^{\mathrm{ne}}$ \\ cyclist}} 
 & $\hat{N}$ & - & 5 & 5\\
\cline{2-5}
& $\sigma_{\alpha}$ [rad] &-& 2e-4& 3e-4\\
\cline{2-5}
& $\sigma_{c}$ [m] &- & 0.027& 0.027\\
\cline{2-5}
& \multirow{3}{*}{\thead{$\sigma_{r}$ [m] \\  $r=10\,\mathrm{m}, 20\,\mathrm{m}, 40\,\mathrm{m}$}} &-&0.014& 0.010\\
&&-&0.022& 0.015\\
&&-&0.039& 0.026\\
\hline
\hline
\multirow{5}{*}{\thead{$\boldsymbol{\theta}^{\mathrm{ne}}$ \\ pedestrian}} 
 & $\hat{N}$ & - & 5 & 5\\
\cline{2-5}
& $\sigma_{\alpha}$ [rad] &-& 3e-4& 2e-5\\
\cline{2-5}
& $\sigma_{c}$ [m] &- & 0.018& 0.015\\
\cline{2-5}
& \multirow{3}{*}{\thead{$\sigma_{r}$ [m] \\  $r=10\,\mathrm{m}, 20\,\mathrm{m}, 40\,\mathrm{m}$}} &-&0.007& 4e-4\\
&&-&0.010& 6e-4\\
& & -&0.017& 0.001\\
\Xhline{3\arrayrulewidth}
\vspace{-1em}
\label{tab: observation noise}
\end{tabularx}
\end{table}

Among the datasets, the agent trajectories in A1 are notably noisier than those in A2 and WO, with the highest values in $\boldsymbol{\theta}^{\mathrm{ego}}$ and $\boldsymbol{\theta}^{\mathrm{ne}}$. In contrast, the WO dataset demonstrates the lowest estimated observation noise for ego, vehicle, and pedestrian trajectories, likely due to its offline tracking algorithm \cite{ettinger_waymo_2021}. This finding is consistent with the outlier results in Table \ref{tab: outlier}, where the WO dataset shows the lowest percentage of outliers detected by the RTS-smoother.

\subsection{Optimal Polynomial Degree}
\label{sec: optimal polynomial degree}
\begin{figure}[thb]
\centering
\includegraphics[width=\textwidth]{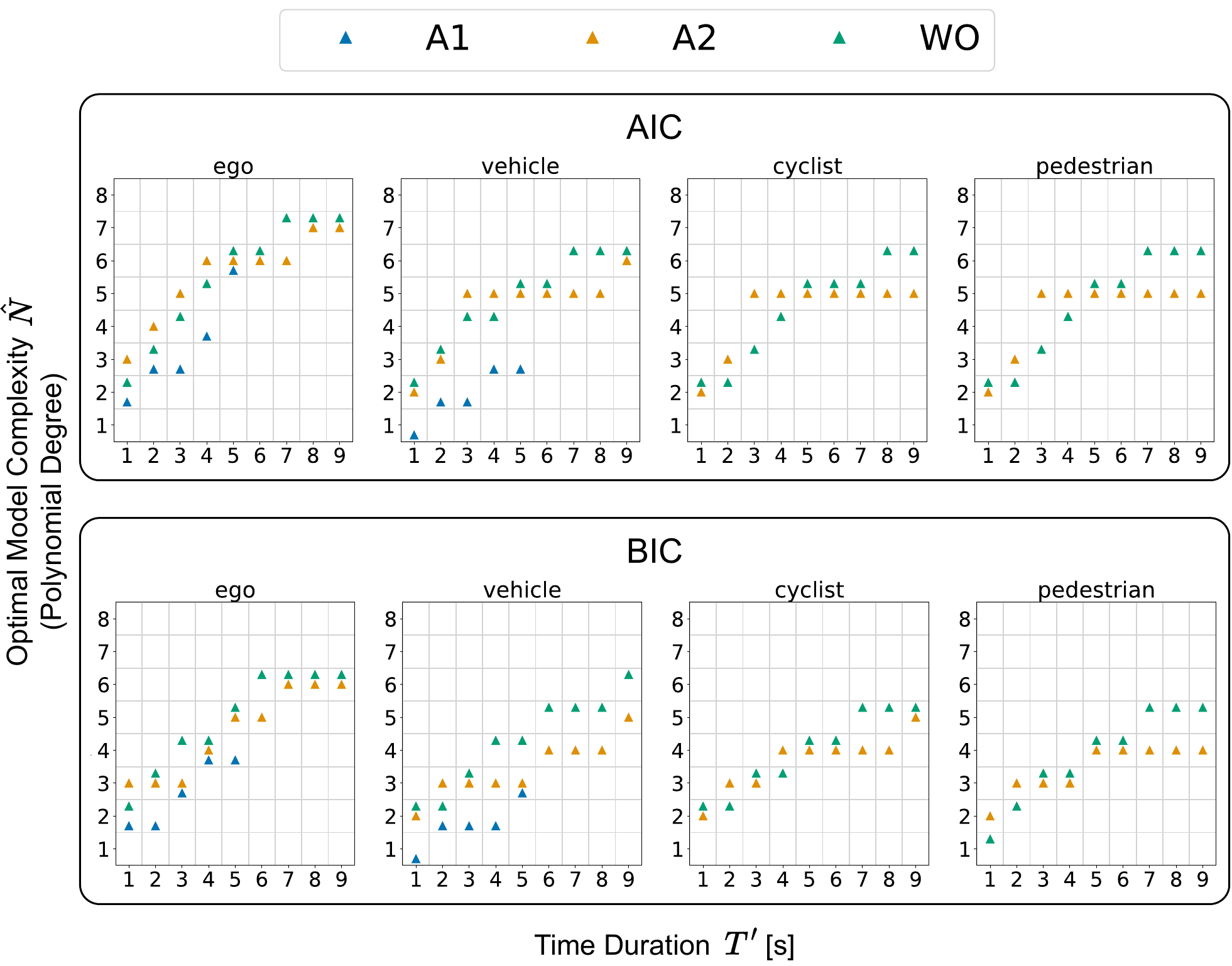}
\caption[Optimal polynomial degree]{Optimal polynomial degree based on AIC and BIC}
\label{fig: aic and bic}
\end{figure}

Figure \ref{fig: aic and bic} visualizes the optimal polynomial degree for trajectory durations from 1s to 9s, as determined by AIC and BIC. As expected, longer trajectories generally require higher polynomial degrees. However, both criteria consistently favor moderate degrees -- even for long trajectories. For example, AIC recommends at most a 7-degree polynomial for ego trajectories and a 6-degree polynomial for non-ego trajectories.

Despite variations in dataset characteristics and motion patterns of different agent types, the estimated optimal degrees remain relatively stable across trajectories of the same duration. For instance, AIC consistently selects a 6-degree polynomial for 5-second ego trajectories across all datasets, and a 5-degree polynomial for 5-second non-ego trajectories in both A2 and WO. In contrast, A1 exhibits higher measurement noise for non-ego trajectories, which is less amenable to polynomial fitting. As a result, lower-degree polynomials are preferred for non-ego trajectories in A1.

These results suggest that polynomial representations generalize well across different datasets and agent types.

\subsection{Trade-off between Polynomial Degree and Fit Error}
\label{sec 3.6.5: Trade-off between Model Complexity and Fit Error}
\begin{figure}[!thb]
\centering
\includegraphics[width=1.\textwidth]{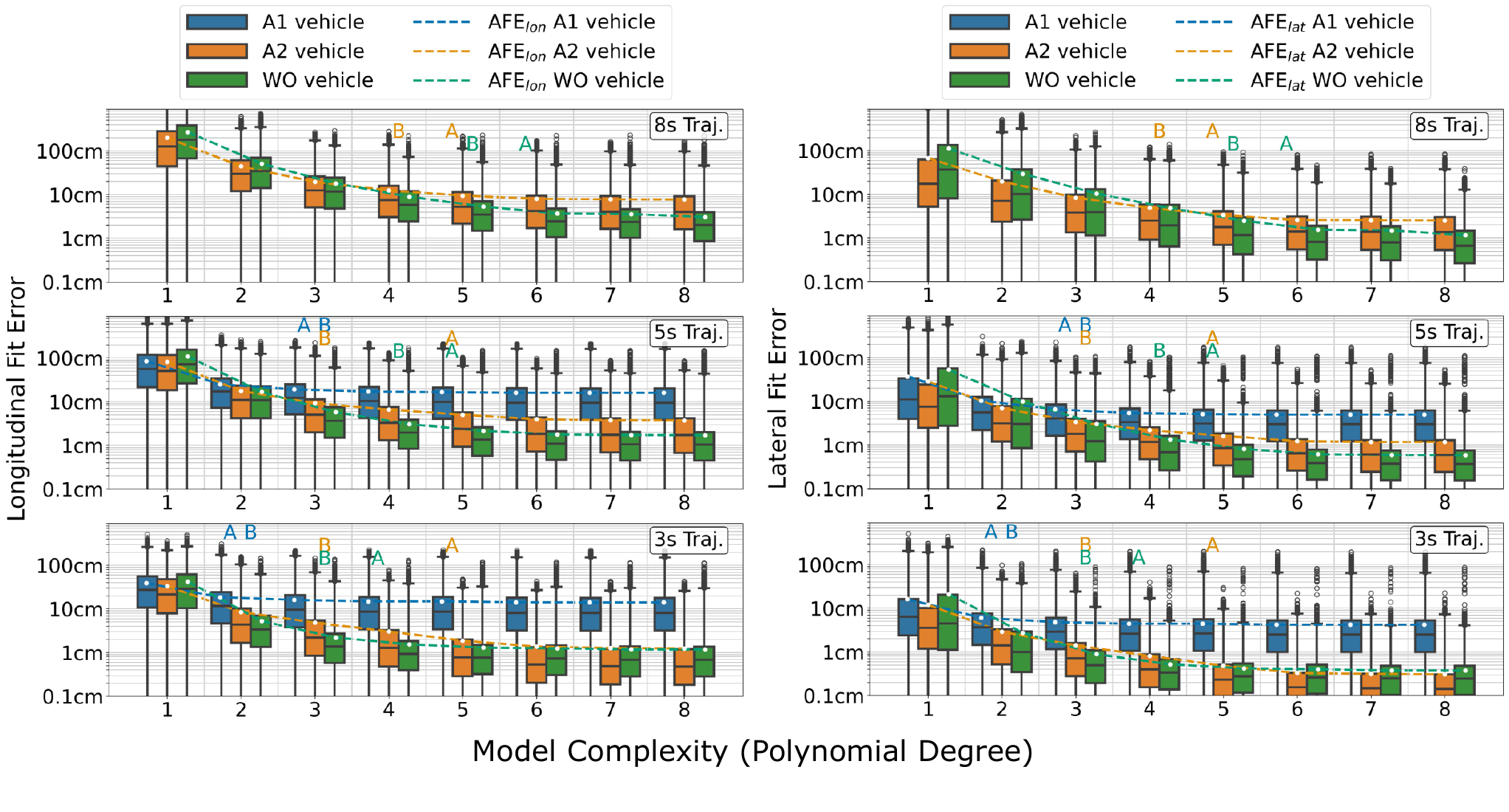}
\caption[Fit error and polynomial degree (vehicle)]{The longitudinal (left) and lateral (right) fit error of models for vehicle trajectories in A1, A2 and WO with $T' = 3\,\mathrm{s}, 5\,\mathrm{s}, 8\,\mathrm{s}$. ``A, B'' denote the model complexity $\hat{N}$ that maximizes AIC and BIC, respectively. The dashed lines indicate the mean values, and the upper whisker denotes the 99.9\% percentile.}
\label{fig: tradeoff vehicle}
\vspace{-1.0em}
\end{figure}

\begin{figure}[!hb]
\centering
\includegraphics[width=1.\textwidth]{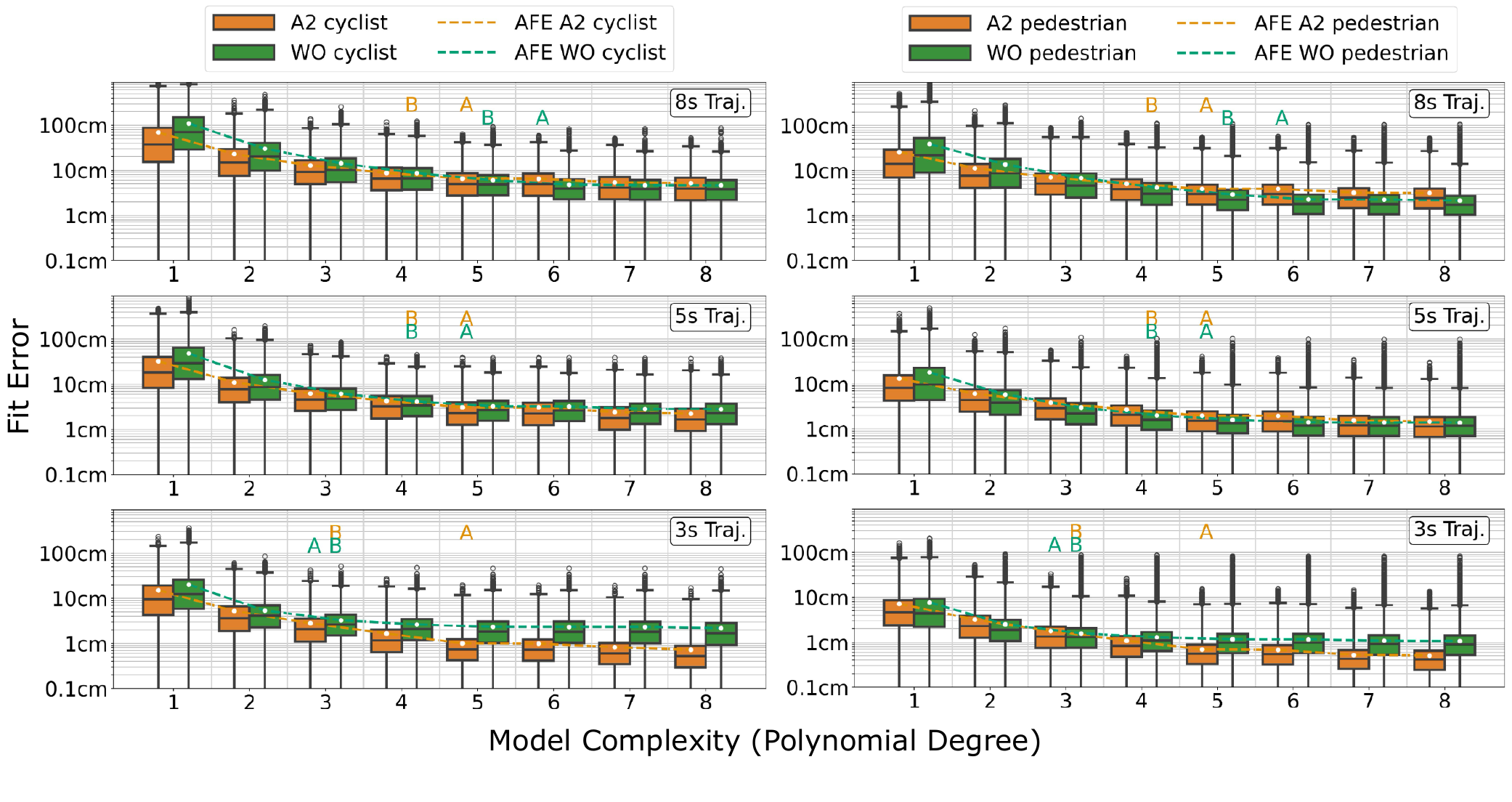}
\caption[Fit error and polynomial degree (cyclist and pedestrian)]{The fit error of models for cyclist (left) and pedestrian (right) trajectories in A2 and WO with $T' = 3\,\mathrm{s}, 5\,\mathrm{s}, 8\,\mathrm{s}$. ``A, B'' denote the model complexity $\hat{N}$ that maximizes AIC and BIC, respectively. The dashed lines indicate the mean values, and the upper whisker denotes the 99.9\% percentile.}
\label{fig: tradeoff cyclist and pedestrian}
\end{figure}

Figure \ref{fig: tradeoff vehicle} presents box-plots of longitudinal and lateral fit errors for vehicle trajectories with $T' = 3\,\mathrm{s}, 5\,\mathrm{s}, 8\,\mathrm{s}$. Figure \ref{fig: tradeoff cyclist and pedestrian} presents the fit errors for cyclist and pedestrian trajectories over the same durations. Detailed longitudinal and lateral fit errors, evaluated at the polynomial degree $N=\hat{N}$ that maximizes AIC, are summarized in Table \ref{tab: result summary}. 

These results show that the longer trajectories generally require higher model complexity for accurate representation. However, the benefits of increasing $N$ diminish beyond an optimal model $\hat{N}$, as identified by AIC or BIC. Overall, Bernstein polynomials of moderate model complexity can represent trajectories with very high fidelity. For example, a $6$-degree polynomial can approximate 8-second vehicle trajectories in WO with only $\SI{3.7}{cm}$ longitudinal and $\SI{1.6}{cm}$ lateral average fit error. 

Despite this, we observe large deviations (greater than 1 meter) in some samples, visualized as points beyond the upper whisker in Figures \ref{fig: tradeoff vehicle} and \ref{fig: tradeoff cyclist and pedestrian}. Inspecting these samples, we find they correspond to physically implausible measurements due to timing jitter or tracking issues that are not excluded by the RTS smoother. Several examples of trajectories with large deviations, along with randomly selected examples from all datasets, are visualized in Appendix \ref{appendix: examples of fitted trajectories}.

\begin{table}[!t]
\caption[Longitudinal and lateral fit error]{Longitudinal and lateral fit error [m]. $\hat{N}$ denotes the best polynomial degree according to AIC. $99.9\%$ means the 99.9 percentile of the fit error. 

\centering (Lon. | Lat.)}
\centering
\begin{tabularx}{\textwidth}{c c c!{\vrule width 1.5pt}>{\centering\arraybackslash}X|>{\centering\arraybackslash}X|>{\centering\arraybackslash}X}
\Xhline{2\arrayrulewidth}
&\multirow{2}{*}{$T'$ [s]}&& \multicolumn{3}{c}{Dataset} \\
\cline{4-6}
&& & A1 & A2& WO \\
\Xhline{3\arrayrulewidth}

\multirow{9}{*}{ego} 
&\multirow{3}{*}{3} & $\hat{N}$ & 3 & 5 &4 \\
&&AFE& 0.022\:\vline\:0.002& 0.004\:\vline\:0.001 & 0.004\:\vline\:0.001\\
&&99.9\%&  0.173\:\vline\:0.120 & 0.094\:\vline\:0.012 & 0.066\:\vline\:0.021\\
\cline{2-6}

&\multirow{3}{*}{5} & $\hat{N}$ & 5 & 6 &6 \\
&&AFE& 0.022\:\vline\:0.004& 0.007\:\vline\:0.002 & 0.005\:\vline\:0.002\\
&&99.9\%&  0.157\:\vline\:0.111 &0.143\:\vline\:0.030 & 0.071\:\vline\:0.025\\
\cline{2-6}

&\multirow{3}{*}{8} & $\hat{N}$ & - & 7 & 7 \\ &&AFE& -& 0.019\:\vline\:0.006 & 0.013\:\vline\:0.004\\
&&99.9\%&  - & 0.251\:\vline\:0.090 & 0.114\:\vline\:0.069\\

\hline
\hline
\multirow{9}{*}{vehicle} 
&\multirow{3}{*}{3} & $\hat{N}$ & 2 & 5 &4 \\
&&AFE& 0.185\:\vline\:0.060& 0.019\:\vline\:0.005 & 0.016\:\vline\:0.005\\
&&99.9\%&  1.753\:\vline\:0.835 & 0.312\:\vline\:0.114 & 0.359\:\vline\:0.056\\
\cline{2-6}

&\multirow{3}{*}{5} & $\hat{N}$ & 3 & 5 &5 \\
&&AFE& 0.191\:\vline\:0.065& 0.051\:\vline\:0.016 & 0.022\:\vline\:0.008\\
&&99.9\%&  1.774\:\vline\:0.848 &0.735\:\vline\:0.284 & 0.408\:\vline\:0.090\\
\cline{2-6}

&\multirow{3}{*}{8} & $\hat{N}$ & - & 5 & 6 \\ &&AFE& -& 0.093\:\vline\:0.033 & 0.037\:\vline\:0.016\\
&&99.9\%&  - & 1.115\:\vline\:0.461 & 0.483\:\vline\:0.186\\

\hline
\hline
\multirow{9}{*}{cyclist} 
&\multirow{3}{*}{3} & $\hat{N}$ & - & 5 &3 \\
&&AFE& -& 0.008\:\vline\:0.004 & 0.025\:\vline\:0.016\\
&&99.9\%&  - & 0.098\:\vline\:0.054 & 0.179\:\vline\:0.130\\
\cline{2-6}

&\multirow{3}{*}{5} & $\hat{N}$ & - & 5 &5 \\
&&AFE& - & 0.022\:\vline\:0.016 & 0.025\:\vline\:0.016\\
&&99.9\%&  - &0.216\:\vline\:0.183 & 0.174\:\vline\:0.123\\
\cline{2-6}

&\multirow{3}{*}{8} & $\hat{N}$ & - & 5 & 6 \\ &&AFE& -& 0.042\:\vline\:0.040 & 0.031\:\vline\:0.028\\
&&99.9\%&  - & 0.364\:\vline\:0.358 & 0.217\:\vline\:0.242\\

\hline
\hline
\multirow{9}{*}{pedestrian} 
&\multirow{3}{*}{3} & $\hat{N}$ & - & 5 &3 \\
&&AFE& - & 0.005\:\vline\:0.004 & 0.011\:\vline\:0.009\\
&&99.9\%&  - & 0.058\:\vline\:0.046 & 0.091\:\vline\:0.070\\
\cline{2-6}

&\multirow{3}{*}{5} & $\hat{N}$ & - & 5 &5 \\
&&AFE& - & 0.012\:\vline\:0.012 & 0.011\:\vline\:0.009\\
&&99.9\%&  - &0.135\:\vline\:0.126 & 0.086\:\vline\:0.066\\
\cline{2-6}

&\multirow{3}{*}{8} & $\hat{N}$ & - & 5 & 6 \\ &&AFE& -& 0.023\:\vline\:0.026 & 0.016\:\vline\:0.012\\
&&99.9\%&  - & 0.257\:\vline\:0.255 & 0.133\:\vline\:0.108\\
\Xhline{2\arrayrulewidth}
\label{tab: result summary}
\vspace{-1em}
\end{tabularx}
\vspace{-1em}
\end{table}

\subsection{Fit Error vs. Displacement Error}
Although polynomials can represent real-world trajectories with high fidelity, a key question addressed in this chapter is whether the representation error (bias) introduced by polynomial models can fundamentally limit the performance of predictors. To investigate this, we compare the fit error -- serving as an upper bound on the representation error -- with the displacement error of SotA methods that use unbiased, sequence-based trajectory representations. 

We select MultiPath++ \cite{varadarajan_multipath++_2022} and Wayformer \cite{nayakanti_wayformer_2022} for the A1 and WO datasets for $T' = 3\,\mathrm{s}, 5\,\mathrm{s}, 8\,\mathrm{s}$, and QCNet \cite{zhou_query_2023} and Forecast-MAE \cite{cheng_forecast_2023} for A2 dataset with $T'=6\,\mathrm{s}$. These selections align with the prediction tasks and evaluation setups defined in their respective benchmark competitions, ensuring a fair and meaningful comparison. Since the displacement error measured by $\text{minADE}_6$ (with 6 prediction modes) is typically much lower than $\text{minADE}_1$ (single best prediction), we use $\text{minADE}_6$ to assess whether fit error fundamentally limits the multi-modal prediction performance.

\begin{figure}[thb]
\centering
\includegraphics[width=0.9\textwidth]{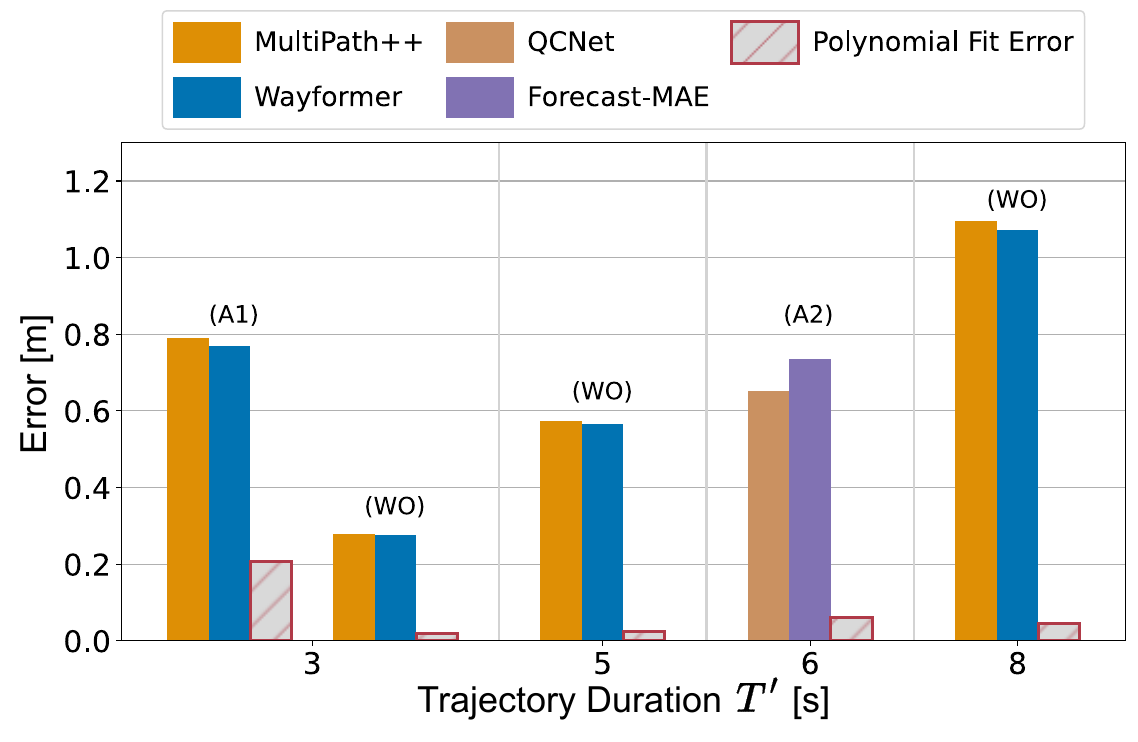}
\caption[Fit error vs. displacement error]{Comparison of the average fit error (AFE) of polynomial representations at $\hat{N}$ maximizing AIC and the average displacement error ($\text{minADE}_{6}$) of SotA predictors for vehicle trajectories.}
\label{fig: representation vs prediction}
\end{figure}

Figure \ref{fig: representation vs prediction} visualizes the fit error from the polynomial representation with optimal polynomial degree $\hat{N}$ that maximizes AIC, and the displacement error from SotA predictors for vehicle trajectories. Note that both fit error and displacement error are dependent on the observation noise level, as shown in Figure \ref{fig: components of displacement error}. If the observation noise level is low, as in A2 and WO, the representation error of the polynomial representation, upper bounded by the fit error, is negligible in the prediction task. Even in noisier datasets such as A1, the fit error remains significantly lower than the displacement error.

While we focus on vehicle trajectories for this comparison due to consistent evaluation across datasets, our results in Section \ref{sec 3.6.5: Trade-off between Model Complexity and Fit Error} demonstrate similarly low fit errors for cyclist and pedestrian trajectories in A2 and WO.

These results suggest that the bias introduced by polynomial representation does not fundamentally limit the performance of prediction models.

\section{Conclusion and Discussion}
\label{sec 3.7: Conclusion and Discussion}
This chapter presented a comprehensive analysis of polynomial trajectory representations using Bernstein polynomials on three large-scale motion datasets. The primary objective was to evaluate the suitability of these representations for agent trajectory modeling in traffic scene prediction, focusing on fit fidelity, optimal model complexity, and explicit consideration of observation noise.

Using a principled empirical Bayes framework, we quantified the trade-off between polynomial degree and fit error across various agent types (vehicles, cyclists, and pedestrians) and trajectory durations. Our results show that polynomial representations can approximate real-world trajectories with high fidelity using moderate polynomial degrees and do not pose a fundamental limitation to prediction performance. Moreover, polynomial representations generalize well across diverse agent types and datasets. The estimated priors and observation noise parameters can effectively regularize the measurement uncertainty and dataset-specific biases, offering valuable guidance for prediction models -- an aspect that will be discussed in detail in Chapter \ref{cpt: improving generalization}.

These findings support the central hypothesis that polynomial representations -- such as Bernstein polynomials -- are a practical and effective choice for trajectory modeling in autonomous driving. The enforced continuity and compactness of the polynomial form offer distinct advantages over discrete sequence-based representations, particularly in terms of computational efficiency and generalization potential.

Additionally, we notice several issues in the A1 dataset -- such as high noise levels, temporal inconsistencies (as discussed in Section \ref{sec 3.5: outlier filtering}), and a relatively short prediction horizon of 3 seconds -- that can significantly impact prediction performance. As a result, we exclude the A1 dataset from the analyses in Chapter \ref{cpt: improving generalization} and Chapter \ref{cpt: generative model}.

Nonetheless, several limitations remain. While information criteria such as AIC and BIC help identify optimal polynomial degrees per dataset, these values are context-dependent and may require adaptive selection based on scene characteristics or system constraints. Additionally, the assumption of Gaussian priors and observation noise may not fully capture the complexity of real-world distributions.

In conclusion, this chapter provides a foundational justification for using polynomial trajectory representations in learning-based traffic scene prediction systems. The insights gained here directly inform the design of robust and efficient models developed in Chapters \ref{cpt: improving generalization} and \ref{cpt: generative model}, with an emphasis on improving out-of-distribution generalization and multi-agent scene coherence.

\chapter{Improving Prediction Generalization via Polynomial Representations}
\label{cpt: improving generalization}

This chapter builds upon the author's conference paper published at the \textit{IEEE/RSJ International Conference on Intelligent Robots and Systems (IROS)} 2024 \cite{yao_improving_2024}\footnote{\copyright~2024 IEEE. Reprinted, with permission, from Y. Yao, S. Yan, D. Goehring, W. Burgard, and J. Reichardt, ``Improving Out-of-Distribution Generalization of Trajectory Prediction for Autonomous Driving via Polynomial Representations,'' in \textit{Proc. IEEE/RSJ Int. Conf. Intelligent Robots and Systems (IROS)}, pp.~488--495, 2024.}, co-authored with Shengchao Yan, Prof. Dr. Daniel Goehring (advisor), Prof. Dr. Wolfram Burgard, and Dr. Joerg Reichardt. The work has been significantly extended and refined for this chapter.

The author was responsible for the conceptual development and implementation
of the methods presented in Sections \ref{sec 4.3: data homogenization},
\ref{sec 4.4: represent data with polynomials}, and
\ref{sec 4.5: model design}. The author further conducted all experiments
and evaluations of the proposed models (EP variants) reported in
Section \ref{sec 4.6: experiments}. Shengchao Yan conducted the experiments and evaluations of the benchmark models presented in Section \ref{sec 4.6: experiments}.
Prof. Dr. Daniel Goehring, Prof. Dr. Wolfram Burgard, and
Dr. Joerg Reichardt contributed through scientific discussions and supervision.


The corresponding implementation is publicly available at: \url{https://github.com/aumovio/everything-polynomial}.

\section{Introduction and Motivation}
\label{sec 4.1: introduction and motivation}
Chapter \ref{cpt: Empirical Bayes Analysis} demonstrated that the representation error of Bernstein polynomials is significantly smaller than the prediction errors observed in SotA prediction models. These results provide strong justification for adopting polynomial representations in practical prediction systems. Building on this insight, this chapter addresses the key question: How can Bernstein polynomials be effectively integrated into prediction systems, and what impact do they have on prediction performance? We first focus on the more straightforward \emph{marginal} prediction task to evaluate this approach, and extend it to the more challenging \emph{joint} prediction task in Chapter~\ref{cpt: generative model}.

To rigorously assess the impact of integrating Bernstein polynomials into trajectory prediction systems, it is essential to understand the evaluation landscape. Prediction performance is typically benchmarked through structured competitions that define specific metrics and test protocols (cf. Section \ref{sec: advancements in motion datasets}). Therefore, we begin by reviewing the evaluation frameworks and limitations of current benchmarks, with particular focus on model generalization.

Motion datasets such as Argoverse 2 (A2) \cite{wilson_argoverse2_2021} and Waymo Open (WO) \cite{ettinger_waymo_2021}, along with their associated competitions, have significantly advanced the development of traffic scene prediction models. These competitions provide a set of standardized metrics (cf. Section \ref{sec: evaluation metrics}) and test protocol that scores prediction systems on withheld test data. This enables objective comparison of model performance and assessment of generalization ability.

However, there are still inherent similarities between training and test samples -- such as shared sensor configurations, map formats, data processing pipelines, and biases in geographic and data selection. As an example of data selection bias, the WO dataset explicitly favors the selection of prediction targets that deviate from constant-velocity or straight-line motion \cite{ettinger_waymo_2021} to increase the diversity of trajectory patterns \cite{feng_unitraj_2024}. These similarities introduce dataset-specific biases that affect both training and test partitions, challenging claims of generalization. Consequently, the test performance reported from these benchmarks primarily reflects \emph{In-Distribution} (ID) performance, rather than evidence of true generalization.

\begin{figure}[!t]
\centering
\includegraphics[width=0.9\textwidth]{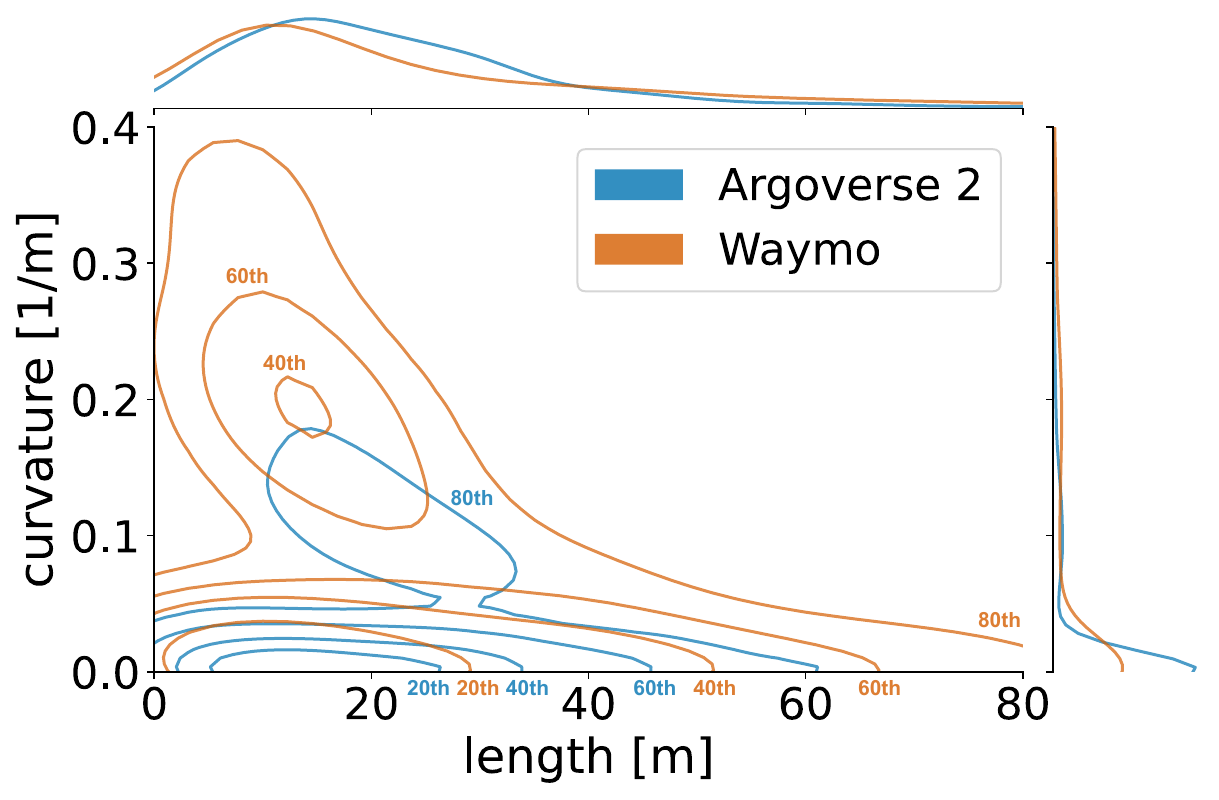}
\caption[Distribution shift in map representations across datasets]{Kernel density plot of the maximum absolute curvature and length for $5000$ random lane segments in A2 and WO. Contours indicate the $20$-th, $40$-th, $60$-th, and $80$-th percentiles, respectively. The plot reveals a clear distribution shift in map geometry between A2 and WO.}
\label{fig: data_distribution_shift}
\end{figure}

The development of SotA traffic scene prediction models is largely driven by their ID performance on individual public datasets -- models are often tuned explicitly to their respective benchmarks. This raises concerns about overfitting to specific data distributions. In contrast, for practical deployment, traffic scene prediction models must perform reliably across a wide range of conditions, independent of dataset-specific biases. This requires evaluation under \emph{Out-of-Distribution} (OoD) settings, for example, by testing models across different datasets. As an example, Figure
\ref{fig: data_distribution_shift} clearly visualizes the distribution shift caused by different map generation processes between A2 and WO. However,
the efforts for cross-dataset evaluation are hampered by
the heterogeneous data format and task specifications among datasets and competitions. Specifically, models trained on one dataset are difficult to apply to other datasets due to variations in traffic scene duration, observation history length, and prediction horizon.

To address these challenges, this chapter introduces a homogenization protocol that aligns data formats and prediction tasks across two large-scale motion datasets: A2 and WO. This homogenized framework enables systematic OoD testing and evaluation of prediction models. The Argoverse 1 (A1) dataset is \emph{not} included in this chapter due to its high noise levels and time inconsistencies, as noted in Chapter \ref{cpt: Empirical Bayes Analysis}.

Using the homogenized evaluation framework across datasets, this chapter further explores a fundamental question: How can prediction models be designed to enhance generalization across diverse datasets?

A common strategy to improve robustness in deep learning models is to increase the amount of training data -- data augmentation. As demonstrated in \cite{bahari_vehicle_2022}, generalization performance can be improved by augmenting training data with programmatically created variations. However, this approach increases computational requirements and may require careful data selection to ensure diversity and representativeness \cite{buda_systematic_2018}.

A complementary strategy involves constraining model expressiveness through limited input and output representations, thereby reducing the risk of overfitting to dataset-specific biases. Polynomial representations, in particular, offer a compact and physically interpretable approach to modeling agent motion and map structure. In addition, they also help regularize observation noise, as discussed in Chapter \ref{cpt: Empirical Bayes Analysis}, and enable consistent map format alignment across datasets without reliance on sample density. These properties provide theoretical advantages, including enhanced generalization, smaller model size, reduced training effort, and faster inference times. 

In this chapter, our primary experiment trains models on the smaller A2 and tests them on the larger WO dataset. This setup is intended to isolate generalization ability from training data advantages. Accordingly, we select benchmark models that are originally designed for A2. As a complementary experiment, we reverse this setup -- training on WO and testing on A2 -- where improved performance is expected for all models due to the larger training data volume.

This chapter is organized as follows: Section \ref{sec 4.2: related work} reviews recent marginal prediction models and their design decisions regarding data representation and augmentation strategies. We select two SotA benchmark models
and detail how the dataset and competition characteristics have
shaped their model designs. Section \ref{sec 4.3: data homogenization} details our dataset homogenization protocol that enables systematic OoD testing across different motion datasets. Section \ref{sec 4.4: represent data with polynomials} details our approach for representing both input and output data using polynomial representations. Section \ref{sec 4.5: model design} presents the complete model design and training procedures. Section \ref{sec 4.6: experiments} reports comprehensive experimental results, comparing ID and OoD performance across all models. Our results demonstrate the impact of different augmentation strategies and highlight the improved OoD robustness achieved through polynomial representations. Our analysis also uncovers counterintuitive results regarding cross-dataset generalization, leading to a discussion of underlying factors that may influence model robustness beyond architectural design choices. Section \ref{sec 4.7: conclusion} concludes with a discussion of the results and their implications for robust traffic scene prediction systems.

\section{Related Work and Benchmark Models}
\label{sec 4.2: related work}
\subsection{Data Representation}
\label{sec: model data representation}

The formats of individual datasets and competition rules greatly influence the design decisions of the models proposed in the literature. Deep learning-based models typically process data in its original form, often using \emph{sequence-based} representations -- i.e., sequences of data points -- as both model input and output \cite{cheng_forecast_2023, zhou_query_2023, liang_learning_2020, varadarajan_multipath++_2022, nayakanti_wayformer_2022}. This representation aligns the format of measurements in datasets and effectively captures various information, e.g., agent trajectories and road geometries. However, it also comes with notable drawbacks: it introduces high redundancy and variance, and it is sensitive to measurement noise and outliers, which can result in physically implausible predictions. Moreover, the computational complexity of this approach increases with the length and temporal or spatial resolution of the sequences.

Some previous studies explored the possibility of using \emph{polynomial representations} for predictions, achieving competitive ID results with relatively low-degree polynomials as outputs \cite{buhet_plop_2020, su_temporally_2021}. However, these studies primarily focus on \emph{output} representations and ID performance evaluation. In contrast, representing \emph{inputs} with polynomials and examining their impact on OoD generalization remain relatively underexplored in the current literature. Motivated by the theoretical and practical advantages of polynomial representations, this chapter investigates both aspects -- input representation and generalization performance -- in greater depth.

\subsection{Data Augmentation}
\label{sec: model data augmentation}
Competitions typically designate one or more agents in a scene as \emph{target} agents and only score predictions for these target agents. As a result, it is common practice to train models using data from target agents. However, training the model exclusively with the target agent's behavior fails to exploit all available data. To address this, predicting the future motion of \emph{non-target} agents is a typical strategy for augmenting training data. As there are many more non-target agents than target agents, another important design decision relates to how authors interpret the importance of target vs.\ non-target agent data.

\begin{figure}[!thb]
\centering
\includegraphics[width=1.0\textwidth]{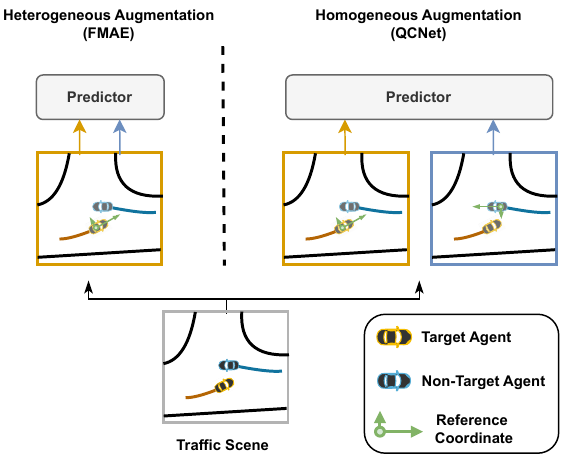}
\caption[Augmentation strategies]{The two augmentation strategies for non-target agent data employed in benchmark models. \textbf{Left}: FMAE employs heterogeneous augmentation, representing information in the target agent's coordinate frame and only making uni-modal predictions for non-target agents. \textbf{Right}: QCNet employs homogeneous augmentation, encoding information in each agent's individual coordinate frame and making multi-modal predictions for both target and non-target agents alike.}
\label{fig:augmentation_strategy}
\end{figure}

To investigate these design choices, we select two open-sourced and thoroughly documented SotA models on the A2 dataset: Forecast-MAE \cite{cheng_forecast_2023} (FMAE) and QCNet \cite{zhou_query_2023}, with approximately \SI{1.9}{M} and \SI{7.6}{M} parameters, respectively. As summarized in Table \ref{tab: model difference in non-target}, both models employ sequence-based representation but exhibit different augmentation strategies in dealing with non-target agents:
\begin{itemize}[leftmargin=*]
    \item[] \textbf{Heterogeneous Augmentation}: target and non-target agents are treated differently. FMAE follows the prediction competition protocol and prioritizes target agent prediction. Thus, agent history and map information are computed within the \emph{target agent's} coordinate frame, as shown in the left panel of Figure \ref{fig:augmentation_strategy}. Compared to the \emph{multi-modal} prediction of the target agent, FMAE considers the prediction for non-target agents as an auxiliary task and only outputs \emph{uni-modal} prediction for them. The displacement error of the target agent also weighs higher in the loss function than non-target agents.
    \item[] \textbf{Homogeneous Augmentation}: all agents are treated identically. QCNet does not focus on the selected target agent and proposes a more generalized approach. It encodes the information of agents and map elements in each agent's \emph{individual} coordinate frame, as shown in the right panel of Figure \ref{fig:augmentation_strategy}. It outputs \emph{multi-modal} predictions for target and non-target agents alike, ensuring a consistent prediction task for all agents. The loss of target agent prediction shares the same weight as non-target agents.
\end{itemize}

\begin{table}[!tbh]
\caption[Model variants under study]{Model variants under study based on augmentation strategies and data representations.}
\vspace{-0.0em}
\centering
\begin{tabularx}{0.8\textwidth}{l >{\centering\arraybackslash}X |>{\centering\arraybackslash}X }
\Xhline{3\arrayrulewidth}
augmentation & \multicolumn{2}{!{\vrule width 1pt}c}{input and output representation} \\
\cline{2-3}
strategy&\multicolumn{1}{!{\vrule width 1pt}c|}{sequence-based} &  polynomial-based (ours) \\
\Xhline{3\arrayrulewidth}
heterogeneous & \multicolumn{1}{!{\vrule width 1pt}c|}{\multirow{1}{*}{FMAE \cite{cheng_forecast_2023}}}&\multirow{1}{*}{EP-F}\\
\cline{1-3}
homogeneous & \multicolumn{1}{!{\vrule width 1pt}c|}{\multirow{1}{*}{QCNet \cite{zhou_query_2023}}} & \multirow{1}{*}{EP-Q} \\
\cline{1-3}
\multirow{2}{*}{w/o augmentation} & \multicolumn{1}{!{\vrule width 1pt}c|}{FMAE-noAug} & \multirow{2}{*}{EP-noAug} \\
&\multicolumn{1}{!{\vrule width 1pt}c|}{QCNet-noAug}& \\
\Xhline{3\arrayrulewidth}
\end{tabularx}
\label{tab: model difference in non-target}
\end{table}

To study the influence of these augmentation strategies independently of data representation, we add a third augmentation strategy for comparison: \textbf{no augmentation}. This can be achieved by removing the loss of non-target agents in each model, thus limiting the model to learning from target agent behavior only. We denote the benchmark models without augmentation as ``FMAE-noAug'' and ``QCNet-noAug''. For our own model proposed in Section \ref{sec 4.5: model design}, we will implement all three augmentation strategies for comparison.

\section{Data Homogenization}
\label{sec 4.3: data homogenization}
Motion datasets of different origins utilize distinct data collection processes and sensor platforms. This diversity creates an opportunity to perform OoD testing on truly independent data samples. However, this also requires overcoming substantial challenges in aligning inconsistent data formats and prediction task specifications between datasets.

In our case, there are multiple notable distinctions between A2 and WO, as listed in Table \ref{tab: dataset difference}. First, scene durations differ significantly: a scene in A2 is $11$-second long and consists of  $5$-second observation history and $6$-second future to predict, while WO only provides $1.1$-second observation history but requires an $8$-second prediction horizon. 

Second, map representations vary considerably between datasets. A2 provides lane boundaries for junction lanes and includes annotations for them, which are absent in WO. Additionally, A2 describes straight lane segments using only the start and end points, as visualized in Figure \ref{fig: A2 fitted}, whereas WO provides interpolated points along the entire segment.

Third, the datasets also differ in their target agent selection strategies. For marginal prediction, A2 selects a single target agent per scene, excluding the ego vehicle. Additionally, A2 ensures that the target agent is fully observed at all timesteps throughout the prediction period. In contrast, WO takes a broader approach, evaluating up to 8 target agents per scene -- potentially including the ego vehicle -- without guaranteeing complete observability during the prediction horizon.

\begin{table}[!th]
\caption[Datasets comparison and homogenization protocol]{Datasets comparison and homogenization protocol for out-of-distribution testing.}
\centering
\begin{tabularx}{\textwidth}{l l >{\centering\arraybackslash}X >{\centering\arraybackslash}X >{\centering\arraybackslash}X}
\Xhline{3\arrayrulewidth}
\multicolumn{2}{c}{\multirow{2}{*}{} } & \multirow{2}{*}{A2 ~\cite{wilson_argoverse2_2021}} & \multirow{2}{*}{WO~\cite{ettinger_waymo_2021}} & homogenized dataset \\ 
\Xhline{3\arrayrulewidth}
\multicolumn{2}{l}{scenario length}& 11s & 9.1s & - \\
\hline
\multicolumn{2}{l}{sampling rate}& 10 Hz & 10 Hz & 10 Hz \\
\hline
\multicolumn{2}{l}{history length}& 5s & 1.1s & 5s \\
\hline
\multirow{2}{*}{prediction horizon} & train (A2 | WO)& \multirow{2}{*}{6s}  & \multirow{2}{*}{8s} & 6s | 4.1s \\
&test&&& 4.1s  \\
\hline
\multirow{5}{*}{\thead{map\\information}}  & lane centerline & yes & \multirow{1}{*}{yes} & \multirow{1}{*}{yes} \\
\cline{2-5}
& \multirow{3}{*}{lane boundary} & \multirow{3}{*}{yes} & not for junction lanes& \multirow{3}{*}{no}\\
\cline{2-5}
& junction lanes labeled& yes& no& no \\
\hline
\multirow{4}{*}{\thead{prediction\\target}}  & \# target agents & 1 & up to 8 & 1\\
\cline{2-5}
& ego included & no & yes & no\\
\cline{2-5}
& \multirow{2}{*}{fully observed}& \multirow{2}{*}{yes} & not guaranteed & \multirow{2}{*}{yes} \\
\Xhline{3\arrayrulewidth}
\end{tabularx}
\label{tab: dataset difference}
\end{table}

To address these challenges and enable systematic cross-dataset evaluation, we develop a homogenization protocol that standardizes data formats and prediction tasks across datasets, as detailed in Table \ref{tab: dataset difference}. Since both benchmark models were originally designed for A2, our homogenization protocol is designed to align the specifications of WO with A2's configuration: 
\begin{itemize}[leftmargin=*]
    \item \textbf{History Length:} We standardize the 5-second (50 steps) history for both datasets. This extends WO’s original 1.1-second history to match A2's specification.
    \item \textbf{Prediction Horizon:}
    After reserving 5-second history, A2 has 6-second future data available while WO has only 4.1-second future remaining. During training, we maintain these dataset-specific horizons: 6-second for A2-trained models and 4.1-second for WO-trained models. For evaluation, we standardize to a 4.1-second horizon to ensure fair comparison across all models.
    \item \textbf{Map Information:} We retain only map elements available in both datasets: lane centerlines and crosswalks. Lane boundary information and junction lane labels are excluded due to their absence in WO.
    \item \textbf{Target Agent Selection:} We adopt A2's marginal prediction approach, using the designated single target agent. From WO, only the first fully observed, non-ego agent in the list of target agents is chosen. As the list of target agents is unordered, this corresponds to randomly sampling a single fully observed target agent without introducing additional bias. Traffic scenes without any fully observed, non-ego target agents are considered as invalid samples and excluded.
\end{itemize}

We apply this homogenization protocol to both the train and validation sets of each dataset. As the agents' future trajectories in the test sets of the A2 and WO competitions are not accessible, they cannot be included in our homogenization protocol. 

This homogenization protocol yields \num{199908} training samples from A2 and \num{472235} from WO, with \num{24988} and \num{42465} validation samples, respectively.


\section{Representing Data via Polynomials}
\label{sec 4.4: represent data with polynomials}
As outlined in Section \ref{sec: model data representation}, our primary motivation is to explore the impact of using polynomial representations in prediction models. To achieve this, we must first transform the sequence-based data format provided in motion datasets into polynomial representations suitable for our model (cf. Section \ref{sec 4.5: model design}).

In the following sections, we present our approach for representing both the input data (agent history and map geometry) and the output data (predicted trajectories) as polynomials.

\subsection{Agent History as Polynomial}
\label{sec 4.4.1: Agent History as Polynomial}
To transform sequence-based data into polynomial representations, we must first determine the polynomial degree. Our results in Section \ref{sec: optimal polynomial degree} provide strong guidance for selecting the appropriate polynomial degree. Based on the Akaike Information Criterion (AIC) \cite{akaike_AIC_1973}, the 5-second history trajectories of vehicles, cyclists, and pedestrians in both A2 and WO are represented optimally with 5-degree Bernstein polynomials. Although AIC recommends a 6-degree polynomial for ego vehicle trajectories, we adopt a 5-degree polynomial for the ego vehicle as well, to maintain consistency across all agent types.

To transform agent history from sequences to Bernstein polynomials, one approach is to apply Bayesian regression (cf. Equation \ref{eqn:posterior_mean} in Section \ref{sec 3.3: Bayesian regression}). However, applying Bayesian regression to sequence data is commonly associated with the \emph{boundary effect}\footnote{Bayesian regression often exhibits increased uncertainty and reduced stability at the boundaries of the input domain. This is due to weaker constraints from surrounding data and can lead to unstable estimates near the sequence edges, particularly at the start and end.}, even with incorporating prior distributions \cite[p. 158]{murphy_machine_2012}. As a result, kinematic estimates at the reference timestep $t_0$ can become unstable, potentially degrading the model’s prediction performance.

To address this, we employ a modified Kalman filter proposed in \cite{reichardt_trajectories_2022}. This approach follows the same predict-update process as the Kalman filter, described in Section \ref{sec 3.5.2: tracking issues}. However, it deviates from the typical Kalman filter, which tracks the state at a single timestep, by instead tracking the control points of the complete trajectory. This fundamentally changes the transition and observation models of the Kalman filter. We now present our implementation details.

\subsubsection{State Transition Model}
The state vector of agent $i$ comprises all control points defining its trajectory and is expressed as:
\begin{equation}
    \mathbf{x}_{i,t} = \boldsymbol{\omega}_{i,t}=
[\mathbf{w}_0; \dots; \mathbf{w}_5]_{i,t} \in \mathbb{R}^{12}
\end{equation}
where each $\mathbf{w}_n$ denotes a control point at timestep $t$. This state vector represents the agent's trajectory over the preceding 5 seconds up to the timestep $t$. For clarity, we omit the agent index $i$ in the following discussion of the Kalman filter implementation.

The state transition model follows Equation \ref{eqn: state transition model}. To build the state transition matrix $\mathbf{F}_t$, we define two temporally equally spaced input variable vectors: 
\begin{equation}
\tau_{t,n} = \frac{n\left(1 - \frac{\delta_t}{\delta_{T}}\right)}{N}, \quad
\tau_{t+1,n} = \tau_{t,n} + \frac{\delta_t}{\delta_{T}}, \quad \text{for } n = 0, \dots, N.
\end{equation}
where $N=5$ is the selected polynomial degree and $\delta_{T}=5s$ is the 5-second time length of agent history. $\delta_t$ represents the time difference between successive timesteps $t$ and $t+1$. 

\begin{figure}[!thb]
\centering
\includegraphics[width=1.0\textwidth]{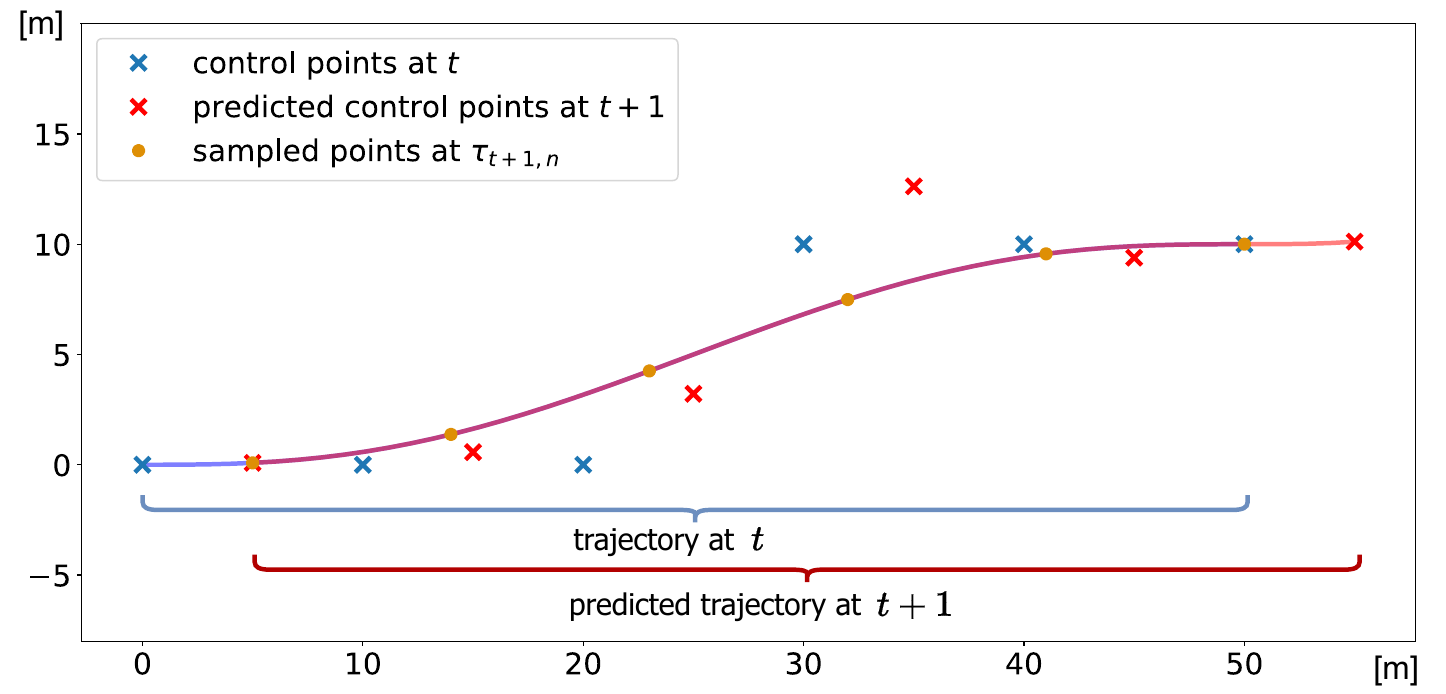}
\caption[Transition model]{Example of predicted control points based on the transition model in Equation \ref{eqn: ep transition model}. Curves show modeled trajectories at consecutive timesteps.}
\label{fig: transition model}
\end{figure}

To construct the state transition matrix, trajectory points are sampled at time $\tau_{t+1,n}$, as shown in Figure \ref{fig: transition model}. Following the formulation in \cite{reichardt_trajectories_2022}, we then define two $(N+1) \times (N+1)$ matrices:
\begin{equation}
    \mathbf{B}_t = [\boldsymbol{\phi}(\tau_{t,0}), \dots, \boldsymbol{\phi}(\tau_{t,N})], \quad
     \mathbf{B}_{t+1} = [\boldsymbol{\phi}(\tau_{t+1,0}), \dots, \boldsymbol{\phi}(\tau_{t+1,N})], \\
\end{equation}
where $\boldsymbol{\phi}(\cdot)$ is the column vector of Bernstein bases defined in Section \ref{sec: 3.2 parametric representation}. The state transition matrix $\mathbf{F}_t$ can then be formulated as:
\begin{equation}
\label{eqn: ep transition model}
    \mathbf{F}_t = (\mathbf{B}_t^{\top}\mathbf{B}_t + \mathbf{\Sigma}^{-1}_\text{P})^{-1} \mathbf{B}_t^{\top}\mathbf{B}_{t+1} \otimes \mathbf{I}_{2}
\end{equation}
where $\mathbf{\Sigma}_\text{P}$ defines the prior covariance of agent motion between consecutive timesteps. Since we did not analyze the agent motion between consecutive timesteps in Chapter \ref{cpt: Empirical Bayes Analysis}, we set $\mathbf{\Sigma}_\text{P}$ to an uninformative prior as a diagonal matrix, with high variance entries of $2 \times 10^5$.

For the process noise covariance matrix $\mathbf{Q}_t$, we empirically define it as a diagonal matrix as: $\mathbf{Q}_t=\mathrm{diag}(0,\ 0,\ 0,\ 0.3,\ 0.2,\ 0.1) \otimes \mathbf{I}_2$. The first three entries are set to zero based on the assumption that the new trajectory at timestep $t+1$ will move along the previous trajectory at timestep $t$ in the overlapping region, as shown in Figure \ref{fig: transition model}. The non-zero entries (0.3, 0.2, 0.1) correspond to the later control points beyond the overlap region, where the agent's future motion is less predictable. This descending order reflects the interpolation property of Bernstein polynomials: the final control point coincides with the last trajectory position and thus has direct influence, whereas interior control points only approximate the curve. A given variance on an interior control point therefore produces less trajectory-level change than the same variance on the endpoint, so the chosen values yield uncertainty that grows along the trajectory despite decreasing in magnitude.

\subsubsection{Observation Model}
At each timestep $t$, we use the observation at this timestep to update the state vector. The observation model follows Equation \ref{eqn: observation model}. As both A2 and WO have measurements of agent position and velocity, the observation vector is modeled as: 
\begin{equation}
    \mathbf{z}_{t} = 
\begin{bmatrix}
p_x \\ p_y \\ v_x \\ v_y 
\end{bmatrix}_{t}
\end{equation}
where $(p_x, p_y)$ denotes position and $(v_x, v_y)$ denotes velocity. The observation matrix is constructed as:
\begin{equation}
    \mathbf{H} = [\boldsymbol{\phi}^{\top}(1); \boldsymbol{\dot\phi}^{\top}(1)] \otimes \mathbf{I}_2
\end{equation}
where $\boldsymbol{\phi}(1)$ and $\boldsymbol{\dot\phi}(1)$ are the Bernstein bases and their derivatives (cf. Equation \ref{eqn: basis function derivative}), respectively. The input variable $\tau$ is set to 1, corresponding to the latest observation at timestep $t$.

Since we extensively analyzed the observation noise characteristics in Chapter \ref{cpt: Empirical Bayes Analysis} (cf. Sections \ref{sec 3.4.2: observation noise modeling} and \ref{sec: 3.6.3: observation noise estimation}), we set the observation noise covariance $\mathbf{R}_t = \mathbf{\Sigma}_{o,i,t}$ using the values estimated in that chapter.

\subsubsection{Prediction and Update Steps}
The prediction and update steps follow the standard Kalman filter procedure, as described in Equation \ref{eq:kalman_prediction} and Equation \ref{eq:kalman_update}. One remaining challenge is estimating the initial state of the control points, denoted as $\hat{\mathbf{x}}_1$, and corresponding covariance matrix $\mathbf{P}_1$ at the moment the agent is first observed. 

Given that both the A2 and WO datasets provide position and velocity measurements for each observed agent, we assume that the agent has been moving at a constant velocity over the 5 seconds preceding its first observation. Based on this assumption, we reconstruct a straight-line trajectory as the agent's motion history and use it to initialize the control points $\hat{\mathbf{x}}_1$.


However, this initialization is inherently uncertain due to two factors: first, the simplifying constant velocity assumption, and second, potential inaccuracies or outliers in the dataset, as discussed in Section \ref{sec 3.5: outlier filtering}. To account for this uncertainty, we initialize the covariance matrix $\mathbf{P}_1$ using the prior covariance $\mathbf{\Sigma}_{\boldsymbol{\omega}}$ estimated for each agent type and dataset in Chapter \ref{cpt: Empirical Bayes Analysis}. This provides a principled, dataset-specific prior over control point distributions without making overconfident assumptions about their initial values.

The Kalman filter is then applied iteratively to each agent in the traffic scene over the history timesteps to estimate the control point states.

Since most modern tracking systems in autonomous vehicles already employ Kalman 
filtering or similar state estimation techniques \cite{chang_argoverse_2019, caesar_nuscenes_2020, ettinger_waymo_2021, chen_interactive_2022}, the polynomial parameterization of agent history described above can be integrated directly into these systems with minimal computational 
overhead. This integration ensures that the polynomial representations are available 
in real-time without introducing additional latency in the prediction pipeline.


\subsection{Map as Polynomial}
Map elements, such as the centerlines of lane segments
and crosswalks, are represented with 3-degree Bernstein polynomials to maintain consistency with OpenDRIVE standards \cite{opendrive}. Given that map elements in motion datasets are typically constructed with high precision \cite{wilson_argoverse2_2021, ettinger_waymo_2021} and exhibit minimal observation noise compared to dynamic tracking systems, linear regression would seem to provide a straightforward approach for fitting these polynomial representations.

 However, a key challenge arises in parameterizing the input variable $\boldsymbol{\tau}$ for map elements, which differs fundamentally from trajectory modeling. While trajectories benefit from explicit temporal sampling that enables straightforward normalization of the input variable $\boldsymbol{\tau}$, the sample points of map elements lack temporal information entirely. The sampling points along map elements are spatial rather than temporal, making direct estimation of $\tau$ non-trivial.

\begin{figure}[thb]
\centering
\includegraphics[width=0.8\textwidth]{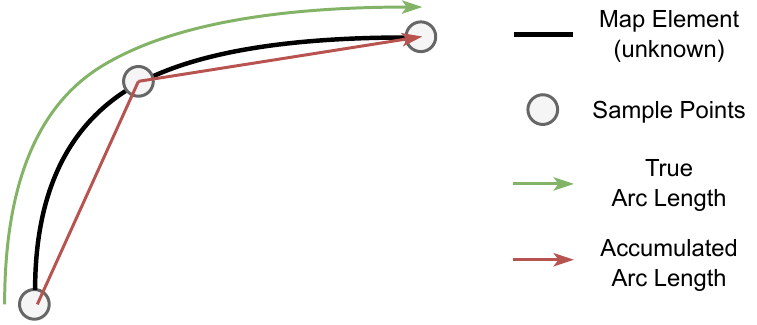}
\caption[Arc length error]{Accumulated arc length introduces errors for curved map elements.}
\label{fig: arc length error}
\end{figure}

A natural approach is to define $\boldsymbol{\tau}$ based on the arc length. For a map element with $L$ sample points, we denote $l'$ and $L'$ as the continuous-distance equivalents (in meters) at the $l$-th discrete sample point and the last sample point, respectively. This distinction parallels the temporal case, where discrete sampling points may not be uniformly distributed along the actual geometric curve. The normalized input variable is then: 
\begin{equation}
    \tau_l = \frac{l'}{L'}, \quad \text{so that} \quad \boldsymbol{\tau} = [\tau_1, \tau_2, \dots, \tau_L].
\end{equation}
In practice, the true arc length is typically unknown and is approximated by the accumulated Euclidean distance (chord-length) between consecutive sample points. While this approximation works well for nearly straight segments, it can deviate significantly from the actual arc length for curved map elements, as illustrated in Figure \ref{fig: arc length error}. This deviation leads to parameterization errors and significantly degrades the quality of the polynomial fit, suggesting that treating $\boldsymbol{\tau}$ as error-free (as in standard linear regression) may not be appropriate for curved map elements.

To address this, we fit the sample points of map elements using the total-least-squares (TLS) method by Borges and Pastva \cite{borges_2002_total}. Unlike standard least squares, TLS does not assume a fixed parameterization $\boldsymbol{\tau}$ for the curve. Instead, it iteratively optimizes the best-fitting $\boldsymbol{\tau}$ values for each sample point, resulting in a more accurate geometric fit. Next, we detail the TLS method.

Following \cite{borges_2002_total}, we separate the control point vector  $\boldsymbol{\omega}$ into two column vectors: $\boldsymbol{\omega}_{x}$ and   $\boldsymbol{\omega}_{y}$, corresponding to the x- and y-components of the control points, respectively. Given a set of sample points of a map element $\mathcal{P} = \{\mathbf{p}_l=(p_{x,l}, p_{y,l})\}_{l=1}^L$, the $x$ and $y$ positions are defined as: 
\begin{equation}
    \mathbf{p}_x=[p_{x,1}, \dots, p_{x,L}]^{\top}, \quad \mathbf{p}_y=[p_{y,1}, \dots, p_{y,L}]^{\top}.
\end{equation}
Same as in Section \ref{sec: 3.2 parametric representation}, let $\boldsymbol{\Phi}_{\text{B}} = [\boldsymbol{\phi}(\tau_{1}), \dots,\boldsymbol{\phi}(\tau_{L})] \in \mathbb{R}^{(N+1)\times L}$ be the Bernstein bases at each $\tau_l$. Then the residual (fit error) is:
\begin{equation}
\label{eqn: borges 1}
    \epsilon(\boldsymbol{\tau}) =\left\| 
\begin{bmatrix}
\mathbf{p}_x \\
\mathbf{p}_y
\end{bmatrix}
-
\begin{bmatrix}
\boldsymbol{\Phi}_{\text{B}}^{\top} & \\
 & \boldsymbol{\Phi}_{\text{B}}^{\top} 
\end{bmatrix}
\begin{bmatrix}
\boldsymbol{\omega}_x \\
\boldsymbol{\omega}_y
\end{bmatrix}
\right\|_2^2
\end{equation}
For a fixed set of $\boldsymbol{\tau}$-values, the estimates of $\boldsymbol{\omega}$ can be computed by solving the linear least squares problem:
\begin{equation}
\label{eqn: borges 2}
    \begin{bmatrix}
\boldsymbol{\omega}_x \\
\boldsymbol{\omega}_y
\end{bmatrix} = \begin{bmatrix}
\boldsymbol{\Phi}_{\text{B}}^{+\top} & \\
 & \boldsymbol{\Phi}_{\text{B}}^{+\top} 
\end{bmatrix}
\begin{bmatrix}
\mathbf{p}_x \\
\mathbf{p}_y
\end{bmatrix},
\end{equation}
where $\boldsymbol{\Phi}_{\text{B}}^{+\top}$ denotes the pseudo-inverse of $\boldsymbol{\Phi}_{\text{B}}^{\top}$. Substituting Equation \ref{eqn: borges 2} into Equation \ref{eqn: borges 1} yields a residual that depends only on $\boldsymbol{\tau}$:
\begin{equation}
\label{eqn: borges 3}
\epsilon(\boldsymbol{\tau})=  \left\| 
\begin{bmatrix}
\mathbf{p}_x \\
\mathbf{p}_y
\end{bmatrix}
-
\begin{bmatrix}
\boldsymbol{\Phi}_{\text{B}}^{\top}\boldsymbol{\Phi}_{\text{B}}^{+\top} & \\
 & \boldsymbol{\Phi}_{\text{B}}^{\top}\boldsymbol{\Phi}_{\text{B}}^{+\top}
\end{bmatrix}
\begin{bmatrix}
\mathbf{p}_x \\
\mathbf{p}_y
\end{bmatrix}
\right\|_2^2
\end{equation}
Let $\mathbf{P}_{\boldsymbol{\Phi}_{\text{B}}} = \boldsymbol{\Phi}_{\text{B}}^{\top}\boldsymbol{\Phi}_{\text{B}}^{+\top}$ and $\mathbf{P}^{\perp}_{\boldsymbol{\Phi}_{\text{B}}}=\mathbf{I} - \mathbf{P}_{\boldsymbol{\Phi}_{\text{B}}}$, The residual $\epsilon(\boldsymbol{\tau})$ can be expressed as:
\begin{equation}
   \epsilon(\boldsymbol{\tau}) = \boldsymbol{\epsilon}^{\top}(\boldsymbol{\tau})\boldsymbol{\epsilon}(\boldsymbol{\tau}), \quad \text{where} \quad \boldsymbol{\epsilon}(\boldsymbol{\tau}) = \begin{bmatrix}
\mathbf{P}^{\perp}_{\boldsymbol{\Phi}_{\text{B}}}\mathbf{p}_x \\
\mathbf{P}^{\perp}_{\boldsymbol{\Phi}_{\text{B}}}\mathbf{p}_y
\end{bmatrix}
\end{equation}
When performing QR decomposition with column pivoting \cite{golub_matrix_1996} on $\boldsymbol{\Phi}^{\top}_{\text{B}}$ :
\begin{equation}
    \boldsymbol{\Phi}_{\text{B}}^{\top} \boldsymbol{\Pi} = 
    \begin{bmatrix}
    \hat{\mathbf{Q}}_{1} & \hat{\mathbf{Q}}_{2}
    \end{bmatrix}
    \begin{bmatrix}
    \hat{\mathbf{R}}_{1,1} & \hat{\mathbf{R}}_{1,2} \\
    \mathbf{0} & \mathbf{0}
    \end{bmatrix}
\end{equation}
where $\boldsymbol{\Pi}$ is a permutation matrix. $\begin{bmatrix} \hat{\mathbf{Q}}_{1} & \hat{\mathbf{Q}}_{2} \end{bmatrix}$ is an $L \times L$ orthogonal matrix. $\hat{\mathbf{R}}_{1,1}$ is upper-triangular matrix and $\hat{\mathbf{R}}_{1,2}$ vanishes when $\boldsymbol{\Phi}_{\text{B}}$ is of full rank. 

As shown in \cite{borges_2002_total}, the projection matrices can be computed efficiently as:
\begin{equation}
\begin{aligned}
    \mathbf{P}_{\boldsymbol{\Phi}_{\text{B}}} &= \hat{\mathbf{Q}}_{1} \hat{\mathbf{Q}}^{\top}_{1}, \\
    \mathbf{P}^{\perp}_{\boldsymbol{\Phi}_{\text{B}}} &= \hat{\mathbf{Q}}_{^2}\hat{\mathbf{Q}}^{\top}_{2}
\end{aligned}
\end{equation}
This allows the residual to be simplified to:
\begin{equation}
 \epsilon(\boldsymbol{\tau}) = \left\| \hat{\mathbf{Q}}_{2}^{\top} \mathbf{p}_x \right\|_2^2 + \left\| \hat{\mathbf{Q}}_{2}^{\top} \mathbf{p}_y \right\|_2^2.
\end{equation}
To iteratively minimize this residual, we apply the Gauss–Newton update:
\begin{equation}
\label{eqn: borges update}
\boldsymbol{\tau}^{(r+1)} = \boldsymbol{\tau}^{(r)} - \alpha^{(r)} \boldsymbol{J}^{+}(\boldsymbol{\tau}^{(r)}) \, \boldsymbol{\epsilon}(\boldsymbol{\tau}^{(r)})
\end{equation}
where $\alpha^{(r)}$ is the step size and $\boldsymbol{J}^{+}(\boldsymbol{\tau}^{(r)})$ is the pseudo-inverse of the Jacobian matrix $\boldsymbol{J}$ at optimizing iteration $r$. The step size $\alpha^{(r)}$ is adaptively selected according to \cite{mathworks_borgespastva}. Borges et al. propose to compute the Jacobian matrix as:
\begin{equation}
\boldsymbol{J}=
\begin{bmatrix}
\hat{\mathbf{Q}}_{2} \hat{\mathbf{Q}}_{2}^{\top} \, \mathrm{diag}(\mathbf{P} \mathbf{p}_x) + \mathbf{P}^{\top} \, \mathrm{diag}(\hat{\mathbf{Q}}_{2} \hat{\mathbf{Q}}_{2}^{\top} \mathbf{p}_x) \\
\hat{\mathbf{Q}}_{2} \hat{\mathbf{Q}}_{2}^{\top} \, \mathrm{diag}(\mathbf{P} \mathbf{p}_y) + \mathbf{P}^{\top} \, \mathrm{diag}(\hat{\mathbf{Q}}_{2} \hat{\mathbf{Q}}_{2}^{\top} \mathbf{p}_y)
\end{bmatrix},
\end{equation}
where 
\begin{equation}
\mathbf{P} = \mathbf{\dot\Phi}^{\top}_{\text{B}} \boldsymbol{\Pi} 
\begin{bmatrix}
\hat{\mathbf{R}}_{1,1}^{-1} \hat{\mathbf{Q}}_1^{\top} \\
\mathbf{0}
\end{bmatrix}.
\end{equation}
Here $\mathbf{\dot\Phi}_\text{B}$ is the derivative of the Bernstein basis with respect to $\tau$ and can be computed based on Equation \ref{eqn: basis function derivative}. 

We iterate Equation~\ref{eqn: borges update} up to three iterations or until the residual $\epsilon(\boldsymbol{\tau})$ falls below $10^{-5}$ meters. After estimating $\boldsymbol{\tau}$ using TLS, the final control points are computed via standard least squares using Equation \ref{eqn: borges 2}.

\begin{figure}[!thb]
\centering
\includegraphics[width=0.9\textwidth]{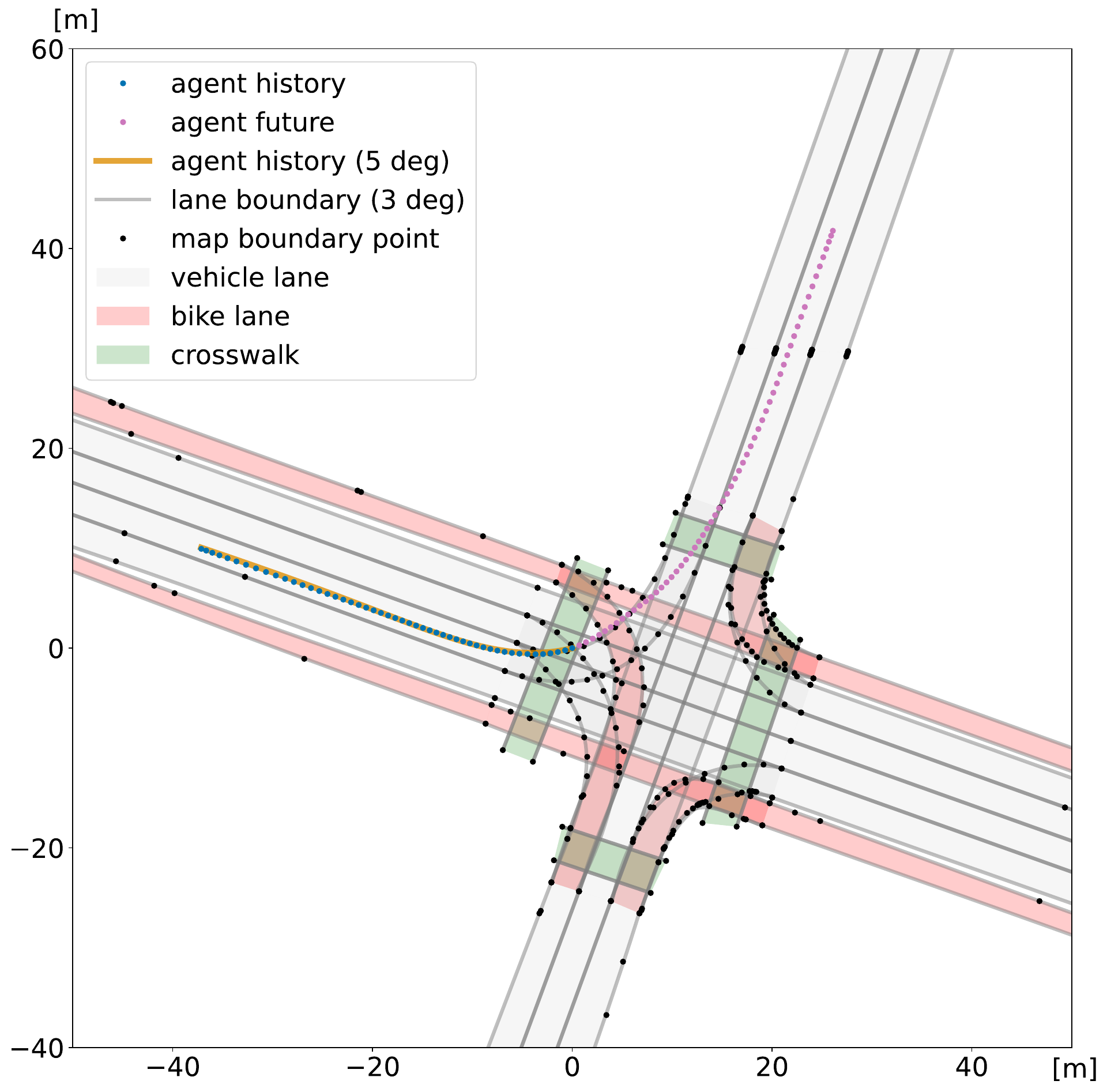}
\caption[Traffic scene with polynomial representation]{One traffic scene in A2 represented with polynomials. Only boundaries of map elements are shown for visual clarity, while actual map inputs are detailed in Section \ref{sec 4.5.1: model inputs}. Points represent sequence-based data provided in A2.}
\label{fig: A2 fitted}
\end{figure}

The WO dataset contains longer and more curved lane segments than those in A2, as visualized in Figure \ref{fig: data_distribution_shift}, posing a challenge for fitting them with 3-degree polynomials. To ensure fit quality, we iteratively split the lane segments in half and apply TLS fitting until the average fit error for each segment falls below \SI{0.1}{\meter}.

Figure \ref{fig: A2 fitted} visualizes a traffic scene with both sequence and polynomial representations from the A2 dataset. The polynomial representation requires only $40.8 \%$ and $8.7 \%$ of data space compared to the sequence representations provided in A2 and WO, respectively. This difference arises because A2 uses fewer sample points to represent straight map elements, whereas WO retains a higher sampling density, even for relatively straight segments.

The TLS fitting process for map elements can be executed offline during map preprocessing, avoiding any runtime computational cost. Moreover, if high-definition maps are already available in OpenDRIVE format -- which natively represents lane geometry using polynomial parameterizations -- the fitting step can be bypassed entirely. Therefore, the map preprocessing does not significantly increase computational overhead in practical deployment scenarios.

\subsection{Prediction as Polynomial}
The 6-second predicted trajectories are formulated as 6-degree polynomials -- one degree higher than recommended by AIC in Chapter \ref{cpt: Empirical Bayes Analysis} (cf. Section \ref{sec: optimal polynomial degree}) -- to capture more complex motion patterns. As these trajectories constitute our model output, we detail this approach in Section \ref{sec 4.5.4: model outputs}.

\section{Model Design}
\label{sec 4.5: model design}
\begin{figure}[thb]
\centering
\includegraphics[width=\textwidth]{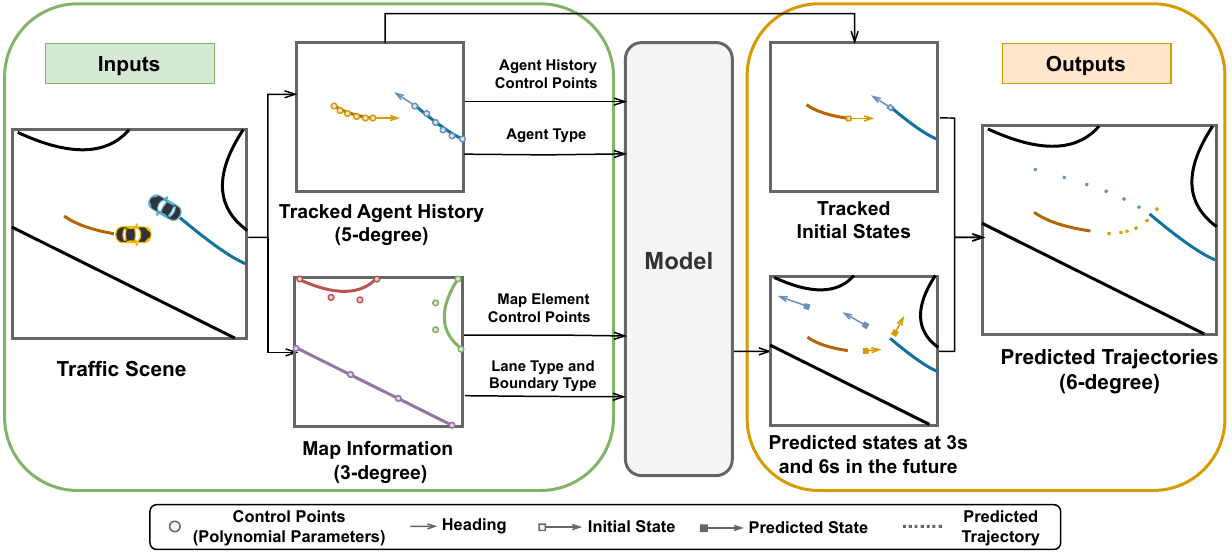}
\caption[EP pipeline]{Our proposed model pipeline. \textbf{Inputs}: Agent histories and road geometry are both represented via polynomials. \textbf{Outputs}: The tracked initial states and predicted states are fused into one polynomial trajectory prediction, ensuring continuity of past observation and future prediction.}
\label{fig:pipeline}
\end{figure}

Figure \ref{fig:pipeline} shows the pipeline of our model. Given that polynomial inputs and outputs constitute our key innovation, we refer to our contribution as \emph{Everything Polynomial} (\emph{EP}).

Like many SotA models and consistent with our benchmark models, EP employs the popular encoder-decoder architecture. The encoder encodes context information, including agent history and map data, while modeling agent interactions. The decoder generates outputs based on the encoded representations from the encoder.

To enable fair comparison with benchmark models, we implemented three EP variants following different data augmentation strategies: (\lowerromannumeral 1) EP-F employs the heterogeneous augmentation of FMAE.
(\lowerromannumeral 2) EP-Q employs the homogeneous augmentation of QCNet.
(\lowerromannumeral 3) EP-noAug has no augmentation. Table \ref{tab: model difference in non-target} summarizes the design choices across all models under study.

For multi-modal prediction, EP variants follow the benchmark model design in terms of prediction mode count: EP-F and EP-noAug predict 6 modes for target agents, while EP-Q outputs 6 modes for all agents.


In the following sections, we begin by introducing the inputs to EP models, followed by the detailed architectures of the encoder and decoder. Finally, we present the training settings.

\subsection{Inputs}
\label{sec 4.5.1: model inputs}
EP-F and EP-Q employ different augmentation strategies, representing data in different coordinate frames, as shown in Figure \ref{fig:augmentation_strategy}. EP-noAug follows the same coordinate representation as EP-F since it only considers target agent data. To clearly explain the input structure, we begin with EP-F, which encodes all data in a single target agent coordinate frame, then describe how EP-Q extends this representation to individual agent coordinates.


\subsubsection{Agent Inputs}
The history of all $A$ agents in the traffic scene is represented using 5-degree polynomials. The control points of agents' history are denoted as $\textbf{W}^{\text{a,hist}}_{n} \in \mathbb{R}^{A \times 2}$ with $n =0,1,\dots,5$, where the notation here represents the target agent coordinate frame case. Inspired by FMAE and QCNet, which compute vectors between consecutive sequence points, we represent the kinematics of agents by computing vectors between consecutive control points as $\boldsymbol{\Delta}^{\text{a,hist}}_{n} = \textbf{W}^{\text{a,hist}}_{n} - \textbf{W}^{\text{a,hist}}_{n-1}\in \mathbb{R}^{A \times 2}$.

While these vectors can encode motion patterns, they are not sufficient to describe agents' spatial relationships. To address this, we employ the last control point $\textbf{W}^{\text{a,hist}}_{5}$ as the reference position and the normalized vector $\boldsymbol{\Gamma}^{\text{a,hist}}_{5} = \boldsymbol{\Delta}^{\text{a,hist}}_{5} / \lVert\boldsymbol{\Delta}^{\text{a,hist}}_{5}\rVert_2 \in \mathbb{R}^{A \times 2}$ as the reference heading. 

Another point to consider with the polynomial representation is that it does not inherently indicate valid timesteps. Therefore, we define a time window $\boldsymbol{TW} \in \mathbb{R}^{A \times 2}$ to specify the first and last timesteps during which the agent is observed.

The input features of agents provided to the encoder include: 
\begin{itemize}[leftmargin=*]
    \item \textbf{Control point vectors:} $\boldsymbol{\Delta}^{\text{a,hist}} = \text{concat}(\boldsymbol{\Delta}^{\text{a,hist}}_{1}, \boldsymbol{\Delta}^{\text{a,hist}}_{2}, \dots, \boldsymbol{\Delta}^{\text{a,hist}}_{5}) \in \mathbb{R}^{A\times10}$,
    \item \textbf{Agent positional information:} $\boldsymbol{PI}^{\text{a}}=\text{concat}(\textbf{W}^{\text{a,hist}}_{5}, \boldsymbol{\Gamma}^{\text{a,hist}}_{5}) \in \mathbb{R}^{A\times4}$,
    \item \textbf{Time window:} $\boldsymbol{TW} \in \mathbb{R}^{A \times 2}$, indicating the observed historical period,
    \item \textbf{Agent type:} $\boldsymbol{AT} \in \mathbb{Z}^A$, where each entry is an integer label indicating the agent type  (e.g., vehicle, pedestrian, cyclist).
\end{itemize}

For EP-Q, the agent features are transformed into each agent's individual coordinate frame, requiring an additional data dimension. Specifically, the control point vectors become $\boldsymbol{\Delta}^{\text{a,hist}} \in \mathbb{R}^{A\times A\times10}$, agent positional information becomes $\boldsymbol{PI}^{\text{a}}\in \mathbb{R}^{A\times A \times4}$, where the first dimension represent $A$ agents' coordinate frames. The non-geometric features (time windows and agent types) are simply replicated for each coordinate frame.

\begin{figure}[thb]
\centering
\includegraphics[width=1.0\textwidth]{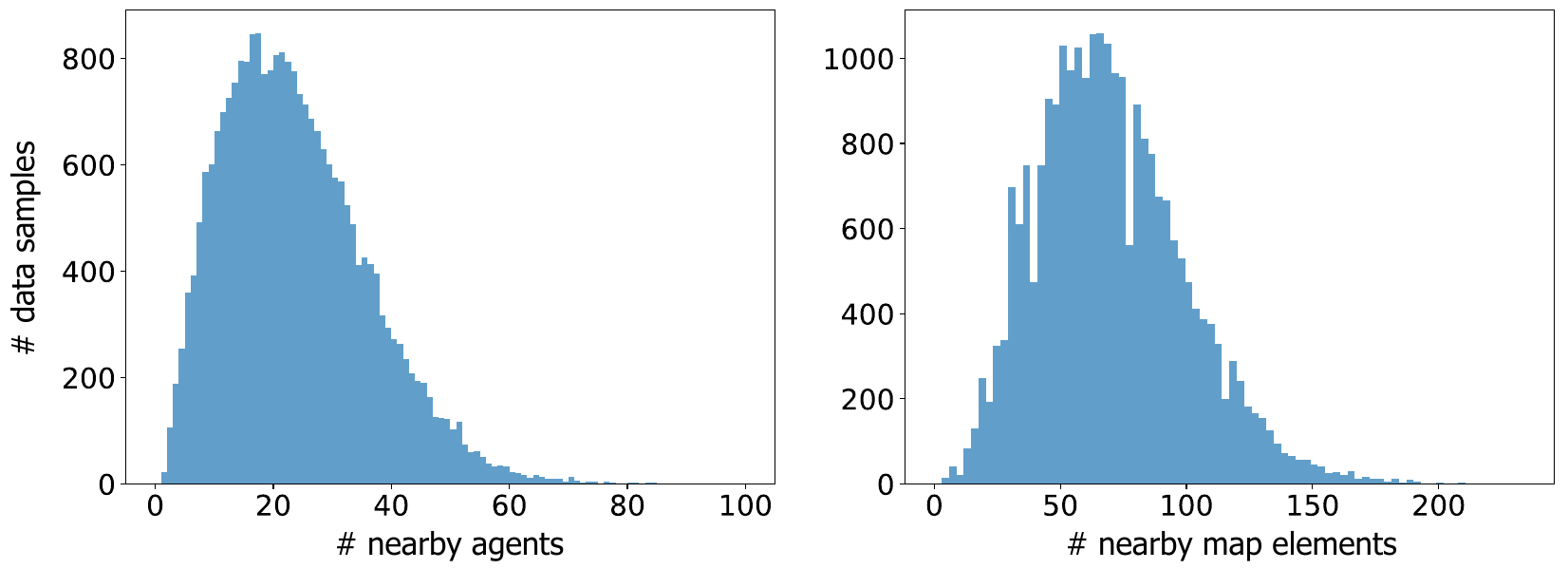}
\caption[Statistics of nearby agents and map elements]{Histogram of nearby agents and map elements within a 100-meter range of the target agent on the Argoverse 2 validation set.}
\label{fig: statistics of nearby agents and map elements}
\end{figure}

To enable batch processing with fixed-size data tensors, all EP variants consider up to 50 agents closest to the target agent in each traffic scene. Scenes containing fewer than 50 agents are zero-padded accordingly. The choice of 50 is based on an analysis of the distribution of nearby agents around the target agent, as visualized in the left panel of Figure~\ref{fig: statistics of nearby agents and map elements}, ensuring that the most relevant neighboring agents are included for modeling interaction and prediction. 

\subsubsection{Map Inputs}
The geometry of $M$ map elements in the traffic scene is represented using 3-degree polynomials. Features are computed similarly to agents:
\begin{itemize}[leftmargin=*]
    \item \textbf{Map control point vectors:} $\boldsymbol{\Delta}^{\text{m}} = \text{concat}(\boldsymbol{\Delta}^{\text{m}}_{1}, \boldsymbol{\Delta}^{\text{m}}_{2}, \boldsymbol{\Delta}^{\text{m}}_{3}) \in \mathbb{R}^{M\times6}$,
    \item \textbf{Map positional information:} $\boldsymbol{PI}^{\text{m}}=\text{concat}(\textbf{W}^{\text{m}}_{0}, \boldsymbol{\Gamma}^{\text{m}}_{1}) \in \mathbb{R}^{M\times4}$,
    \item \textbf{Map element type:} $\boldsymbol{MT} \in \mathbb{Z}^M$,  where each entry is an integer label indicating the map element type (e.g., vehicle lane, bike lane, crosswalk).
\end{itemize}
The control point vectors $\boldsymbol{\Delta}^{\text{m}}$ and positional information $\boldsymbol{PI}^{\text{m}}$ represent the centerlines of lane segments and crosswalks, following our dataset homogenization protocol. We use $\textbf{W}^{\text{m}}_{0}$ (initial control point) and $\boldsymbol{\Gamma}^{\text{m}}_{1}$ (normalized vector between the first two control points, representing the initial direction) as reference features for map elements.

For EP-Q, map features are transformed into each agent's coordinate frame, resulting in $\boldsymbol{\Delta}^{\text{m}} \in \mathbb{R}^{A\times M\times6}$ and $\boldsymbol{PI}^{\text{m}} \in \mathbb{R}^{A\times M\times4}$, where each agent has its own transformed representation of all map elements. Map element types are replicated across coordinate frames.

To ensure consistent input sizes during training, each scene includes up to 150 map elements closest to the target agent. This limit is based on the distribution of nearby map elements around the target agent, as shown in the right panel of Figure~\ref{fig: statistics of nearby agents and map elements}.

\subsection{Encoder}
\begin{figure}[!thb]
\centering
\includegraphics[width=0.9\textwidth]{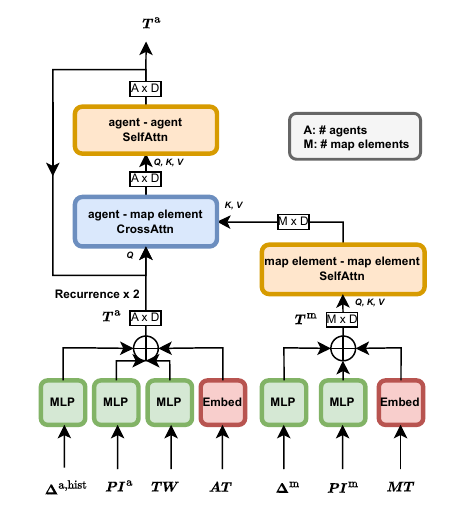}
\caption[Encoder architecture of EP]{Encoder architecture shared by EP variants. The inputs are defined in Section \ref{sec 4.5.1: model inputs}. $\boldsymbol{T}^{\text{a}}, \boldsymbol{T}^{\text{m}}$ denote agent and map-element tokens, respectively. The notations $Q, K, V$ denote whether the features serve as queries, keys, or values in the attention mechanism, respectively.}
\label{fig: ep encoder architecture}
\end{figure}

All EP variants share the same encoder architecture, as visualized in Figure \ref{fig: ep encoder architecture}. Although the encoder architecture is identical across variants, their inputs -- $\boldsymbol{\Delta}^{\text{a,hist}}$, $\boldsymbol{\Delta}^{\text{m}}$, $\boldsymbol{PI}^{\text{a}}$, and $\boldsymbol{PI}^{\text{m}}$ -- are computed in different coordinate frames: the target agent’s coordinate for EP-F and EP-noAug, and the individual agent’s coordinate for EP-Q. For clarity, we use the EP-F dimensional notation in the following description.

Numerical features, such as control point vectors $\boldsymbol{\Delta}^{\text{a,hist}}$ and $\boldsymbol{\Delta}^{\text{m}}$, are encoded via simple 3-layer MLPs. Categorical attributes, such as $\boldsymbol{AT}$ and $\boldsymbol{MT}$, are encoded via individual embedding layers \cite{mikolov_efficient_2013}. Embedded features are added and fused to agent tokens $\boldsymbol{T}^{\text{a}} \in \mathbb{R}^{A \times D}$ and map element tokens $\boldsymbol{T}^{\text{m}} \in \mathbb{R}^{M \times D}$, where $D$ denotes the hidden dimension. Multiple attention blocks (cf. Section \ref{sec 2.2.2: Environment Encoding and Modeling}) with ``pre-layer normalization'' \cite{xiong_layer_2020} perform the \emph{map}-\emph{map}, \emph{agent}-\emph{map} and \emph{agent}-\emph{agent} attentions sequentially and update $\boldsymbol{T}^{\text{a}}$.

After these attention blocks, the updated agent tokens $\boldsymbol{T}^{\text{a}}$ form the output of the encoder: each token aggregates the kinematic history of its agent together with its interactions with map elements and other agents. The agent tokens are then passed to the decoder (Section \ref{sec: ep decoder}) to produce the predictions.

\subsection{Decoder}
\label{sec: ep decoder}
\begin{figure}[thb]
\centering
\includegraphics[width=1.0\textwidth]{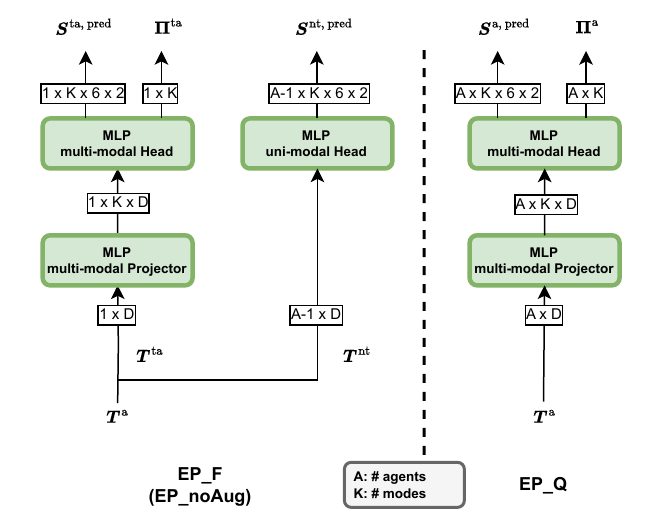}
\caption[Decoder architecture of EP]{Decoder architectures of EP-F, EP-Q and EP-noAug.}
\label{fig: ep decoder architecture}
\end{figure}

Figure \ref{fig: ep decoder architecture} visualizes the two decoder architectures of EP-F (EP-noAug) and EP-Q. Although both variants operate on computed agent tokens $\boldsymbol{T}^{\text{a}}$ from the encoder, they produce different outputs due to their distinct augmentation strategies. 

The decoder for EP-F and EP-noAug, shown in the left panel of Figure~\ref{fig: ep decoder architecture}, first splits the agent tokens $\boldsymbol{T}^{\text{a}}$ into target agent tokens $\boldsymbol{T}^{\text{ta}}$ and non-target agent tokens $\boldsymbol{T}^{\text{nt}}$. For multi-modal prediction, the target agent tokens $\boldsymbol{T}^{\text{ta}}$ is first projected into 6 modes and subsequently decoded to predictions $\boldsymbol{S}^{\text{ta, pred}}$ along with their corresponding mode probabilities $\boldsymbol{\Pi}^{\text{ta}}$. The non-target agent tokens $\boldsymbol{T}^{\text{nt}}$ are directly decoded into uni-modal predictions $\boldsymbol{S}^{\text{nt, pred}}$. 

In contrast, the decoder for EP-Q does not distinguish between target and non-target agents, as shown in the right panel of Figure \ref{fig: ep decoder architecture}. Instead, all agent tokens $\boldsymbol{T}^{\text{a}}$ are projected into 6 modes and decoded to predictions $\boldsymbol{S}^{\text{a, pred}}$ and mode probabilities $\boldsymbol{\Pi}^{\text{a}}$ for all agents.


\subsection{Outputs}
\label{sec 4.5.4: model outputs}
All EP variants predict 6-second future trajectories. Rather than directly predicting trajectory control points, EP variants predict each agent's future kinematic states at specific time horizons. These predicted kinematic states are then combined with the tracked states at the reference timestep (cf. Section \ref{sec 4.4.1: Agent History as Polynomial}) to determine the trajectory's control points through least squares fitting. This approach provides flexibility in balancing the reliability of tracking data versus prediction quality, while ensuring temporal continuity between observed and predicted motion.

For all EP variants, we use the tracked position at the reference timestep ($t=t_0$) along with the predicted positions, velocities, and accelerations at two future horizons: 3 seconds ($t=t_0+30$) and 6 seconds ($t=t_0+60$).

For the $k$-th mode of the $i$-th agent, we concatenate the tracked and predicted states into a single column vector:
\begin{equation}
    \mathbf{s}_{i, k} = [\mathbf{p}^{\text{track}}_{i, t_0}, \mathbf{p}^{\text{pred}}_{i, t_0+30, k}, \mathbf{v}^{\text{pred}}_{i, t_0+30, k}, \mathbf{a}^{\text{pred}}_{i, t_0+30, k}, \mathbf{p}^{\text{pred}}_{i, t_0+60, k}, \mathbf{v}^{\text{pred}}_{i, t_0+60, k}, \mathbf{a}^{\text{pred}}_{i, t_0+60, k}]^{\top} \in \mathbb{R}^{14}
\end{equation}
Here, $\mathbf{p}^{\text{track}}_{i, t_0}$ is the tracked 2D position at reference timestep $t_0$, while $\mathbf{p}^{\text{pred}}_{i, t, k}, \mathbf{v}^{\text{pred}}_{i, t, k}, \mathbf{a}^{\text{pred}}_{i, t, k}$ represent the predicted 2D position, velocity, and acceleration at timestep $t$ for mode $k$, respectively. The kinematic state vector $\mathbf{s}_{i, k}$ provides sufficient constraints to determine the future trajectory's control points. 

Following the formulation in Section \ref{sec: 3.2 parametric representation}, the Bernstein basis functions for 6-degree future trajectories are expressed as:
\begin{equation}
    \boldsymbol{\phi}(\tau_t)=[\phi_0(\tau_t);\dots;\phi_6(\tau_t)]
\end{equation}
where $\tau_t = \frac{t - t_0}{T - t_0}$ denotes the normalized future timestamp. 
We employ discrete timesteps rather than continuous time to compute $\tau_t$, since actual future sampling times are unavailable during inference. Furthermore, as discussed in Section \ref{sec: outlier filtering results}, both the A2 and WO datasets exhibit consistent sampling rates. We therefore assume uniform, temporally equidistant sampling over the prediction horizon when computing $\tau_t$.

To determine the trajectory control points from the kinematic state vector $\mathbf{s}_{i, k}$. we construct the matrix $\mathbf{H} \in \mathbb{R}^{14\times14}$ as:
\begin{equation}
    \mathbf{H}=[\boldsymbol{\phi}(\tau_{t_0}), \boldsymbol{\phi}(\tau_{t_0+30}), \dot {\boldsymbol{\phi}}(\tau_{t_0+30}), \ddot{\boldsymbol{\phi}}(\tau_{t_0+30}), \boldsymbol{\phi}(\tau_{t_0+60}), \dot{\boldsymbol{\phi}}(\tau_{t_0+60}), \ddot{\boldsymbol{\phi}}(\tau_{t_0+60})]^{\top}\otimes\mathbf{I}_2
\end{equation}
where $\mathbf{I}_2$ is the $2\times2$ identity matrix and $\otimes$ denotes the Kronecker product. The terms $\dot {\boldsymbol{\phi}}(\tau_{t})$ and $\ddot{\boldsymbol{\phi}}(\tau_{t})$ represent the first and second derivatives of the Bernstein basis functions at timestep $t$, respectively, as defined in Equation \ref{eqn: basis function derivative}.

The predicted control points $\boldsymbol{\omega}^{\text{pred}}_{i, k}$ for the $k$-th mode of the $i$-th agent are then estimated via least squares:
\begin{equation}
    \boldsymbol{\omega}^{\text{pred}}_{i, k}=(\mathbf{H}^{\top}\mathbf{H})^{-1}\mathbf{H}^{\top}\mathbf{s}_{i, k}.
\end{equation}
Finally, the predicted future positions can then be reconstructed using the polynomial trajectory representation defined in Equation \ref{eqn: trajectory representation}.

\subsection{Training Loss}
Our training objective employs different loss formulations depending on the prediction modality and EP variant.

\textbf{Multi-modal} prediction utilizes two complementary loss components:
\begin{itemize}[leftmargin=*]
    \item \textbf{Regression loss $\ell_{\mathrm{mul}}$}: minimum Average Displacement Error over 6 modes ($\mathrm{minADE}_6$).
    \item \textbf{Mode-weighted loss $\ell_{\mathrm{mode}}$}: weighted average displacement error using predicted mode probabilities as weights across all predicted modes.
\end{itemize}
\textbf{Uni-modal} prediction employs a single loss component:
\begin{itemize}[leftmargin=*]
    \item \textbf{Regression loss $\ell_{\mathrm{uni}}$}: standard average displacement error.
\end{itemize}


The overall training losses for each EP variant are defined as:
\begin{equation}
\label{eqn:EP loss}
\begin{aligned}
\ell_{\text{EP-F}} = & \ell^{\mathrm{ta}}_{\mathrm{mul}} + \ell^{\mathrm{ta}}_\mathrm{{mode}} + \ell^{\mathrm{nt}}_{\mathrm{uni}}\\
\ell_{\text{EP-Q}} = & \ell^{\mathrm{a}}_{\mathrm{mul}} + \ell^{\mathrm{a}}_{\mathrm{mode}}\\
\ell_{\text{EP-noAug}} = &  \ell^{\mathrm{ta}}_{\mathrm{mul}} + \ell^{\mathrm{ta}}_{\mathrm{mode}}\\
\end{aligned}
\end{equation}
where the superscripts indicate the agent scope: ``ta'' for target agents, ``nt'' for non-target agents, and ``a'' for all agents.

\subsection{Training Settings}
\label{sec: EP Training Settings}
We report the training settings of EP variants in Table \ref{tab: setting for EP}. EP-Q is trained with a reduced batch size and fewer epochs due to its higher computational complexity. Note that the batch size and number of epochs are adjusted to ensure the same training iterations for all EP variants. 
\begin{table}[!tbh]
\caption[Training settings for EP variants]{Training settings for EP variants}
\vspace{-0.0em}
\centering
\begin{tabularx}{\textwidth}{c  >{\centering\arraybackslash}X |>{\centering\arraybackslash}X }
\Xhline{3\arrayrulewidth}
model& EP-F \& EP-noAug & EP-Q \\
\Xhline{3\arrayrulewidth}
hidden dimension $\mathrm{D}$ &  \multicolumn{2}{c}{64}\\
\hline
optimizer &  \multicolumn{2}{c}{Adam \cite{kingma_adam_2014}} \\
\hline
learning rate & 1e-3 & 5e-4\\
\hline
learning rate schedule &  \multicolumn{2}{c}{cosine \cite{loshchilov_sgdr_2016}}\\
\hline
batch size &  64 & 32\\
\hline
training/warmup epochs &  128/20 & 64/10\\
\hline
dropout &  \multicolumn{2}{c}{0.1} \\
\Xhline{3\arrayrulewidth}
\vspace{-1em}
\end{tabularx}
\label{tab: setting for EP}
\end{table}

\section{Experiments}
\label{sec 4.6: experiments}
\subsection{Experimental Setup}
\label{sec 4.6.1: experimental setup}
We design three \emph{training}$\rightarrow$\emph{testing} configurations to comprehensively evaluate all models for both ID performance and OoD robustness, as summarized in Table \ref{tab: experiment setting}.

\subsubsection{\TOA{}}
This configuration follows the official A2 competition protocol: all models are trained from scratch on the original A2 training set using their respective official implementations and hyperparameters, and evaluated on the original A2 test set. Both training and testing use a 6-second prediction horizon. We refer to this configuration as \textbf{\TOA{}} in the remainder of this chapter.

\subsubsection{\TA{}}
Asterisk (*) indicates homogenized datasets. In this configuration, all models are trained from scratch on the homogenized A2 training set using the hyperparameters originally reported for the A2 competition and a 6-second prediction horizon. Models are evaluated on:
\begin{itemize}
    \item the homogenized A2 validation set for ID performance, and
    \item the homogenized WO validation set for OoD robustness (4.1-second prediction horizon).
\end{itemize}
Although ID performance can be assessed up to a 6-second horizon, results are reported at the homogenized 4.1-second horizon by default unless otherwise specified.

Dataset homogenization requires modifications to certain model components. For example, QCNet's preprocessing and encoding layers are adjusted to remove junction-lane label encodings, while FMAE remains unaffected. We refer to this configuration as \textbf{\TA{}} in the remainder of this chapter.


\subsubsection{\TWO{}}
Asterisk (*) indicates homogenized datasets. This configuration is the reverse of \TA{}: all models are trained from scratch on the homogenized WO training set using their default hyperparameters and a 4.1-second prediction horizon. Evaluation is performed on:
\begin{itemize}
    \item the homogenized WO validation set for ID performance, and
    \item the homogenized A2 validation set for OoD robustness (both with 4.1-second horizon).
\end{itemize}
Compared to A2, WO exhibits substantially higher map complexity, with more lane segments and denser map points. This complexity causes GPU memory exhaustion when training QCNet with its default configuration, even on a cluster with 192~GB GPU memory. To fit within hardware constraints while ensuring fairness across models, we match QCNet’s scenario complexity to that of the EP variants, limiting each scene to at most 50 agents and the 150 map elements closest to the target agent.

As in \TA{}, QCNet’s preprocessing and encoding layers are adjusted to remain compatible with the homogenized dataset. We refer to this configuration as \textbf{\TWO{}} in the remainder of this chapter.

\begin{table}[!tbh]
\caption[Summary of experimental setups]{Summary of experimental setups. Asterisk (*) indicates homogenized datasets.}
\vspace{-0.0em}
\centering
\begin{tabularx}{1\textwidth}{c c >{\centering\arraybackslash}X >{\centering\arraybackslash}X >{\centering\arraybackslash}X}
\Xhline{3\arrayrulewidth}
& & \TOA & \TA & \TWO \\
\Xhline{3\arrayrulewidth}
& data split & \multicolumn{1}{!{\vrule width 1pt}c}{\multirow{1}{*}{A2 train}}&\multirow{1}{*}{A2* train} & WO* train\\
\cmidrule(lr){2-5}
training & pred. horizon & \multicolumn{1}{!{\vrule width 1pt}c}{\multirow{1}{*}{\SI{6}{s}}} & \SI{6}{s} & \SI{4.1}{s}\\
\cmidrule(lr){2-5}
& $\#$data samples &\multicolumn{1}{!{\vrule width 1pt}c}{\multirow{1}{*}{\num{199908}}} & \num{199908}& \num{472235}\\
\hhline{=====}
& data split & \multicolumn{1}{!{\vrule width 1pt}c}{\multirow{1}{*}{A2 test}}&\multirow{1}{*}{A2* val} & WO* val\\
\cmidrule(lr){2-5}
ID testing & pred. horizon & \multicolumn{1}{!{\vrule width 1pt}c}{\multirow{1}{*}{\SI{6}{s}}} & \SI{6}{s} | \SI{4.1}{s} & \SI{4.1}{s}\\
\cmidrule(lr){2-5}
& $\#$data samples &\multicolumn{1}{!{\vrule width 1pt}c}{\multirow{1}{*}{\num{24984}}} & \num{24988}& \num{42465}\\
\hhline{=====}
& data split & \multicolumn{1}{!{\vrule width 1pt}c}{\multirow{1}{*}{-}}&\multirow{1}{*}{WO* val} & A2* val\\
\cmidrule(lr){2-5}
OoD testing & pred. horizon & \multicolumn{1}{!{\vrule width 1pt}c}{\multirow{1}{*}{-}} & \SI{4.1}{s} & \SI{4.1}{s}\\
\cmidrule(lr){2-5}
& $\#$data samples &\multicolumn{1}{!{\vrule width 1pt}c}{\multirow{1}{*}{-}} & \num{42465}& \num{24988}\\
\Xhline{3\arrayrulewidth}
\end{tabularx}
\label{tab: experiment setting}
\end{table}

\subsection{Metrics}
For ID testing, we use the official benchmark metrics, including minimum Average Displacement Error ($\text{minADE}_{K}$) and minimum Final Displacement Error
($\text{minFDE}_{K}$), for evaluation. The metric $\text{minADE}_{K}$ calculates the Euclidean distance in meters between the observed trajectory and the best of $K$ predicted trajectories as an average of all future time steps. Conversely, $\text{minFDE}_{K}$ focuses solely on the displacement error at the final time step, emphasizing long-term performance. 

For OoD testing, we also propose $\Delta \text{minADE}_{K}$ and $\Delta \text{minFDE}_{K}$ as the difference of displacement error between ID  and OoD testing to measure model robustness. Following standard practice in the A2 and WO competitions, $K$ is selected as 1 and 6.

\subsection{Impact of Data Homogenization}
\label{sec 4.6.3: impact of data homogenization}
Our data homogenization protocol introduces two major modifications compared to the original competition setups: (\lowerromannumeral{1}) exclusion of certain map information (lane boundaries and junction lane labels), and (\lowerromannumeral{2}) extending WO's history length from \SI{1.1} {s} to \SI{5}{s}.

To assess these impacts, we evaluate benchmark models trained on original versus homogenized A2 data to quantify the effect of reduced map information. Additionally, we analyze how the extended history length affects prediction task complexity in the WO dataset.

\subsubsection{Impact of Reduced Map Information}
\begin{figure}[!th]
\centering
\includegraphics[width=\textwidth]{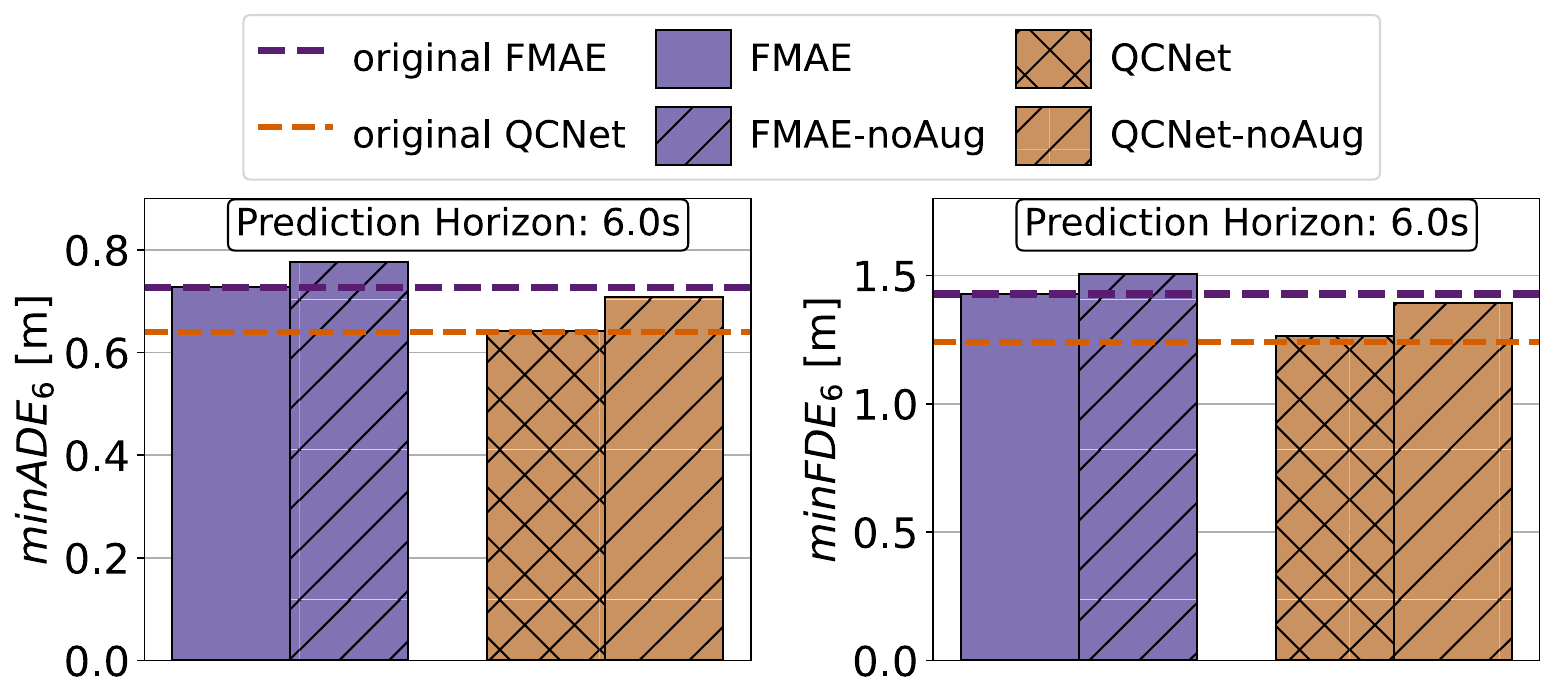}
\caption[In-distribution performance of benchmarks on Argoverse 2]{ID performance of benchmarks on the A2 test set. \textbf{Dashed Lines}: The published results for benchmark models trained on original A2 \cite{cheng_forecast_2023, zhou_query_2023}. \textbf{Bars}: Results for benchmark models trained on homogenized A2.}
\label{fig:ID_results}
\end{figure}

Figure \ref{fig:ID_results} compares the prediction performance of benchmark models on the A2 test set (6-second prediction horizon) when trained with original versus homogenized A2 data. FMAE shows no performance change, as its architecture does not utilize the removed map features. QCNet exhibits only marginal differences from its original performance, demonstrating that our homogenization protocol has minimal impact on its ID performance. 

These results validate our approach and justify using both benchmark models for subsequent ID and OoD comparisons on homogenized datasets. Furthermore, the close agreement with the published results confirms the correctness of our training procedure.

\subsubsection{Impact of Extended WO History Length}
Creators of both datasets mined their recordings to identify challenging scenarios and target agents for prediction \cite{wilson_argoverse2_2021, ettinger_waymo_2021}. For example, WO explicitly notes that their target selection is ``biased to include objects that do not follow a constant velocity model or
straight paths.'' Deviations from constant velocity typically arise due to interactions
with other agents or non-trivial map geometry (e.g., turns, stops, merges). In such cases, the agent trajectory is more dependent on context, making prediction more difficult for learned models. Our homogenization protocol extends WO's history length from \SI{1.1}{s} to \SI{5}{s}, which effectively shifts the reference time $t'_{0}$ and may alter the original selection bias.

To investigate this effect, we quantify prediction complexity in WO under different history lengths by measuring target agents’ deviation from a constant velocity model during the prediction horizon. To ensure fair comparison, we use a consistent 4.1-second prediction horizon at different starting time $t'_{0}$ and measure the normalized longitudinal and lateral deviations between the start position $\mathbf{p}_{t'_0}$ and the end position $\mathbf{p}_{t'_0+4.1}$ of the observed future trajectory:
\begin{equation}
\begin{aligned}
\label{eqn:normlized distance}
\mathbf{d} &= \frac{(\mathbf{p}_{t'_0+4.1}-\mathbf{p}_{t'_0})\boldsymbol{R}^{\mathrm{rot}}_{t'_0}}{|\mathbf{v}_{t'_0}|* 4.1}\\
\end{aligned}
\end{equation}
Here, $\boldsymbol{R}^{\mathrm{rot}}_{t'_0}$ is the $2\times 2$ rotation matrix determined by the agent's heading at $t'_0$, which aligns the agent's motion with the x-axis direction at time $t'_0$. The term $|\mathbf{v}_{t'_0}|$ denotes the velocity magnitude at time ${t'_0}$. The normalized distance $\mathbf{d}$ is a 2D vector of normalized longitudinal and lateral distances. A value of $\mathbf{d}= (1,0)$ indicates that agent's motion strictly follows the constant velocity model. In contrast, larger deviations from $(1,0)$ reflect greater divergence from constant velocity, signifying higher prediction complexity.

\begin{figure}[!tbh]
\centering
\includegraphics[width=0.8\textwidth]{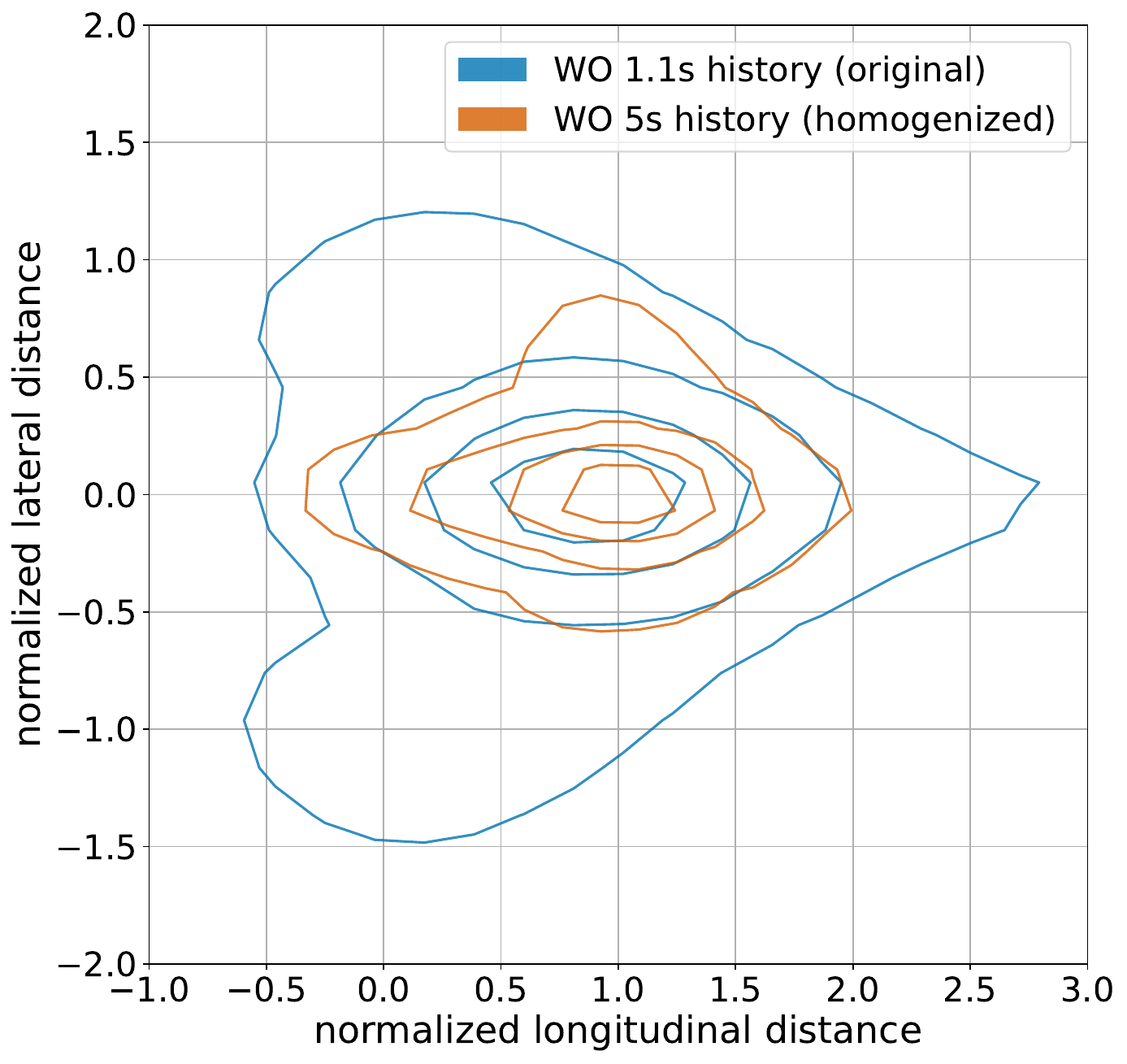}
\caption[Complexity of prediction tasks]{Kernel density plot of normalized longitudinal and lateral distances with 1.1-second and 5-second history lengths for WO validation set. Contours indicate the 20-th, 40-th, 60-th and 80-th
percentiles, respectively, illustrating a broader spread for the prediction task with the original 1.1-second history.}
\label{fig:prediction_complexity}
\vspace{0.0em}
\end{figure}

Figure \ref{fig:prediction_complexity} visualizes the distribution of $\mathbf{d}$ for $t'_{0} = 1.1\: \text{s}$ and \SI{5}{s} in the WO validation set. Compared to the distribution at $t'_{0}=5\:\text{s}$, the distribution at $t'_{0}=1.1\:\text{s}$ is broader, indicating that the prediction task with the original 1.1-second history is more complex than with the 5-second history in the homogenized WO dataset. 

This simplified prediction task, resulting from data homogenization, may introduce advantages for models tested on the homogenized WO dataset, but may also introduce disadvantages due to reduced training data variance. We discuss this point in Sections \ref{sec 4.6.4: ID results} and \ref{sec 4.6.5: OoD results}.

\begin{figure}[!th]
\centering
\includegraphics[width=1.0\textwidth]{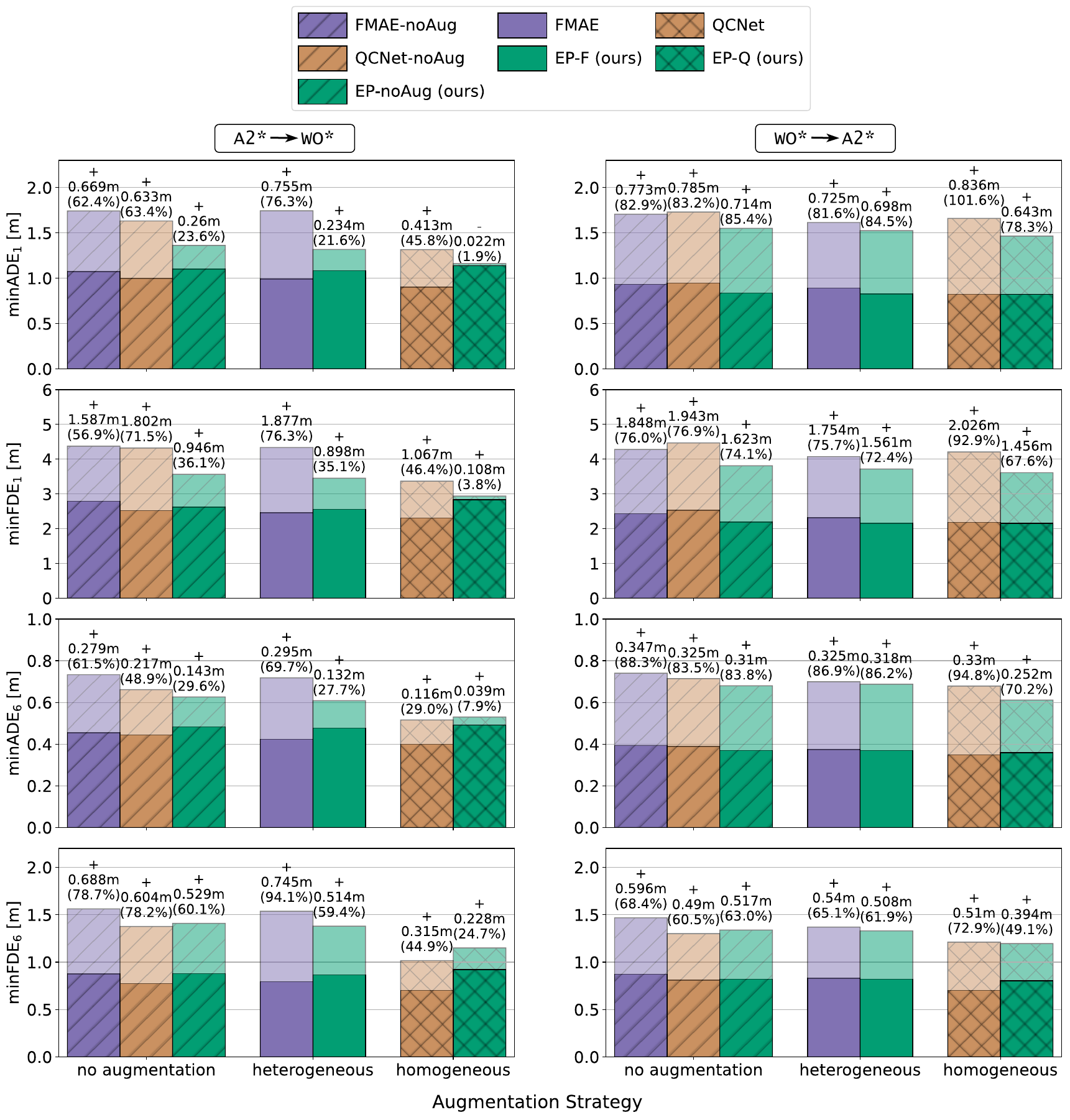}
\caption[Statistics of nearby agents and map elements]{The OoD testing results of FMAE, QCNet, EP and their variants. The \emph{solid} bars represent ID results reported in Table \ref{tab: in-distribution result combined}, while the \emph{transparent} bars indicate the increase in displacement error during OoD testing results reported in Table \ref{tab: out-of-distribution result combined}. We indicate the \emph{absolute} and \emph{relative} difference in displacement error between ID  and OoD results. \textbf{Left (\TA{})}: Models trained on the homogenized A2 training set and tested on the homogenized A2 (ID) and WO (OoD) validation sets. \textbf{Right (\TWO{})}: Models trained on the homogenized WO training set and tested on the homogenized WO (ID) and A2 (OoD) validation sets.}
\label{fig: OoD_results}
\end{figure}

\subsection{In-Distribution Results}
\label{sec 4.6.4: ID results}
While ID performance is not the primary focus of this chapter, we follow standard practice by reporting it first. We begin with results from models trained on A2 (\TOA{} and \TA{} setups), followed by results from models trained on WO (\TWO{} setup).

As visualized in Figure \ref{fig:ID_results}, removing data augmentation leads to a small but consistent negative effect on ID performance for both benchmark models. Since the ``noAug'' variants are primarily designed to analyze the impact of augmentation strategies in OoD evaluation (cf. Section \ref{sec 4.6.5: OoD results}), we exclude them from the ID results to maintain clarity and focus on the core comparison between our polynomial-based approach and the established benchmarks.

\subsubsection{\TOA{} Results}
The ID results under the \TOA{} setup with a 6-second prediction horizon are presented in Table \ref{tab: in-distribution result (6s)}, with benchmark results as reported by the original authors. We use QCNet as the reference point (100$\%$) and report the relative performance of all models accordingly.

EP achieves near SotA performance. For instance, the $\text{minFDE}_{1}$ of EP-F is only $6.0\%$ higher compared to QCNet. Although EP-F exhibits a considerable gap compared to QCNet in terms of multi-modal prediction ($K=6$), it reports only \SI{0.074}{m}  higher $\text{minADE}{6}$ (\SI{0.801}{m} vs. \SI{0.727}{m}) and \SI{0.099}{m} higher $\text{minFDE}_{6}$ (\SI{1.526}{m} vs. \SI{1.427}{m}) relative to FMAE. Notably, EP-F outperforms EP-Q in ID performance, suggesting that heterogeneous augmentation is more effective than homogeneous augmentation for ID optimization on the A2 dataset.

\begin{table}[!b]
\caption[In-distribution results with a 6-second prediction horizon]{ID results in competition setting on Argoverse 2 test set with a 6-second prediction horizon (\TOA{} setup). The best and second-best results for each metric across models are highlighted in \textbf{bold} and \underline{underline}, respectively.}
\centering
    \begin{tabularx}{\textwidth}{l |*{4}{>{\centering\arraybackslash}X}}
    \toprule
    \multirow{2}{*}{\shortstack{model}} & minADE$_{1}$ & minFDE$_{1}$ & minADE$_{6}$ & minFDE$_{6}$ \\
        & [m] $\downarrow$ &[m] $\downarrow$ & [m] $\downarrow$ & [m] $\downarrow$ \\
        \midrule
        \multirow{2}{*}{QCNet \cite{zhou_query_2023}} & \textbf{1.702} & \textbf{4.309} & \textbf{0.643}& \textbf{1.244}\\
        &\footnotesize (\textbf{100.0\%})& \footnotesize (\textbf{100.0\%})&\footnotesize (\textbf{100.0\%})& \footnotesize (\textbf{100.0\%})\\
        \cmidrule(lr){2-5} 
        \multirow{2}{*}{FMAE \cite{cheng_forecast_2023}} &\underline{1.845} & 4.602& \underline{0.727} & \underline{1.427} \\
        &\footnotesize (\underline{108.4\%})&\footnotesize (106.8\%)&\footnotesize (\underline{113.1\%})&\footnotesize (\underline{114.7\%})\\
        \cmidrule(lr){2-5} 
        \multirow{2}{*}{EP-Q (ours)} &2.134 & 5.415 & 0.841 & 1.683 \\
        &\footnotesize (125.4\%)& \footnotesize (125.7\%)& \footnotesize (131.0\%)& \footnotesize (135.3\%)\\
        \cmidrule(lr){2-5}
        \multirow{2}{*}{EP-F (ours)} & 1.887 & \underline{4.567}  & 0.801 & 1.526  \\ 
        &\footnotesize (110.9\%) & \footnotesize (\underline{106.0\%}) & \footnotesize (124.6\%) & \footnotesize (122.7\%) \\
        \bottomrule
    \end{tabularx}
    \label{tab: in-distribution result (6s)}
\end{table}


\subsubsection{\TA{} ID Results}
The ID results under the \TA{} setup with a 4.1-second prediction horizon are presented in the upper section of Table~\ref{tab: in-distribution result combined}.

Despite the reduced prediction horizon and data homogenization modifications, the relative performance patterns remain consistent with \TOA{}. QCNet and FMAE again demonstrate superior ID performance, while EP-F outperforms EP-Q and achieves performance comparable to benchmark models -- for example, a $\text{minADE}_{6}$ of \SI{0.476}{m} versus \SI{0.400}{m} for QCNet.

\begin{table}[t]
\caption[In-distribution results with a 4.1-second prediction horizon]{ID results evaluated with a 4.1-second prediction horizon. The best and second-best results for each metric across models are highlighted in \textbf{bold} and \underline{underline}, respectively. For clarity, the results of models without augmentation are not reported. \textbf{Upper Section (\TA)}: ID results of models trained on the homogenized A2 training set and tested on the homogenized A2 validation set. \textbf{Lower Section (\TWO)}: ID results of models trained on the homogenized WO training set and tested on the homogenized WO validation set.}

\centering

    \begin{tabularx}{\textwidth}{>{\centering\arraybackslash}X c | >{\centering\arraybackslash}X >{\centering\arraybackslash}X >{\centering\arraybackslash}X >{\centering\arraybackslash}X }
    \toprule
    \multirow{2}{*}{\shortstack{Setup}} & \multirow{2}{*}{\shortstack{Models}} & \multicolumn{1}{c}{minADE$_{1}$} & \multicolumn{1}{c}{minFDE$_{1}$} & \multicolumn{1}{c}{minADE$_{6}$} & \multicolumn{1}{c}{minFDE$_{6}$} \\
        &&[m] $\downarrow$&[m] $\downarrow$&[m] $\downarrow$&[m] $\downarrow$ \\
        \midrule

        \multirow{8}{*}{\shortstack{\thead{\TA \\ ID}}} & \multirow{2}{*}{QCNet \cite{zhou_query_2023}} &\textbf{0.902} & \textbf{2.301} & \textbf{0.400}& \textbf{0.701} \\
        &&\footnotesize (\textbf{100.0\%})&\footnotesize (\textbf{100.0\%})&\footnotesize (\textbf{100.0\%})& \footnotesize (\textbf{100.0\%})\\
        \cmidrule(lr){3-6} 
        & \multirow{2}{*}{FMAE \cite{cheng_forecast_2023}} &\underline{0.989} & \underline{2.459}& \underline{0.423} & \underline{0.792} \\
        &&\footnotesize (\underline{109.4\%})&\footnotesize (\underline{106.9\%})&\footnotesize (\underline{105.8\%})& \footnotesize (\underline{113.0\%})\\
        \cmidrule(lr){3-6} 
        & \multirow{2}{*}{EP-Q (ours)} & 1.161 & 2.830 & 0.491 & 0.922\\
        &&\footnotesize (128.7\%)&\footnotesize (123.0\%)&\footnotesize (122.8\%)&\footnotesize (131.5\%)\\
        \cmidrule(lr){3-6} 
        & \multirow{2}{*}{EP-F (ours)} & 1.082 & 2.555  & 0.476 & 0.865\\ 
        &&\footnotesize (120.0\%) & \footnotesize (111.0\%) & \footnotesize (119.0\%) & \footnotesize (123.4\%)\\
        \bottomrule
        \multirow{8}{*}{\shortstack{\thead{\TWO \\ ID}}} & \multirow{2}{*}{QCNet \cite{zhou_query_2023}} &\underline{0.823} & 2.182 & \textbf{0.348}& \textbf{0.700} \\
        & &\footnotesize (\underline{100.0\%})& \footnotesize (100.0\%)& \footnotesize (\textbf{100.0\%})& \footnotesize (\textbf{100.0\%})\\
        \cmidrule(lr){3-6} 
        & \multirow{2}{*}{FMAE \cite{cheng_forecast_2023}} &0.889 & 2.318& 0.374 & 0.829 \\
        &&\footnotesize (108.0\%)&\footnotesize (106.2\%)&\footnotesize (107.5\%)&\footnotesize (118.4\%)\\
        \cmidrule(lr){3-6} 
        & \multirow{2}{*}{EP-Q (ours)} & \textbf{0.821} & \textbf{2.155} & \underline{0.359} & \underline{0.802}\\
        &&\footnotesize (\textbf{99.8\%})& \footnotesize (\textbf{98.8\%})& \footnotesize (\underline{103.2\%})& \footnotesize (\underline{114.6\%})\\
        \cmidrule(lr){3-6} 
        & \multirow{2}{*}{EP-F (ours)} & 0.826 & \underline{2.156}  & 0.369 & 0.821\\ 
        &&\footnotesize (100.4\%) & \footnotesize (\underline{98.8\%}) & \footnotesize (106.0\%) & \footnotesize (117.3\%)\\
        \bottomrule
    \end{tabularx}
    \label{tab: in-distribution result combined}
\end{table}

\subsubsection{\TWO{} ID Results}
In contrast to the previous A2-based evaluations, we now assess ID performance on the WO dataset, where models are both trained and tested on the homogenized WO dataset. The ID results are presented in the lower section of Table~\ref{tab: in-distribution result combined}.

Although the larger volume and higher variance of the WO dataset \cite{feng_unitraj_2024} might be expected to pose challenges for smaller models, both EP-F and EP-Q outperform FMAE and achieve performance closer to QCNet -- marking a notable shift from the relative rankings observed on A2. Remarkably, EP-F and EP-Q even outperform QCNet in $\text{minFDE}_1$, achieving lower errors of \SI{2.156}{m} and \SI{2.155}{m}, respectively. Despite our computational constraints limiting the number of considered agents and map elements for QCNet (cf. Section \ref{sec 4.6.1: experimental setup}), it still achieves the best ID performance for multi-modal prediction ($K=6$)

Interestingly, unlike the A2 results, EP-F and EP-Q perform comparably, rather than EP-F demonstrating clear superiority. This suggests that the heterogeneous augmentation strategy is less effective on the homogenized WO dataset than on A2.

Additionally, we observe that most models achieve better absolute performance on WO than on A2 across ID tests, except the $\text{minFDE}_6$ of FMAE increases from \SI{0.792}{m} to \SI{0.829}{m}. This overall improvement on WO can be attributed to the larger size of the WO dataset, the relative simplicity of homogenized WO scenes (cf. Section \ref{sec 4.6.3: impact of data homogenization}), the lower noise levels in WO compared to A2 (cf. Section \ref{sec: 3.6.3: observation noise estimation}), or a combination of these factors.

\subsection{Out-of-Distribution Results}
\label{sec 4.6.5: OoD results}
The key question now is whether the results achieved in ID testing will translate to OoD testing. Figure \ref{fig: OoD_results} summarizes the results, where solid bars represent the ID results and transparent extensions indicate the performance drop in OoD testing. Both the relative and absolute increases in displacement error are shown and serve as our measures of model generalization. Detailed results are reported in Table \ref{tab: out-of-distribution result combined}.

Following the same structure as the ID evaluation, we begin with the results under the \TA{} setup, followed by the results under the \TWO{} setup.

\begin{table}[t]
\caption[Out-of-distribution results with a 4.1-second prediction horizon]{OoD results evaluated with a 4.1-second prediction horizon. The best and second-best results for each metric across models are highlighted in \textbf{bold} and \underline{underline}, respectively. For clarity, the results of models without augmentation are not reported.  \textbf{Upper Section (\TA)}: OoD results of models trained on the homogenized A2 training set and tested on the homogenized WO validation set. \textbf{Lower Section (\TWO)}: OoD results of models trained on the homogenized WO training set and tested on the homogenized A2 validation set.}

\centering

    \begin{tabularx}{\textwidth}{>{\centering\arraybackslash}X c | >{\centering\arraybackslash}X >{\centering\arraybackslash}X >{\centering\arraybackslash}X >{\centering\arraybackslash}X }
    \toprule
    \multirow{2}{*}{\shortstack{Setup}} & \multirow{2}{*}{\shortstack{Models}} & \multicolumn{1}{c}{minADE$_{1}$} & \multicolumn{1}{c}{minFDE$_{1}$} & \multicolumn{1}{c}{minADE$_{6}$} & \multicolumn{1}{c}{minFDE$_{6}$} \\
        &&[m] $\downarrow$&[m] $\downarrow$&[m] $\downarrow$&[m] $\downarrow$ \\
        \midrule

        \multirow{8}{*}{\shortstack{\thead{\TA{} \\ OoD}}} & \multirow{2}{*}{QCNet \cite{zhou_query_2023}} &\underline{1.315} & \underline{3.368} & \textbf{0.516}& \textbf{1.016} \\
        &&\footnotesize (\underline{100.0\%})&\footnotesize (\underline{100.0\%})&\footnotesize (\textbf{100.0\%})&\footnotesize (\textbf{100.0\%})\\
        \cmidrule(lr){3-6} 
        & \multirow{2}{*}{FMAE \cite{cheng_forecast_2023}} &1.744 & 4.336& 0.718 & 1.537 \\
        &&\footnotesize (132.6\%)&\footnotesize (128.7\%)&\footnotesize (139.1\%)&\footnotesize (151.3\%)\\
        \cmidrule(lr){3-6} 
        & \multirow{2}{*}{EP-Q (ours)} & \textbf{1.139} & \textbf{2.938} & \underline{0.530} & \underline{1.150}\\
        &&\footnotesize (\textbf{86.6\%})&\footnotesize (\textbf{87.2\%})&\footnotesize (\underline{102.7\%})&\footnotesize (\underline{113.2\%})\\
        \cmidrule(lr){3-6} 
        & \multirow{2}{*}{EP-F (ours)} & 1.316 & 3.453  & 0.608 & 1.379\\ 
        &&\footnotesize (120.0\%) & \footnotesize (100.1\%) & \footnotesize (117.8\%) & \footnotesize (135.7\%)\\
        \bottomrule
        \multirow{8}{*}{\shortstack{\thead{\TWO{} \\ OoD}}} & \multirow{2}{*}{QCNet \cite{zhou_query_2023}} &1.659 & 4.208 & \underline{0.678}& \underline{1.210} \\
        & &\footnotesize (100.0\%)&\footnotesize (100.0\%)&\footnotesize (\underline{100.0\%})&\footnotesize (\underline{100.0\%})\\
        \cmidrule(lr){3-6} 
        & \multirow{2}{*}{FMAE \cite{cheng_forecast_2023}} &1.614 & 4.072& 0.699 & 1.369 \\
        &&\footnotesize (97.3\%)&\footnotesize (96.8\%)&\footnotesize (103.1\%)&\footnotesize (113.1\%)\\
        \cmidrule(lr){3-6} 
        & \multirow{2}{*}{EP-Q (ours)} &\textbf{1.464} & \textbf{3.611} & \textbf{0.611} & \textbf{1.196}\\
        &&\footnotesize (\textbf{88.2\%})&\footnotesize (\textbf{85.8\%})& \footnotesize (\textbf{90.1\%})&\footnotesize (\textbf{98.8\%})\\
        \cmidrule(lr){3-6} 
        & \multirow{2}{*}{EP-F (ours)} & \underline{1.524} & \underline{3.717}  & 0.687 & 1.329\\ 
        &&\footnotesize (\underline{91.9\%}) & \footnotesize (\underline{88.3\%}) & \footnotesize (101.3\%) & \footnotesize (109.8\%)\\
        \bottomrule
    \end{tabularx}
    \label{tab: out-of-distribution result combined}
\end{table}

\subsubsection{\TA{} OoD Results}
Models' OoD performance when trained on the A2 dataset is visualized in the left panel of Figure \ref{fig: OoD_results} and reported in the upper section of Table \ref{tab: out-of-distribution result combined}. We present OoD results under the \TA{} setup from three perspectives: (\lowerromannumeral{1}) data representation, (\lowerromannumeral{2}) augmentation strategy, and (\lowerromannumeral{3}) contrast results between ID  and OoD testing:
\begin{enumerate}[label=\roman*.]
    \item \textbf{Data Representation}: As visualized in the left panel of Figure \ref{fig: OoD_results}, EP variants exhibit improved robustness compared to benchmarks using sequence-based data, independent of the augmentation strategy employed. For instance, EP-F outperforms FMAE by demonstrating lower displacement errors and smaller error increases in OoD testing. Similarly, EP-Q demonstrates improved robustness compared to QCNet, even achieving a slight reduction in displacement error of -\SI{0.022}{m} $(-1.9\%)$ in $\Delta \text{minADE}_1$.

    \item \textbf{Augmentation Strategy}: We observe that without augmentation, i.e., excluding the non-target agents in the loss function, all models generalize poorly on all metrics, but our model has the smallest relative and absolute increase in error in all cases. 

    Compared to FMAE-noAug, heterogeneous augmentation in FMAE does not improve OoD generalization, e.g., +\SI{0.279}{m} $(61.5\%)$ for FMAE-noAug vs. +\SI{0.295}{m} $(69.7\%)$ for FMAE in $\Delta \text{minADE}_{6}$. 
    
    
    On the other hand, the robustness of QCNet significantly benefits from homogeneous augmentation, with $\Delta \text{minFDE}_{1}$ reduced from +\SI{1.802}{m} $(71.5\%)$ to +\SI{1.067}{m} $(46.4\%)$. Compared to EP-noAug, EP-F and EP-Q replicate the behavior observed from FMAE and QCNet, showing only marginal improvement from heterogeneous augmentation and a significant improvement from homogeneous augmentation.

    \item \textbf{ID vs OoD results}: We observe multiple performance reversals between ID and OoD testing results under the \TA{} setup: 
    \begin{itemize}
        \item Despite FMAE demonstrating comparable ID performance to QCNet, it exhibits significantly lower robustness than QCNet in OoD testing, e.g, +\SI{0.755}{m} $(+76.3\%)$ compared to +\SI{0.413}{m} $(+45.8\%)$ in $\Delta \text{minADE}_1$.
        \item While FMAE and QCNet outperform EP variants on ID samples, EP variants show lower displacement errors than both benchmarks across multiple metrics in OoD testing. For example, in the upper section of Table \ref{tab: out-of-distribution result combined}, EP-F outperforms FMAE with the lower $\text{minADE}_1$ (\SI{1.316}{m} vs \SI{1.744}{m}). Similarly, EP-Q outperforms QCNet in $\text{minADE}_1$ and achieved nearly identical $\text{minADE}_6$.
        \item There is a performance reversal between EP-F and EP-Q across ID and OoD testing. For example, EP-F demonstrates the lower $\text{minFDE}_1$ than EP-Q in ID testing (\SI{2.555}{m} vs \SI{2.830}{m}). However, EP-Q exhibits significantly improved robustness than EP-F in OoD testing, with much lower $\text{minFDE}_1$ (\SI{2.938}{m} vs \SI{3.453}{m}) and $\Delta \text{minFDE}_1$ (+\SI{0.108}{m} vs +\SI{0.898}{m} as visualized in Figure \ref{fig: OoD_results}).
    \end{itemize}
    The reversals indicate that the ID results under the \TA{} setup do not fully reflect models' generalization capabilities and underscore the importance of OoD testing for a thorough evaluation of prediction models.
\end{enumerate}

\subsubsection{\TWO{} OoD Results}
The OoD results under the \TWO{} setup are visualized in the right panel of Figure \ref{fig: OoD_results} and reported in the lower section of Table \ref{tab: out-of-distribution result combined}. 


Following the same analytical framework as the \TA{} results, we present the \TWO{} OoD results from three perspectives: (\lowerromannumeral{1}) data representation, (\lowerromannumeral{2}) augmentation strategy, and (\lowerromannumeral{3}) comparison between \TA{} and \TWO{} setups.
\begin{enumerate}[label=\roman*.]
    \item \textbf{Data Representation}: The advantage of using polynomial representation for model generalization remains evident for EP-Q, with a substantial decrease in $\Delta \text{minFDE}_6$ from +\SI{0.516}{m} $(74.1\%)$ to +\SI{0.394}{m} $(49.1\%)$ compared to QCNet. Moreover, EP-Q achieves the lowest displacement error across all benchmarks.

    In contrast, EP-F and EP-noAug still demonstrate only marginal improvements over sequence-based models, with similar $\Delta \text{minFDE}_6$ between EP-F (+\SI{0.508}{m} $(61.9\%)$) and FMAE (+\SI{0.540}{m} $(65.1\%)$).
    

    \item \textbf{Augmentation Strategy}:
        Our \TA{} OoD results showed that heterogeneous augmentation yields only marginal robustness improvements for OoD testing. This trend continues under the \TWO{} setup for both EP-F and FMAE, confirming consistent results across datasets.

        For homogeneous augmentation, we expected notable improvements in generalization. EP-Q aligns with this expectation, showing reduced $\Delta \text{minADE}_6$ from +\SI{0.310}{m} $(83.8\%)$ to +\SI{0.252}{m} $(70.2\%)$ compared to EP-noAug. However, QCNet shows slightly degraded robustness compared to QCNet-noAug, with $\Delta \text{minADE}_6$ increasing from +\SI{0.325}{m} $(83.5\%)$ to +\SI{0.330}{m} $(94.8\%)$. 

    \item \textbf{Comparison between \TA{} and \TWO{}}:
    As discussed in Section \ref{sec 4.1: introduction and motivation}, the larger WO training dataset should theoretically improve model generalization. However, comparing OoD results between \TA{} and \TWO{} setups (left and right panels of Figure \ref{fig: OoD_results}), our observations diverge significantly from this expectation. 
    
    FMAE exhibits no notable improvement in robustness, with nearly identical $\Delta \text{minADE}_1$ values across both setups. Moreover, QCNet and EP variants exhibit significantly poorer generalization under \TWO{}. Notably, EP-Q's $\Delta \text{minADE}_1$ increases from  -\SI{0.022}{m} $(1.9\%)$ in \TA{} to +\SI{0.643}{m} $(78.3\%)$ in \TWO{}. This indicates that factors beyond our studied model design choices significantly influence performance.
    
    Drawing from our dataset analysis in Chapter \ref{cpt: Empirical Bayes Analysis} and experimental observations in Section \ref{sec 4.6.3: impact of data homogenization}, we identify three potential explanations:
    \begin{itemize}
        \item \textbf{Noise Level in Datasets}: As established in Chapter \ref{cpt: Empirical Bayes Analysis}, WO features lower noise levels than A2. Homogeneous augmentation and polynomial representation can be effective when training on noisy data (A2) and testing on clean data (WO). Conversely, when the test data exhibits higher noise levels than the training data, both methods have only a minimal improvement in model robustness.
        \item \textbf{Complexity of Prediction Task}: As discussed in Section \ref{sec 4.6.3: impact of data homogenization}, homogenization makes WO scenes less challenging by shifting complex behaviors into the historical window. Models trained on these simplified WO tasks struggle when tested on the more demanding A2 scenes.
        \item \textbf{Hyper-parameter Optimization}: All models use hyperparameters originally tuned for A2, which may be suboptimal for WO training, potentially limiting the benefits of increased training data.

    \end{itemize}
    Validating these hypotheses would require controlled experiments, such as systematic noise injection studies or comprehensive hyperparameter optimization across datasets. However, the dataset-related factors are inherently tied to data collection and processing pipelines, making it challenging to isolate and control variables for definitive conclusions. Additionally, exhaustive hyperparameter optimization for multiple models across datasets would require substantial computational resources beyond the scope of this work. We leave these important questions for future investigation.
\end{enumerate}
These results indicate that OoD robustness depends on a complex interplay of factors including model design choices (e.g., augmentation strategy and data representation), dataset characteristics (e.g., noise levels and task complexity), and experimental configurations. Improving generalization requires understanding these multifaceted relationships rather than simply increasing training data volume.


\begin{table}[!b]
\caption[Model efficiency analysis]{Model size and inference time across models. Inference time is tested on one Tesla T4 GPU with a traffic scene of 50 agents and 150 map elements. The best and second-best results for each metric across models are highlighted in \textbf{bold} and \underline{underline}, respectively.}
\centering
    \begin{tabularx}{\textwidth}{l | >{\centering\arraybackslash}X >{\centering\arraybackslash}X}
    \toprule
    \multirow{2}{*}{\shortstack{model}} & \# parameters& inference time\\
        & [M] $\downarrow$ & [ms] $\downarrow$ \\
        \midrule
        \multirow{2}{*}{QCNet \cite{zhou_query_2023}} &7.6& 120.89\\
        &\footnotesize (100.0\%)&\footnotesize (100.0\%)\\
        \cmidrule(lr){2-3}
        \multirow{2}{*}{Forecast-MAE \cite{cheng_forecast_2023}} & 1.9 &12.63\\
        &\footnotesize (25.0\%)&\footnotesize (10.4\%)\\
        \cmidrule(lr){2-3}
        \multirow{2}{*}{EP-Q (ours)} & \textbf{0.3} & \underline{5.72}\\
        &\footnotesize (\textbf{3.9\%})&\footnotesize (\underline{4.7\%})\\
        \cmidrule(lr){2-3}
        \multirow{2}{*}{EP-F (ours)}  & \textbf{0.3} & \textbf{4.66} \\ 
        &\footnotesize (\textbf{3.9\%})&\footnotesize (\textbf{3.9\%})\\
        \bottomrule
    \end{tabularx}
    \label{tab: model efficiency}
\end{table}

\subsection{Model Efficiency}
Beyond prediction accuracy, computational efficiency is essential for real-time autonomous driving applications. Table \ref{tab: model efficiency} compares model parameters and inference time across different architectures.

EP variants achieve significantly smaller model sizes compared to established benchmarks. For instance, EP-Q and EP-F require only $3.9\%$ of QCNet's parameters, demonstrating the efficiency of polynomial-based trajectory representations.

EP variants also exhibit substantially faster inference times. EP-Q processes predictions in just $4.7\%$ of QCNet's inference time and $45.3\%$ of FMAE's inference time, making it particularly well-suited for real-time autonomous driving applications.

These efficiency gains stem from the compact polynomial representations. The reduced representation dimensionality leads to fewer model parameters and, consequently, lower computational cost during inference.

\section{Conclusion and Discussion}
\label{sec 4.7: conclusion}
In this chapter, we emphasized the importance of evaluating traffic scene prediction models beyond In-Distribution (ID) performance, focusing on Out-of-Distribution (OoD) robustness through cross-dataset testing between Argoverse 2 and Waymo Open. We systematically compared the ID and OoD performance of our polynomial-based models against two SotA sequence-based models.

A key finding of this chapter is the significant disconnect between ID and OoD performance. Models that achieve superior ID performance do not necessarily generalize better to unseen datasets. For instance, while QCNet achieves the strongest ID performance, EP-Q surpasses it on the \TA{} OoD split (Figure \ref{fig: OoD_results}). This highlights a critical limitation in current evaluation practices that rely primarily on ID metrics and underscores the necessity of OoD evaluation for robust model assessment.


Our results further demonstrate that polynomial representations improve OoD robustness while substantially reducing model size. In the \TA{} setup, EP-Q achieves the strongest OoD performance among all models tested (Table \ref{tab: out-of-distribution result combined}, upper section), using much fewer parameters than the sequence-based benchmarks. In the reverse \TWO{} setup, EP-Q remains competitive with the strongest benchmark at the same parameter advantage (lower section). Across both directions, no sequence-based model matches this combination of robustness and efficiency.

At the same time, our findings reveal that robustness depends on factors beyond architectural and representational choices. Contrary to our initial hypothesis, models trained on the larger WO dataset did not generalize better to A2 than in the reverse direction, indicating that increasing training data volume alone does not ensure robustness. We do not attribute this asymmetry to a single cause: the two directions differ in prediction task difficulty, noise levels, and hyperparameter settings, any of which may contribute. We report it as an open observation and a direction for future investigation.

Beyond these factors, the input representation itself bounds what the models can capture. It relies on a lane-level map and past agent states, both encoded as polynomials -- compact and benchmark-consistent, but with limits. Arbitrarily shaped static structures, such as barriers or construction layouts, are not part of the model inputs. Furthermore, abstracting the scene to agents and lane geometry discards cues present in raw sensor data but tied to neither, such as turn signals or pedestrian pose, which can be informative for intent estimation. Addressing these, for instance through a dedicated occupancy representation or a learned sensor-level encoding, is left to future work.

\chapter{Diffusion-based Traffic Scene Generation and Prediction via Polynomials}
\label{cpt: generative model}
This chapter builds upon the author’s journal letter published in the \textit{IEEE Robotics and Automation Letters} (RA-L) 2025 \cite{yao_ep_2025}\footnote{\copyright~2025 IEEE. Reprinted, with permission, from Y. Yao, M.-K.\ Bouzidi, D. Goehring, and J. Reichardt, ``EP-Diffuser: An Efficient Diffusion Model for Traffic Scene Generation and Prediction via Polynomial Representations,'' \textit{IEEE Robotics and Automation Letters}, vol.~10, no.~9, pp.~9478--9485, 2025.}, co-authored with Mohamed-Khalil Bouzidi, Prof. Dr. Daniel Goehring (advisor), and Dr. Joerg Reichardt. The work has been significantly extended and refined for this chapter.

The author was responsible for the conceptual development and complete implementation of methods, as well as for all experiments and evaluations presented in the original paper and in this chapter. Mohamed-Khalil Bouzidi contributed to the training of the benchmark model OptTrajDiff \cite{wang_optimizing_2025}. Prof. Dr. Daniel Goehring and Dr. Joerg Reichardt contributed through scientific discussions and supervision.

The corresponding implementation is publicly available at: \url{https://github.com/aumovio/EP-Diffuser}.

\section{Introduction and Motivation}
\label{sec 5.1: introduction and motivation}

In Chapter \ref{cpt: improving generalization}, we focused on the marginal prediction task and demonstrated the improved model efficiency and Out-of-Distribution (OoD) generalization using polynomial representations while maintaining competitive In-Distribution (ID) performance. In this chapter, we extend our approach to address the more challenging \emph{joint} prediction task.

Compared to the marginal prediction that focuses on forecasting individual agent trajectories independently, joint prediction requires modeling the complex inter-dependencies between multiple agents sharing the same traffic environment (cf. Section \ref{sec 2.1.2: prediction competitions}). This fundamental shift from individual to collective behavior modeling introduces new challenges that go beyond those addressed in the previous chapter.

Public motion datasets, such as Argoverse 2 (A2) \cite{wilson_argoverse2_2021} and
Waymo Open (WO) \cite{ettinger_waymo_2021}, support the joint prediction task by annotating multiple target agents in each traffic scene and introduce scene-level evaluation metrics. The evolution of traffic scenes over long time horizons is governed by an inherently multi-modal probability distribution, as multiple plausible futures exist depending on road topology and agent interactions. However, motion datasets can only record a \emph{single} observed future sample per scene. Therefore, motion prediction competitions typically frame the problem as a regression problem, where models are trained to estimate the most likely outcome by minimizing the deviation from
future observations. As summarized in \cite{konstantinidis_marginal_2025}, many studies follow this regression setting, addressing joint prediction either by recombining independent marginal predictions \cite{gilles_thomas_2021, sun_m2i_2022, shi_mtr++_2023} or by training models with scene-level losses such as minSADE \cite{cheng_forecast_2023, zhou_qcnext_2023}. Both approaches focus on producing predictions that closely match the observed future and are typically evaluated using metrics such as minADE, minFDE, or their scene-level variants. 


\begin{figure}[thb]
\centering
\includegraphics[width=0.8\textwidth]{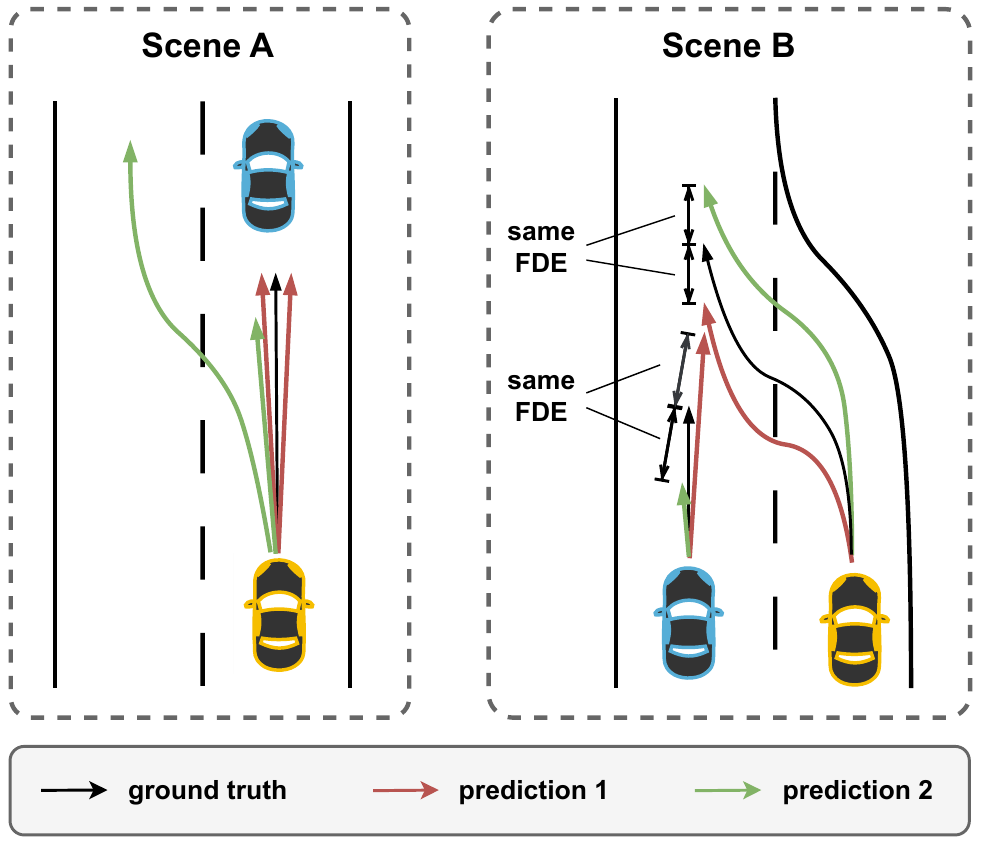}
\caption[Limitations of regression-based metrics]{Two limitations of regression-based metrics. \textbf{Scene A}: An example of \emph{multi-modal marginal} prediction. While Prediction 1 yields a lower Final Displacement Error (FDE), it only captures one possible behavior and ignores other plausible maneuvers. \textbf{Scene B}: An example of \emph{uni-modal joint} prediction. While both predictions yield the same FDE, Prediction 1 is less plausible due to the agent collision.}
\label{fig: regression metrics issue}
\end{figure}

However, this regression-based evaluation approach presents two key limitations illustrated in Figure \ref{fig: regression metrics issue}: First, it does not measure the diversity of predictions. Second, it fails to account for plausibility, both of which are essential for safe trajectory planning. Furthermore, focusing solely on the most probable future evolution while ignoring other plausible possibilities may induce overconfident and risky behavior, making it insufficient for planning algorithms, as highlighted in recent trajectory planning studies \cite{chen_interactive_2022, bouzidi_motion_2024, mustafa_racp_2024, bouzidi_closing_2025}.

Regression-based approaches have attempted to improve prediction diversity using discrete trajectory sets \cite{phan_covernet_2020} or modified training objectives \cite{xu_annealed_2024, lidard_nashformer_2023}. In parallel, other studies have explored \emph{generative} modeling frameworks. These include methods based on GANs \cite{gupta_social_2018, roy_vehicle_2019}, VAEs \cite{salzmann_trajectron_2020}, and diversity-enhancing sampling methods \cite{huang_diversitygan_2020, ma_diverse_2020}, which primarily aim to produce diverse marginal predictions for individual agents. 

Building on this trend, recent studies have begun employing generative approaches and shifting toward \emph{traffic scene generation} \cite{seff_motionlm_2023, wang_optimizing_2025, jiang_motiondiffuser_2023} -- the joint modeling of multiple interacting agents within a generative framework. Unlike regression-based joint prediction, which optimizes models to reproduce the most likely future, traffic scene generation reframes the task as learning the full distribution of plausible future scenes. Conditioned on environmental context (i.e., observed agent history and road layout), generative models enable direct sampling from the modeled joint distribution, capturing not only the most likely traffic scene evolution but also a rich set of plausible alternatives that reflect the inherent uncertainty of long-term, interactive driving scenarios. 

However, employing generative models for traffic scene generation faces several key challenges. First, evaluating generative models is non-trivial, since it involves assessing the learned distribution rather than the matching to a single observed traffic scene. Many recent generative studies choose to inherit evaluation metrics from regression-based approaches, despite fundamental differences in modeling objectives \cite{choi_dice_2024, wang_optimizing_2025, jiang_motiondiffuser_2023}. This risks biasing models towards a narrow subset of plausible futures.

Second, assessing model generalization remains challenging. Prior studies typically evaluate model performance using a test split from the same dataset used for training. As highlighted in Chapter \ref{cpt: improving generalization}, this setup fails to reflect true generalization capability. Moreover, generative models are known to exhibit memorization -- the ability to produce near-replicas of training data \cite{brown_language_2020, somepalli_understanding_2023}. Consequently, models tested only on in-distribution samples may appear to perform well by leveraging their memorizing capability instead of learning robust, transferable representations of traffic patterns. 

Finally, computational efficiency presents a significant practical challenge. Recent generative models are computationally expensive \cite{seff_motionlm_2023, jiang_motiondiffuser_2023} and potentially unsuitable for real-time applications in autonomous vehicles where prediction latency is critical.

Given these challenges, we investigate how polynomial representations can enhance traffic scene generation. Specifically, we extend the EP-Q model from Chapter \ref{cpt: improving generalization} to incorporate the diffusion paradigm, conditioned on road layout and observed agent history. This design exploits the efficiency and generalization advantages of polynomial representations to model the inherently multi-modal distribution of traffic scenes.

Our evaluation framework moves beyond traditional regression-based metrics to provide a comprehensive assessment across three key dimensions: plausibility, diversity, and accuracy. We benchmark our approach against two state-of-the-art models, incorporating Waymo's ``Sim Agents'' metrics \cite{montali_waymo_2024} to evaluate plausibility. Building on the efficiency and OoD generalization advantages that polynomial representations demonstrated in Chapter \ref{cpt: improving generalization}, we examine whether these benefits extend to the more complex domain of traffic scene generation.

Beyond in-distribution evaluation, we systematically assess how the model generalizes to unseen domains. We adopt the homogenization protocol (cf. Section \ref{sec 4.3: data homogenization}) with minor adaptations and the experimental setup \TA{} from Section \ref{sec 4.6.1: experimental setup}. In this setup, models are trained on the A2 dataset and tested on OoD samples from WO. This choice is motivated by our findings in Chapter \ref{cpt: improving generalization}, where the \TA{} setup demonstrated clearer differentiation between model design choices, particularly showing the benefits of polynomial representations and augmentation strategies. The alternative \TWO{} setup (training on WO, OoD testing on A2), while valuable, revealed that model robustness depends on complex interactions between dataset characteristics and model design that extend beyond architectural choices alone, making it less suitable for isolating the impact of specific design decisions. Consequently, we select benchmark models originally developed for the A2 dataset to ensure fair comparison in this chapter.

This chapter is organized as follows: Section \ref{sec 5.2: related work} reviews recent traffic scene (joint) prediction and generation models, along with their evaluation metrics. From these models, we introduce two benchmark models. We further introduce the ``Sim Agents'' metrics, highlighting their key differences for evaluating traffic scene plausibility from regression-based metrics. Section \ref{sec 5.3: model design} presents our diffusion-based approach with polynomial representations. Section \ref{sec 5.4: experiments} evaluates the proposed model against benchmarks, analyzing plausibility, diversity, and regression-based accuracy. Furthermore, we extend our evaluation to OoD scenes to assess generalization beyond the training domain. Finally, Section \ref{sec 5.5: conclusion} concludes with a summary of findings and their implications.

\section{Related Work and Preliminaries}
\label{sec 5.2: related work}

\subsection{Traffic Scene Prediction and Benchmark Models}

Benchmark datasets and associated prediction competitions have significantly shaped research in traffic scene prediction by framing it as a regression task, evaluating the most likely predicted traffic scenes. Recent studies have followed this competition framework and implemented regression-based deep learning models for traffic scene prediction  \cite{cheng_forecast_2023, zhou_qcnext_2023, luo_jfp_2023}. Although these models can output multiple modes for potential traffic scenes, they are primarily scored and ranked using regression-based metrics such as minADE, minFDE, and their variants for scene-level evaluation. These metrics are tailored to multi-modal predictors and measure the minimum displacement error among all predicted modes.

Regression-based approaches have significantly influenced the development of generative models in the field \cite{seff_motionlm_2023, jiang_motiondiffuser_2023, wang_optimizing_2025}. Notably, many diffusion-based models have incorporated regression model backbones to output initial predictions that closely align with the observed future \cite{mao_leapfrog_2023, wang_optimizing_2025}. Although this design choice effectively optimizes for competition results, it may not fully capture the inherent uncertainty of real-world traffic.

\begin{table}[!t]
\caption{Summary of models under study}
\centering
    \begin{tabularx}{\columnwidth}{c| *{3}{>{\centering\arraybackslash}X} }
    \toprule
    model & FMAE-MA \cite{cheng_forecast_2023}& OptTrajDiff \cite{wang_optimizing_2025} & EP-Diffuser (ours)\\
    \midrule
    input \& output & \multirow{2}{*}{sequence} & \multirow{2}{*}{sequence} & \multirow{2}{*}{polynomial} \\
    representation & && \\
     \midrule
    model type & regression & diffusion & diffusion \\
    \midrule
    \# output samples & 6& inf & inf \\
    \midrule
    \# model parameters &   \multirow{2}{*}{1.9} & \multirow{2}{*}{12.5} & \multirow{2}{*}{3.0} \\
    $[$million$]$ & &  & \\
    \bottomrule
    \end{tabularx}
    \label{tab: model summary}
    \vspace{-10pt}
\end{table}

As representatives of the two model classes, we select two recently open-sourced and well-documented SotA models as benchmarks: Forecast-MAE-multiagent (FMAE-MA) \cite{cheng_forecast_2023} and OptTrajDiff \cite{wang_optimizing_2025}. As summarized in Table \ref{tab: model summary}, both models use sequence-based representations but follow different methodological approaches. FMAE-MA extends the benchmark model FMAE from Chapter \ref{cpt: improving generalization} to the joint prediction task. It follows a regression-based approach and predicts 6 distinct modes of future traffic scenes, with a relatively lightweight architecture of $1.9$ million parameters. OptTrajDiff, by contrast, applies a diffusion-based framework and integrates the QCNet, introduced in Chapter \ref{cpt: improving generalization}, as its regression backbone \cite{zhou_query_2023}, resulting in a larger architecture with $12.5$ million parameters.

\subsection{``Sim Agents'' Metrics}
\label{sec: sim agents metrics}
In contrast to regression-based tasks, ``Sim Agents'' frames traffic scene prediction as a multi-agent generative task, emphasizing the importance of capturing the diversity and plausibility of traffic behaviors \cite{montali_waymo_2024}. Rather than focusing solely on minimizing displacement errors, ``Sim Agents'' evaluates a model by comparing the distribution of its 32 generated samples against the observed future. This comparison is expressed through a set of metrics, each a normalized score in $[0,1]$ (1 is best) measuring how well the generated distribution matches the real one along a particular aspect of driving behavior. The metrics are organized into three groups and an aggregate, as summarized in Table \ref{tab: sim agent metrics}:
\begin{itemize}[leftmargin=*]
    \item \textbf{Agent Kinematic Metrics}: Evaluate individual motion properties such as linear speed, linear acceleration, angular speed, and angular acceleration, assessing adherence to realistic dynamics.

    \item \textbf{Agent Interaction Metrics}: Quantify social behavior through collision ratio, time to collision (TTC), and distance to nearest agent, indicating the plausibility of interactions.

    \item \textbf{Map Adherence Metrics}: Evaluate whether trajectories conform to road layouts, using off-road ratio and distance to road edges.

    \item \textbf{Realism Meta Metric}: A weighted average of the above component scores, forming a single holistic measure of scene plausibility.
\end{itemize}
This encourages models to replicate the variability and interaction patterns observed in actual traffic scenes.

\begin{table}[b]
\caption{Structure of the ``Sim Agents'' metrics for plausibility evaluation}
\centering
    \begin{tabularx}{\columnwidth}{c *{2}{>{\centering\arraybackslash}X} }
    \toprule
    meta group & metric group  & metric \\
    \midrule
    \multirow{10}{*}{realism meta} & \multirow{4}{*}{agent kinematic} & linear speed \\
    \hhline{~~-}
    & & linear acceleration \\
    \hhline{~~-}
    & & angular speed \\
    \hhline{~~-}
    & & angular acceleration \\
    \hhline{~==}
    & \multirow{3}{*}{agent interaction} & collision \\
    \hhline{~~-}
    & & TTC \\
    \hhline{~~-}
    & & distance to agent \\
    \hhline{~==}
    & \multirow{2}{*}{map adherence} & offroad \\
    \hhline{~~-}
    & & distance to road edge\\
    \bottomrule
    \end{tabularx}
    \label{tab: sim agent metrics}
    \vspace{-10pt}
\end{table}

The ``Sim Agents'' framework offers several advantages over regression-based metrics. First, it evaluates the plausibility of predicted traffic scenes using multiple high-level heuristics rather than emphasizing positional accuracy alone. Second, it reduces target agent selection bias. While agent kinematic and map adherence metrics are computed for the designated target agent, achieving high agent interaction scores requires consistency of predictions across all agents in the scene. For example, if a moving non-target agent collides with a parked target agent, the interaction score will be low -- even if the parked target agent itself is predicted with zero positional error.

In addition to these plausibility measures, the ``Sim Agents'' benchmark also reports $\text{minSADE}_{32}$ (cf. Section \ref{sec: evaluation metrics}) as an accuracy metric. However, this value is not emphasized in the official scoring and serves only as a supplementary reference. We refer to it as $\textbf{minSADE}$ in the remainder of this chapter for clarity.

In this chapter, we use the \emph{realism meta} metric as the primary measure of predicted scene plausibility.

\section{Data Representation and Model Design}
\label{sec 5.3: model design}
Our findings in Chapter \ref{cpt: improving generalization} show that combining polynomial representations with homogeneous augmentation yields the strongest generalization. Building on this, we extend the marginal prediction model EP-Q by employing these two design choices within a generative diffusion framework for multi-agent traffic scene generation. Unlike prior diffusion models that rely on sequence-based representations \cite{wang_optimizing_2025, mao_leapfrog_2023, choi_dice_2024}, our approach employs polynomial representations for both map elements and trajectories. We refer to this model as the \emph{Everything Polynomial Diffuser} (\emph{EP-Diffuser}). 

\begin{figure*}[tbh]
\centering
\includegraphics[width=\textwidth]{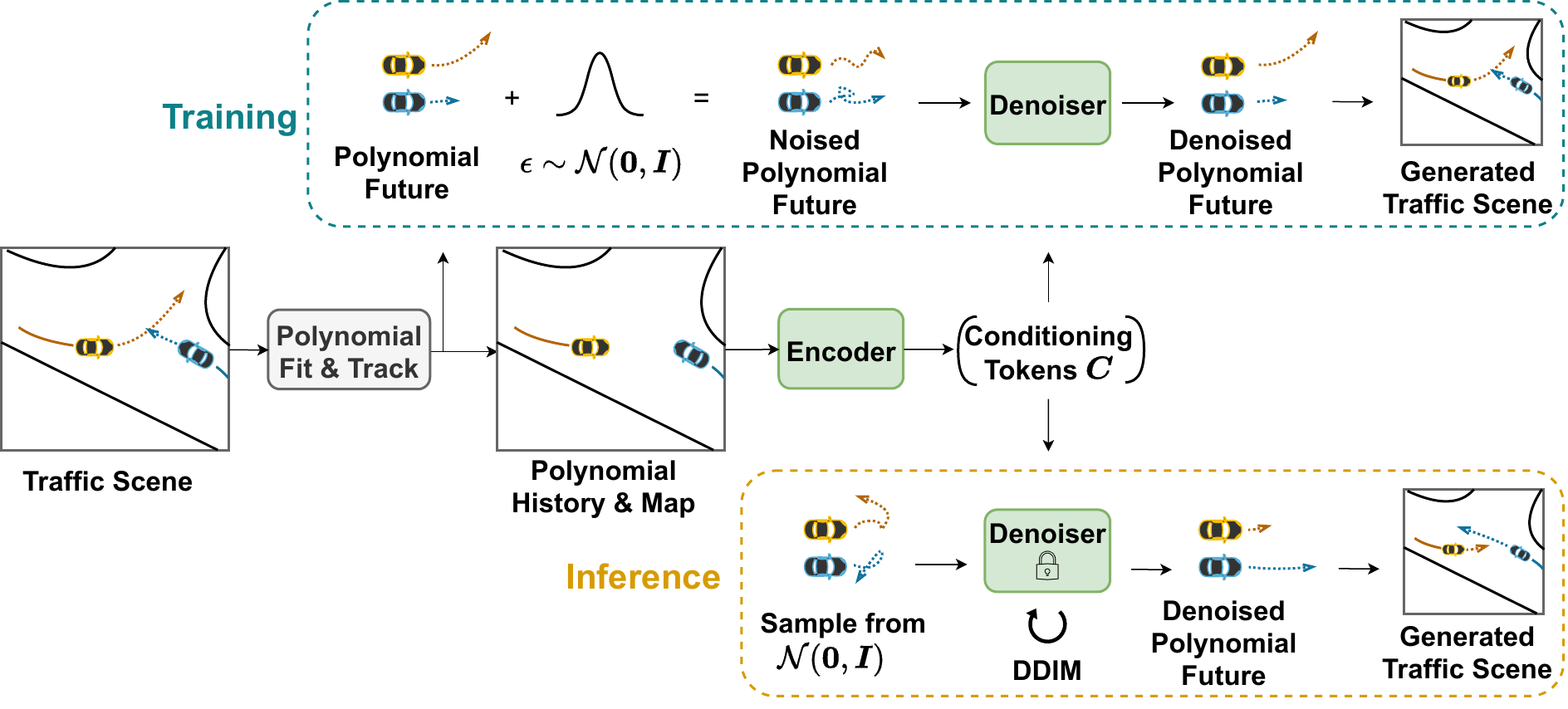}
\caption[Overview of EP-Diffuser for traffic scene generation]{Overview of EP-Diffuser for traffic scene generation. For clear presentation, we adopt a similar visual layout to \cite{jiang_motiondiffuser_2023}, while the technical implementation extends the EP-Q model from Chapter \ref{cpt: improving generalization}. The traffic scene comprised of agent history and map elements as degree $5$ and $3$ polynomials, respectively, is encoded via the encoder into a set of conditioning tokens $\boldsymbol{C}$. The observed future trajectories are represented as polynomials of degree $6$. During \textbf{training}, a random set of noise is sampled i.i.d. from a standard normal distribution and added to the parameters of the observed future trajectory. The denoiser, while attending to the conditioning tokens, jointly predicts the denoised polynomial parameters of trajectories corresponding to each agent. During \textbf{inference}, a set of trajectory parameters for each agent is initially sampled from a standard normal distribution, and iteratively denoised using a DDIM schedule \cite{song_denoising_2021} to produce plausible future trajectories.}
\label{fig: diffusion  pipeline}
\end{figure*}

This polynomial design provides three key benefits within diffusion frameworks. First, it provides a low-dimensional, truncated Taylor-like approximation that captures key motion patterns without modeling complex high-order residuals. This improves the efficiency of the diffusion-denoising process and simplifies the overall learning task. Second, since polynomial basis functions possess well-defined derivatives, training with positional loss implicitly influences kinematic terms such as velocity and acceleration. This yields smooth and physically plausible trajectories when appropriately regularized with moderate polynomial degrees. Third, it enhances model generalization by mitigating dataset-specific bias and noise commonly present in sequence-based representations, as highlighted in Chapter \ref{cpt: improving generalization}.

The EP-Diffuser pipeline, illustrated in Figure \ref{fig: diffusion pipeline}, follows an encoder–denoiser architecture. The encoder embeds the environmental context -- comprising agent histories and map elements -- into a set of conditioning tokens $\boldsymbol{C}$. The denoiser learns to predict noise from trajectory samples conditioned on $\boldsymbol{C}$. At training time, it operates on noised future trajectories; at inference time, it iteratively denoises samples starting from a prior Gaussian distribution to generate future trajectories.


In the following, we first describe our approach for representing diverse data types using polynomial representations, followed by the implementation details of EP-Diffuser.

\subsection{Data as Polynomials}
We employ the same approaches from Section \ref{sec 4.4: represent data with polynomials} to represent 5-second agent histories (5-degree) and map elements (3-degree) with Bernstein polynomials. 

For training the EP-Diffuser, 6-second agent futures are also represented as polynomials before being noised. They are represented by 6-degree Bernstein polynomials -- one degree higher than recommended by AIC in Chapter \ref{cpt: Empirical Bayes Analysis} (cf. Section \ref{sec: optimal polynomial degree}) -- to capture more complex motion patterns. Since WO is used only for testing in this chapter, we do not fit polynomials to its 4.1-second futures after homogenization.

To transform agent future trajectories from the sequence-based format into Bernstein polynomials, a natural approach would be to apply the Kalman filter method from Section \ref{sec 4.4.1: Agent History as Polynomial}, as done for agent histories. However, the Kalman filter inherently assigns higher weight to recent observations. As a result, the tracked polynomials align well with later observations but show worse alignment with initial observations, creating a mismatch between agent history and future. The RTS smoother introduced in Section \ref{sec 3.5.2: tracking issues} could alleviate this by improving alignment with initial observations, but its two-pass process introduces computational overhead by requiring storage of past estimated states.

To address these limitations, we employ Bayesian regression (Section \ref{sec 3.3: Bayesian regression}) to fit future trajectories while including the observed position at $t_0$, following the priors and observation noise models established in Chapter \ref{cpt: Empirical Bayes Analysis}. To mitigate the boundary effect of Bayesian regression (cf. Section \ref{sec 4.4.1: Agent History as Polynomial}) and ensure stronger consistency with history, we explicitly reduce the observation noise covariance at $t_0$ by a factor of 9.

\subsection{Traffic Scene Graph}
\label{sec 5.3.2: traffic scene graph}
EP-Q employs the homogeneous augmentation and computes the environmental information in each agent's individual coordinate. As discussed in Section \ref{sec 4.4: represent data with polynomials}, this approach transforms traffic scenes into fixed-size data tensors for batch processing. For example, EP-Q considers up to 50 agents and 150 map elements closest to the target agent in each traffic scene. This design is effective in the marginal prediction setting from the previous chapter, where a single target agent is defined and surrounding agents and map elements can be selectively included. It provides a practical balance between prediction performance and training efficiency.

In contrast, in the traffic scene generation setting -- where all agents must be modeled jointly -- this fixed-size approach can omit important agents and map elements in complex scenes, while also introducing computational overhead from padded entries in simpler cases.

\begin{figure}[!thb]
\centering
\includegraphics[width=\textwidth]{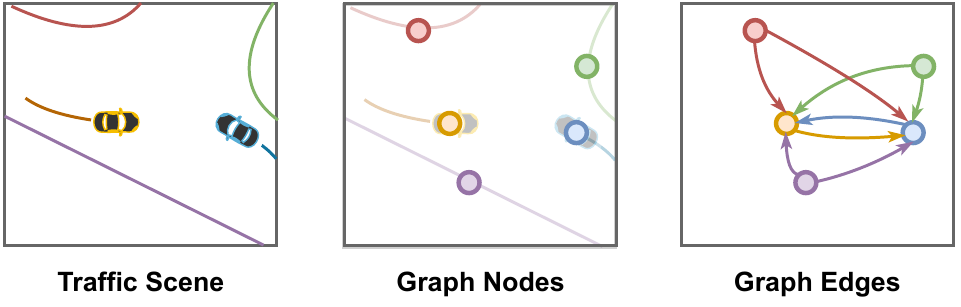}
\caption[Traffic scene graph]{An example of a traffic scene graph. \textbf{Middle}: agents and map elements as nodes. \textbf{Right}: interactions (e.g., relative positions and motions) as directed edges. Edges between map elements are omitted for clarity. Directed edges describe the spatial relationship from a \emph{source} node with respect to a \emph{target} node.}
\label{fig: traffic scene graph}
\end{figure}

To address this issue while retaining the generalization benefits of homogeneous augmentation, we adopt the \emph{graph}-based approach \cite{fey_fast_2019} for structuring traffic scenes in EP-Diffuser, following recent studies \cite{liang_learning_2020, cui_gorela_2022, zhou_query_2023}. Graph-based structures naturally model the relational structure of traffic scenes by treating agents and map elements as \emph{nodes} and their interactions (e.g., relative positions and motions) as \emph{edges}, as visualized in Figure \ref{fig: traffic scene graph}. In this formulation, each traffic scene corresponds to an individual graph, allowing the model to flexibly accommodate varying numbers of agents and map elements without resorting to truncation or padding. 

Both nodes and edges include features of traffic scenes. The node features characterize the intrinsic properties of each element in the traffic scene, such as agent kinematics, map geometry, and their type, while remaining independent of spatial relationships to other nodes. In other words, they describe what an individual agent or map element is, but not how it interacts with or is positioned relative to its surroundings.

The edge features, in contrast, describe the relationships between nodes. They encode spatial relationships, such as relative position, distance, or heading between nodes. In this way, edge features provide the model with information about how elements in the scene influence one another, complementing the self-contained descriptions provided by node features.

In EP-Diffuser, we use the following node and edge features:
\begin{itemize}[leftmargin=*]
    \item \textbf{Node features of conditioning tokens}: Environmental context used by the encoder for encoding the conditioning tokens $\boldsymbol{C}$.

    \item \textbf{Node features of future trajectories}: Features of the (noised) future trajectories, serving as input to the denoiser.

    \item \textbf{Edge features}: Relative position and heading derived from the nodes’ reference positions and headings.
\end{itemize}
In the following, we detail each of these features.

\subsubsection{Node Features of Conditioning Tokens}
The node features of conditioning tokens are used in the encoder of EP-Diffuser and include information from both agents and map elements. They are similar to the inputs of EP in Section \ref{sec 4.5.1: model inputs}, and we follow the same notations for consistency. 

Unlike EP-Q, which transforms the entire traffic scene into each agent's individual coordinate frame -- requiring an additional data dimension for individual agent coordinates -- EP-Diffuser transforms each agent's history control points into that agent's local coordinate frame, then computes the history control point vectors $\boldsymbol{\Delta}^\text{a,hist}$. This local frame is defined by the agent’s observed position and heading at the reference timestep $t_0$ as the origin and heading. This design eliminates the need for an additional data dimension while preserving the agent-specific kinematic description. 

For $A$ agents in the traffic scene, the node features consist of:
\begin{itemize}[leftmargin=*]
    \item \textbf{History control point vectors:} $\boldsymbol{\Delta}^{\text{a,hist}} = \text{concat}(\boldsymbol{\Delta}^{\text{a,hist}}_{1}, \boldsymbol{\Delta}^{\text{a,hist}}_{2}, \dots, \boldsymbol{\Delta}^{\text{a,hist}}_{5}) \in \mathbb{R}^{A\times10}$, where each $\boldsymbol{\Delta}^{\text{a,hist}}_{n}$ represents vectors between consecutive control points (cf. Section \ref{sec 4.5.1: model inputs}),
    \item \textbf{Time window:} $\boldsymbol{TW} \in \mathbb{R}^{A \times 2}$, indicating the observed historical period,
    \item \textbf{Agent type:} $\boldsymbol{AT} \in \mathbb{Z}^A$, where each entry is an integer label indicating the agent type  (e.g., vehicle, pedestrian, cyclist).
\end{itemize}

For map elements, control points of centerlines are transformed into map elements' local coordinate frames before computing the control point vectors $\boldsymbol{\Delta}^\text{m}$, similar to the agent history processing. Following QCNet \cite{zhou_query_2023}, the local frame is defined using centerline midpoint's position and heading.


For $M$ map elements in the traffic scene, the node features are:
\begin{itemize}[leftmargin=*]
    \item \textbf{Map control point vectors:} $\boldsymbol{\Delta}^{\text{m}} = \text{concat}(\boldsymbol{\Delta}^{\text{m}}_{1}, \boldsymbol{\Delta}^{\text{m}}_{2}, \boldsymbol{\Delta}^{\text{m}}_{3}) \in \mathbb{R}^{M\times6}$, where each $\boldsymbol{\Delta}^{\text{m}}_{n} \in \mathbb{R}^{M\times2}$ represents vectors between consecutive control points (cf. Section \ref{sec 4.5.1: model inputs}),
    \item \textbf{Map element type:} $\boldsymbol{MT} \in \mathbb{Z}^M$,  where each entry is an integer label indicating the map element type (e.g., vehicle lane, bike lane, crosswalk).
\end{itemize}

In EP-Diffuser's graph-based traffic scene structure, positional information $\boldsymbol{PI}$ (cf. Section \ref{sec 4.5.1: model inputs}) is not included in the node features but is instead handled through edge features, which capture spatial relationships between elements in the traffic scene.

\subsubsection{Node Features of Future Trajectories}
In addition to the conditioning tokens, the denoiser of EP-Diffuser requires the control points of future trajectories as input during training for the reverse denoising process. Before being input to the denoiser, these future trajectories are first perturbed with Gaussian noise through the forward diffusion process (cf. Section \ref{sec 5.3.3: diffusion and denoising}). The node features described below correspond to the clean future trajectories before noise is applied.

The future trajectories of all $A$ agents in the traffic scene are represented using 6-degree polynomials. The control points of these trajectories are denoted as $\textbf{W}^{\text{a,fut}}_{n}\in \mathbb{R}^{A\times2}$ with $n =0,1,\dots,6$. Similar to the agent history processing, these control points are first transformed into each agent's local coordinate frame at the reference timestep $t_0$. From these, the control point vectors are computed as $\boldsymbol{\Delta}^{\text{a,fut}}_{n} = \textbf{W}^{\text{a,fut}}_{n} - \textbf{W}^{\text{a,fut}}_{n-1}$, which capture the kinematic properties of agents' future trajectories. 

For $A$ agents in the traffic scene, the node features of future trajectories consist of:
\begin{itemize}[leftmargin=*]
    \item \textbf{Future control point vectors:} $\boldsymbol{\Delta}^{\text{a,fut}} = \text{concat}(\boldsymbol{\Delta}^{\text{a,fut}}_{1}, \boldsymbol{\Delta}^{\text{a,fut}}_{2}, \dots, \boldsymbol{\Delta}^{\text{a,fut}}_{6}) \in \mathbb{R}^{A\times12}$,
    \item \textbf{Diffusion step:} $\boldsymbol{s} \in \mathbb{R}^{A}$, indicating the diffusion step (noise level) applied to each agent during the forward diffusion process.
\end{itemize}

\subsubsection{Edge Features}
In EP-Diffuser, edge features capture the spatial relationships between nodes, complementing the intrinsic properties captured by node features. Positional information used in Section \ref{sec 4.5.1: model inputs} is intentionally excluded from the node features and instead represented through directed edges. 

Each directed edge describes the spatial relationship from a \emph{source} node with respect to a \emph{target} node at the reference timestep $t_0$ (cf. Figure \ref{fig: traffic scene graph}). Within the attention mechanism framework, source nodes provide features as keys and values, while target nodes provide features as queries. This enables each entity to selectively attend to relevant neighbors based on the edge relationships.

\begin{figure}[!thb]
\centering
\includegraphics[width=0.5\textwidth]{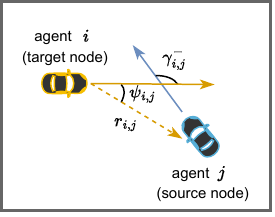}
\caption[Edge features]{The edge features between source and target nodes at the reference timestep, illustrated here for two agents but applied universally to all node pairs (agents and map elements).}
\label{fig: edge features}
\end{figure}

A straightforward way to define edge features is to express node relationships in each other’s Cartesian frames, as done in EP-Q. However, this involves multiple coordinate transformations including translations and rotations. To reduce complexity, we instead adopt the approaches used in studies \cite{zhou_query_2023, cui_gorela_2022} and represent the spatial relationships in polar frames. As visualized in Figure \ref{fig: edge features}, the edge feature for a source node $j$ relative to a target node $i$ is expressed as the triplet $(r_{i,j}, \psi_{i,j}, \gamma_{i,j}^{-})$, where $r_{i,j}$ denotes the radial distance, $\psi_{i,j}$ the relative angle, and $\gamma_{i,j}^{-}$ the heading difference. These quantities can be computed directly without additional translations and rotations.

Following QCNet \cite{zhou_query_2023}, we construct directed edges from each node to all other nodes within a radius of \SI{150}{m}, thereby capturing long-range interactions.

For $P$ directed edges, let $\boldsymbol{r} \in \mathbb{R}^{P}$, $\boldsymbol{\psi} \in \mathbb{R}^{P}$, and $\boldsymbol{\gamma}^{-} \in \mathbb{R}^{P}$ denote the sets of radial distances, relative angles, and heading differences, respectively. The edge features are summarized as:
\begin{itemize}[leftmargin=*]
    \item \textbf{Spatial relationship:} $\boldsymbol{SR} = \text{concat}(\boldsymbol{r}, \cos{(\boldsymbol{\psi})},\sin{(\boldsymbol{\psi})}, \cos{(\boldsymbol{\gamma^{-}})}, \sin{(\boldsymbol{\gamma^{-}})}) \in \mathbb{R}^{P\times5}$, where trigonometric functions transform the angular values into normalized Cartesian components.
\end{itemize}

\subsection{Diffusion and Denoising}
\label{sec 5.3.3: diffusion and denoising}
Following the practice of Denoising Diffusion Probabilistic Models (DDPM) \cite{ho_denoising_2020} (cf. Section \ref{sec: generative model introduction}), we perform diffusion-denoising on the agents' future control point vectors introduced in Section \ref{sec 5.3.2: traffic scene graph}.

For a \textbf{single} agent $i$ at diffusion step $s$, we denote the (possibly noised) future control point vectors as $\boldsymbol{\delta}^{\text{a,fut}}_{s,i} \in \mathbb{R}^{12}$, where $s=0$ corresponds to the future control point vectors without added noise. Following Equation \ref{eqn: forward diffusion}, we can write $\boldsymbol{\delta}^{\text{a,fut}}_{s,i}$ at an arbitrary diffusion step as a linear combination of the clean future control point vectors $\boldsymbol{\delta}^{\text{a,fut}}_{0,i}$ and a Gaussian noise $\boldsymbol{\epsilon}_i$:
\begin{equation}
\begin{aligned}
\label{eqn: diffusion linear}
\boldsymbol{\delta}^{\text{a,fut}}_{s,i} &= \sqrt{\Bar{\alpha}_s}\boldsymbol{\delta}^{\text{a,fut}}_{0,i} + \sqrt{1-\Bar{\alpha}_s}\boldsymbol{\epsilon}_i, \quad \text{where } \boldsymbol{\epsilon}_i \sim \mathcal{N}(\mathbf{0}, \mathbf{I}).
\end{aligned}
\end{equation}
\noindent Here, $\Bar{\alpha}_{s}$ is the pre-defined noise-scheduling parameter at diffusion step $s$ that controls the diffusion process.

Aggregating over all $A$ agents under a common diffusion step $s$ yields:
\begin{equation}
    \boldsymbol{\Delta}^{\text{a,fut}}_s = \{\boldsymbol{\delta}^{\text{a,fut}}_{s,i}\}_{i=1}^A \in \mathbb{R}^{A \times 12}.
\end{equation}
Note that the diffusion step index $s$ is distinct from the control point index $n$ defined in Section \ref{sec 5.3.2: traffic scene graph}.

Unlike other diffusion models that apply a uniform diffusion
step across all agents in a scene \cite{wang_optimizing_2025}, EP-Diffuser allows each agent in a traffic scene to have its own diffusion step $s_i$ during training, denoted as 
\begin{equation}
    \boldsymbol{\Delta}^{\text{a,fut}}_{\boldsymbol{s}} = \{\boldsymbol{\delta}^{\text{a,fut}}_{s_i,i}\}_{i=1}^A 
, \quad \text{with } \boldsymbol{s} = [s_1, \dots, s_A].
\end{equation}
Empirically, we observe that this provides modest performance improvements over uniform diffusion steps, although a systematic investigation of this effect is left for future work.

For the forward diffusion process, we apply a total of $S=1000$ diffusion steps, as in the original DDPM \cite{ho_denoising_2020}, to gradually transition from the data distribution $q(\boldsymbol{\delta}^{\text{a,fut}}_{0,i})$ to the target prior distribution $\mathcal{N}(\mathbf{0}, \mathbf{I})$. 

\begin{figure}[!thb]
\centering
\includegraphics[width=1.\textwidth]{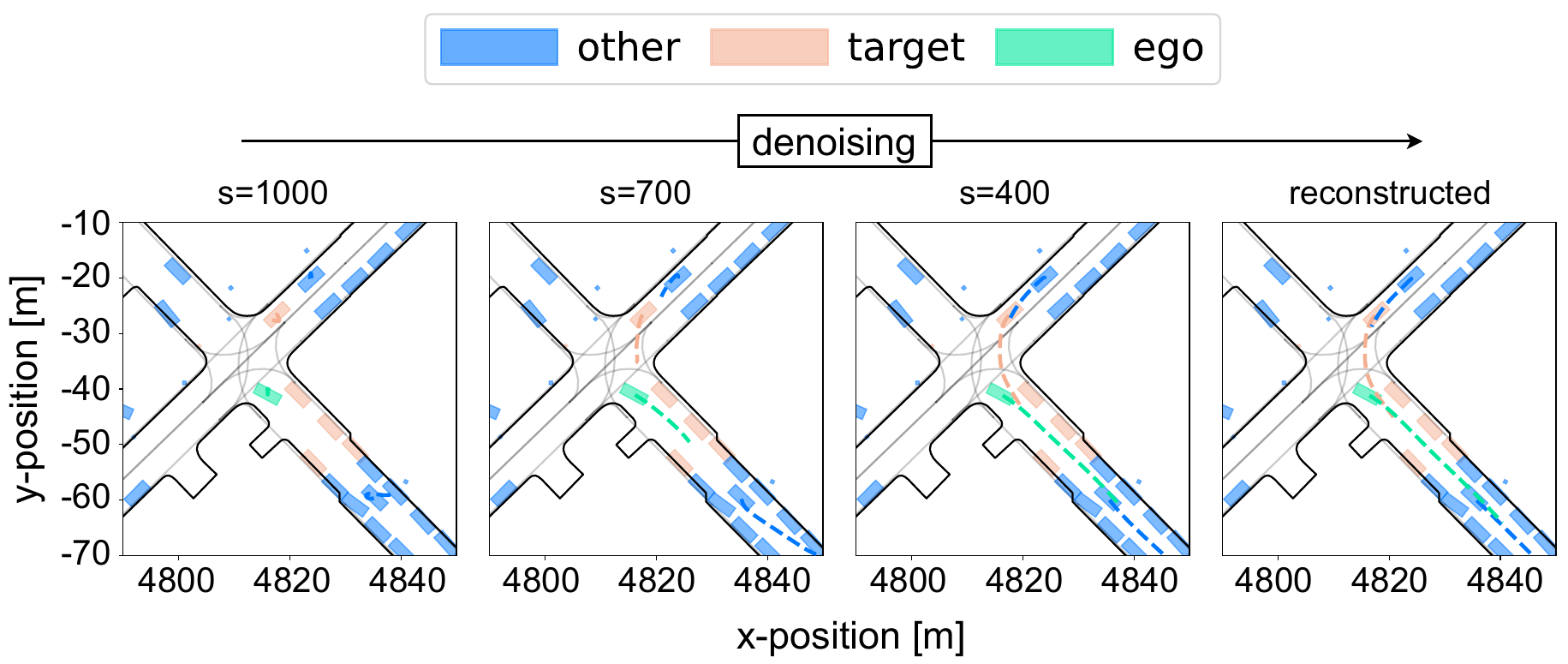}
\caption[Denoising process of EP-Diffuser]{Illustration of the EP-Diffuser denoising process on Argoverse 2. \textbf{Dashed lines} represent the generated trajectories of selected highly interactive agents at different denoising steps.}
\label{fig: denoising process}
\end{figure}

For the reverse denoising process, we adopt Denoising Diffusion Implicit Models (DDIM) \cite{song_denoising_2021} with 10 denoising steps -- consistent with the configuration of OptTrajDiff \cite{wang_optimizing_2025}. This mitigates variability in sampling procedures and accelerates the inference process. Following Algorithm \ref{alg: ddim-sampling} in Section \ref{sec: generative model introduction}, the future trajectories of agents are iteratively denoised by predicting the added noise $\hat{\boldsymbol{\epsilon}}_i$ for each agent and subtracting $\hat{\boldsymbol{\epsilon}}_i$ from $\boldsymbol{\delta}^{\text{a,fut}}_{s,i}$ at each step. Figure \ref{fig: denoising process} visualizes how the traffic scene evolves during the denoising process.

\subsection{Encoder}
\label{sec 5.3.4: encoder}
In this section, we describe how the encoder of EP-Diffuser processes the node and edge features defined in Section \ref{sec 5.3.2: traffic scene graph}.

\begin{figure}[!thb]
\centering
\includegraphics[width=0.7\textwidth]{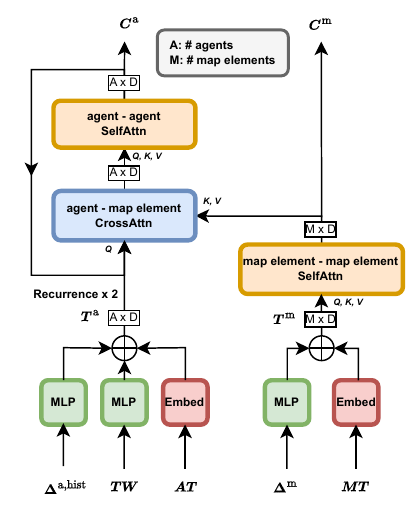}
\caption{Encoder architecture of EP-Diffuser}
\label{fig: ep-diffuser encoder}
\end{figure}

EP-Diffuser adopts the encoder architecture of the EP-Q model and adapts it to the graph-based structure of traffic scenes. As visualized in Figure \ref{fig: ep-diffuser encoder}, node features of conditioning tokens are encoded in the same way as in EP-Q. Numerical features, such as control point vectors $\boldsymbol{\Delta}^{\text{a,hist}}$ and $\boldsymbol{\Delta}^{\text{m}}$, are processed with simple 3-layer MLPs. Categorical attributes, such as $\boldsymbol{AT}$ and $\boldsymbol{MT}$, are encoded via individual embedding layers. Agent features are summarized into agent tokens $\boldsymbol{T}^{\text{a}}$, while map features are summarized into map tokens $\boldsymbol{T}^{\text{m}}$.

Edge features $\boldsymbol{SR}$, which describe map–map, agent–map, and agent–agent relations, are encoded using individual 3-layer MLPs. These encoded edge features are incorporated into the attention computation (cf.  Equation \ref{eqn: attention}) within the corresponding attention blocks.

After multiple attention blocks that model the interactions between agents and map elements, the encoder outputs two types of conditioning tokens: (\lowerromannumeral{1}) agent conditioning tokens $\boldsymbol{C}^{\text{a}} \in \mathbb{R}^{A\times D}$, and (\lowerromannumeral{2}) map conditioning tokens $\boldsymbol{C}^{\text{m}} \in \mathbb{R}^{M\times D}$, where $D$ is the hidden dimension.

\subsection{Denoiser}
\begin{figure}[!thb]
\centering
\includegraphics[width=0.7\textwidth]{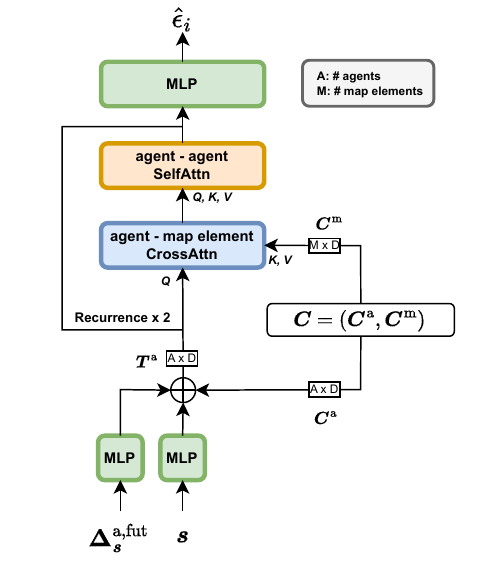}
\caption{Denoiser architecture of EP-Diffuser}
\label{fig: ep-diffuser denoiser}
\end{figure}

While the encoder extracts contextual conditioning tokens from the traffic scene, the denoiser uses them to refine the noised future trajectories. Figure \ref{fig: ep-diffuser denoiser} visualizes the denoiser architecture of EP-Diffuser. The denoiser takes as input the noised future control point vectors $\boldsymbol{\Delta}^{\text{a,fut}}_{\boldsymbol{s}} \in \mathbb{R}^{A \times 12}$, together with the diffusion step indices for each agent $\boldsymbol{s} \in \mathbb{R}^{A}$. These inputs are encoded with 3-layer MLPs and combined with the agent conditioning tokens $\boldsymbol{C}^{\text{a}}$ to form the agent tokens $\boldsymbol{T}^{\text{a}} \in \mathbb{R}^{A \times D}$ within the denoiser. 

Similar to the encoder, multiple attention blocks model the agent-map and agent-agent interactions, incorporating edge features $\boldsymbol{SR}$ to sequentially update $\boldsymbol{T}^{\text{a}}$. In particular, to integrate map information during denoising, the denoiser performs cross attention between agent tokens $\boldsymbol{T}^{\text{a}}$ and map conditioning tokens $\boldsymbol{C}^{\text{m}}$. Finally, the updated agent tokens $\boldsymbol{T}^{\text{a}}$ are passed through a 3-layer MLP to predict the added noise $\hat{\boldsymbol{\epsilon}}_i$ for each agent, which is then used during the reverse denoising process (cf. Section \ref{sec 5.3.3: diffusion and denoising} and Algorithm \ref{alg: ddim-sampling}).

\begin{table}[ht]
\caption{Training settings for EP-Diffuser}
\vspace{-0.0em}
\centering
\begin{tabularx}{\textwidth}{c  >{\centering\arraybackslash}X }
\Xhline{3\arrayrulewidth}
hidden dimension $D$ &  128\\
\hline
\multirow{2}{*}{$\beta_s$} &  $ \frac{s}{S}\beta^{\text{end}}- (1-\frac{s}{S})\beta^{\text{start}}, \quad \text{with}$\\
&  $\beta^{\text{start}}=1e-5, \beta^{\text{end}}=0.2, S=1000$ \\
\hline
optimizer &  AdamW \cite{loshchilov_decoupled_2017}\\
\hline
learning rate & 5e-4\\
\hline
learning rate schedule & cosine \cite{loshchilov_sgdr_2016}\\
\hline
batch size &  32\\
\hline
training / warmup epochs  & 64 / 10\\
\hline
dropout &  0.1 \\
\Xhline{3\arrayrulewidth}

\end{tabularx}
\label{tab: setting for EP-Diffuser}
\end{table}

\subsection{Training Loss}
Following DDPM \cite{ho_denoising_2020}, the EP-Diffuser is trained to minimize the mean squared error (MSE) between the added noise $\boldsymbol{\epsilon}_i$ and predicted noise $\hat{\boldsymbol{\epsilon}}_i$ averaged across all agents: 
\begin{equation}
    \ell = \frac{1}{A}\sum_{i=1}^A || \boldsymbol{\epsilon}_i - \hat{\boldsymbol{\epsilon}}_i ||^2_2.
\end{equation}

\subsection{Training Settings}
Table \ref{tab: setting for EP-Diffuser} reports the training settings of EP-Diffuser. The noise scheduling parameter is expressed as $\bar{\alpha}_s = \Pi^s_k \alpha_k$, where $\alpha_s = 1 - \beta_s$. Following DDPM \cite{ho_denoising_2020},
$\beta_s$ denotes the variance schedule of the forward diffusion
process at diffusion step $s$.

\subsection{Post-processing}
\label{sec: post processing}
The observed trajectories of stationary agents often exhibit minor positional shifts and unrealistic rotations, which can lead to unnatural behaviors in predicted scenes across all models. To address this, we adopt a lightweight post-processing step inspired by prior works \cite{zeng_dsdnet_2020, hu_gaia_2023}:
\begin{itemize}[leftmargin=*]
    \item \textbf{Stationary Agent Correction}: For each predicted trajectory, if an agent moves less than \SI{1}{\meter} over the prediction horizon, we classify it as non-moving and retain its last measured position and heading.
\end{itemize}
\noindent This ensures physically consistent behavior for stationary agents without modifying the model’s core predictions. For fairness, we apply this post-processing step uniformly across both our model and benchmark models. The effect of this post-processing step will be discussed in Section \ref{sec 5.4.3: impact of post processing}.

\section{Experiments}
\label{sec 5.4: experiments}
\subsection{Experimental Setup}
\subsubsection{Data Homogenization}
To enable cross-dataset evaluation, we follow the homogenization protocol introduced in Section \ref{sec 4.3: data homogenization}, with minor adaptations in the selection of target agents.

For A2, unlike the marginal prediction setting with a single target agent per scene, the dataset designates multiple target agents for joint prediction. We therefore retain all annotated target agents provided in A2. 

For WO, which specifies up to 8 target agents per scene for both marginal and joint prediction tasks, we likewise preserve all annotated target agents as given in the dataset.

This procedure ensures comparability across datasets while preserving the dataset-specific definitions of target agents.

Although the evaluation focuses on designated target agents, each traffic scene contains additional non-target agents. The computation of the ``Sim Agents'' metrics requires coherent scene-level generation, including trajectories for these non-target agents. Consequently, the models predict trajectories for all agents within the scene rather
than only the annotated targets.

\subsubsection{Training and Testing}
As highlighted in Section \ref{sec 4.6: experiments}, the \TA{} setup demonstrated clearer differentiation between model design choices, particularly showing the benefits of polynomial representations and augmentation strategies. We therefore adopt the \TA{} setup introduced in Section \ref{sec 4.6.1: experimental setup} for all experiments in this chapter. In this setup, models are trained from scratch on the homogenized A2 training set using their default hyperparameters. They are evaluated on the homogenized A2 validation set for ID testing and on the homogenized WO validation set for OoD testing. Each model receives a 5-second history and predicts 6-second future traffic scenes.


Following the ``Sim Agents'' protocol, models are required to output 32 modeled traffic scene samples for metric computation. For generative models, we randomly sample 32 traffic scenes. For FMAE-MA, which by design outputs 6 predictions, we construct an ensemble of 6 independently trained models, each initialized with a different seed, predicting 36 traffic scenes in total. From these, we select the 32 predictions with the highest predicted probabilities for evaluation.

For ID testing, we report model performance on the 6-second prediction horizon. For OoD testing, traffic scene continuations are evaluated on the first 4.1 seconds, since WO recordings are shorter and limited to 9.1 seconds in total.

\subsubsection{Test Data Selection}
Due to the computational cost of ``Sim Agents'' metric calculations, we subsample the validation data from both the A2 and WO datasets under two distinct settings:
\begin{itemize}[leftmargin=*]
    \item \textbf{20\% Random Subsample (R20P):} We randomly select $20\%$ of the validation set from each dataset, corresponding to 5{,}000 samples from A2 and 8{,}400 samples from WO. This setting aims to reflect the model's overall performance across typical scenes presented by the datasets. 

    \item \textbf{500 Most Challenging Scenes (C500):} We identify and select the 500 most difficult traffic scenes in each validation set based on the largest deviations between observed future trajectories and those predicted by a constant velocity model, as measured by the realism meta metric. This setting emphasizes model performance under more complex and demanding conditions.
\end{itemize}
We refer to these subsets as \textbf{R20P} and \textbf{C500}, respectively, in the remainder of the chapter.

\subsection{Metrics}
We use the ``Sim Agents'' metrics to assess the plausibility of traffic scene continuations. In addition, we report $\text{minSADE}$ from the ``Sim Agents'' benchmark as an accuracy-oriented measure. 

To quantify the diversity of future samples, we adopt an approach inspired by the inter-policy diversity metric proposed in \cite{shiroshita_behaviorally_2020}. Specifically, we compute the pairwise distance between the final positions of the 32 sampled trajectories averaged over all agents to measure the trajectory diversity. This diversity metric serves as an \textbf{auxiliary} indicator rather than a standalone performance measure -- since high diversity is only meaningful when accompanied by high plausibility.

For OoD robustness evaluation, we report absolute metric values in OoD testing to enable direct comparison across evaluation dimensions. Additionally, we calculate performance changes between ID and OoD test sets (e.g., $\Delta
\text{minSADE}$) where relevant to highlight robustness patterns, consistent with Chapter \ref{cpt: improving generalization}.

\subsection{Impact of Post-processing}
\label{sec 5.4.3: impact of post processing}
The post-processing step proposed in Section~\ref{sec: post processing} stabilizes the position and heading of stationary agents. We evaluate its effect by testing all models on the A2 validation set with a 6-second prediction horizon using the C500 subset.

\begin{figure}[thb]
\centering
\includegraphics[width=0.7\textwidth]{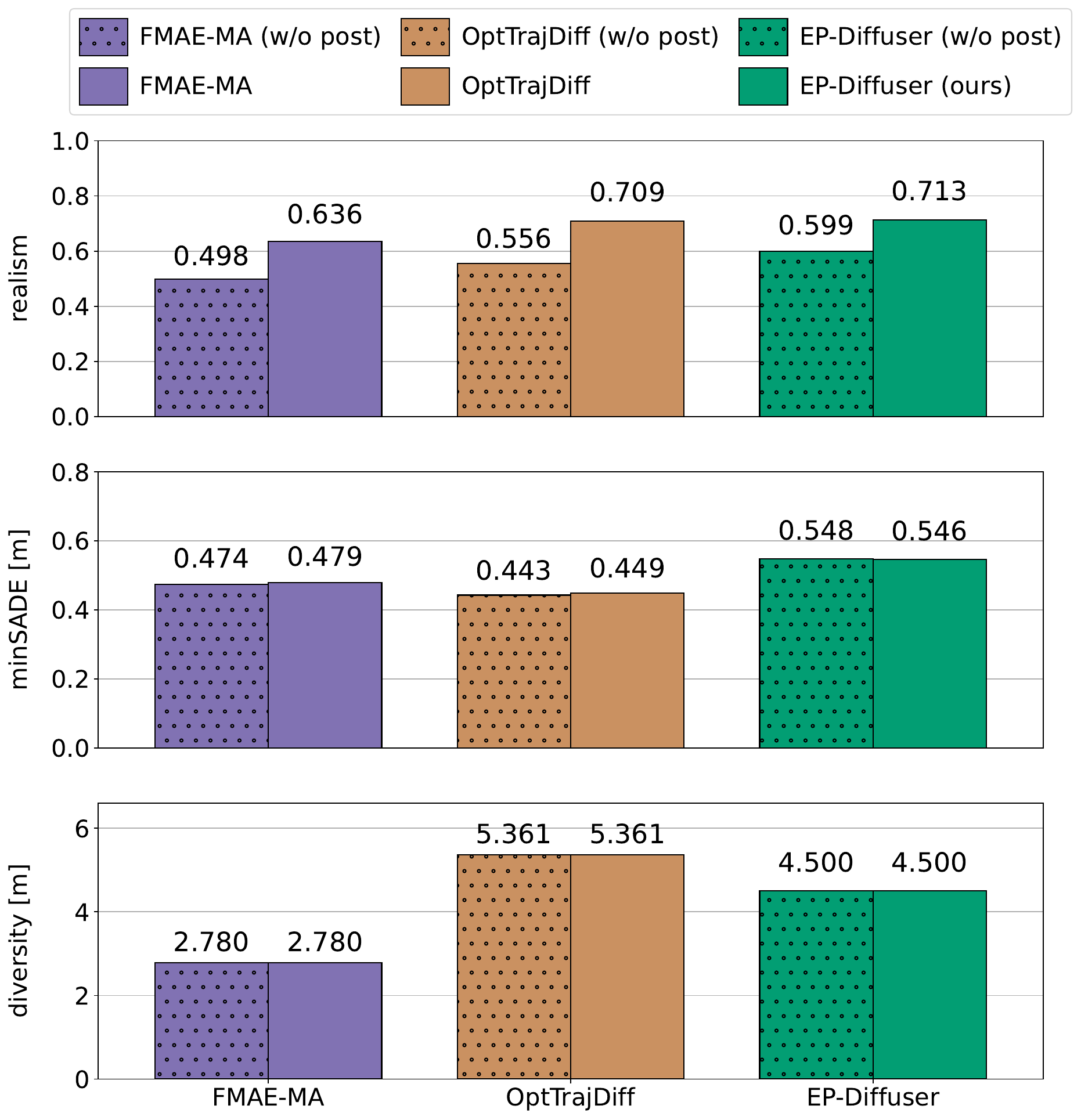}
\caption[Impact of post-processing]{Impact of post-processing on realism score, accuracy (minSADE), and diversity metrics, tested on A2 validation set with a 6-second prediction horizon using the C500 subset. Dotted bars show results without post-processing, solid bars show results with post-processing.}
\label{fig: impact of post processing}
\end{figure}

As shown in Figure~\ref{fig: impact of post processing}, post-processing consistently improves realism scores across all models, with gains of $0.138$ for FMAE-MA, $0.153$ for OptTrajDiff, and $0.114$ for EP-Diffuser. The slightly larger improvements observed for sequence-based benchmarks suggest that their raw predictions contain more instances of unrealistic stationary agent behavior that benefit from post-processing.

Importantly, post-processing has minimal impact on prediction accuracy and diversity, with minor changes for minSADE and no difference for diversity across all models. This selective improvement -- enhancing plausibility while preserving accuracy and diversity -- demonstrates a clear disconnect between regression-based accuracy metrics and plausibility as measured by the ``Sim Agents'' framework.

These improvements in prediction plausibility, coupled with minimal impact on accuracy and diversity, were observed consistently across ID and OoD testing on both the R20P and C500 test subsets.


\begin{figure}[!t]
\centering
\includegraphics[width=0.7\textwidth]{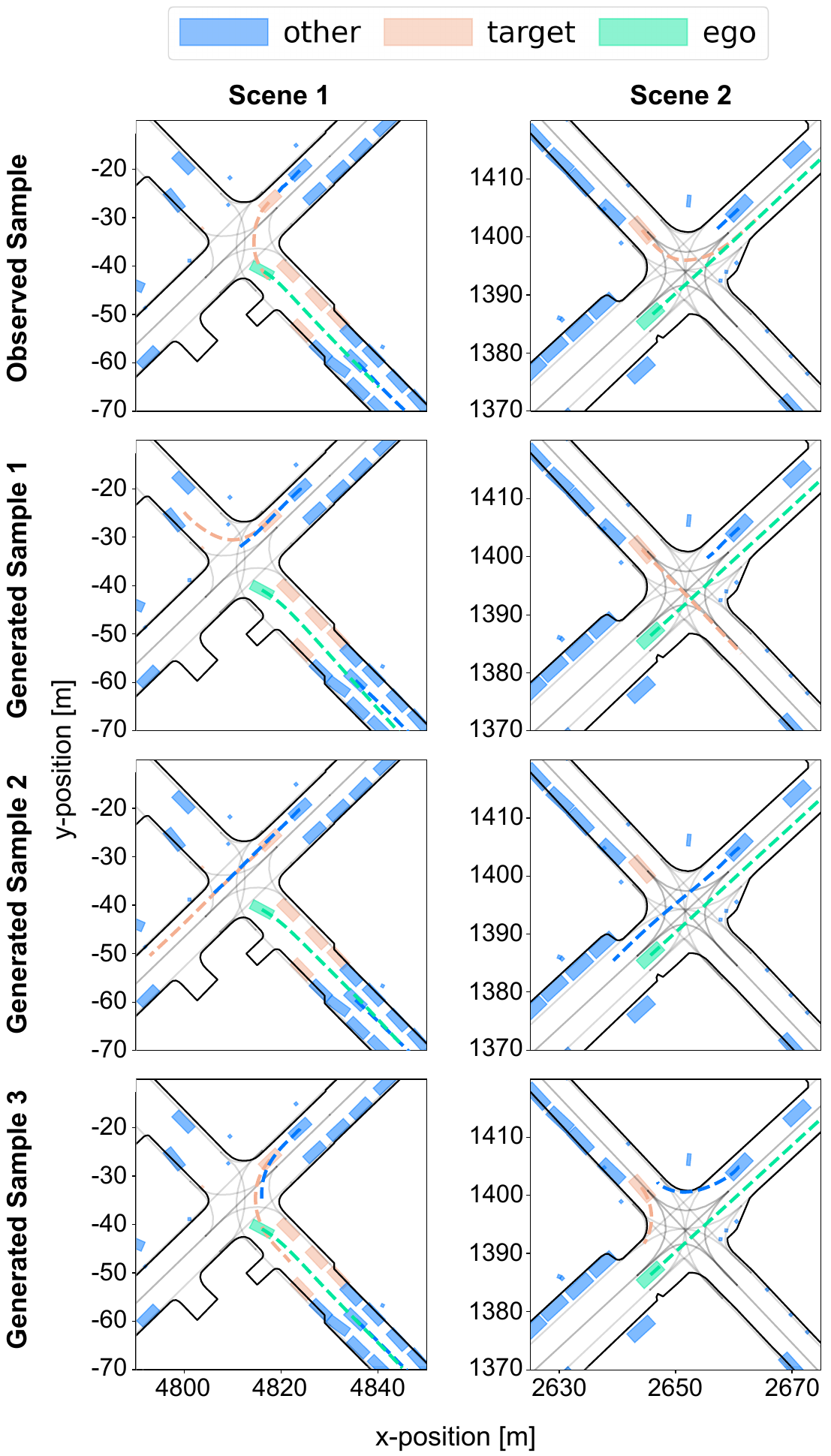}
\caption[Generated traffic scene samples from EP-Diffuser]{Observed sample and generated traffic scene samples from EP-Diffuser for two traffic scenes in Argoverse 2, demonstrating EP-Diffuser's capability to generate diverse and plausible traffic scenes. \textbf{Dashed lines} represent the future trajectories of selected highly interactive agents.}
\label{fig: simulated traffic scenes}
\end{figure}

\subsection{In-Distribution Results}
Table \ref{tab: sim agent 6s} presents the ID results for all models under a 6-second prediction horizon. The evaluation is conducted on the A2 validation set using both the R20P and C500 subsets. Figure \ref{fig: simulated traffic scenes} visualizes multiple samples generated by EP-Diffuser.

\begin{table*}[!thb]

\caption[In-distribution results with a 6-second prediction horizon]{ID results on the A2 dataset with a 6-second prediction horizon. Reported metrics include plausibility (``Sim Agents'' metrics), accuracy (minSADE), and diversity, evaluated on the R20P and C500 subsets. The best and second-best results for each metric across models are highlighted in \textbf{bold} and \underline{underline}, respectively.}
\centering
    \begin{tabularx}{\textwidth}{c c c | *{6}{>{\centering\arraybackslash}X}}
    \toprule
    \multicolumn{2}{c}{\multirow{2}{*}{Subsets}} &\multirow{2}{*}{model} & $\scriptstyle \text{realism}$ & $\scriptstyle \text{kinematic}$ & $\scriptstyle \text{interaction}$ & $\scriptstyle \text{map}$ & $\scriptstyle \text{minSADE}$ & $\scriptstyle \text{diversity}$ \\
    &&& $\scriptstyle \text{meta}\:\uparrow$ &  $\scriptstyle \text{metrics}\:\uparrow$ & $\scriptstyle \text{metrics}\:\uparrow$ & $\scriptstyle \text{metrics}\:\uparrow$ & $\scriptstyle \text{[m]} \: \downarrow$ & $\scriptstyle \text{[m]} \: \uparrow$ \\
        \midrule
        \multicolumn{2}{c}{\multirow{5}{*}{R20P}}&$\scriptstyle \text{FMAE-MA \cite{cheng_forecast_2023}}$ &0.749 & 0.459 & 0.786 & 0.867 & 0.333 & 2.000\\
        \cmidrule(lr){3-9}
        &&$\scriptstyle \text{OptTrajDiff \cite{wang_optimizing_2025}}$ & \underline{0.795}& \underline{0.564}  & \underline{0.805} & \textbf{0.915}& \textbf{0.325} & \textbf{4.447} \\
        \cmidrule(lr){3-9}
        & &$\scriptstyle \text{Seq-Diffuser (ablation)}$ & 0.748 & 0.414 & 0.789& 0.885& \underline{0.331} & 1.718\\
         \cmidrule(lr){3-9}
        && $\scriptstyle \text{EP-Diffuser (ours)}$ & \textbf{0.809} & \textbf{0.632} & \textbf{0.808}& \underline{0.913}& 0.398 & \underline{4.178}\\
        \bottomrule
        \multicolumn{2}{c}{\multirow{5}{*}{C500}} & $\scriptstyle \text{FMAE-MA \cite{cheng_forecast_2023}}$ &0.636 & 0.320 & 0.679 & 0.760 & 0.479 & 2.780\\
        \cmidrule(lr){3-9}
        &&$\scriptstyle \text{OptTrajDiff \cite{wang_optimizing_2025}}$ & \underline{0.709} & \underline{0.459}  & \textbf{0.717} & \textbf{0.841} & \textbf{0.449} & \textbf{5.361} \\
        \cmidrule(lr){3-9}
        &&$\scriptstyle \text{Seq-Diffuser (ablation)}$ & 0.634 & 0.279 & 0.674 & 0.786 & \underline{0.467} & 2.020\\
        \cmidrule(lr){3-9}
        &&$\scriptstyle \text{EP-Diffuser (ours)}$ & \textbf{0.713} & \textbf{0.507} & \underline{0.707} & \underline{0.838}& 0.546 & \underline{4.500}\\
        \bottomrule
    \end{tabularx}
    \label{tab: sim agent 6s}
\end{table*}

\begin{figure}[th]
\centering
\includegraphics[width=0.7\textwidth]{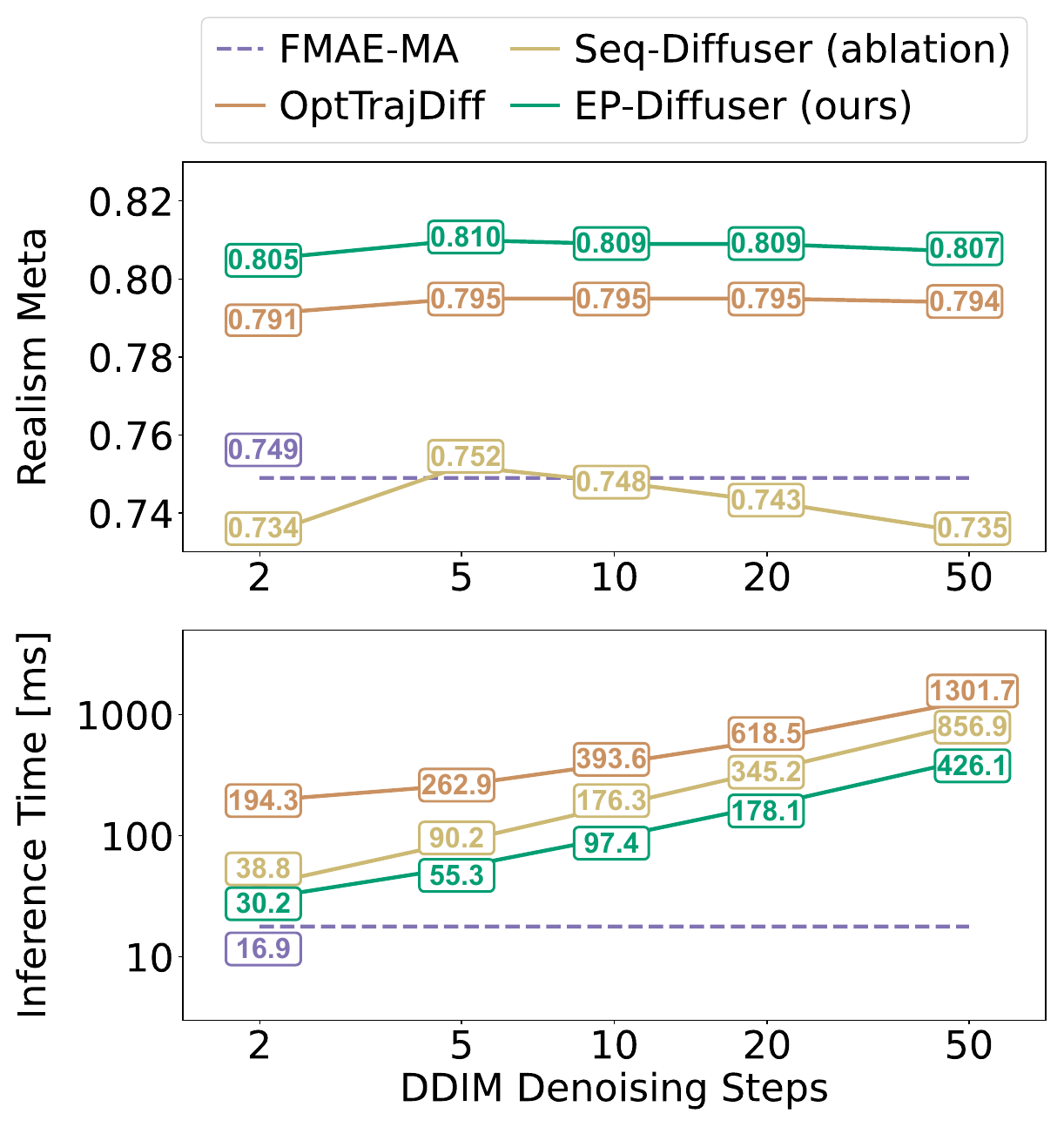}
\caption[Realism meta score and inference time]{Realism meta score on A2 R20P and inference time for diffusion models across different DDIM denoising steps, with FMAE-MA as the regression-based baseline
for comparison. Both the denoising steps and inference time are shown on \emph{log scale}. Inference time is measured by predicting 6 samples of an Argoverse 2 traffic scene with 50 agents and 150 map elements, using a single A10G GPU.}
\label{fig: influence of DDIM steps}
\end{figure}

We observe a reversed ranking when evaluating prediction plausibility and $\text{minSADE}$. For example, on R20P, despite its largest $\text{minSADE}$ (\SI{0.398}{\meter}), EP-Diffuser achieves the highest realism meta score at $0.809$ and performs comparably to OptTrajDiff in terms of agent interaction and map adherence, while also maintaining high diversity (\SI{4.178}{\meter}). Additionally, EP-Diffuser consistently outperforms OptTrajDiff in term of plausibility across different DDIM denoising steps, as visualized in the upper panel of Figure \ref{fig: influence of DDIM steps}. Notably, EP-Diffuser demonstrates superior agent kinematics ($0.632$), outperforming OptTrajDiff ($0.564$) and FMAE-MA ($0.456$).

\begin{figure}[thb]
\centering
\includegraphics[width=0.7\textwidth]{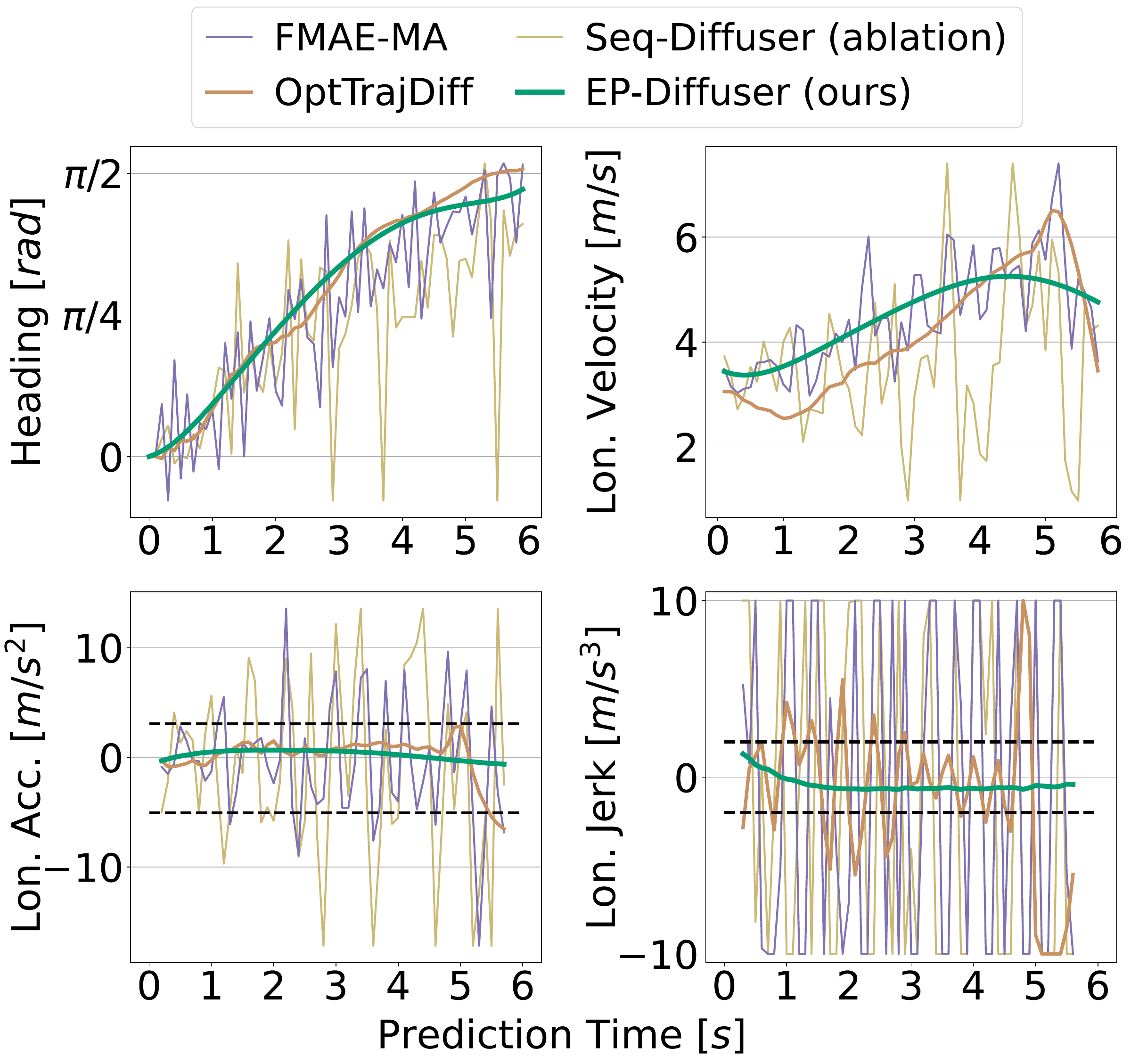}
\caption[Predicted agent kinematic states]{Predicted heading, longitudinal velocity, acceleration, and jerk for a vehicle's left turn maneuver over a 6-second time horizon in A2. For clarity, the results of Seq-Diffuser, as well as the jerk outputs of FMAE-MA and OptTrafDiff, are clipped. \textbf{Dashed lines} indicate the ranges of longitudinal acceleration and jerk for aggressive human drivers based on \cite{bae_self_2020}, suggesting that the agent kinematics of EP-Diffuser are the most plausible.}
\label{fig: simulated agent heading}
\end{figure}

As an example, Figure \ref{fig: simulated agent heading} visualizes the vehicle kinematics during a left-turn maneuver in A2. The trajectory generated by EP-Diffuser exhibits the smoothest kinematic profiles and remains within the plausible range of human driver behavior, according to the measures in \cite{bae_self_2020}.

We observe consistent ID results under the shorter 4.1-second prediction horizon, as summarized in the \textbf{``A2/A2'' sections} of Table \ref{tab: sim agent 4s}. EP-Diffuser again achieves the highest realism score while exhibiting the largest minSADE, further highlighting the disconnect between plausibility and accuracy.

\subsection{Out-of-Distribution Results}
In the OoD setting, models trained on A2 are evaluated on independent scenes from the WO dataset, thereby eliminating any shared bias between training and test data. The corresponding results are presented in the \textbf{``A2/WO'' sections} of Table \ref{tab: sim agent 4s}.

EP-Diffuser maintains the top scores in realism meta and agent kinematics. Additionally, it also demonstrates improved performance by achieving the best agent interaction and map adherence scores on both subsets. Furthermore, EP-Diffuser outperforms FMAE-MA in $\text{minSADE}$ and closely matches OptTrajDiff (\SI{0.372}{\meter} vs. \SI{0.357}{\meter} on C500, respectively) with high diversity (\SI{2.786}{\meter}). 

In addition, when comparing the ID (A2/A2) and OoD (A2/WO) results in Table \ref{tab: sim agent 4s}, EP-Diffuser exhibits the smallest performance degradation under distribution shift. For example, on the C500 subset, the $\text{minSADE}$ of EP-Diffuser increases from \SI{0.265}{\meter} to \SI{0.372}{\meter}, achieving a substantially lower $\Delta \text{minSADE}$ (+\SI{0.107}{\meter}) compared to FMAE-MA (+\SI{0.201}{\meter}) and OptTrajDiff (+\SI{0.146}{\meter}). This finding corroborates the \TA{} setup results from Chapter \ref{cpt: improving generalization}, further confirming that polynomial representations enhance model generalization in OoD test cases.

Overall, these results highlight EP-Diffuser's ability to learn robust, transferable representations from training data, demonstrating its enhanced generalization beyond dataset-specific patterns.

\begin{table*}[!t]

\caption[In-distribution and out-of-distribution results with a 4.1-second prediction horizon]{ID and OoD results with a 4.1-second prediction horizon. Reported metrics include plausibility (“Sim Agents” metrics), accuracy (minSADE), and diversity, evaluated on the R20P (upper section) and C500 (lower section) subsets. The best and second-best results for each metric across models are highlighted in \textbf{bold} and \underline{underline}, respectively.}
\centering
    \begin{tabularx}{\textwidth}{c c c | *{6}{>{\centering\arraybackslash}X}}
    \toprule
    \multicolumn{2}{c}{\multirow{2}{*}{train / test}} &\multirow{2}{*}{model} & $\scriptstyle \text{realism}$ & $\scriptstyle \text{kinematic}$ & $\scriptstyle \text{interaction}$ & $\scriptstyle \text{map}$ & $\scriptstyle \text{minSADE}$ & $\scriptstyle \text{diversity}$ \\
    &&& $\scriptstyle \text{meta}\:\uparrow$ &  $\scriptstyle \text{metrics}\:\uparrow$ & $\scriptstyle \text{metrics}\:\uparrow$ & $\scriptstyle \text{metrics}\:\uparrow$ & $\scriptstyle \text{[m]} \: \downarrow$ & $\scriptstyle \text{[m]} \: \uparrow$ \\
        \midrule
        \multirow{11}{*}{\rotatebox{90}{R20P}}&\multirow{5}{*}{$\scriptstyle \text{A2 / A2}$} &$\scriptstyle \text{FMAE-MA \cite{cheng_forecast_2023}}$ &0.796 & 0.514 & 0.834 & 0.907 & 0.156 & 1.001\\
        \cmidrule(lr){3-9}
        &&$\scriptstyle \text{OptTrajDiff \cite{wang_optimizing_2025}}$ & \underline{0.835}& \underline{0.613}  & \textbf{0.850} & \textbf{0.941}& \textbf{0.152} & \textbf{2.696} \\
        \cmidrule(lr){3-9}
        & &$\scriptstyle \text{Seq-Diffuser (ablation)}$ & 0.787 & 0.442 & 0.841& 0.915& \underline{0.155} & 1.019\\
         \cmidrule(lr){3-9}
        && $\scriptstyle \text{EP-Diffuser (ours)}$ & \textbf{0.847} & \textbf{0.687} & \underline{0.848}& \underline{0.937}& 0.188 & \underline{2.476}\\
        \hhline{~========}
        &\multirow{5}{*}{$\scriptstyle \text{A2 / WO}$} &$\scriptstyle \text{FMAE-MA \cite{cheng_forecast_2023}}$&0.709  & 0.309 & 0.800 & 0.819 & 0.363& 3.560\\
        \cmidrule(lr){3-9}
        &&$\scriptstyle \text{OptTrajDiff \cite{wang_optimizing_2025}}$ & \underline{0.768} & \underline{0.420}  & \underline{0.827} & \underline{0.890} & $\textbf{0.310}$ & \textbf{3.496} \\
        \cmidrule(lr){3-9}
        &&$\scriptstyle \text{Seq-Diffuser (ablation)}$ &  0.688 & 0.212& 0.801& 0.814& 0.458& 2.446\\
        \cmidrule(lr){3-9}
        &&$\scriptstyle \text{EP-Diffuser (ours)}$ & \textbf{0.788} & \textbf{0.491} & \textbf{0.834} & \textbf{0.900} & \underline{0.348} & \underline{3.224}\\
        \bottomrule
        \multirow{11}{*}{\rotatebox{90}{C500}}&\multirow{5}{*}{$\scriptstyle \text{A2 / A2}$}&$\scriptstyle \text{FMAE-MA \cite{cheng_forecast_2023}}$ &0.701 & 0.359 & 0.751 & 0.832 & 0.225 & 1.447\\
        \cmidrule(lr){3-9}
        &&$\scriptstyle \text{OptTrajDiff \cite{wang_optimizing_2025}}$ & 0.759 & 0.494  & \textbf{0.774} & \textbf{0.891} & \textbf{0.211} & \textbf{3.269} \\
        \cmidrule(lr){3-9}
        &&$\scriptstyle \text{Seq-Diffuser (ablation)}$ & 0.696 & 0.302 & 0.754 & 0.846 & \underline{0.221} & 1.204\\
        \cmidrule(lr){3-9}
        &&$\scriptstyle \text{EP-Diffuser (ours)}$ & \textbf{0.767} & \textbf{0.553} & 0.772 & 0.881& 0.265 & \underline{2.683}\\
        \hhline{~========}
        &\multirow{5}{*}{$\scriptstyle \text{A2 / WO}$}& $\scriptstyle \text{FMAE-MA \cite{cheng_forecast_2023}}$ &0.630 & 0.271 & 0.716 & 0.723 & 0.426 & \textbf{3.855}\\
        \cmidrule(lr){3-9}
        &&$\scriptstyle \text{OptTrajDiff \cite{wang_optimizing_2025}}$  & 0.721 & 0.403  & 0.771 & 0.839 & \textbf{0.357} & \underline{3.189}\\
        \cmidrule(lr){3-9}
        &&$\scriptstyle \text{Seq-Diffuser (ablation)}$ & 0.630 & 0.210 & 0.739 & 0.730 & 0.448& 1.980\\
        \cmidrule(lr){3-9}
        &&$\scriptstyle \text{EP-Diffuser (ours)}$ & \textbf{0.742} & \textbf{0.456} & \textbf{0.782} & \textbf{0.854} & \underline{0.372} & 2.786\\
        \bottomrule
    \end{tabularx}
    \label{tab: sim agent 4s}
\end{table*}

\subsection{Model Efficiency}
\label{sec 5.4.6: efficiency}
Beyond performance and generalization, we further analyze the computational efficiency of EP-Diffuser. Leveraging polynomial representations, EP-Diffuser has a substantially smaller model size -- only $\SI{3.0}{M}$ parameters -- compared to OptTrajDiff ($\SI{12.5}{M}$) and the FMAE-MA ensemble ($\SI{11.4}{M}$).

Figure \ref{fig: influence of DDIM steps} reports the realism meta scores and inference time of diffusion-based models over a range of DDIM denoising steps, with FMAE-MA as the regression-based baseline for comparison. Across all denoising step configurations, EP-Diffuser consistently outperforms OptTrajDiff in both metrics, achieving higher realism scores with lower inference time. While EP-Diffuser's inference time exceeds FMAE-MA's, it reaches peak performance with 5 denoising steps in just \SI{55.3}{\milli\second}, making it viable for real-time applications operating at \SI{10}{Hz} \cite{wilson_argoverse2_2021, ettinger_waymo_2021}. 

Overall, these findings further highlight the efficiency gains enabled by compact polynomial representations.

\subsection{Ablation Study}
To evaluate the impact of polynomial representations, we introduce a model variant, \textbf{Seq-Diffuser}, which removes the ``Polynomial Fit and Track'' module from EP-Diffuser (Figure \ref{fig: diffusion  pipeline}) and operates directly on sequence-based representations. Only minor modifications to the model architecture are required to accommodate the increased dimensionality of the sequence-based representation. The training and diffusion settings are identical for both EP-Diffuser and Seq-Diffuser. Comparative results are presented in Table \ref{tab: sim agent 6s} and \ref{tab: sim agent 4s}.

According to Table \ref{tab: sim agent 4s}, while Seq-Diffuser achieves better ID accuracy with a lower $\text{minSADE}$ (e.g., \SI{0.155}{m} vs. \SI{0.188}{m} on R20P), it exhibits a notable degradation in both the plausibility and diversity of the generated traffic scenes across both subsets. Moreover, Seq-Diffuser generalizes poorly, as reflected by its lower performance in plausibility and higher OoD $\text{minSADE}$ (e.g., \SI{0.458}{m} vs. \SI{0.348}{m} on R20P). It also incurs a higher inference time, as shown in Figure \ref{fig: influence of DDIM steps}.

These results indicate that, although sequence-based representations may capture finer-grained details beneficial for pointwise accuracy, polynomial representations lead to better plausibility, diversity, OoD generalization, and efficiency in multi-agent scene generation.

\subsection{Failure Cases and Limitations}
\label{sec:failure_cases}

\begin{table*}[!thb]

\caption[Collision rate and off-road rate]{Collision rate and off-road rate tested on the A2 C500 subset with a 6-second prediction horizon. Observed future trajectories are included as reference. The best and second-best results for each metric across models are highlighted in \textbf{bold} and \underline{underline}, respectively.}

\centering
    \begin{tabularx}{\textwidth}{c | *{2}{>{\centering\arraybackslash}X}}
    \toprule
    model & collision rate $\:\downarrow$ & off-road rate $\:\downarrow$ \\
        \midrule
        observed future &0.046 & 0.145 \\
        \hhline{===}
        FMAE-MA \cite{cheng_forecast_2023} &0.129 & 0.208 \\
        \cmidrule(lr){2-3}
        OptTrajDiff \cite{wang_optimizing_2025} & \textbf{0.124} & \underline{0.157} \\
        \cmidrule(lr){2-3}
        Seq-Diffuser (ablation) & 0.165 & 0.212\\
        \cmidrule(lr){2-3}
        EP-Diffuser (ours) & \underline{0.128} & \textbf{0.150}\\
        \bottomrule
    \end{tabularx}
    \label{tab: collision and offroad}
\end{table*}

While EP-Diffuser demonstrates strong performance across multiple metrics, we identify occasional implausible behaviors in generated scenes, such as off-road trajectories and agent collisions.

\begin{figure}[th]
\centering
\includegraphics[width=0.8\textwidth]{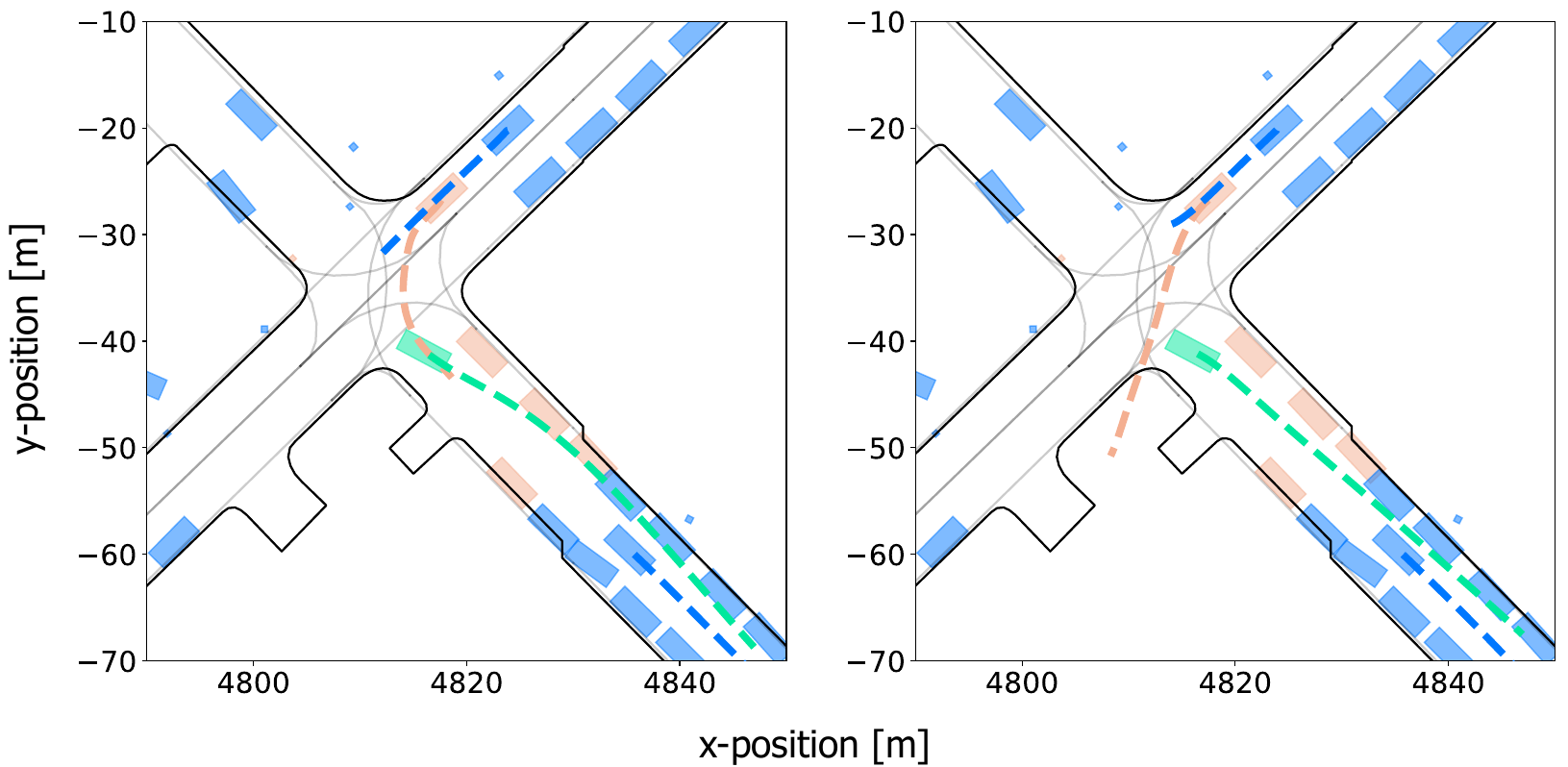}
\caption[Collision and off-road examples]{Examples of failure cases from EP-Diffuser: (left) agent collision where generated ego trajectory intersects with parked vehicles, and (right) off-road trajectory that violates map boundaries.}
\label{fig: collision and offroad}
\end{figure}

Table \ref{tab: collision and offroad} reports collision and off-road rates on the A2 C500 subset. While EP-Diffuser achieves comparable results to OptTrajDiff, it exhibits a higher off-road rate than observed ground-truth trajectories and a considerably higher collision rate (0.128 vs. 0.046 for observed futures). Representative examples of these failure cases are visualized in Figure \ref{fig: collision and offroad}, which illustrates generated samples that violate map boundaries or result in agent collisions.

This indicates that accurately modeling complex multi-agent interactions and map constraints remains challenging for EP-Diffuser.

It is important to note that the collision rate (\num{0.046}) of the observed future trajectories in Table \ref{tab: collision and offroad} primarily stems from perception and tracking inaccuracies inherent in the data (cf. Section \ref{sec: outlier filtering results} and \ref{sec: post processing}). In particular, overlapping agent bounding boxes may result from noisy detections rather than actual physical collisions. Similarly, the off-road rate (\num{0.145}) is largely attributable to incomplete spatial coverage in the provided HD maps. 

\section{Conclusion and Discussion}
\label{sec 5.5: conclusion}
This chapter extended the polynomial-based prediction framework introduced in Chapter \ref{cpt: improving generalization} to the more complex joint prediction task. We proposed EP-Diffuser, a diffusion-based generative framework that leverages polynomial representations of trajectories and map geometries to capture distribution of traffic scene continuations. EP-Diffuser employs the graph-based scene structure that enables efficient interaction modeling across agents and map elements, addressing a key limitation of EP-Q proposed in Chapter \ref{cpt: improving generalization}.

Comprehensive experiments on the Argoverse 2 (A2) and Waymo Open (WO) datasets demonstrate four key findings. First, EP-Diffuser consistently achieves higher scene plausibility and more realistic agent kinematics compared to both regression-based (FMAE-MA) and diffusion-based (OptTrajDiff) baselines using sequence data. Second, the model retains strong OoD generalization, exhibiting the smallest performance degradation between A2 and WO evaluation domains -- confirming that the generalization benefits of polynomial representations observed in Chapter \ref{cpt: improving generalization} extend to the joint prediction task. Third, despite the iterative nature of diffusion sampling, EP-Diffuser attains favorable computational efficiency, reaching real-time inference and requiring significantly fewer model parameters. Finally, the ablation study with Seq-Diffuser underscores the benefits of polynomial representations: replacing them with sequence-based representations improves accuracy but substantially degrades plausibility, diversity, generalization, and efficiency.

Despite these advances, EP-Diffuser exhibits several limitations that warrant further investigation. While achieving the highest realism scores among evaluated models, the generated scenes occasionally contain implausible behaviors such as off-road trajectories or agent collisions. Additionally, unlike regression-based approaches that provide explicit probability estimates for each predicted mode, diffusion-based sampling does not naturally yield scene-level likelihood scores. The absence of probabilistic confidence measures may complicate integration with downstream planning systems that rely on uncertainty quantification for safe decision-making.

Our results also highlight a fundamental disconnect between traditional regression-based metrics and measures of prediction plausibility. The performance reversal between minSADE rankings and realism scores demonstrates that optimizing for positional accuracy alone may not yield the most plausible predictions for downstream tasks. This finding reinforces the importance of adopting evaluation frameworks such as ``Sim Agents'' that assess multiple dimensions of prediction quality beyond mere accuracy.

In summary, this chapter demonstrates that polynomial representations, when integrated into modern diffusion frameworks, provide a powerful approach for traffic scene generation that achieves high plausibility, strong generalization, and computational efficiency. These results validate the core research thesis that polynomial representations offer fundamental advantages for autonomous driving prediction systems, extending the benefits observed in marginal prediction to the more complex domain of multi-agent joint scene generation.
\chapter{Conclusion and Outlook}
\section{Conclusion}
This dissertation investigated whether polynomial representations can provide a principled and efficient foundation for traffic scene prediction in autonomous driving. The central thesis of this work is that polynomial representations offer a compact, generalizable, and physically consistent basis for traffic scene prediction, enabling improved model efficiency, robustness, and prediction plausibility while maintaining competitive predictive performance.

The results presented across Chapters \ref{cpt: Empirical Bayes Analysis} to \ref{cpt: generative model} substantiate this claim from complementary perspectives:

\textbf{Chapter \ref{cpt: Empirical Bayes Analysis}} established a principled foundation through an empirical Bayes analysis of polynomial representations in learning-based prediction. Quantitative evaluation across several large-scale motion datasets demonstrated that moderate-degree Bernstein polynomials can approximate real-world trajectories with high fidelity while maintaining compactness and smoothness. The results show that the representation error introduced by polynomial representations does not constrain the performance of state-of-the-art (SotA) prediction systems. These results provide principled justification for the use of polynomial representations in learning-based prediction and form the basis for the subsequent model designs.

\textbf{Chapter \ref{cpt: improving generalization}} extended this foundation to a prediction model that represents both trajectories and map elements using polynomial parameterizations. The proposed approach achieved competitive In-Distribution (ID) and superior Out-of-Distribution (OoD) generalization while requiring only $3.9\%$ of the parameters of the strongest sequence-based benchmark. This indicates that polynomial representations can strengthen robustness under distribution shift, improve computational efficiency, and maintain competitive predictive performance.


\textbf{Chapter \ref{cpt: generative model}} further demonstrated that polynomial representations can be integrated into a diffusion-based generative framework for multi-agent traffic scene generation. With 3.0 million parameters -- a fraction of the diffusion-based benchmark's 12.5 million -- the proposed EP-Diffuser achieved the highest realism meta score (0.809) among all evaluated models, together with improved OoD generalization. These findings confirm that polynomial representations remain effective and efficient even in complex multi-agent generative settings.






Across these investigations, several broader insights emerged that challenge conventional practices in traffic scene prediction. The consistent gap between ID accuracy and OoD robustness suggests that prevailing benchmark paradigms may inadequately assess real-world generalization. Furthermore, the divergence between regression-based metrics and plausibility-oriented evaluations underscores the need to reconsider how prediction quality is measured in safety-critical systems. Together, these findings highlight that architectural complexity alone does not guarantee robustness; rather, principled data representations aligned with physical motion structure play a decisive role.

In summary, this work demonstrates that polynomial representations provide a structurally grounded alternative to discretized sequence-based representations. By reducing data dimensionality, encouraging motion continuity, and regularizing measurement noise, they improve computational efficiency, promote behavioral plausibility, and strengthen generalization under distribution shift in predicted traffic scenes. These results suggest that future advances in traffic scene prediction may depend as much on representation design as on model architecture, particularly in safety-critical autonomous driving systems where reliable generalization is essential.

\section{Limitations and Outlook}
\label{sec 6.2}
While this dissertation demonstrated the effectiveness of polynomial representations for traffic scene prediction, several limitations remain that suggest promising directions for future research.

\subsubsection{Methodological Limitations}

At the methodological level, some constraints stem from the intrinsic properties of polynomial and diffusion-based formulations.

Although polynomial representations encourage motion continuity and compactness, they do not guarantee physical plausibility. Generated traffic scenes occasionally exhibit implausible behaviors such as off-road trajectories, agent collisions, or kinematically implausible accelerations. Future work could integrate explicit physical constraints into the polynomial framework—for instance, differentiable collision checking, road boundary constraints, or kinematic feasibility filters during training or inference. Hybrid approaches that combine polynomial representations with physics-based models or safety layers could further ensure that predictions remain physically realizable.

While Chapter 3 provided data-driven guidance for selecting polynomial degree via information criteria (AIC, BIC), these optimal degrees were determined offline and globally. Different traffic contexts may benefit from adaptive degree selection -- simpler polynomials for highway scenes and higher-degree terms for complex urban maneuvers. Future work could investigate dynamic, scene-dependent degree selection that adapt complexity to local motion characteristics.

Another key limitation concerns probabilistic confidence and uncertainty quantification. Unlike regression-based approaches that yield explicit likelihoods, the diffusion sampling in EP-Diffuser does not naturally produce scene-level probability estimates. This absence of calibrated confidence measures may complicate integration with downstream planning systems that rely on probability estimates for decision-making. Future research could explore ensemble-based uncertainty estimation, likelihood approximation techniques, or hybrid diffusion-probabilistic models that combine generative diversity with interpretable confidence outputs.

A further limitation concerns the input representation itself. The models rely on a lane-level map and past agent states, both encoded as polynomials -- a representation that is compact and consistent with the evaluation benchmarks, but bounded in what it can express. Arbitrarily shaped static structures, such as barriers or construction layouts, are not part of the model inputs, and abstracting the scene to agents and lane geometry discards cues present in raw sensor data but tied to neither, such as turn signals or pedestrian pose, which can be informative for intent estimation (cf. Section \ref{sec 4.7: conclusion}). Future work could address this through a dedicated occupancy representation for static structure, or a learned representation operating closer to the sensor input, which would retain such cues at the cost of the interpretability that the polynomial abstraction provides.

\subsubsection{Experimental and Practical Limitations}

Beyond model formulation, several experimental constraints also shape current performance.
The OoD evaluation revealed that robustness depends on interactions among dataset characteristics (e.g., noise levels, task complexity), model design choices, and training configurations. For instance, the \TWO{} setup (training on Waymo Open, testing on Argoverse 2) showed that larger training datasets do not always yield better generalization when test data differ in noise or task difficulty. Controlled experiments -- such as synthetic noise injection or systematic ablations -- could help disentangle these factors and provide clearer guidance on when polynomial representations confer the greatest robustness advantages.

In terms of computational efficiency, although EP-Diffuser achieved real-time inference under moderate scene complexity, runtime grows with the number of agents and map elements. Further optimization could involve hierarchical scene decomposition, sparse-attention mechanisms, or progressive-refinement schemes that allocate computation adaptively based on scene importance.

\subsubsection{System-level Outlook}

From a broader system perspective, this work evaluated prediction models in open-loop settings, where predictions are compared against logged trajectories without feedback from planning. However, models optimized for open-loop metrics may not perform optimally in closed-loop planning environments \cite{bouzidi_closing_2025}, where the ego vehicle’s actions influence other agents. The prediction models developed here are also ego-agnostic, in line with the benchmarks used for evaluation, and therefore do not yet close this loop. 

The polynomial formulation is well suited to bridging prediction and planning. Because the derivative of a Bernstein polynomial is itself a lower-degree Bernstein polynomial, dynamic feasibility limits such as velocity and acceleration reduce to closed-form conditions on the control points -- a property widely exploited in trajectory optimization. Furthermore, since EP-Diffuser already generates the trajectories of all agents jointly, it is a natural entry point for joint prediction and planning: by conditioning the ego agent on planning inputs such as a navigation route or goal, the model could generate a route-following ego trajectory together with, and reactive to, the predicted behavior of the surrounding agents. This would treat prediction and planning within a single generative process, capturing the mutual influence between the ego vehicle and other agents that a sequential prediction-then-planning pipeline omits. Realizing such a formulation within a closed-loop stack is a promising direction for future work.



The polynomial framework may also extend to other components of the autonomous driving stack, for instance to perception tasks such as agent tracking, further exploiting the regularizing properties of polynomials.

Beyond simulation- and dataset-based evaluation, validating the predictor in real time on a physical vehicle platform would be a necessary step toward deployment, and remains an important direction for future work.

Together, these directions outline a path toward physically grounded, uncertainty-aware, and computationally scalable prediction frameworks that bridge the gap between interpretable modeling and real-world deployment.


\begin{spacing}{0.9}


\bibliographystyle{ieeetr} 
\cleardoublepage
\bibliography{References/references} 

@string{ICRA = "IEEE International Conference on Robotics and Automation (ICRA)"}

@string{IROS = "IEEE/RSJ International Conference on Intelligent Robots and Systems (IROS)"}

@string{CVPR = "IEEE/CVF Conference on Computer Vision and Pattern Recognition (CVPR)"}

@string{ITSC = "IEEE International Conference on Intelligent Transportation Systems (ITSC)"}

@string{IV = "IEEE Intelligent Vehicles Symposium (IV)"}

@string{ICLR = "International Conference on Learning Representations (ICLR)"}

@string{CORL = "Conference on Robot Learning (CoRL)"}

@string{ICML = "International Conference on Machine Learning (ICML)"}

@string{NeurIPS = "Conference on Neural Information Processing Systems (NeurIPS)"}

@string{ECCV = "European Conference on Computer Vision (ECCV)"}

@string{ICCV = "IEEE/CVF International Conference on Computer Vision (ICCV)"}

@inproceedings{chai_multipath_2019,
  title={Multipath: Multiple probabilistic anchor trajectory hypotheses for behavior prediction},
  author={Chai, Yuning and Sapp, Benjamin and Bansal, Mayank and Anguelov, Dragomir},
  booktitle=CoRL,
  year={2019}
}

@inproceedings{varadarajan_multipath++_2022,
  title={Multipath++: Efficient information fusion and trajectory aggregation for behavior prediction},
  author={Varadarajan, Balakrishnan and Hefny, Ahmed and Srivastava, Avikalp and Refaat, Khaled S and Nayakanti, Nigamaa and Cornman, Andre and Chen, Kan and Douillard, Bertrand and Lam, Chi Pang and Anguelov, Dragomir and others},
  booktitle=ICRA,
  pages={7814--7821},
  year={2022},
}

@inproceedings{liang_learning_2020,
  title={Learning lane graph representations for motion forecasting},
  author={Liang, Ming and Yang, Bin and Hu, Rui and Chen, Yun and Liao, Renjie and Feng, Song and Urtasun, Raquel},
  booktitle=ECCV,
  pages={541--556},
  year={2020},
}

@inproceedings{salzmann_trajectron_2020,
  title = {Trajectron++: {{Dynamically-Feasible Trajectory Forecasting}} with {{Heterogeneous Data}}},
  booktitle = ECCV,
  author = {Salzmann, Tim and Ivanovic, Boris and Chakravarty, Punarjay and Pavone, Marco},
  year = {2020},
  pages = {683--700}
}

@inproceedings{chang_argoverse_2019,
author = {Chang, Ming-Fang and Lambert, John and Sangkloy, Patsorn and Singh, Jagjeet and Bak, Slawomir and Hartnett, Andrew and Wang, De and Carr, Peter and Lucey, Simon and Ramanan, Deva and Hays, James},
title = {Argoverse: 3D Tracking and Forecasting With Rich Maps},
booktitle = CVPR,
year = {2019}
}

@inproceedings{caesar_nuscenes_2020,
  title={nuscenes: A multimodal dataset for autonomous driving},
  author={Caesar, Holger and Bankiti, Varun and Lang, Alex H and Vora, Sourabh and Liong, Venice Erin and Xu, Qiang and Krishnan, Anush and Pan, Yu and Baldan, Giancarlo and Beijbom, Oscar},
  booktitle=CVPR,
  pages={11621--11631},
  year={2020}
}

@article{caesar_nuplan_2021,
  title={nuplan: A closed-loop ml-based planning benchmark for autonomous vehicles},
  author={Caesar, Holger and Kabzan, Juraj and Tan, Kok Seang and Fong, Whye Kit and Wolff, Eric and Lang, Alex and Fletcher, Luke and Beijbom, Oscar and Omari, Sammy},
  journal={arXiv preprint arXiv:2106.11810},
  year={2021}
}

@article{malinin_shifts_2021,
  title={Shifts: A dataset of real distributional shift across multiple large-scale tasks},
  author={Malinin, Andrey and Band, Neil and Chesnokov, German and Gal, Yarin and Gales, Mark JF and Noskov, Alexey and Ploskonosov, Andrey and Prokhorenkova, Liudmila and Provilkov, Ivan and Raina, Vatsal and others},
  journal={arXiv preprint arXiv:2107.07455},
  year={2021}
}

@inproceedings{gao_vectornet_2020,
  title={Vectornet: Encoding hd maps and agent dynamics from vectorized representation},
  author={Gao, Jiyang and Sun, Chen and Zhao, Hang and Shen, Yi and Anguelov, Dragomir and Li, Congcong and Schmid, Cordelia},
  booktitle=CVPR,
  pages={11525--11533},
  year={2020}
}

@inproceedings{wilson_argoverse2_2021,
  title={Argoverse 2: Next Generation Datasets for Self-Driving Perception and Forecasting},
  author={Wilson, Benjamin and Qi, William and Agarwal, Tanmay and Lambert, John and Singh, Jagjeet and Khandelwal, Siddhesh and Pan, Bowen and Kumar, Ratnesh and Hartnett, Andrew and Pontes, Jhony Kaesemodel and others},
  booktitle={Conference on Neural Information Processing Systems (NeurIPS) Datasets and Benchmarks Track (Round 2)},
  year={2021}
}

@article{macadam_understanding_2003,
  title={Understanding and modeling the human driver},
  author={Macadam, Charles C},
  journal={Vehicle system dynamics},
  volume={40},
  number={1-3},
  pages={101--134},
  year={2003},
  publisher={Taylor \& Francis}
}

@article{bae_self_2020,
  title={Self-driving like a human driver instead of a robocar: Personalized comfortable driving experience for autonomous vehicles},
  author={Bae, Il and Moon, Jaeyoung and Jhung, Junekyo and Suk, H and Kim, Taewoo and Park, Hyunbin and Cha, Jaekwang and Kim, Jincheol and Kim, Dohyun and Kim, Shiho},
  journal={arXiv preprint arXiv:2001.03908},
  year={2020}
}

@article{hayati_jerk_2020,
  title={Jerk within the context of science and engineering—A systematic review},
  author={Hayati, Hasti and Eager, David and Pendrill, Ann-Marie and Alberg, Hans},
  journal={Vibration},
  volume={3},
  number={4},
  pages={371--409},
  year={2020},
  publisher={MDPI}
}

@incollection{akaike_AIC_1973,
  title={Information theory and an extension of the maximum likelihood principle},
  author={Akaike, Hirotogu},
  booktitle={Selected papers of hirotugu akaike},
  pages={199--213},
  year={1998},
  publisher={Springer}
}

@book{efron_large_2012,
  title={Large-scale inference: empirical Bayes methods for estimation, testing, and prediction},
  author={Efron, Bradley},
  volume={1},
  year={2012},
  publisher={Cambridge University Press}
}

@article{schwarz_BIC_1978,
  title={Estimating the dimension of a model},
  author={Schwarz, Gideon},
  journal={The annals of statistics},
  pages={461--464},
  year={1978},
  publisher={JSTOR}
}

@inproceedings{bahari_vehicle_2022,
  title={Vehicle trajectory prediction works, but not everywhere},
  author={Bahari, Mohammadhossein and Saadatnejad, Saeed and Rahimi, Ahmad and Shaverdikondori, Mohammad and Shahidzadeh, Amir Hossein and Moosavi-Dezfooli, Seyed-Mohsen and Alahi, Alexandre},
  booktitle=CVPR,
  pages={17123--17133},
  year={2022}
}

@article{dillon_tensorflow_2017,
  title={Tensorflow distributions},
  author={Dillon, Joshua V and Langmore, Ian and Tran, Dustin and Brevdo, Eugene and Vasudevan, Srinivas and Moore, Dave and Patton, Brian and Alemi, Alex and Hoffman, Matt and Saurous, Rif A},
  journal={arXiv preprint arXiv:1711.10604},
  year={2017}
}

@article{nayakanti_wayformer_2022,
  title={Wayformer: Motion Forecasting via Simple \& Efficient Attention Networks},
  author={Nayakanti, Nigamaa and Al-Rfou, Rami and Zhou, Aurick and Goel, Kratarth and Refaat, Khaled S and Sapp, Benjamin},
  journal={arXiv preprint arXiv:2207.05844},
  year={2022}
}

@phdthesis{philipp_perception_2021,
  title={Perception and Prediction of Urban Traffic Scenarios for Autonomous Driving},
  author={Philipp, Andreas},
  year={2021},
  school={Freie Universitaet Berlin}
}

@inproceedings{lee_desire_2017,
  title={Desire: Distant future prediction in dynamic scenes with interacting agents},
  author={Lee, Namhoon and Choi, Wongun and Vernaza, Paul and Choy, Christopher B and Torr, Philip HS and Chandraker, Manmohan},
  booktitle=CVPR,
  pages={336--345},
  year={2017}
}

@article{buhet_plop_2020,
  title={Plop: Probabilistic polynomial objects trajectory planning for autonomous driving},
  author={Buhet, Thibault and Wirbel, Emilie and Bursuc, Andrei and Perrotton, Xavier},
  journal      = {arXiv preprint arXiv:2003.08744},
  year={2020}
}

@phdthesis{kampchen_feature_2007,
  title={Feature-level fusion of laser scanner and video data for advanced driver assistance systems},
  author={K{\"a}mpchen, Nico},
  year={2007},
  school={Universit{\"a}t Ulm}
}

@article{zhan_interaction_2019,
  title={Interaction dataset: An international, adversarial and cooperative motion dataset in interactive driving scenarios with semantic maps},
  author={Zhan, Wei and Sun, Liting and Wang, Di and Shi, Haojie and Clausse, Aubrey and Naumann, Maximilian and Kummerle, Julius and Konigshof, Hendrik and Stiller, Christoph and de La Fortelle, Arnaud and others},
  journal={arXiv preprint arXiv:1910.03088},
  year={2019}
}

@book{murphy_machine_2012,
  title={Machine learning: a probabilistic perspective},
  author={Murphy, Kevin P},
  year={2012},
  publisher={MIT press}
}

@inproceedings{ettinger_waymo_2021,
  title={Large scale interactive motion forecasting for autonomous driving: The waymo open motion dataset},
  author={Ettinger, Scott and Cheng, Shuyang and Caine, Benjamin and Liu, Chenxi and Zhao, Hang and Pradhan, Sabeek and Chai, Yuning and Sapp, Ben and Qi, Charles R and Zhou, Yin and others},
  booktitle=ICCV,
  pages={9710--9719},
  year={2021}
}

@inproceedings{phan_covernet_2020,
  title={Covernet: Multimodal behavior prediction using trajectory sets},
  author={Phan-Minh, Tung and Grigore, Elena Corina and Boulton, Freddy A and Beijbom, Oscar and Wolff, Eric M},
  booktitle=CVPR,
  pages={14074--14083},
  year={2020}
}

@inproceedings{reichardt_trajectories_2022,
	title = {Trajectories as Markov-States for Long Term Traffic Scene Prediction},
	author = {Reichardt, J{\"o}rg},
	booktitle = {14-th UniDAS FAS-Workshop},
	pages = {14},
    year = {2022},
}

@inproceedings{su_temporally_2021,
  title={Temporally-Continuous Probabilistic Prediction using Polynomial Trajectory Parameterization},
  author={Su, Zhaoen and Wang, Chao and Cui, Henggang and Djuric, Nemanja and Vallespi-Gonzalez, Carlos and Bradley, David},
  booktitle=IROS,
  pages={3837--3843},
  year={2021},
  organization={IEEE}
}

@misc{opendrive,
  author = {{Association for Standardization of Automation and Measuring Systems (ASAM)}},
  title = {{ASAM} {O}pen{DRIVE}},
  year = {2023},
  howpublished = {\url{https://www.asam.net/standards/detail/opendrive/}},
}

@article{famiglietti_bicycle_2020,
  title={Bicycle Braking Performance Testing and Analysis},
  author={Famiglietti, Nicholas and Nguyen, Benjamin and Fatzinger, Edward and Landerville, Jon},
  journal={SAE International Journal of Advances and Current Practices in Mobility},
  volume={2},
  pages={3384--3397},
  year={2020}
}

@article{bokare_acceleration_2017,
  title={Acceleration-deceleration behaviour of various vehicle types},
  author={Bokare, Prashant Shridhar and Maurya, Akhilesh Kumar},
  journal={Transportation research procedia},
  volume={25},
  pages={4733--4749},
  year={2017},
  publisher={Elsevier}
}

@inproceedings{yao_empirical_2023,
  title={An Empirical Bayes Analysis of Object Trajectory Representation Models},
  author={Yao, Yue and Goehring, Daniel and Reichardt, Joerg},
  booktitle=ITSC,
  pages={902--909},
  year={2023},
}

@article{borges_2002_total,
  title={Total least squares fitting of B{\'e}zier and B-spline curves to ordered data},
  author={Borges, Carlos F and Pastva, Tim},
  journal={Computer Aided Geometric Design},
  volume={19},
  number={4},
  pages={275--289},
  year={2002},
  publisher={Elsevier}
}

@inproceedings{zhou_query_2023,
  title={Query-Centric Trajectory Prediction},
  author={Zhou, Zikang and Wang, Jianping and Li, Yung-Hui and Huang, Yu-Kai},
  booktitle=CVPR,
  pages={17863--17873},
  year={2023}
}

@inproceedings{cheng_forecast_2023,
  title={Forecast-mae: Self-supervised pre-training for motion forecasting with masked autoencoders},
  author={Cheng, Jie and Mei, Xiaodong and Liu, Ming},
  booktitle=ICCV,
  pages={8679--8689},
  year={2023}
}

@article{shi_mtr++_2023,
title={Mtr++: Multi-agent motion prediction with symmetric scene modeling and guided intention querying},
  author={Shi, Shaoshuai and Jiang, Li and Dai, Dengxin and Schiele, Bernt},
  journal={IEEE Transactions on Pattern Analysis and Machine Intelligence},
  volume={46},
  number={5},
  pages={3955--3971},
  year={2024},
}

@inproceedings{zhou_hivt_2022,
  title={Hivt: Hierarchical vector transformer for multi-agent motion prediction},
  author={Zhou, Zikang and Ye, Luyao and Wang, Jianping and Wu, Kui and Lu, Kejie},
  booktitle=CVPR,
  pages={8823--8833},
  year={2022}
}

@inproceedings{vaswani_attention_2017,
  title={Attention is all you need},
  author={Vaswani, Ashish and Shazeer, Noam and Parmar, Niki and Uszkoreit, Jakob and Jones, Llion and Gomez, Aidan N and Kaiser, {\L}ukasz and Polosukhin, Illia},
  booktitle=NeurIPS,
  volume={30},
  year={2017}
}

@inproceedings{xiong_layer_2020,
  title={On layer normalization in the transformer architecture},
  author={Xiong, Ruibin and Yang, Yunchang and He, Di and Zheng, Kai and Zheng, Shuxin and Xing, Chen and Zhang, Huishuai and Lan, Yanyan and Wang, Liwei and Liu, Tieyan},
  booktitle=ICML,
  pages={10524--10533},
  year={2020},
}

@inproceedings{wang_optimizing_2025,
  title={Optimizing diffusion models for joint trajectory prediction and controllable generation},
  author={Wang, Yixiao and Tang, Chen and Sun, Lingfeng and Rossi, Simone and Xie, Yichen and Peng, Chensheng and Hannagan, Thomas and Sabatini, Stefano and Poerio, Nicola and Tomizuka, Masayoshi and others},
  booktitle=ECCV,
  pages={324--341},
  year={2025},
}

@inproceedings{montali_waymo_2024,
  title={The waymo open sim agents challenge},
  author={Montali, Nico and Lambert, John and Mougin, Paul and Kuefler, Alex and Rhinehart, Nicholas and Li, Michelle and Gulino, Cole and Emrich, Tristan and Yang, Zoey and Whiteson, Shimon and others},
  booktitle=NeurIPS,
  volume={36},
  year={2024}
}

@article{zhou_qcnext_2023,
  title={Qcnext: A next-generation framework for joint multi-agent trajectory prediction},
  author={Zhou, Zikang and Wen, Zihao and Wang, Jianping and Li, Yung-Hui and Huang, Yu-Kai},
  journal={arXiv preprint arXiv:2306.10508},
  year={2023}
}

@inproceedings{song_denoising_2021,
  title={Denoising diffusion implicit models},
  author={Song, Jiaming and Meng, Chenlin and Ermon, Stefano},
  booktitle=ICLR,
  year={2021}
}

@inproceedings{bouzidi_motion_2024,
  title={Motion planning under uncertainty: Integrating learning-based multi-modal predictors into branch model predictive control},
  author={Bouzidi, Mohamed-Khalil and Derajic, Bojan and Goehring, Daniel and Reichardt, Joerg},
  booktitle=ITSC,
  pages={2592--2598},
  year={2024}
}

@ARTICLE{mustafa_racp_2024,
  title={RACP: Risk-aware contingency planning with multi-modal predictions},
  author={Mustafa, Khaled A and Ornia, Daniel Jarne and Kober, Jens and Alonso-Mora, Javier},
  journal={IEEE Transactions on Intelligent Vehicles},
  year={2024},
  }

@article{chen_interactive_2022,
  author={Chen, Yuxiao and Rosolia, Ugo and Ubellacker, Wyatt and Csomay-Shanklin, Noel and Ames, Aaron D.},
  journal={IEEE Robotics and Automation Letters}, 
  title={Interactive Multi-Modal Motion Planning With Branch Model Predictive Control}, 
  year={2022},
  volume={7},
  number={2},
  pages={5365-5372},
  doi={10.1109/LRA.2022.3156648}}

@inproceedings{mao_leapfrog_2023,
  title={Leapfrog diffusion model for stochastic trajectory prediction},
  author={Mao, Weibo and Xu, Chenxin and Zhu, Qi and Chen, Siheng and Wang, Yanfeng},
  booktitle=CVPR,
  pages={5517--5526},
  year={2023}
}

@inproceedings{ho_denoising_2020,
  title={Denoising diffusion probabilistic models},
  author={Ho, Jonathan and Jain, Ajay and Abbeel, Pieter},
  booktitle=NeurIPS,
  volume={33},
  pages={6840--6851},
  year={2020}
}

@inproceedings{yao_improving_2024,
  title={Improving Out-of-Distribution Generalization of Trajectory Prediction for Autonomous Driving via Polynomial Representations},
  author={Yao, Yue and Yan, Shengchao and Goehring, Daniel and Burgard, Wolfram and Reichardt, Joerg},
  booktitle=IROS,
  pages={488--495},
  year={2024},
  organization={IEEE}
}

@inproceedings{jiang_motiondiffuser_2023,
  title={Motiondiffuser: Controllable multi-agent motion prediction using diffusion},
  author={Jiang, Chiyu and Cornman, Andre and Park, Cheolho and Sapp, Benjamin and Zhou, Yin and Anguelov, Dragomir and others},
  booktitle=CVPR,
  pages={9644--9653},
  year={2023}
}

@inproceedings{choi_dice_2024,
  title={Dice: Diverse diffusion model with scoring for trajectory prediction},
  author={Choi, Younwoo and Mercurius, Ray Coden and Shabestary, Soheil Mohamad Alizadeh and Rasouli, Amir},
  booktitle=IV,
  pages={3023--3029},
  year={2024},
  organization={IEEE}
}

@inproceedings{seff_motionlm_2023,
  title={Motionlm: Multi-agent motion forecasting as language modeling},
  author={Seff, Ari and Cera, Brian and Chen, Dian and Ng, Mason and Zhou, Aurick and Nayakanti, Nigamaa and Refaat, Khaled S and Al-Rfou, Rami and Sapp, Benjamin},
  booktitle=ICCV,
  pages={8579--8590},
  year={2023}
}

@inproceedings{rombach_high_2022,
  title={High-resolution image synthesis with latent diffusion models},
  author={Rombach, Robin and Blattmann, Andreas and Lorenz, Dominik and Esser, Patrick and Ommer, Bj{\"o}rn},
  booktitle=ICCV,
  pages={10684--10695},
  year={2022}
}

@inproceedings{somepalli_understanding_2023,
  title={Understanding and mitigating copying in diffusion models},
  author={Somepalli, Gowthami and Singla, Vasu and Goldblum, Micah and Geiping, Jonas and Goldstein, Tom},
  booktitle=NeurIPS,
  volume={36},
  pages={47783--47803},
  year={2023}
}

@inproceedings{brown_language_2020,
  title={Language models are few-shot learners},
  author={Brown, Tom and Mann, Benjamin and Ryder, Nick and Subbiah, Melanie and Kaplan, Jared D and Dhariwal, Prafulla and Neelakantan, Arvind and Shyam, Pranav and Sastry, Girish and Askell, Amanda and others},
  booktitle=NeurIPS,
  volume={33},
  pages={1877--1901},
  year={2020}
}

@article{kesting_enhanced_2010,
  title={Enhanced intelligent driver model to access the impact of driving strategies on traffic capacity},
  author={Kesting, Arne and Treiber, Martin and Helbing, Dirk},
  journal={Philosophical Transactions of the Royal Society A: Mathematical, Physical and Engineering Sciences},
  volume={368},
  number={1928},
  pages={4585--4605},
  year={2010},
  publisher={The Royal Society Publishing}
}

@article{kesting_general_2007,
  title={General lane-changing model MOBIL for car-following models},
  author={Kesting, Arne and Treiber, Martin and Helbing, Dirk},
  journal={Transportation Research Record},
  volume={1999},
  number={1},
  pages={86--94},
  year={2007},
  publisher={SAGE Publications Sage CA}
}

@article{liu_trajectorycnn_2020,
  title={Trajectorycnn: a new spatio-temporal feature learning network for human motion prediction},
  author={Liu, Xiaoli and Yin, Jianqin and Liu, Jin and Ding, Pengxiang and Liu, Jun and Liu, Huaping},
  journal={IEEE Transactions on Circuits and Systems for Video Technology},
  volume={31},
  number={6},
  pages={2133--2146},
  year={2020},
}

@article{makansi_you_2021,
  title={You mostly walk alone: Analyzing feature attribution in trajectory prediction},
  author={Makansi, Osama and Von K{\"u}gelgen, Julius and Locatello, Francesco and Gehler, Peter and Janzing, Dominik and Brox, Thomas and Sch{\"o}lkopf, Bernhard},
  journal={arXiv preprint arXiv:2110.05304},
  year={2021}
}

@inproceedings{krizhevsky_imagenet_2012,
  title={ImageNet classification with deep convolutional neural networks},
  author={Krizhevsky, Alex and Sutskever, Ilya and Hinton, Geoffrey E},
  booktitle=NeurIPS,
  volume={25},
  pages={1097--1105},
  year={2012}
}

@inproceedings{he_deep_2016,
  title={Deep Residual Learning for Image Recognition},
  author={He, Kaiming and Zhang, Xiangyu and Ren, Shaoqing and Sun, Jian},
  booktitle=CVPR,
  year={2016},
  pages={770--778}
}

@article{ba_layer_2016,
  title={Layer Normalization},
  author={Ba, Jimmy Lei and Kiros, Jamie Ryan and Hinton, Geoffrey E},
  journal={arXiv preprint arXiv:1607.06450},
  year={2016}
}

@inproceedings{gupta_social_2018,
  title={Social gan: Socially acceptable trajectories with generative adversarial networks},
  author={Gupta, Agrim and Johnson, Justin and Fei-Fei, Li and Savarese, Silvio and Alahi, Alexandre},
  booktitle=CVPR,
  pages={2255--2264},
  year={2018}
}

@inproceedings{sadeghian_sophie_2019,
  title={Sophie: An attentive gan for predicting paths compliant to social and physical constraints},
  author={Sadeghian, Amir and Kosaraju, Vineet and Sadeghian, Ali and Hirose, Noriaki and Rezatofighi, Hamid and Savarese, Silvio},
  booktitle=CVPR,
  pages={1349--1358},
  year={2019}
}

@inproceedings{wu_smart_2024,
  title={SMART: scalable multi-agent real-time motion generation via next-token prediction},
  author={Wu, Wei and Feng, Xiaoxin and Gao, Ziyan and Kan, Yuheng},
  booktitle=NeurIPS,
  volume={37},
  pages={114048--114071},
  year={2024}
}

@inproceedings{sohl_deep_2015,
  title={Deep unsupervised learning using nonequilibrium thermodynamics},
  author={Sohl-Dickstein, Jascha and Weiss, Eric and Maheswaranathan, Niru and Ganguli, Surya},
  booktitle=ICML,
  pages={2256--2265},
  year={2015},
  organization={pmlr}
}

@article{hu_gaia_2023,
  title={Gaia-1: A generative world model for autonomous driving},
  author={Hu, Anthony and Russell, Lloyd and Yeo, Hudson and Murez, Zak and Fedoseev, George and Kendall, Alex and Shotton, Jamie and Corrado, Gianluca},
  journal={arXiv preprint arXiv:2309.17080},
  year={2023}
}

@article{kullback_information_1951,
  title={On information and sufficiency},
  author={Kullback, Solomon and Leibler, Richard A},
  journal={The annals of mathematical statistics},
  volume={22},
  number={1},
  pages={79--86},
  year={1951},
  publisher={JSTOR}
}

@inproceedings{alahi_social_2016,
    title = {Social {LSTM}: {Human} {Trajectory} {Prediction} in {Crowded} {Spaces}},
    doi = {10.1109/CVPR.2016.110},
    booktitle = CVPR,

    author = {Alahi, Alexandre and Goel, Kratarth and Ramanathan, Vignesh and Robicquet, Alexandre and Fei-Fei, Li and Savarese, Silvio},

    year = {2016},
    pages = {961--971},
}

@inproceedings{krajewski_highd_2018,
  title={The highd dataset: A drone dataset of naturalistic vehicle trajectories on german highways for validation of highly automated driving systems},
  author={Krajewski, Robert and Bock, Julian and Kloeker, Laurent and Eckstein, Lutz},
  booktitle=ITSC,
  pages={2118--2125},
  year={2018},
}

@inproceedings{feng_unitraj_2024,
  title={Unitraj: A unified framework for scalable vehicle trajectory prediction},
  author={Feng, Lan and Bahari, Mohammadhossein and Amor, Kaouther Messaoud Ben and Zablocki, {\'E}loi and Cord, Matthieu and Alahi, Alexandre},
  booktitle=ECCV,
  pages={106--123},
  year={2024},
  organization={Springer}
}

@inproceedings{zhou_behaviorgpt_2024,
  title={Behaviorgpt: Smart agent simulation for autonomous driving with next-patch prediction},
  author={Zhou, Zikang and Haibo, HU and Chen, Xinhong and Wang, Jianping and Guan, Nan and Wu, Kui and Li, Yung-Hui and Huang, Yu-Kai and Xue, Chun Jason},
  booktitle=NeurIPS,
  volume={37},
  pages={79597--79617},
  year={2024}
}

@inproceedings{worrall_harmonic_2017,
  title={Harmonic networks: Deep translation and rotation equivariance},
  author={Worrall, Daniel E and Garbin, Stephan J and Turmukhambetov, Daniyar and Brostow, Gabriel J},
  booktitle=CVPR,
  pages={5028--5037},
  year={2017}
}

@inproceedings{marcos_rotation_2017,
  title={Rotation equivariant vector field networks},
  author={Marcos, Diego and Volpi, Michele and Komodakis, Nikos and Tuia, Devis},
  booktitle=ICCV,
  pages={5048--5057},
  year={2017}
}

@article{hochreiter_long_1997,
  title={Long short-term memory},
  author={Hochreiter, Sepp and Schmidhuber, J{\"u}rgen},
  journal={Neural computation},
  volume={9},
  number={8},
  pages={1735--1780},
  year={1997},
  publisher={MIT press}
}

@inproceedings{roy_vehicle_2019,
  title={Vehicle trajectory prediction at intersections using interaction based generative adversarial networks},
  author={Roy, Debaditya and Ishizaka, Tetsuhiro and Mohan, C Krishna and Fukuda, Atsushi},
  booktitle=ITSC,
  pages={2318--2323},
  year={2019},
}

@article{jia_amp_2024,
  title={Amp: Autoregressive motion prediction revisited with next token prediction for autonomous driving},
  author={Jia, Xiaosong and Shi, Shaoshuai and Chen, Zijun and Jiang, Li and Liao, Wenlong and He, Tao and Yan, Junchi},
  journal={arXiv preprint arXiv:2403.13331},
  year={2024}
}

@inproceedings{lucas_understanding_2019,
  title={Understanding posterior collapse in generative latent variable models},
  author={Lucas, James and Tucker, George and Grosse, Roger and Norouzi, Mohammad},
  booktitle={International Conference on Learning Representations (ICLR) Workshop DeepGenStruct},
  year={2019}
}

@inproceedings{salimans_improved_2016,
  title={Improved techniques for training gans},
  author={Salimans, Tim and Goodfellow, Ian and Zaremba, Wojciech and Cheung, Vicki and Radford, Alec and Chen, Xi},
  booktitle=NeurIPS,
  volume={29},
  year={2016}
}

@article{khan_transformers_2022,
  title={Transformers in vision: A survey},
  author={Khan, Salman and Naseer, Muzammal and Hayat, Munawar and Zamir, Syed Waqas and Khan, Fahad Shahbaz and Shah, Mubarak},
  journal={ACM computing surveys (CSUR)},
  volume={54},
  number={10s},
  pages={1--41},
  year={2022},
  publisher={ACM}
}

@inproceedings{touvron_training_2021,
  title={Training data-efficient image transformers \& distillation through attention},
  author={Touvron, Hugo and Cord, Matthieu and Douze, Matthijs and Massa, Francisco and Sablayrolles, Alexandre and J{\'e}gou, Herv{\'e}},
  booktitle=ICML,
  pages={10347--10357},
  year={2021},
}

@article{lambert_performance_2020,
  title={Performance analysis of 10 models of 3D LiDARs for automated driving},
  author={Lambert, Jacob and Carballo, Alexander and Cano, Abraham Monrroy and Narksri, Patiphon and Wong, David and Takeuchi, Eijiro and Takeda, Kazuya},
  journal={IEEE Access},
  volume={8},
  pages={131699--131722},
  year={2020},
}

@article{zkebala_pedestrian_2012,
  title={Pedestrian acceleration and speeds},
  author={Z{\k{e}}bala, Jakub and Ci{\k{e}}pka, Piotr and Reza, Adam},
  journal={Problems of Forensic Sciences},
  volume={91},
  pages={227--234},
  year={2012}
}

@article{rauch_maximum_1965,
  title={Maximum likelihood estimates of linear dynamic systems},
  author={Rauch, Herbert E and Tung, F and Striebel, Charlotte T},
  journal={AIAA journal},
  volume={3},
  number={8},
  pages={1445--1450},
  year={1965}
}

@article{buda_systematic_2018,
  title={A systematic study of the class imbalance problem in convolutional neural networks},
  author={Buda, Mateusz and Maki, Atsuto and Mazurowski, Maciej A},
  journal={Neural networks},
  volume={106},
  pages={249--259},
  year={2018},
  publisher={Elsevier}
}

@article{mikolov_efficient_2013,
  title={Efficient estimation of word representations in vector space},
  author={Mikolov, Tomas and Chen, Kai and Corrado, Greg and Dean, Jeffrey},
  journal={arXiv preprint arXiv:1301.3781},
  year={2013}
}

@inproceedings{reece_introduction_2010,
  title={An introduction to Gaussian processes for the Kalman filter expert},
  author={Reece, Steven and Roberts, Stephen},
  booktitle={International Conference on Information Fusion},
  pages={1--9},
  year={2010},
}

@article{kalman_new_1960,
  author={Kalman, Rudolph Emil},
  title     = {A New Approach to Linear Filtering and Prediction Problems},
  journal   = {Journal of Basic Engineering},
  volume    = {82},
  number    = {1},
  pages     = {35--45},
  year      = {1960},
  publisher = {ASME}
}

@inproceedings{bouzidi_closing_2025,
  title={Closing the loop: Motion prediction models beyond open-loop benchmarks},
  author={Bouzidi, Mohamed-Khalil and Schlauch, Christian and Scheuerer, Nicole and Yao, Yue and Klein, Nadja and G{\"o}hring, Daniel and others},
  booktitle=ITSC,
  year={2025}
}

@article{jiang_perception_2025,
  title={Perception Characteristics Distance: Measuring Stability and Robustness of Perception System in Dynamic Conditions under a Certain Decision Rule},
  author={Jiang, Boyu and Shi, Liang and Lin, Zhengzhi and Stowe, Loren and Guo, Feng},
  journal={arXiv preprint arXiv:2506.09217},
  year={2025}
}

@inproceedings{huang_uncertainty_2019,
  title={Uncertainty-aware driver trajectory prediction at urban intersections},
  author={Huang, Xin and McGill, Stephen G and Williams, Brian C and Fletcher, Luke and Rosman, Guy},
  booktitle=ICRA,
  pages={9718--9724},
  year={2019}
}

@ARTICLE{yao_ep_2025,
  author={Yao, Yue and Bouzidi, Mohamed-Khalil and Goehring, Daniel and Reichardt, Joerg},
  journal={IEEE Robotics and Automation Letters}, 
  title={EP-Diffuser: An Efficient Diffusion Model for Traffic Scene Generation and Prediction via Polynomial Representations}, 
  year={2025},
  volume={10},
  number={9},
  pages={9478-9485},
  doi={10.1109/LRA.2025.3595042}}

@book{golub_matrix_1996,
  title={Matrix Computations},
  author={Golub, Gene H. and Van Loan, Charles F.},
  year={1996},
  edition={3rd},
  publisher={The Johns Hopkins University Press},
}

@misc{mathworks_borgespastva,
  author = {{MathWorks}},
  title = {{borgespastva.m}},
  year = {2023},
  howpublished = {\url{https://www.mathworks.com/matlabcentral/fileexchange/46406-borgespastva-m}},
}

@inproceedings{sun_m2i_2022,
  title={M2i: From factored marginal trajectory prediction to interactive prediction},
  author={Sun, Qiao and Huang, Xin and Gu, Junru and Williams, Brian C and Zhao, Hang},
  booktitle=CVPR,
  pages={6543--6552},
  year={2022}
}

@inproceedings{gilles_thomas_2021,
  title={Thomas: Trajectory heatmap output with learned multi-agent sampling},
  author={Gilles, Thomas and Sabatini, Stefano and Tsishkou, Dzmitry and Stanciulescu, Bogdan and Moutarde, Fabien},
  booktitle= ICLR,
  year={2022}
}

@inproceedings{xu_annealed_2024,
  title={Annealed Winner-Takes-All for Motion Forecasting},
  author={Xu, Yihong and Letzelter, Victor and Chen, Micka{\"e}l and Zablocki, {\'E}loi and Cord, Matthieu},
  booktitle=ICRA,
  year={2025}
}

@article{lidard_nashformer_2023,
  title={Nashformer: Leveraging local nash equilibria for semantically diverse trajectory prediction},
  author={Lidard, Justin and So, Oswin and Zhang, Yanxia and DeCastro, Jonathan and Cui, Xiongyi and Huang, Xin and others},
  journal={arXiv preprint arXiv:2305.17600},
  year={2023}
}

@article{huang_diversitygan_2020,
  title={DiversityGAN: Diversity-aware vehicle motion prediction via latent semantic sampling},
  author={Huang, Xin and McGill, Stephen G and DeCastro, Jonathan A and Fletcher, Luke and Leonard, John J and Williams, Brian C and others},
  journal={IEEE Robotics and Automation Letters},
  volume={5},
  number={4},
  pages={5089--5096},
  year={2020},
}

@article{ma_diverse_2020,
  title={Diverse sampling for normalizing flow based trajectory forecasting},
  author={Ma, Yecheng Jason and Inala, Jeevana Priya and Jayaraman, Dinesh and Bastani, Osbert},
  journal={arXiv preprint arXiv:2011.15084},
  year={2020}
}

@inproceedings{luo_jfp_2023,
  title={Jfp: Joint future prediction with interactive multi-agent modeling for autonomous driving},
  author={Luo, Wenjie and Park, Cheol and Cornman, Andre and Sapp, Benjamin and Anguelov, Dragomir},
  booktitle=CORL,
  pages={1457--1467},
  year={2023}
}

@inproceedings{fey_fast_2019,
  title={Fast graph representation learning with PyTorch Geometric},
  author={Fey, Matthias and Lenssen, Jan Eric},
  booktitle={International Conference on Learning Representations (ICLR) Workshop},
  year={2019}
}

@article{cui_gorela_2022,
  title={Gorela: Go relative for viewpoint-invariant motion forecasting},
  author={Cui, Alexander and Casas, Sergio and Wong, Kelvin and Suo, Simon and Urtasun, Raquel},
  journal={arXiv preprint arXiv:2211.02545},
  year={2022}
}

@inproceedings{shiroshita_behaviorally_2020,
  title={Behaviorally diverse traffic simulation via reinforcement learning},
  author={Shiroshita, Shinya and Maruyama, Shirou and Nishiyama, Daisuke and Castro, Mario Ynocente and Hamzaoui, Karim and Rosman, Guy and others},
  booktitle=IROS,
  pages={2103--2110},
  year={2020}
}

@article{kingma_adam_2014,
  title={Adam: A method for stochastic optimization},
  author={Kingma, Diederik P and Ba, Jimmy},
  journal={arXiv preprint arXiv:1412.6980},
  year={2014}
}

@article{loshchilov_decoupled_2017,
  title={Decoupled weight decay regularization},
  author={Loshchilov, Ilya and Hutter, Frank},
  journal={arXiv preprint arXiv:1711.05101},
  year={2017}
}

@article{loshchilov_sgdr_2016,
  title={Sgdr: Stochastic gradient descent with warm restarts},
  author={Loshchilov, Ilya and Hutter, Frank},
  journal={arXiv preprint arXiv:1608.03983},
  year={2016}
}

@article{konstantinidis_marginal_2025,
  title={From Marginal to Joint Predictions: Evaluating Scene-Consistent Trajectory Prediction Approaches for Automated Driving},
  author={Konstantinidis, Fabian and Guerreiro, Ariel Dallari and Trumpp, Raphael and Sackmann, Moritz and Hofmann, Ulrich and Caccamo, Marco and Stiller, Christoph},
  journal={arXiv preprint arXiv:2507.05254},
  year={2025}
}

@inproceedings{zeng_dsdnet_2020,
  title={Dsdnet: Deep structured self-driving network},
  author={Zeng, Wenyuan and Wang, Shenlong and Liao, Renjie and Chen, Yun and Yang, Bin and Urtasun, Raquel},
  booktitle=ECCV,
  pages={156--172},
  year={2020},
}

@inproceedings{bahdanau_neural_2015,
  title={Neural machine translation by jointly learning to align and translate},
  author={Bahdanau, Dzmitry and Cho, Kyunghyun and Bengio, Yoshua},
  booktitle=ICLR,
  year={2015}
}

@inproceedings{zhang_closed_2025,
  title = {Closed-Loop Supervised Fine-Tuning of Tokenized Traffic Models},
  author = {Zhang, Zhejun and Karkus, Peter and Igl, Maximilian and Ding, Wenhao and Chen, Yuxiao and Ivanovic, Boris and Pavone, Marco},
  booktitle = CVPR,
  pages={5422--5432},
  year = {2025},
}

@inproceedings{hu_planning_2023,
  title={Planning-oriented autonomous driving},
  author={Hu, Yihan and Yang, Jiazhi and Chen, Li and Li, Keyu and Sima, Chonghao and Zhu, Xizhou and Chai, Siqi and Du, Senyao and Lin, Tianwei and Wang, Wenhai and others},
  booktitle=CVPR,
  pages={17853--17862},
  year={2023}
}

@inproceedings{sun_sparsedrive_2025,
  title={Sparsedrive: End-to-end autonomous driving via sparse scene representation},
  author={Sun, Wenchao and Lin, Xuewu and Shi, Yining and Zhang, Chuang and Wu, Haoran and Zheng, Sifa},
  booktitle=ICRA,
  pages={8795--8801},
  year={2025},
}

@inproceedings{jiang_vad_2023,
  title={Vad: Vectorized scene representation for efficient autonomous driving},
  author={Jiang, Bo and Chen, Shaoyu and Xu, Qing and Liao, Bencheng and Chen, Jiajie and Zhou, Helong and Zhang, Qian and Liu, Wenyu and Huang, Chang and Wang, Xinggang},
  booktitle=ICCV,
  pages={8340--8350},
  year={2023}
}

@inproceedings{trautman_unfreezing_2010,
  title={Unfreezing the robot: Navigation in dense, interacting crowds},
  author={Trautman, Peter and Krause, Andreas},
  booktitle=IROS,
  pages={797--803},
  year={2010},
}

@article{hagedorn_integration_2024,
  title={The integration of prediction and planning in deep learning automated driving systems: A review},
  author={Hagedorn, Steffen and Hallgarten, Marcel and Stoll, Martin and Condurache, Alexandru Paul},
  journal={IEEE Transactions on Intelligent Vehicles},
  volume={10},
  number={5},
  pages={3626--3643},
  year={2024},
}

@article{lefevre_survey_2014,
  title={A survey on motion prediction and risk assessment for intelligent vehicles},
  author={Lef{\`e}vre, St{\'e}phanie and Vasquez, Dizan and Laugier, Christian},
  journal={ROBOMECH journal},
  volume={1},
  number={1},
  pages={1},
  year={2014},
}

@article{karle_scenario_2022,
  title={Scenario understanding and motion prediction for autonomous vehicles—review and comparison},
  author={Karle, Phillip and Geisslinger, Maximilian and Betz, Johannes and Lienkamp, Markus},
  journal={IEEE Transactions on Intelligent Transportation Systems},
  volume={23},
  number={10},
  pages={16962--16982},
  year={2022},
}

@inproceedings{berndt_continuous_2008,
  title={Continuous driver intention recognition with hidden markov models},
  author={Berndt, Holger and Emmert, Jorg and Dietmayer, Klaus},
  booktitle=ITSC,
  pages={1189--1194},
  year={2008},
}

@inproceedings{gindele_probabilistic_2010,
  title={A probabilistic model for estimating driver behaviors and vehicle trajectories in traffic environments},
  author={Gindele, Tobias and Brechtel, Sebastian and Dillmann, R{\"u}diger},
  booktitle=ITSC,
  pages={1625--1631},
  year={2010},
}

@article{huang_survey_2022,
  title={A survey on trajectory-prediction methods for autonomous driving},
  author={Huang, Yanjun and Du, Jiatong and Yang, Ziru and Zhou, Zewei and Zhang, Lin and Chen, Hong},
  journal={IEEE transactions on intelligent vehicles},
  volume={7},
  number={3},
  pages={652--674},
  year={2022},
}

@string{ICRA = "Proc.~of the IEEE International Conference on Robotics and Automation (ICRA)"}

@string{IROS = "Proc.~of the IEEE/RSJ International Conference on Intelligent Robots and Systems (IROS)"}

@string{CVPR = "Proc.~of the IEEE Conference on Computer Vision and Pattern Recognition (CVPR)"}

@string{ITSC = "Proc.~of the IEEE International Conference on Intelligent Transportation Systems (ITSC)"}

@string{IV = "Proc.~of the IEEE Intelligent Vehicles Symposium (IV)"}

@string{ICLR = "Proc.~of the International Conference on Learning Representations (ICLR)"}

@string{CORL = "Proc.~of Conference on Robot Learning (CoRL)"}

@string{ICML = "Proc.~of the International Conference on Machine Learning (ICML)"}

@string{NeurIPS = "Proc.~of the Conference on Neural Information Processing Systems (NeurIPS)"}



\end{spacing}


\begin{appendices} 


\chapter{Examples of Fitted Trajectories}
\label{appendix: examples of fitted trajectories}
This appendix presents examples of trajectories with high fitting errors, along with randomly sampled trajectories from the Argoverse 1, Argoverse 2, and Waymo Open datasets.

Trajectories with high fitting errors are predominantly outliers caused by tracking issues. In contrast, the randomly sampled trajectories demonstrate that the polynomial representation can approximate real-world trajectories with high fidelity.

\subsubsection{Five 5-seconds vehicle trajectories with highest fit error in A1 (Fitted with $\hat{N} = 3$)}
\centering
\includegraphics[width=\textwidth, height=1.8in]{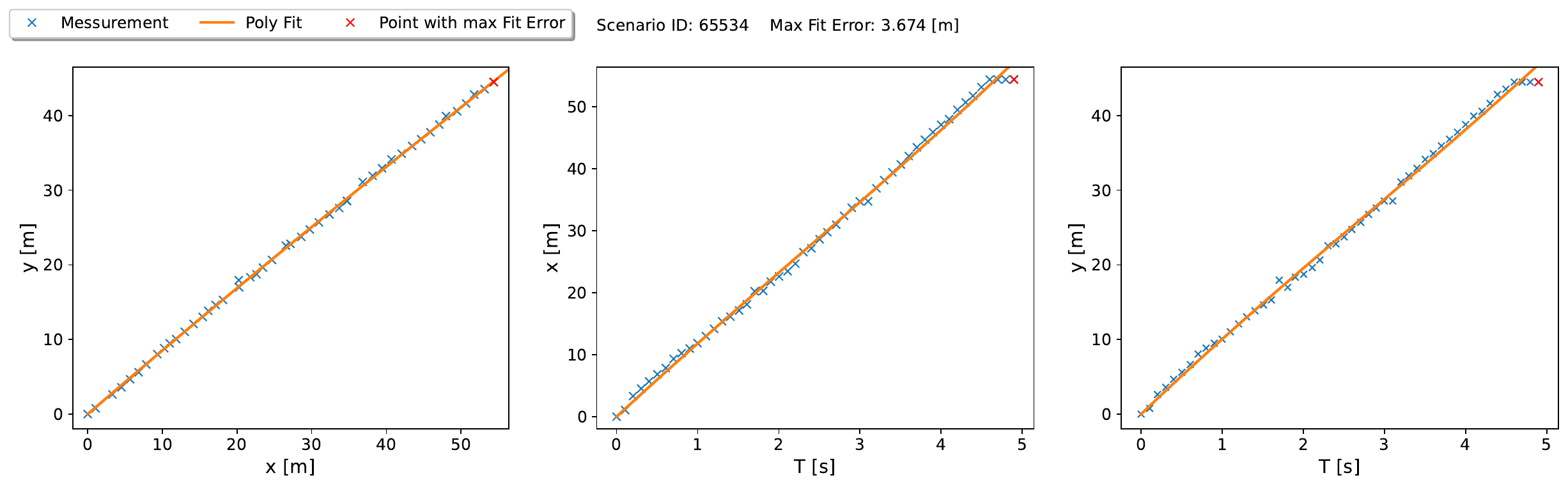} 

\includegraphics[width=\textwidth, height=1.8in]{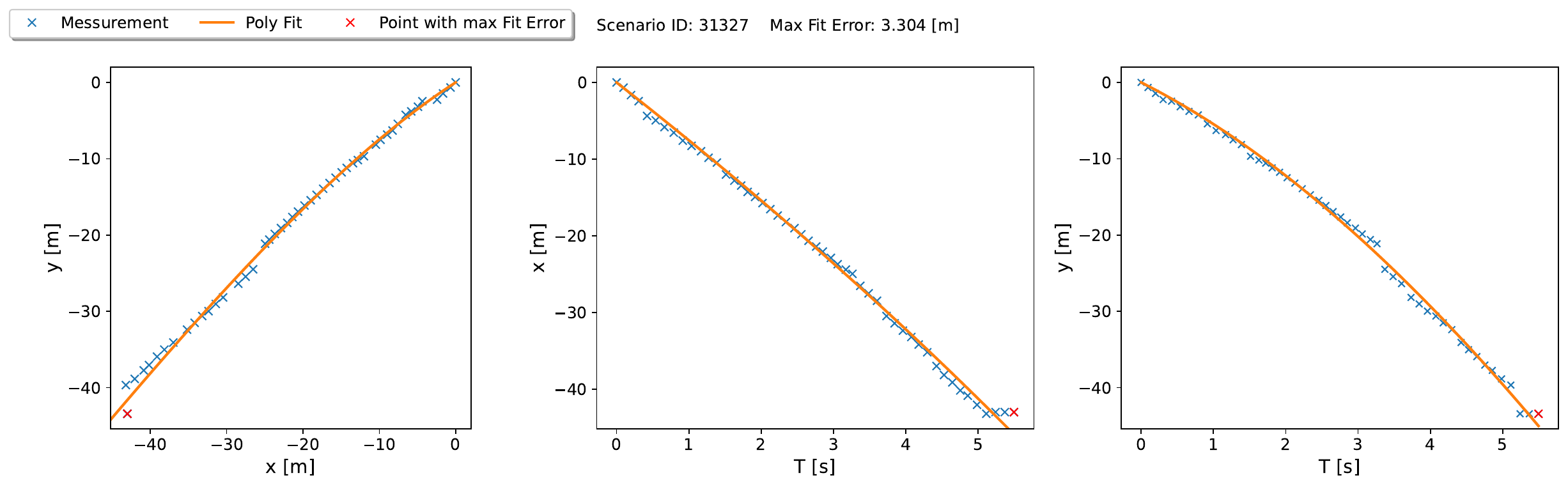} 

\includegraphics[width=\textwidth, height=1.8in]{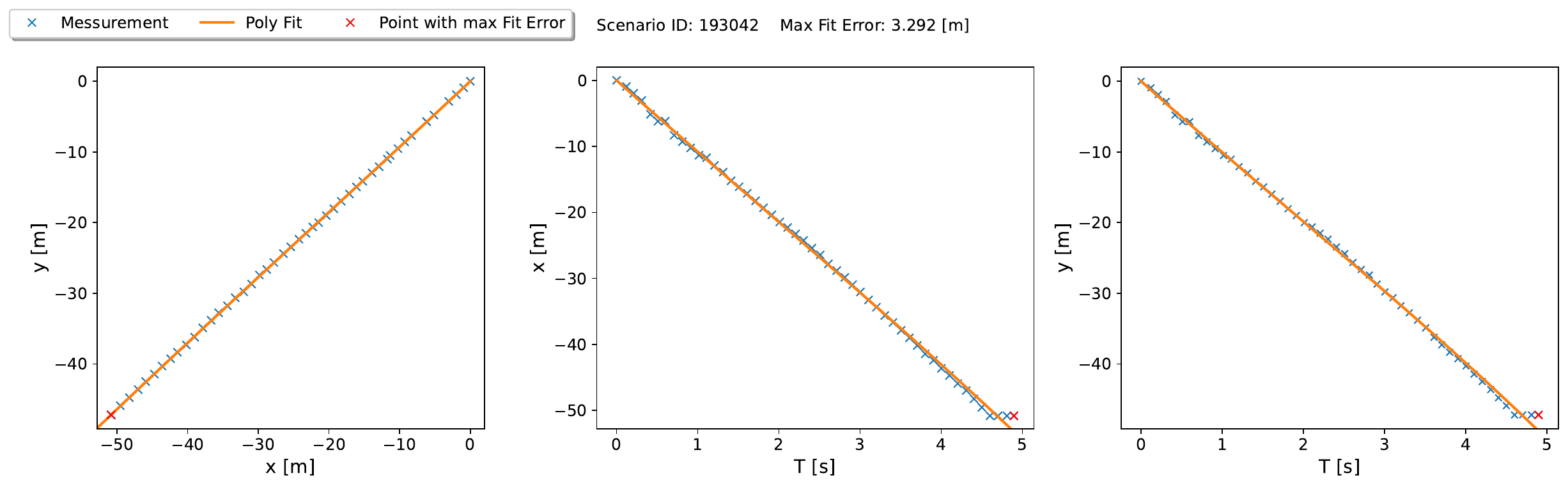} 

\includegraphics[width=\textwidth, height=1.8in]{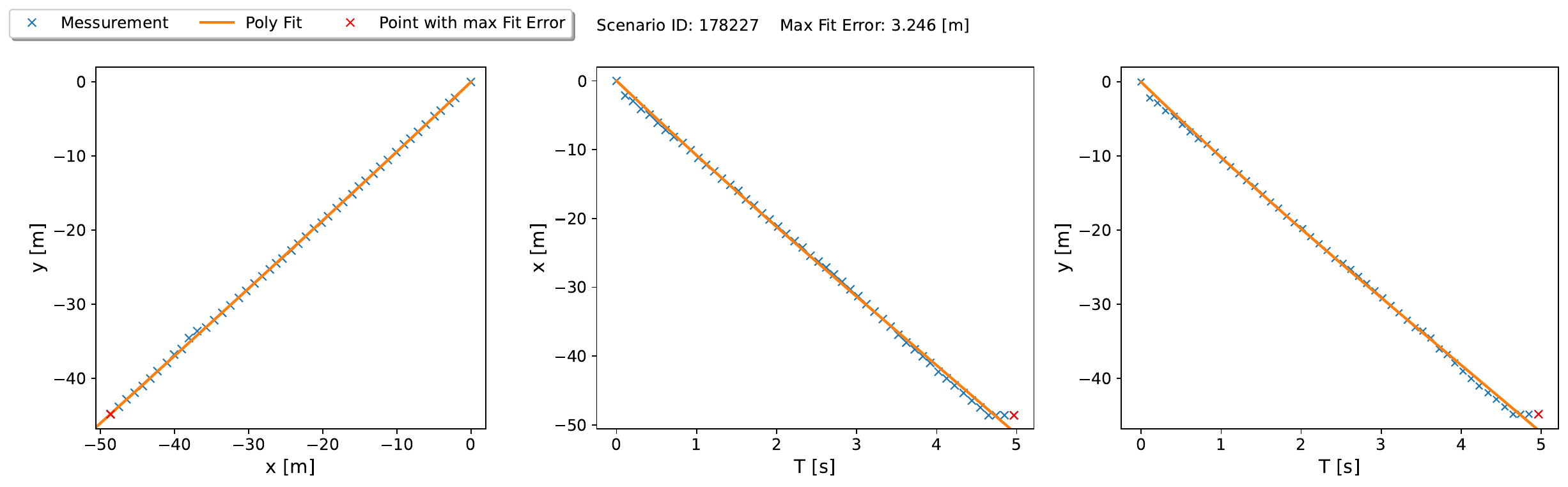} 

\includegraphics[width=\textwidth, height=1.7in]{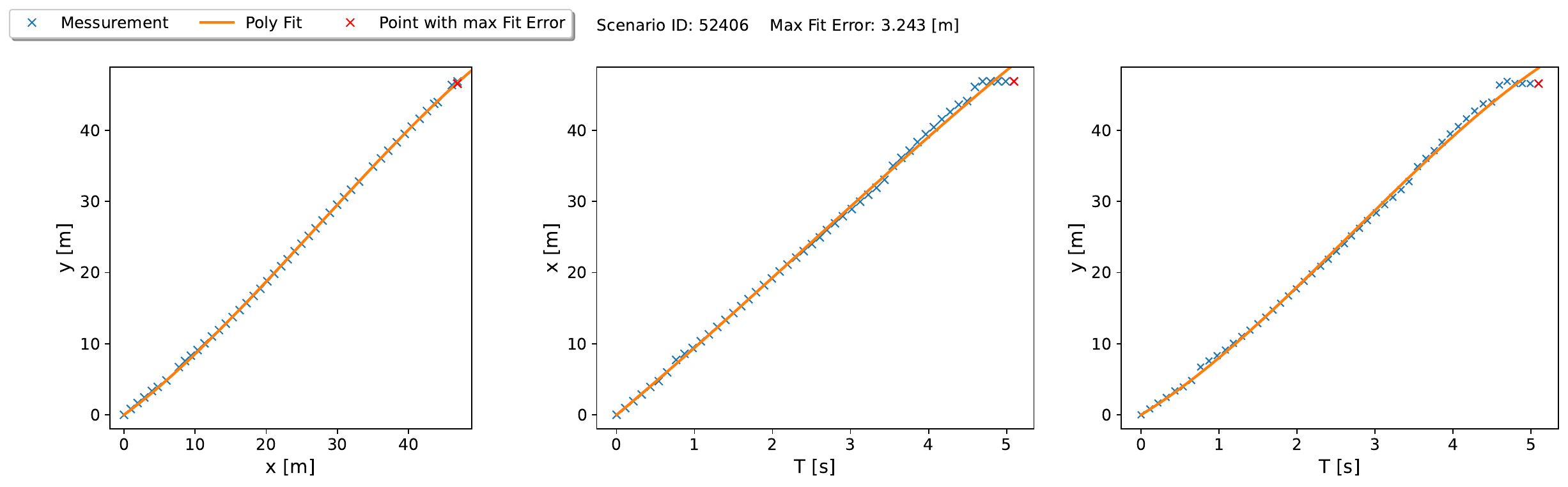}

\subsubsection{Five random 5-seconds vehicle trajectories in A1 (Fitted with $\hat{N} = 3$)}
\centering
\includegraphics[width=\textwidth, height=1.8in]{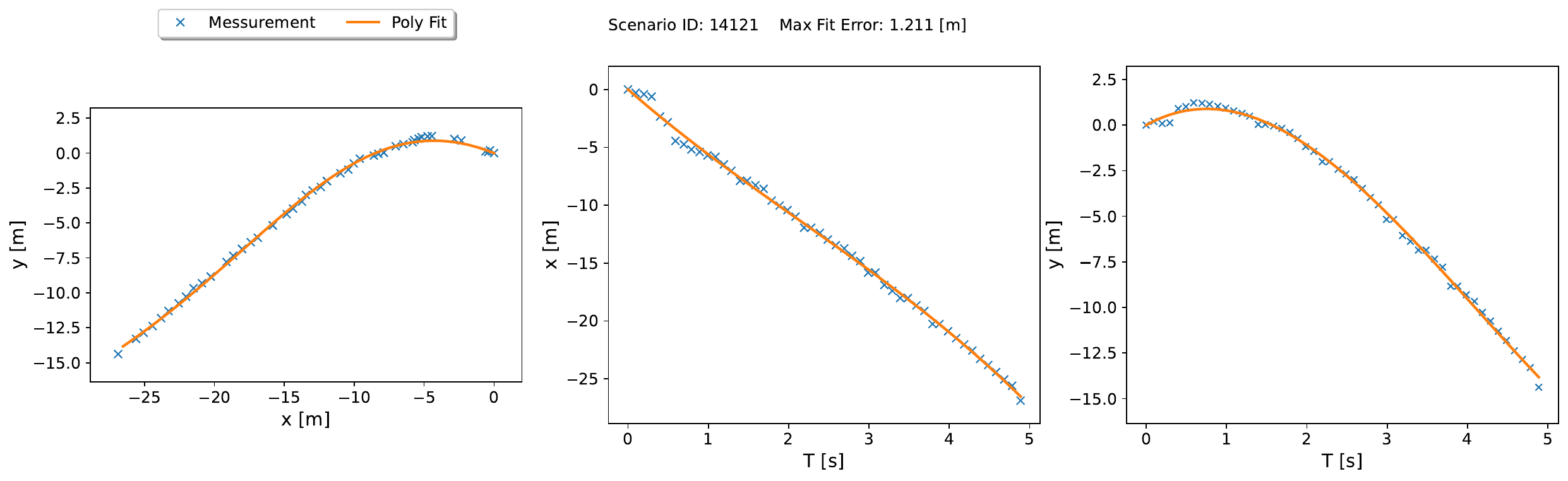} 

\includegraphics[width=\textwidth, height=1.8in]{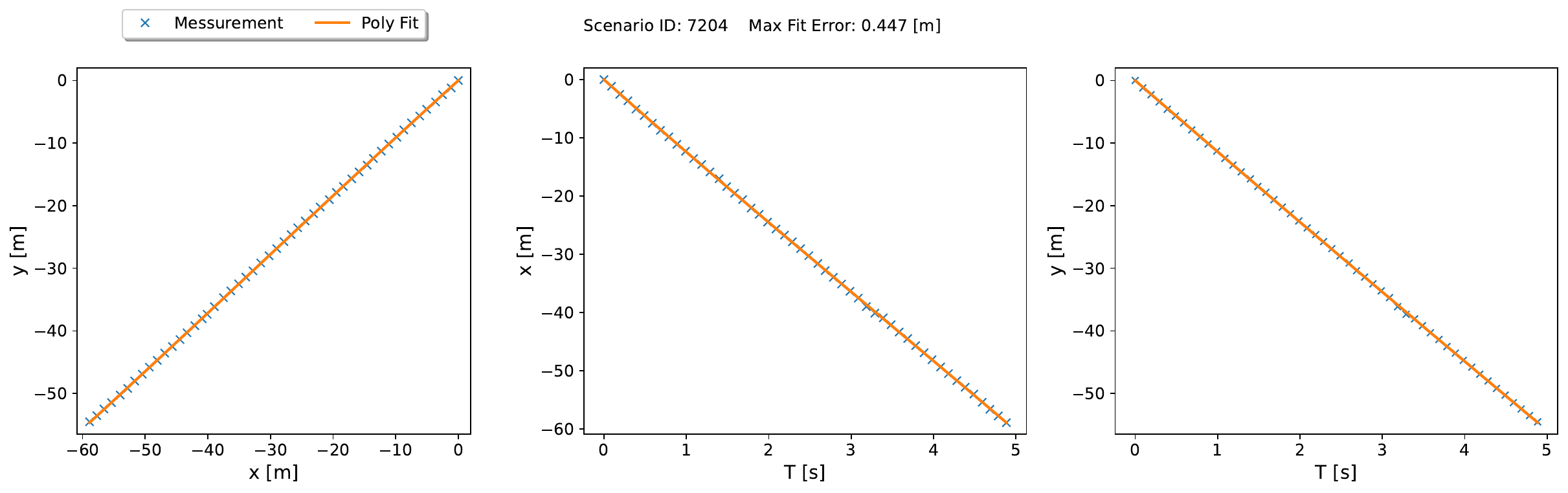} 

\includegraphics[width=\textwidth, height=1.8in]{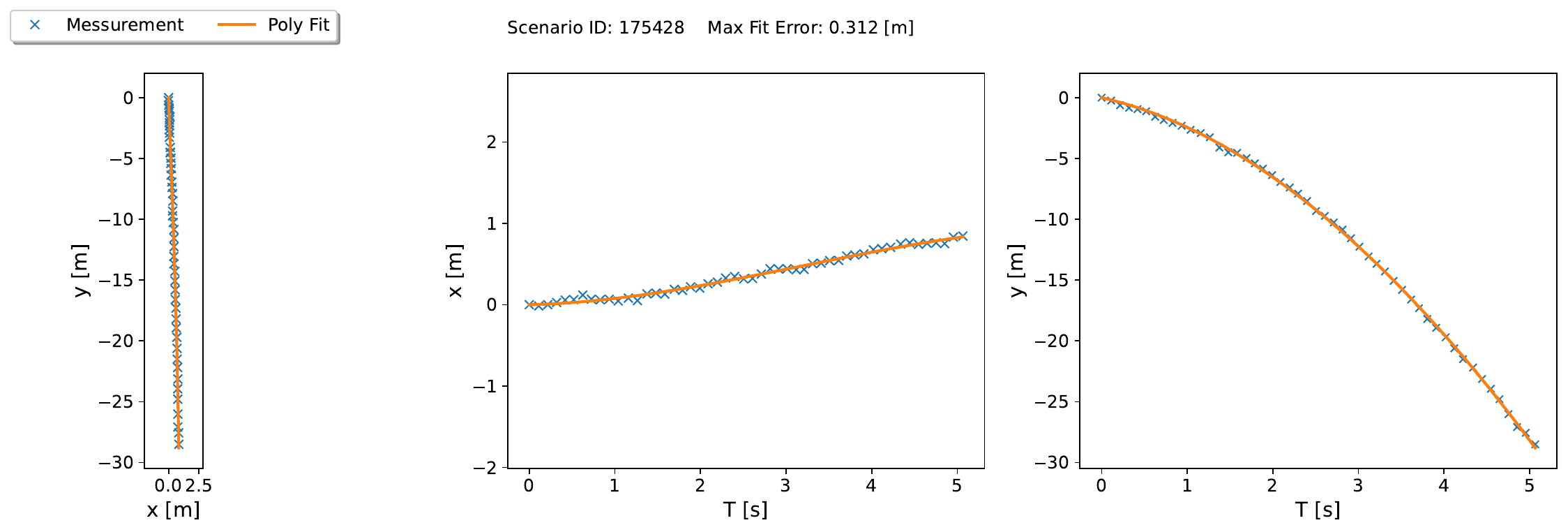} 

\includegraphics[width=\textwidth, height=1.8in]{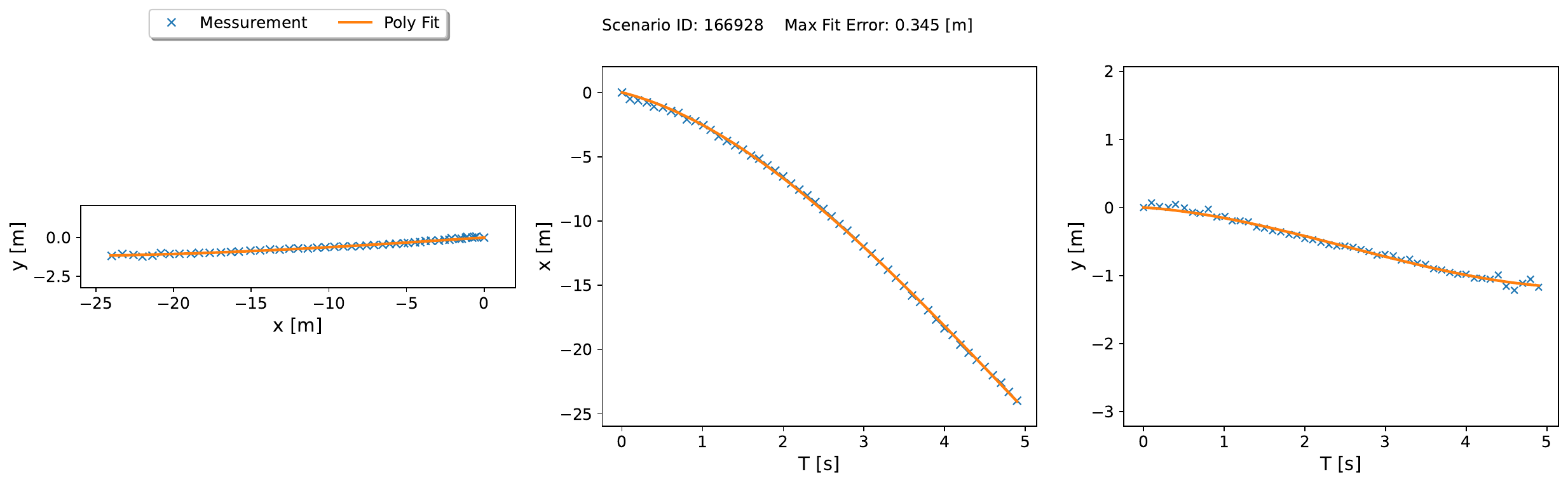} 

\includegraphics[width=\textwidth, height=1.7in]{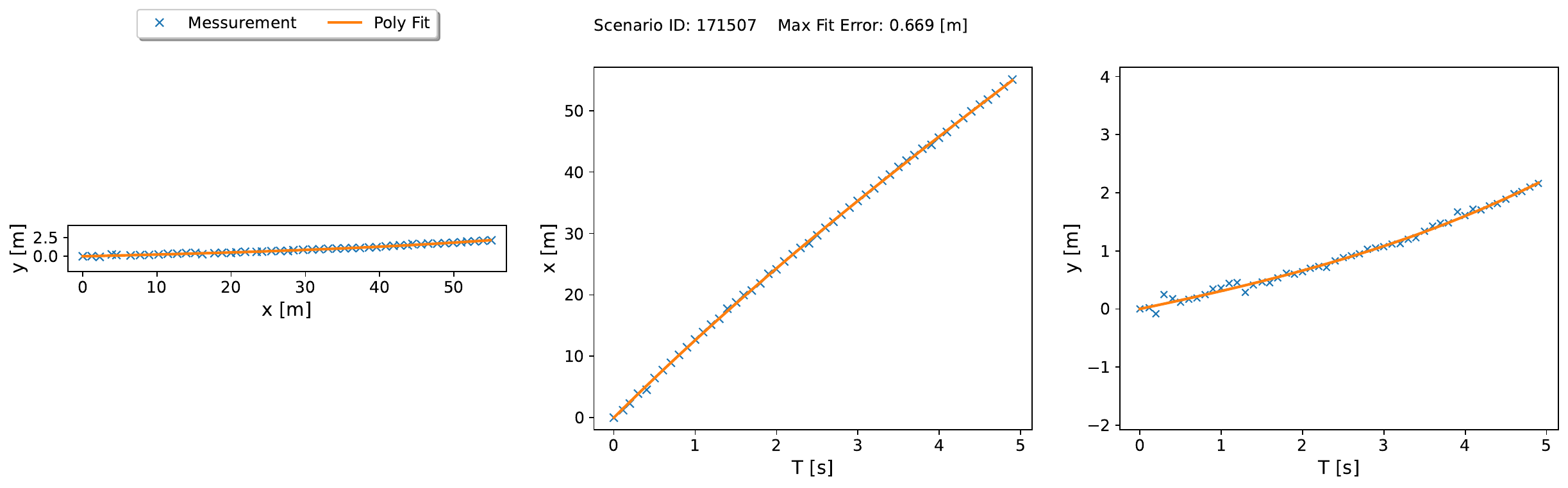} 






\subsubsection{Five 5-seconds vehicle trajectories with highest fit error in A2 (Fitted with $\hat{N} = 5$)}

\centering
\includegraphics[width=\textwidth, height=1.8in]{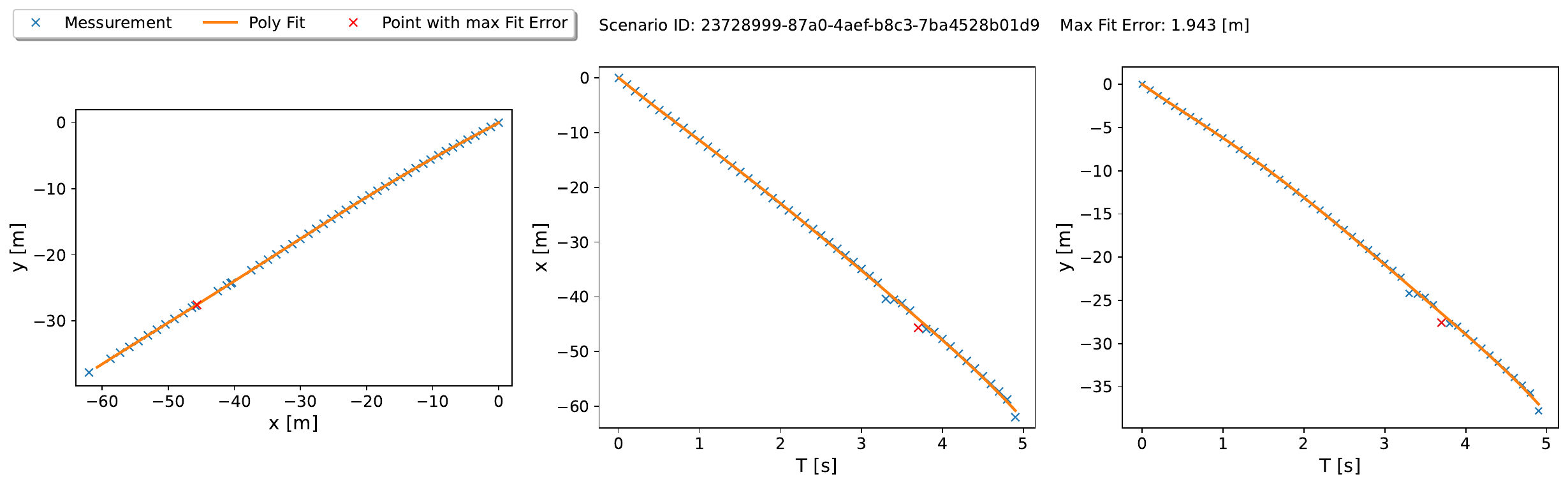} 

\includegraphics[width=\textwidth, height=1.8in]{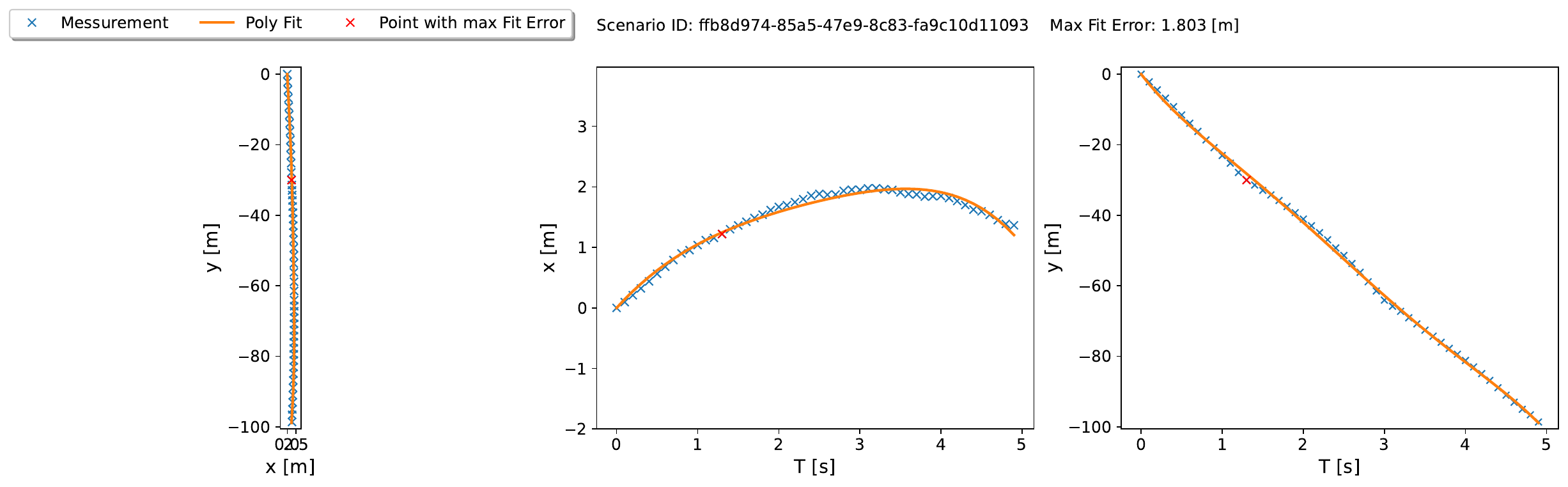}

\includegraphics[width=\textwidth, height=1.8in]{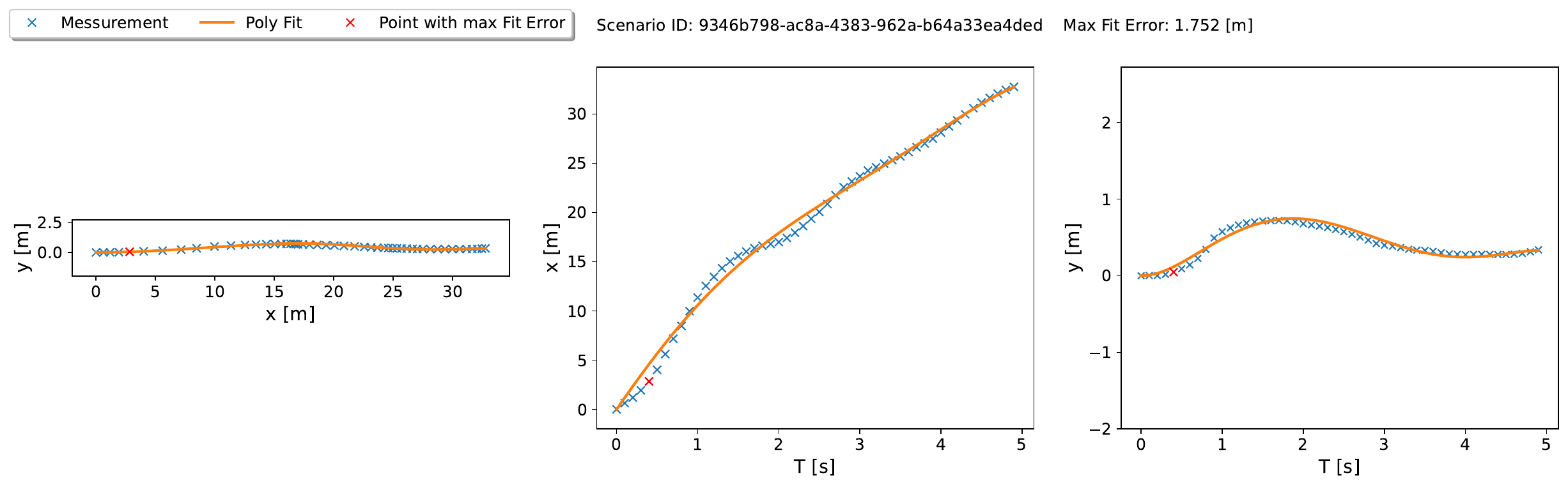} 

\includegraphics[width=\textwidth, height=1.8in]{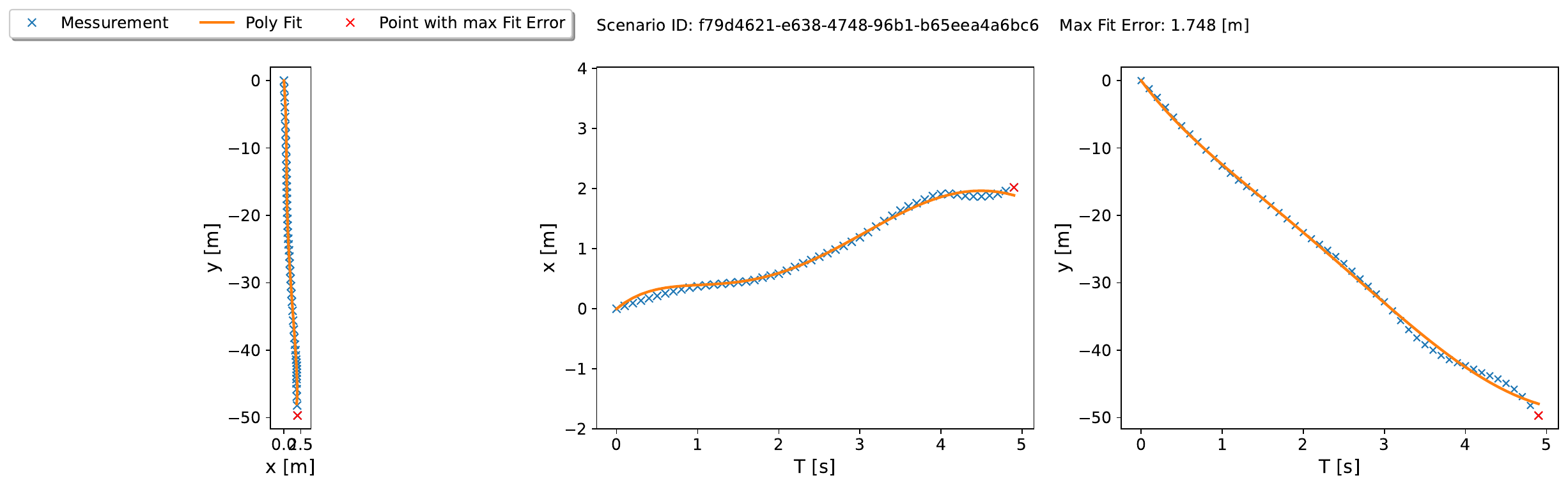} 

\includegraphics[width=\textwidth, height=1.7in]{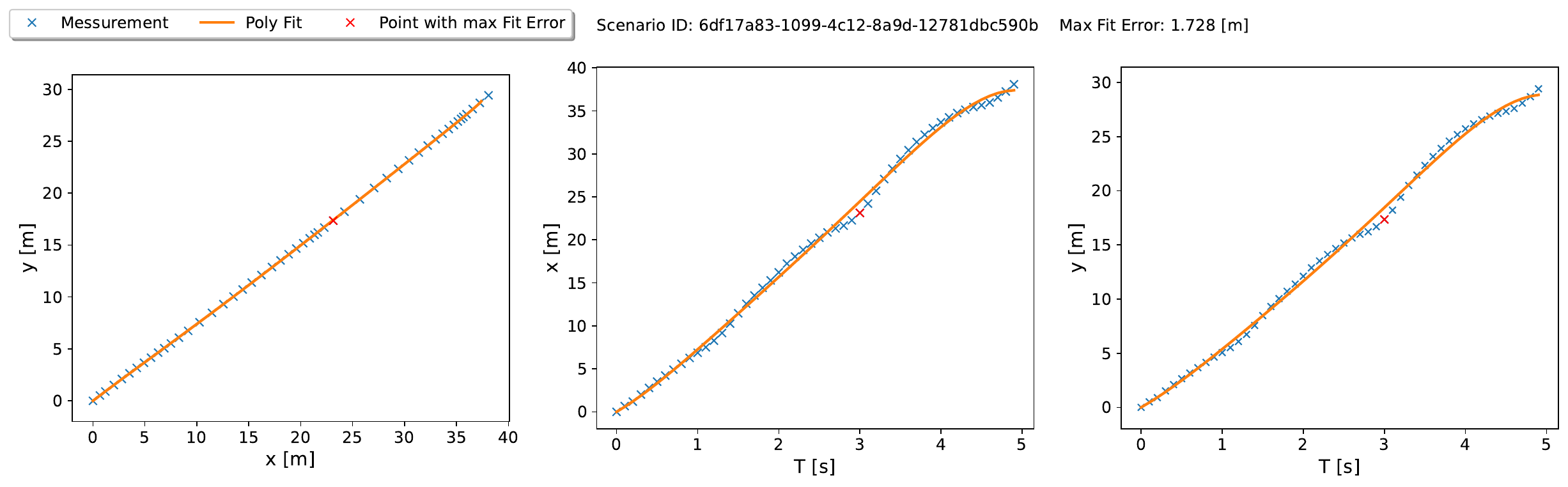}

\subsubsection{Five random 5-seconds vehicle trajectories in A2 (Fitted with $\hat{N} = 5$)}
\centering
\includegraphics[width=\textwidth, height=1.8in]{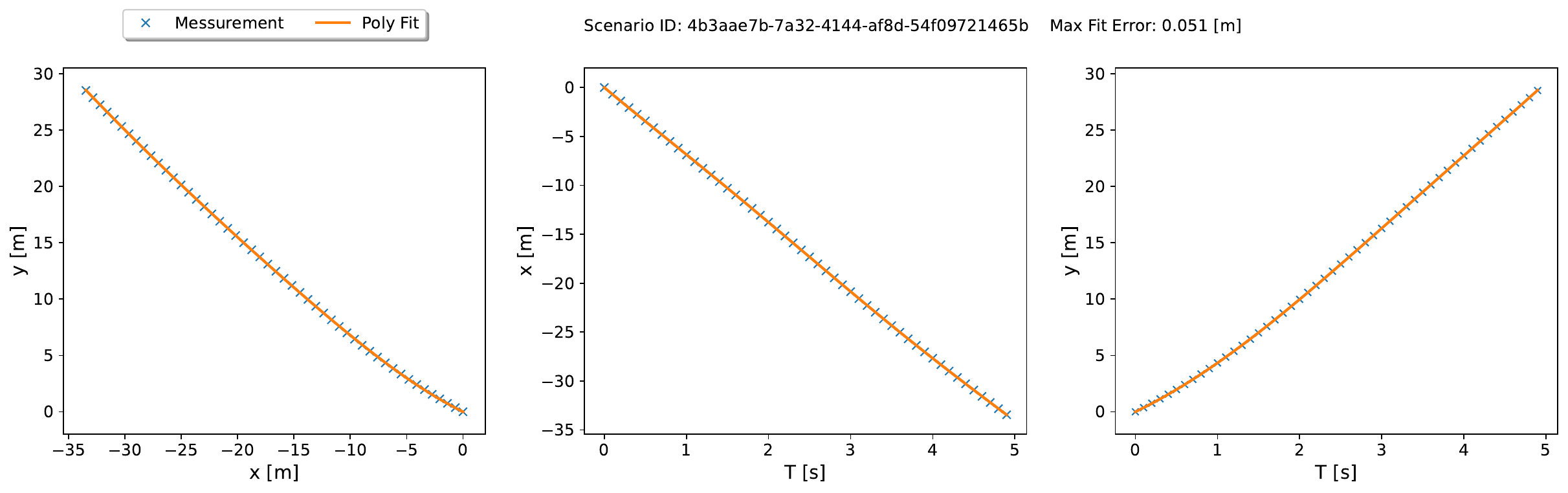} 

\includegraphics[width=\textwidth, height=1.8in]{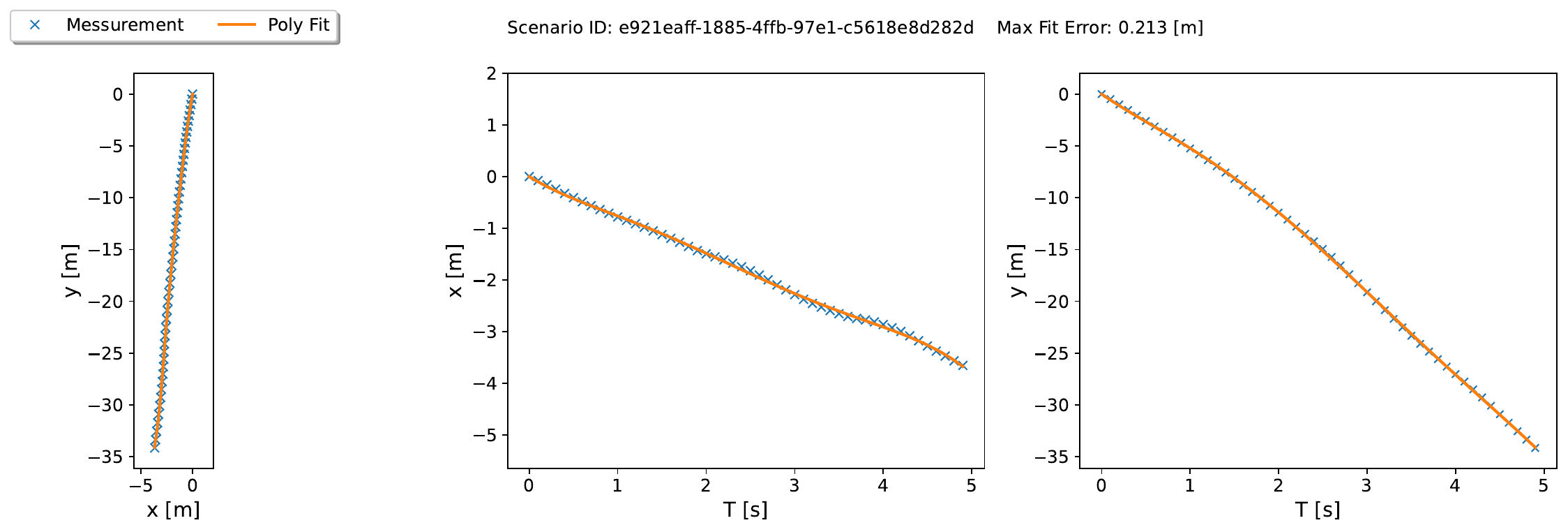} 

\includegraphics[width=\textwidth, height=1.8in]{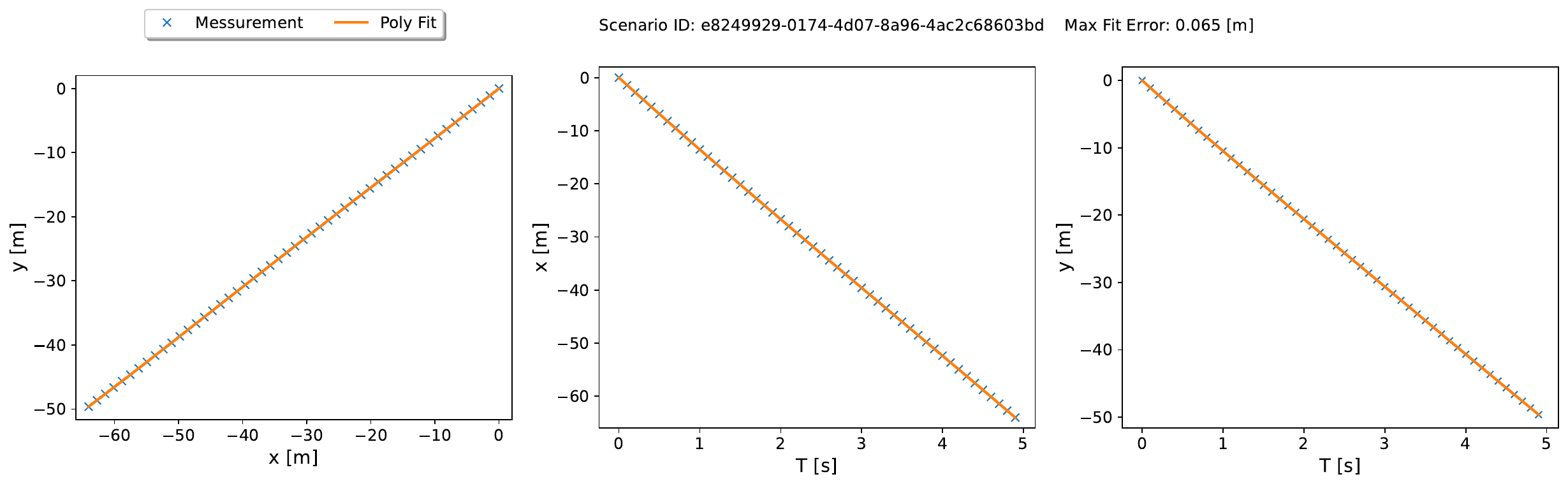} 

\includegraphics[width=\textwidth, height=1.8in]{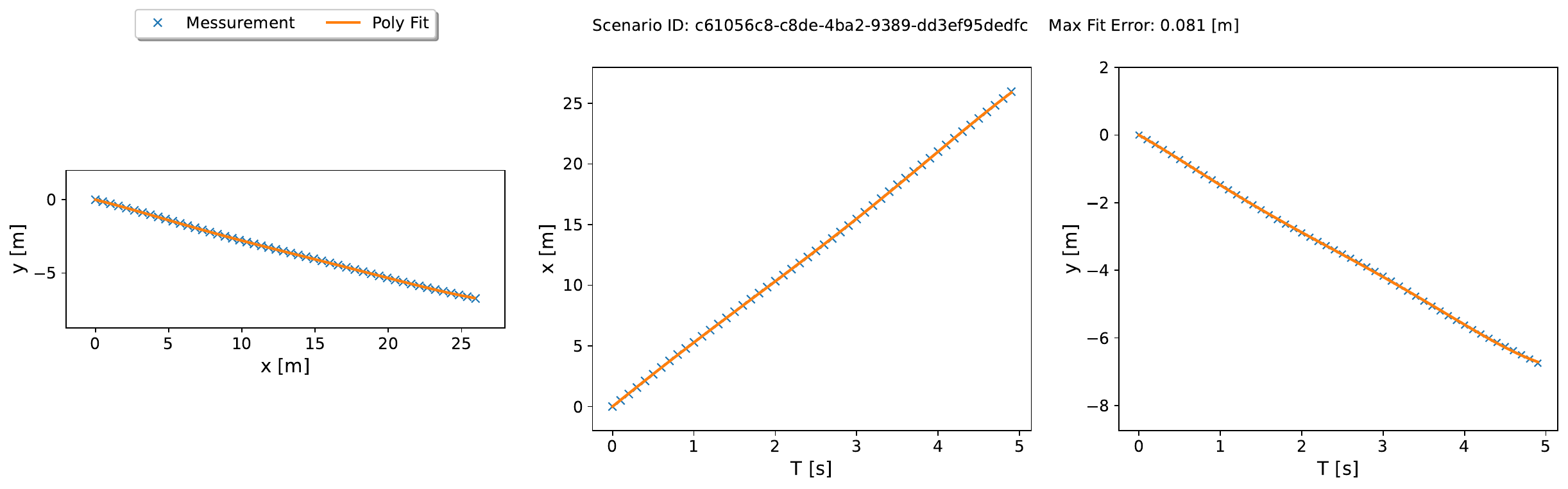} 

\includegraphics[width=\textwidth, height=1.7in]{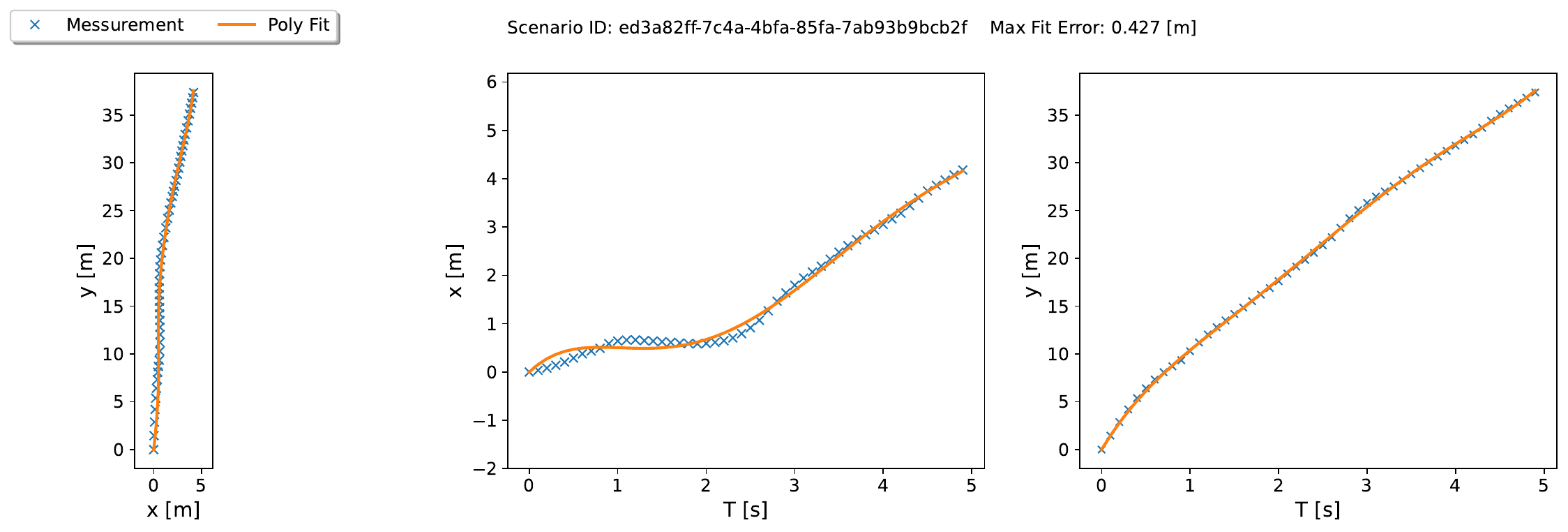}

\subsubsection{Five 5-seconds cyclist trajectories with highest fit error in A2 (Fitted with $\hat{N} = 5$)}
\centering
\includegraphics[width=\textwidth, height=1.8in]{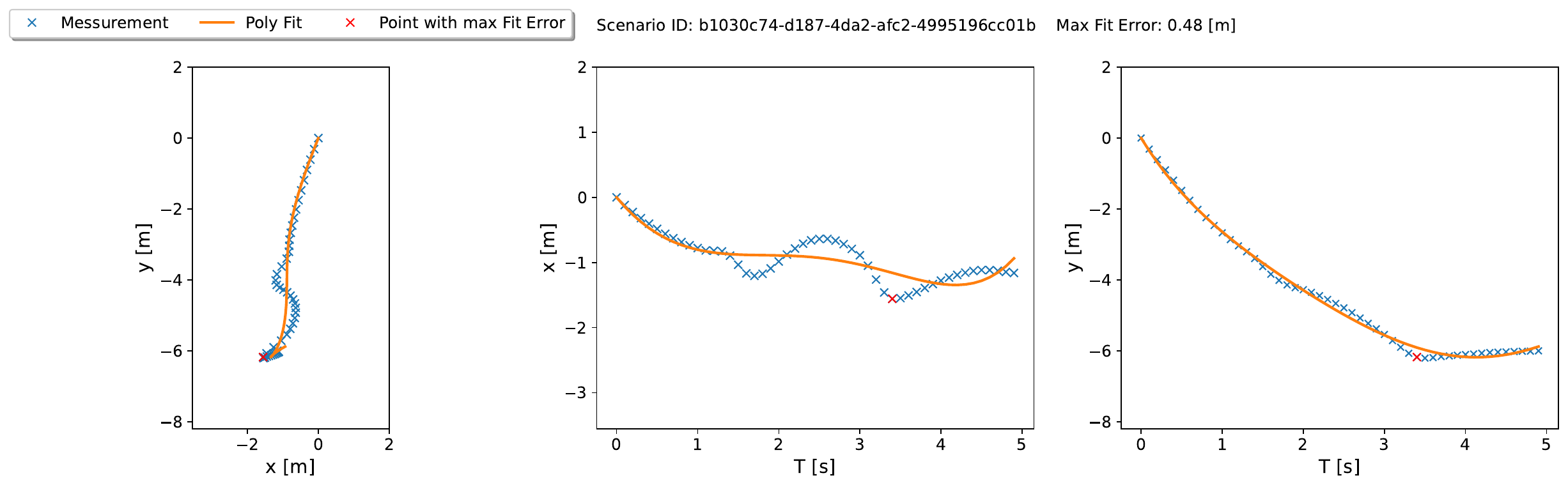} 

\includegraphics[width=\textwidth, height=1.8in]{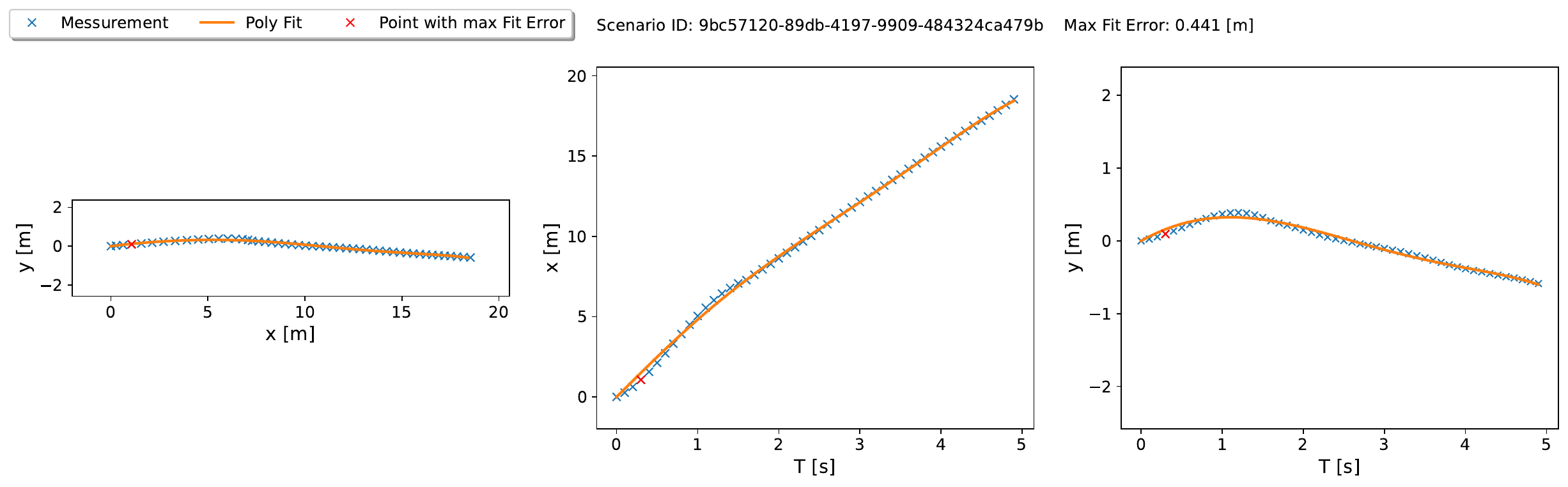} 

\includegraphics[width=\textwidth, height=1.8in]{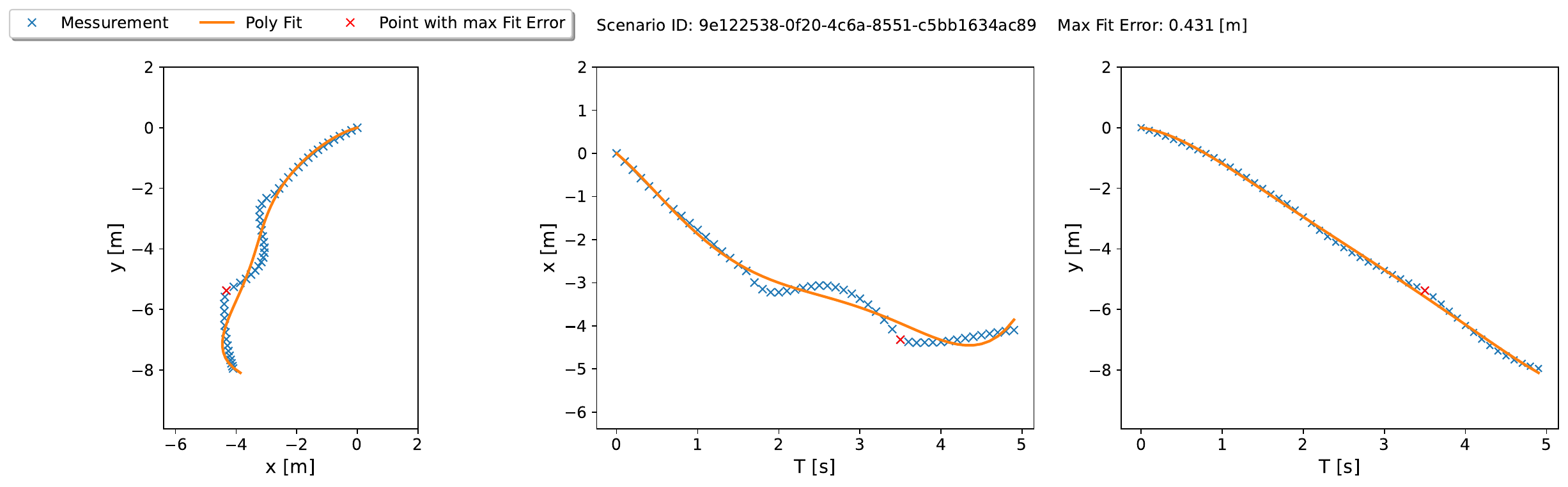} 

\includegraphics[width=\textwidth, height=1.8in]{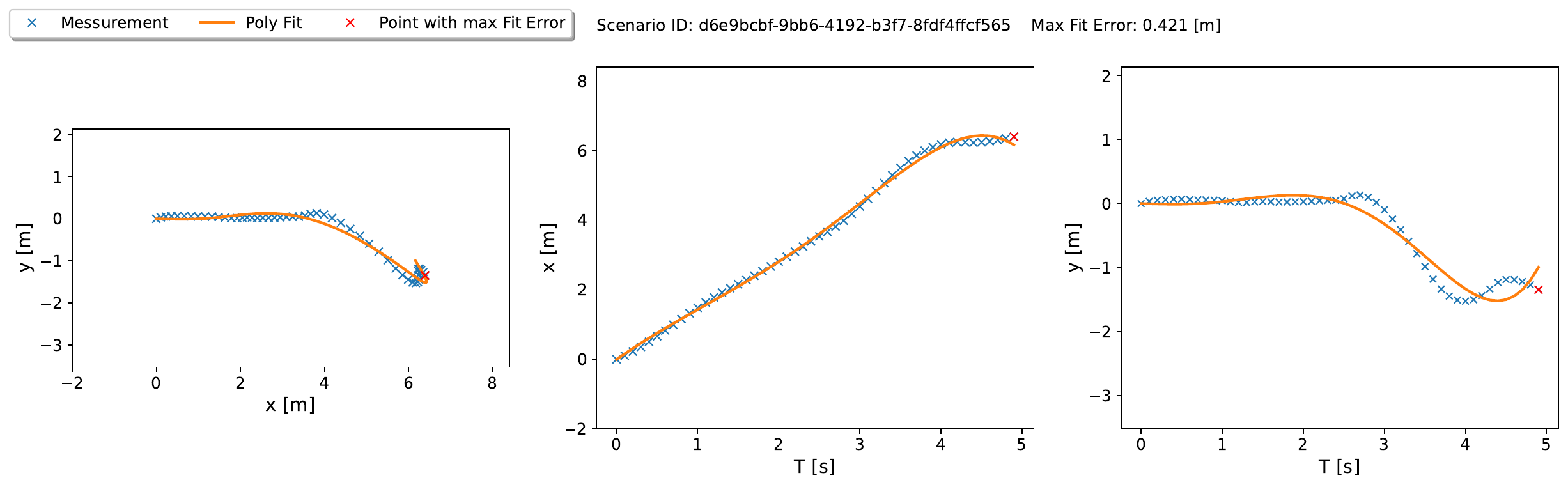} 

\includegraphics[width=\textwidth, height=1.7in]{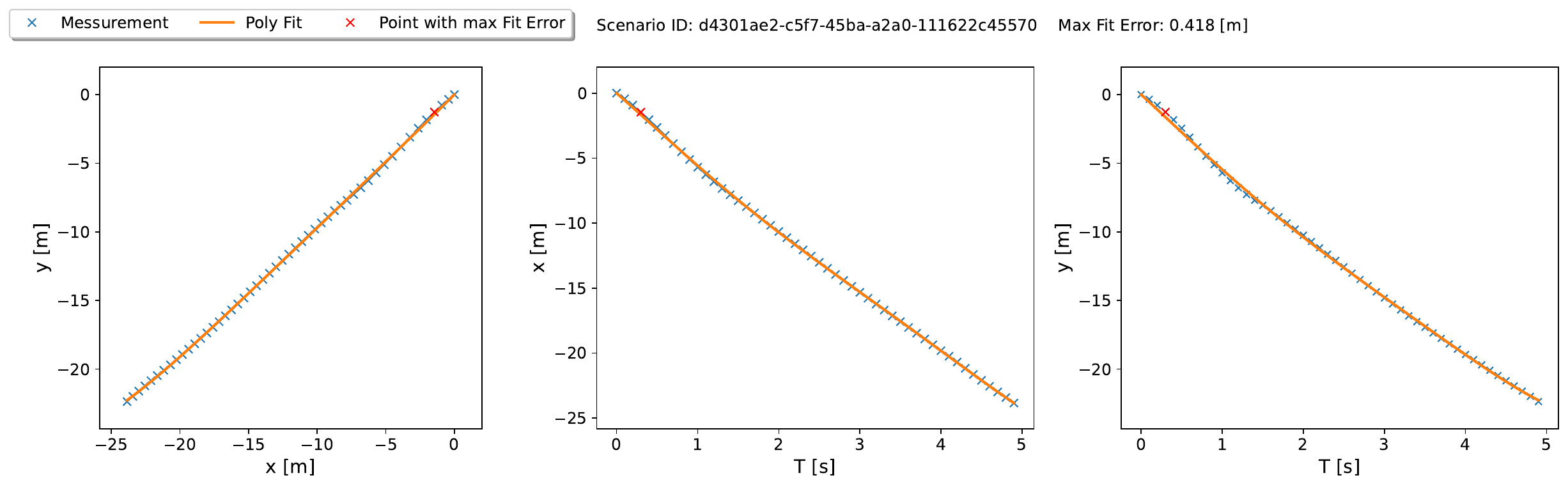}

\subsubsection{Five random 5-seconds cyclist trajectories in A2 (Fitted with $\hat{N} = 5$)}
\centering
\includegraphics[width=\textwidth, height=1.8in]{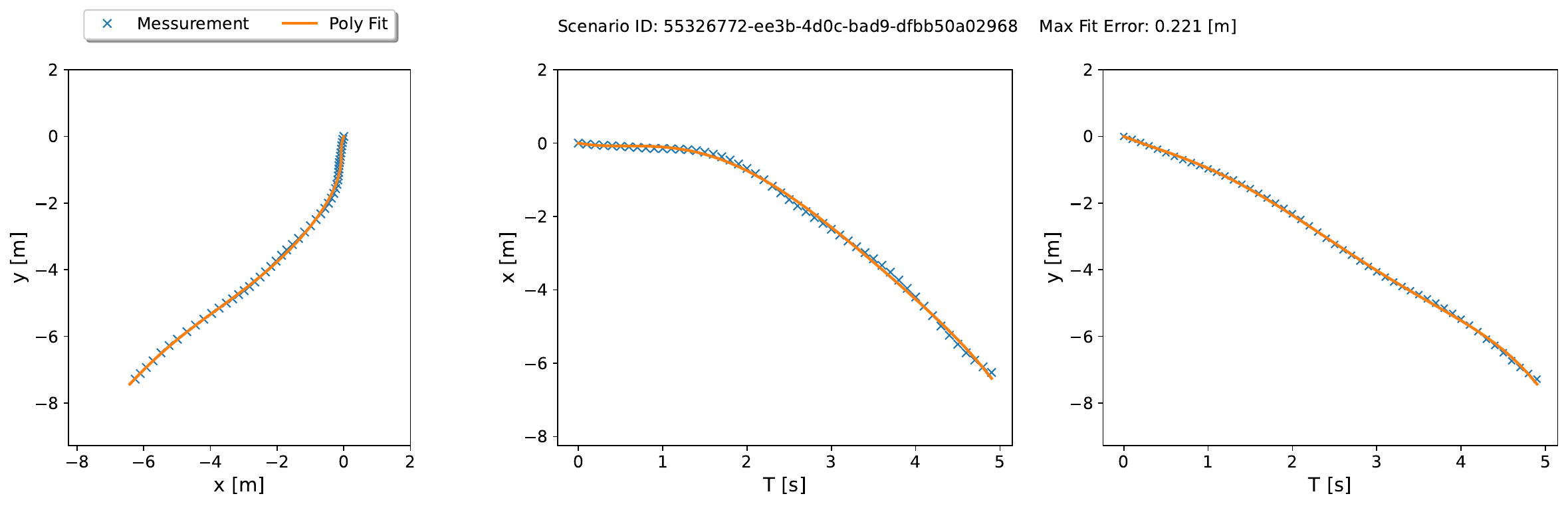} 

\includegraphics[width=\textwidth, height=1.8in]{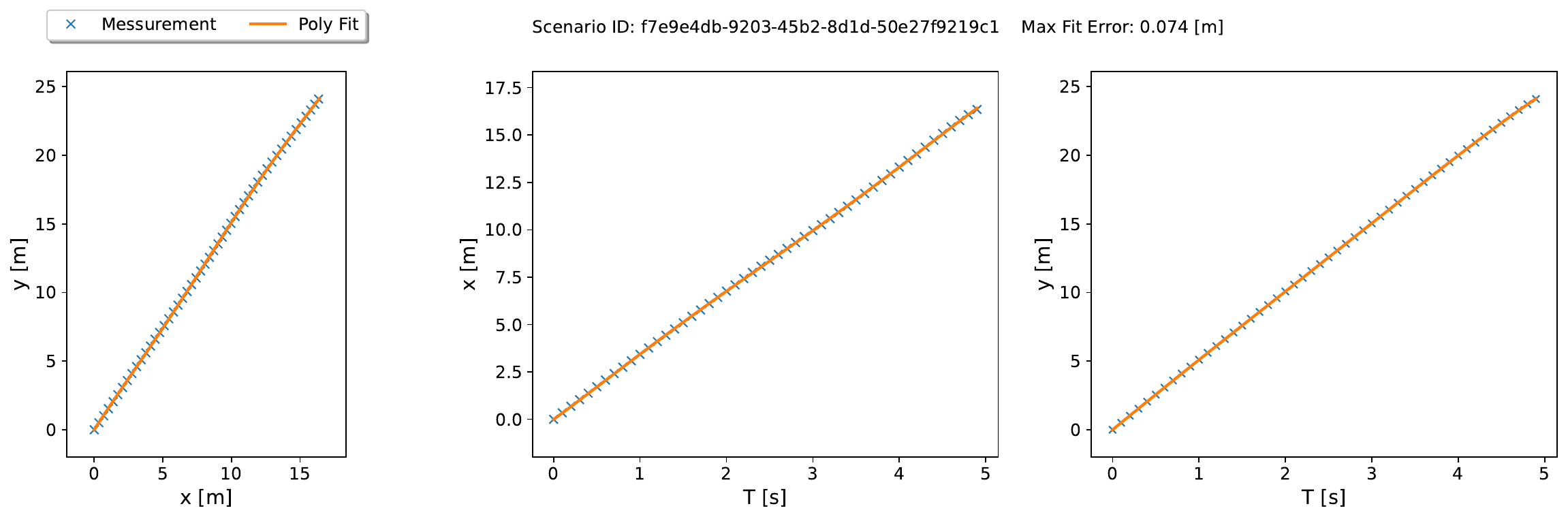} 

\includegraphics[width=\textwidth, height=1.8in]{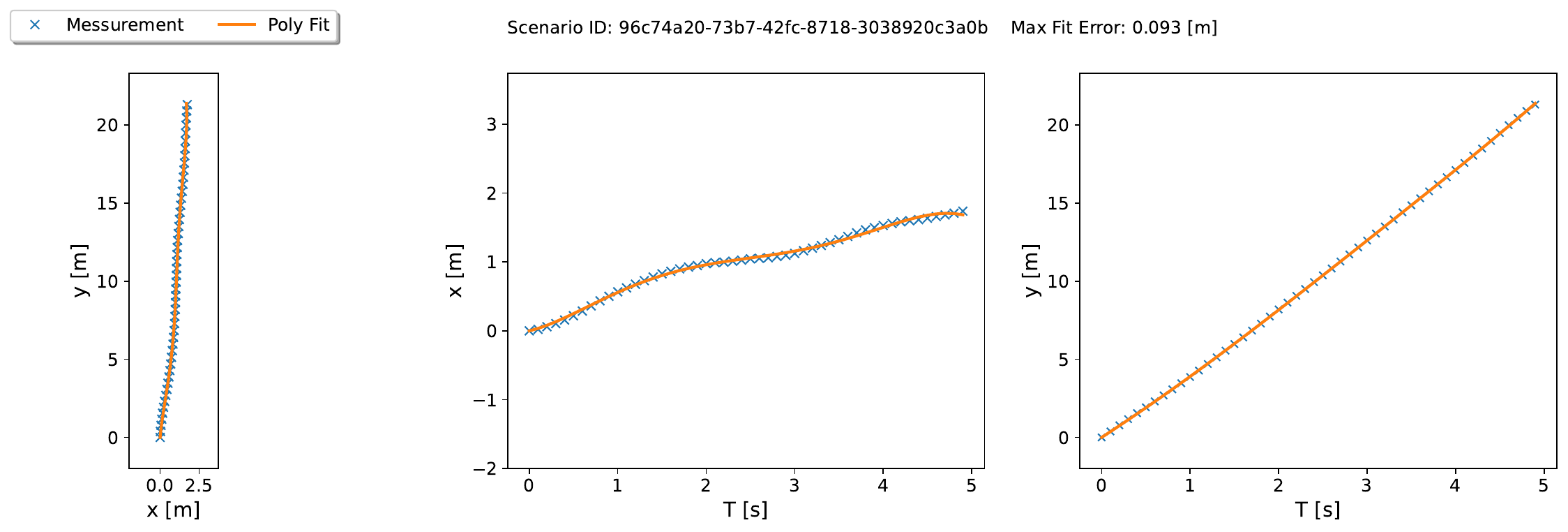} 

\includegraphics[width=\textwidth, height=1.8in]{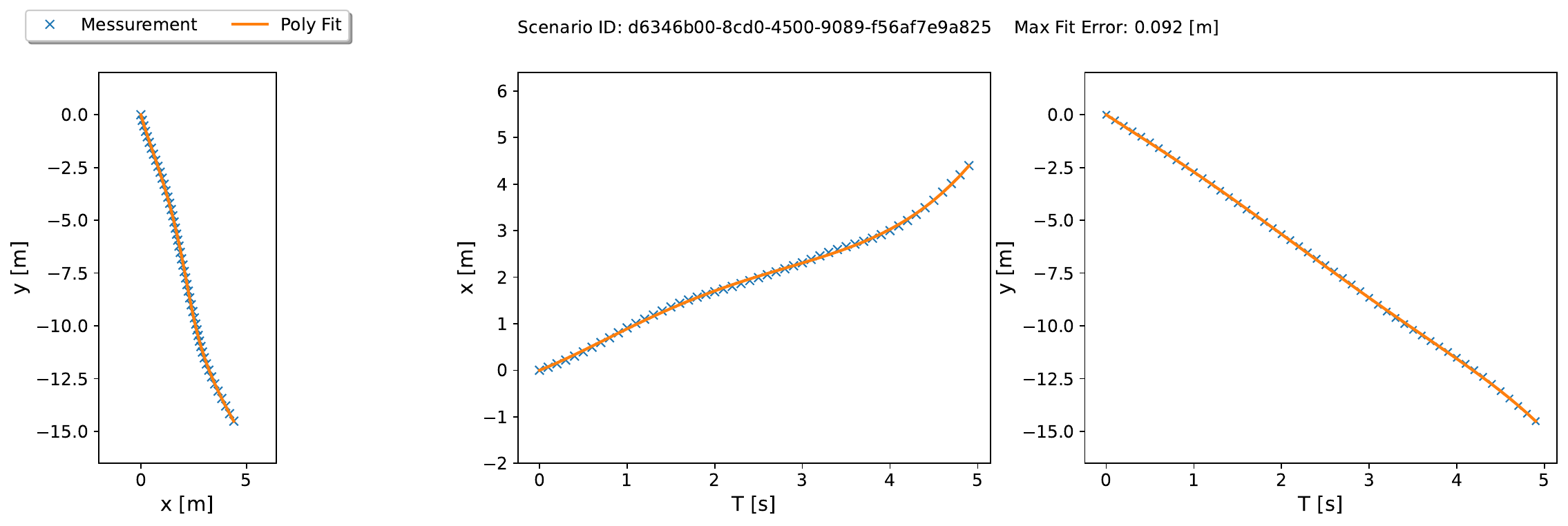} 

\includegraphics[width=\textwidth, height=1.7in]{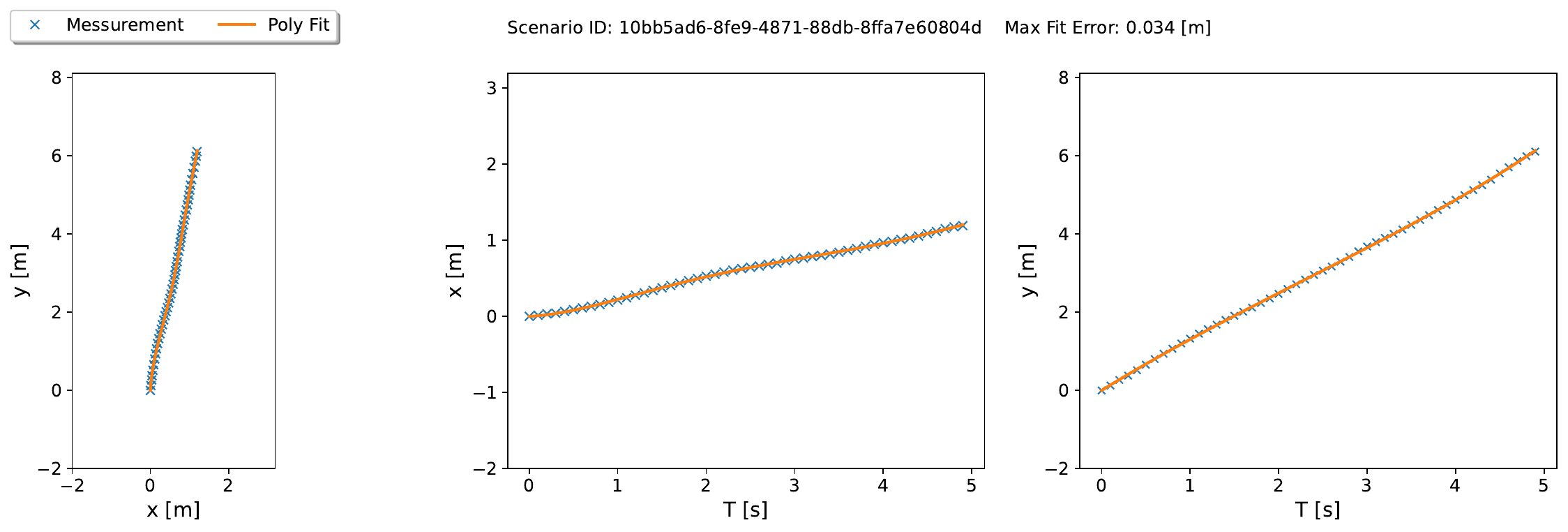}

\subsubsection{Five 5-seconds pedestrian trajectories with highest fit error in A2 (Fitted with $\hat{N} = 5$)}
\centering
\includegraphics[width=\textwidth, height=1.8in]{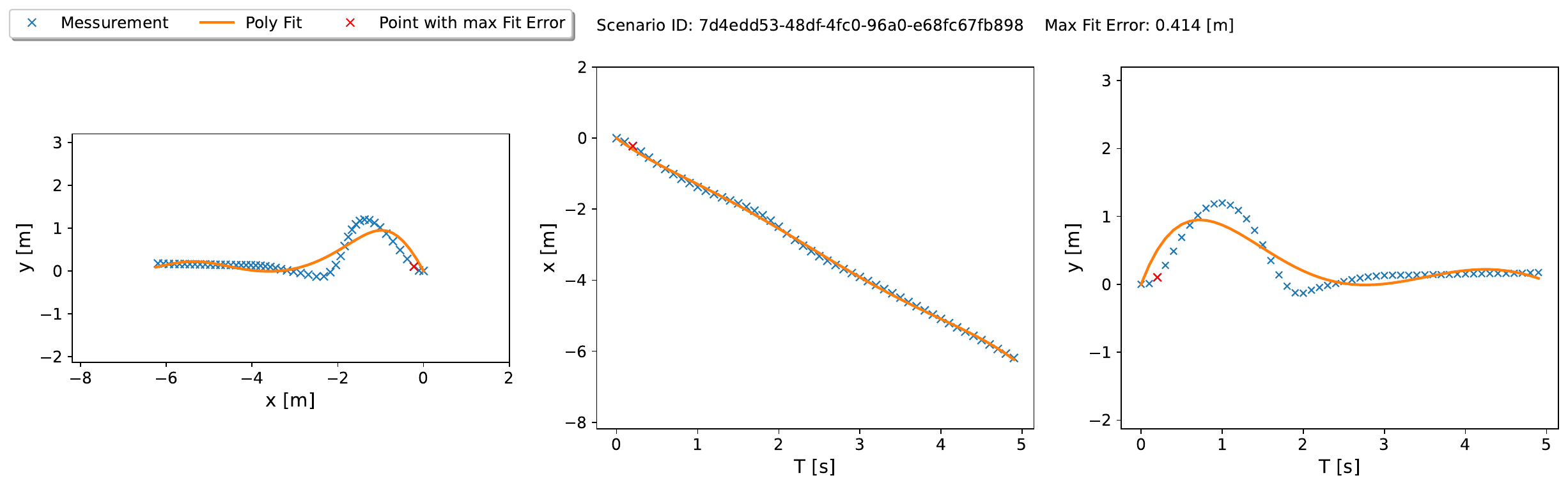} 

\includegraphics[width=\textwidth, height=1.8in]{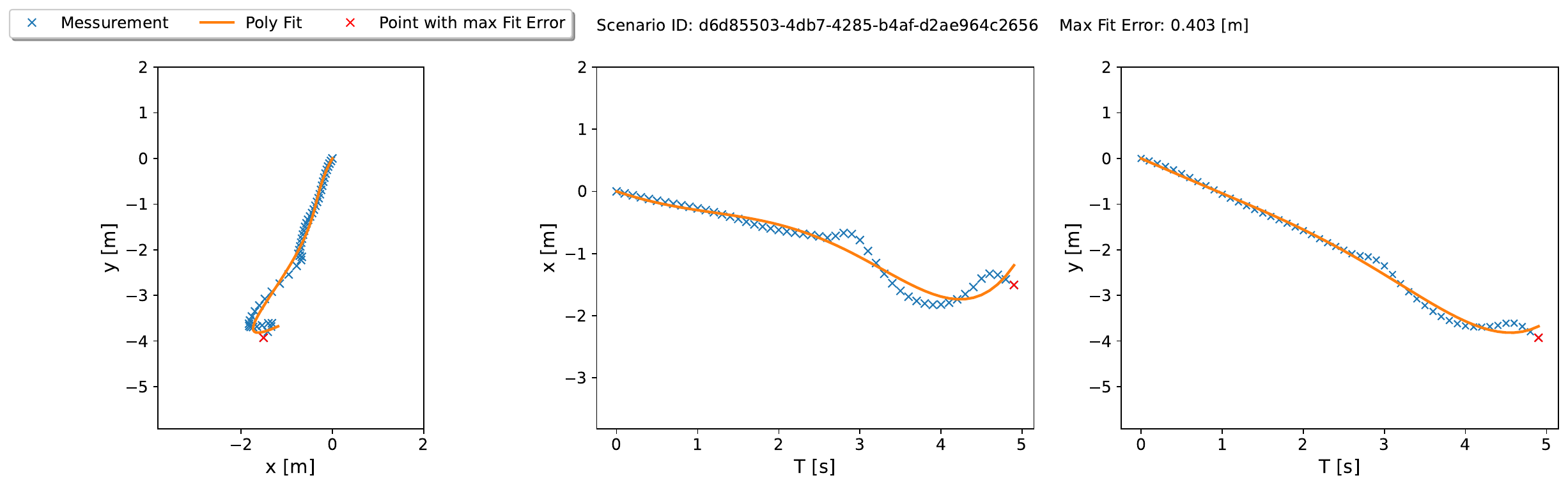} 

\includegraphics[width=\textwidth, height=1.8in]{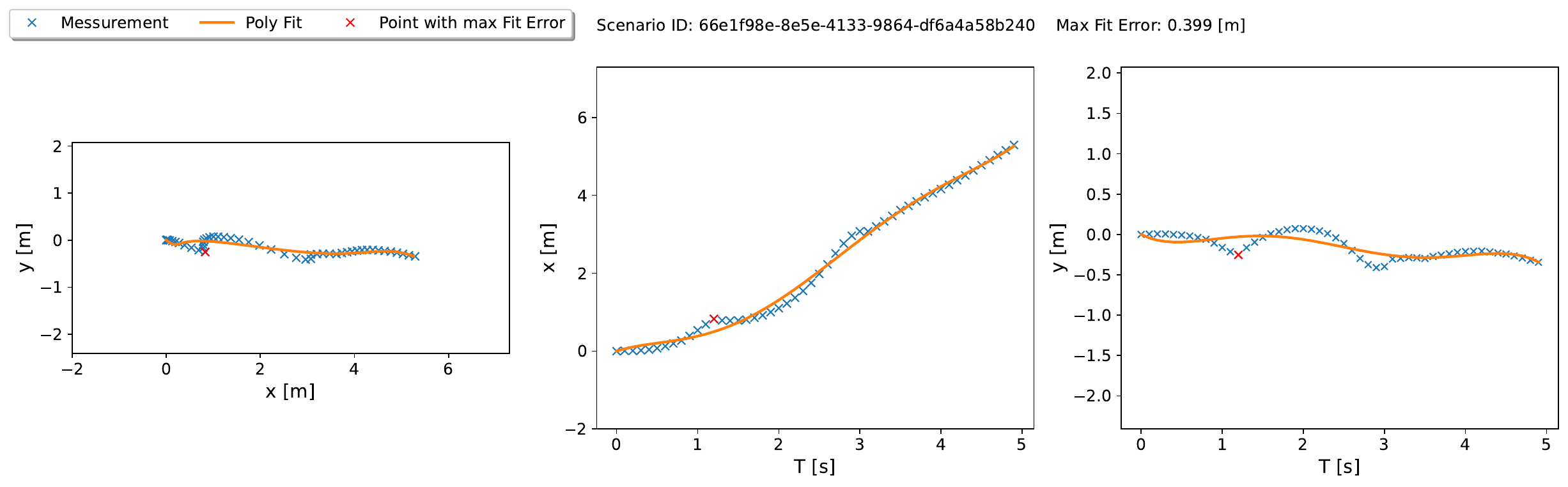} 

\includegraphics[width=\textwidth, height=1.8in]{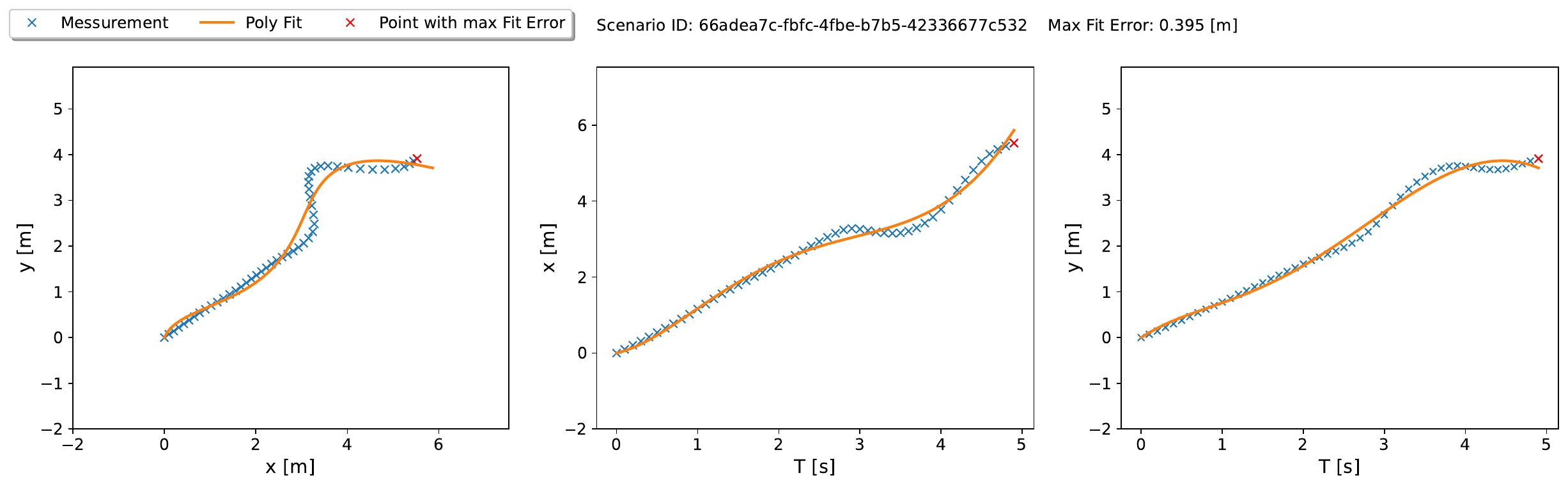} 

\includegraphics[width=\textwidth, height=1.7in]{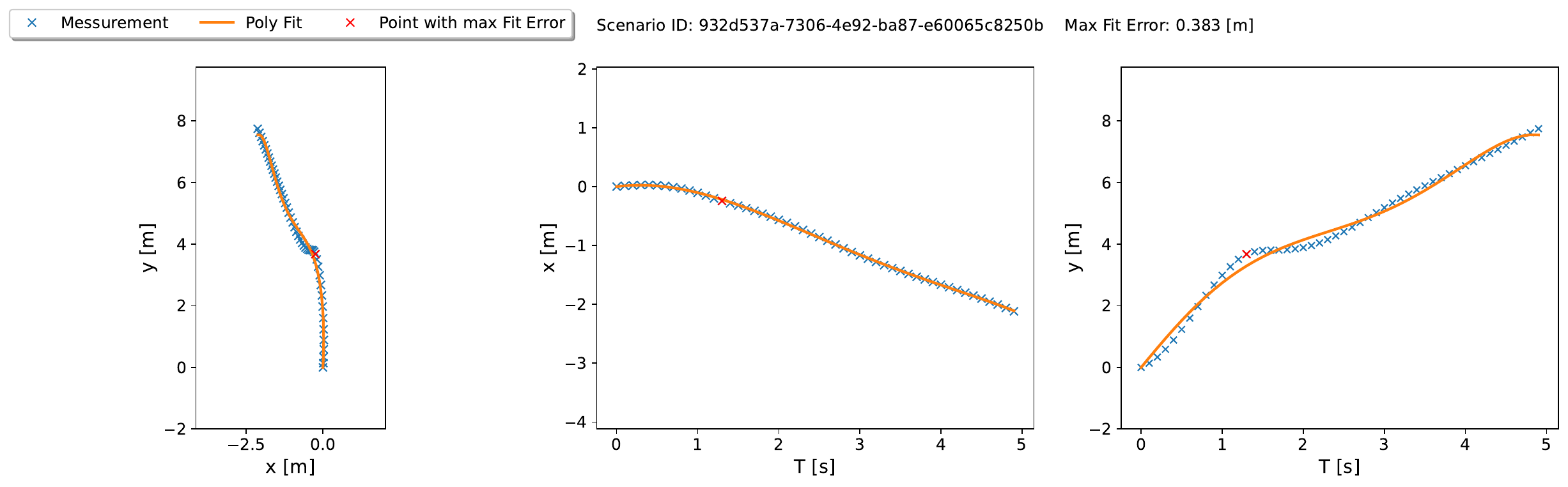}

\subsubsection{Five random 5-seconds pedestrian trajectories in A2 (Fitted with $\hat{N} = 5$)}
\centering
\includegraphics[width=\textwidth, height=1.8in]{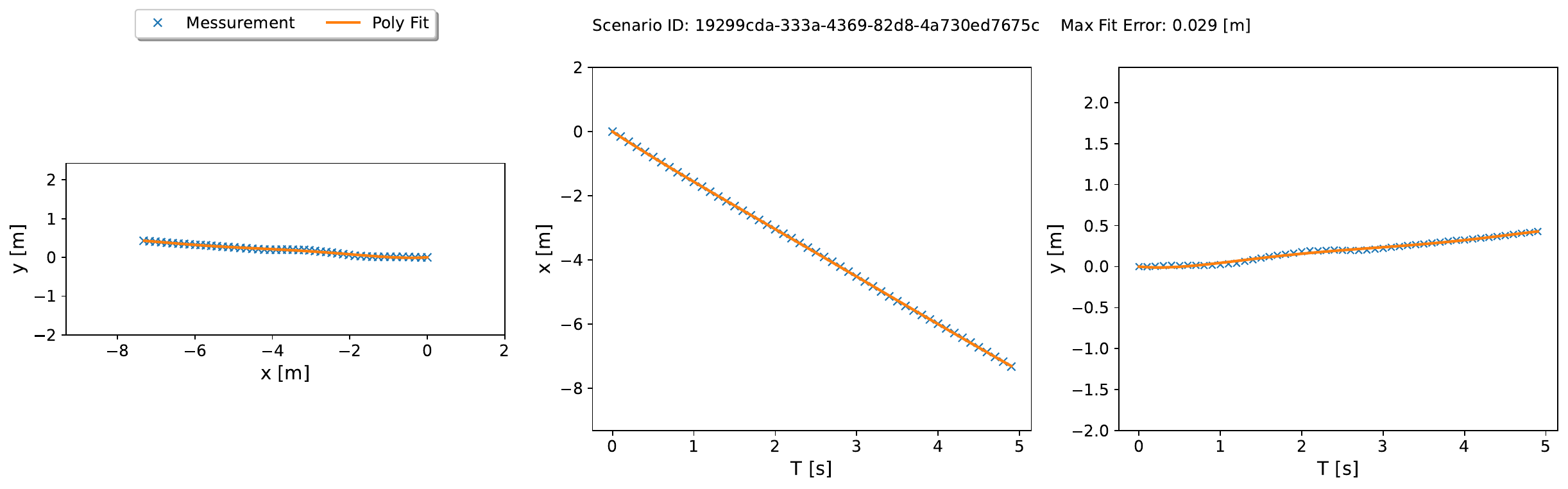} 

\includegraphics[width=\textwidth, height=1.8in]{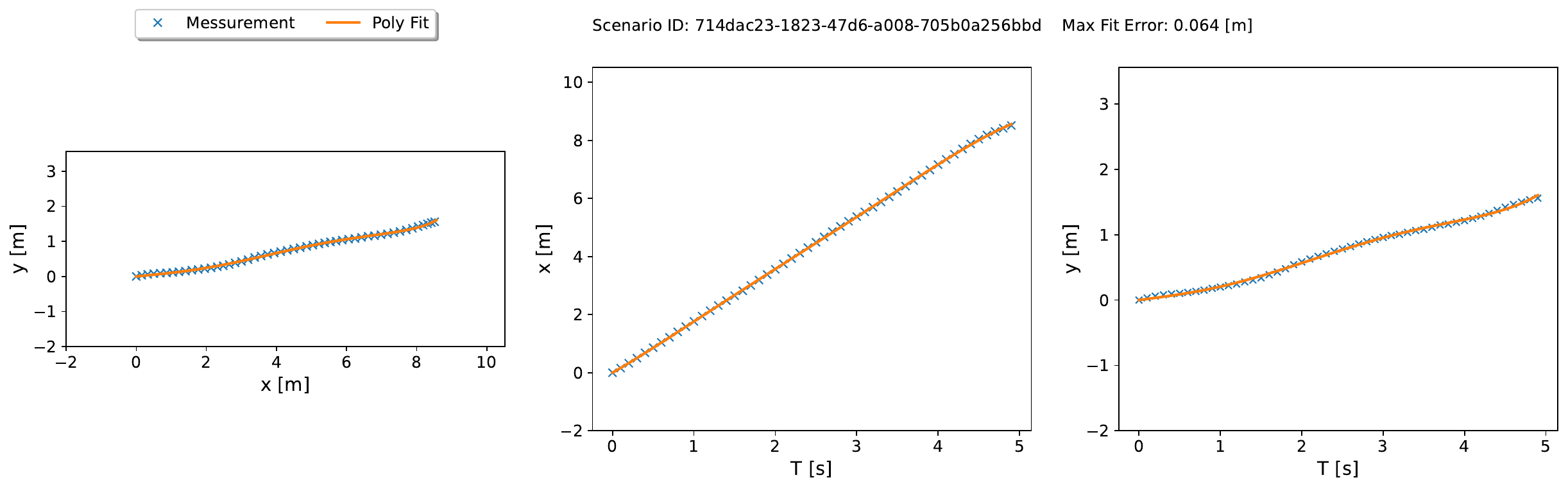} 

\includegraphics[width=\textwidth, height=1.8in]{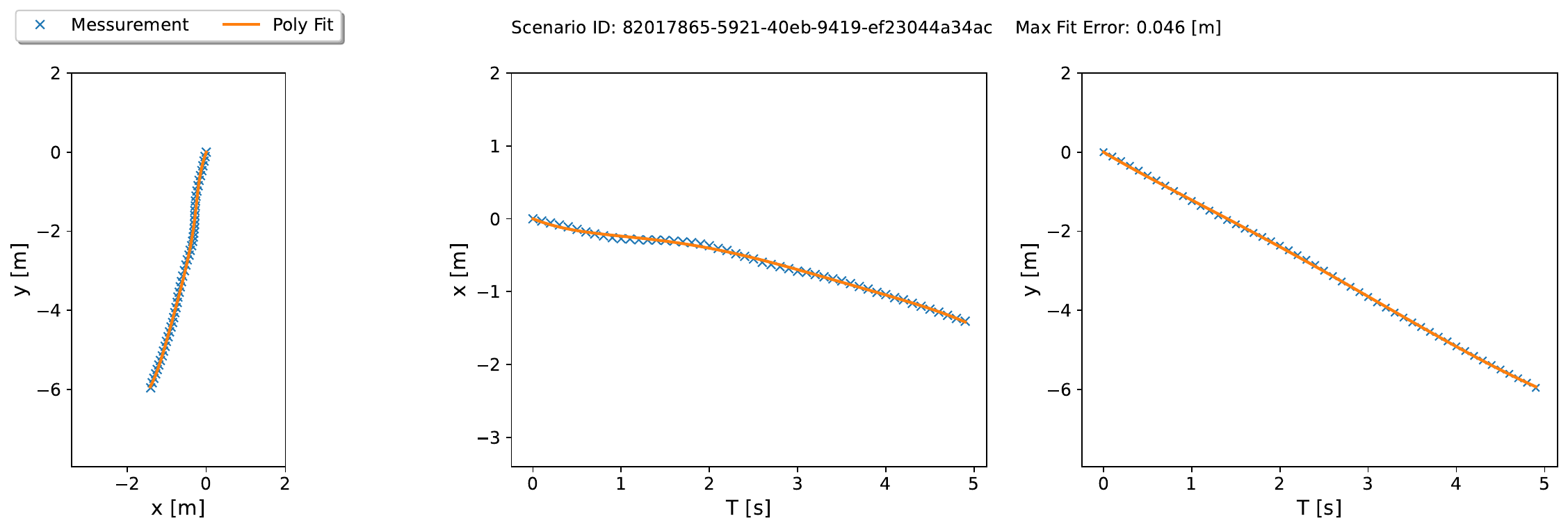} 

\includegraphics[width=\textwidth, height=1.8in]{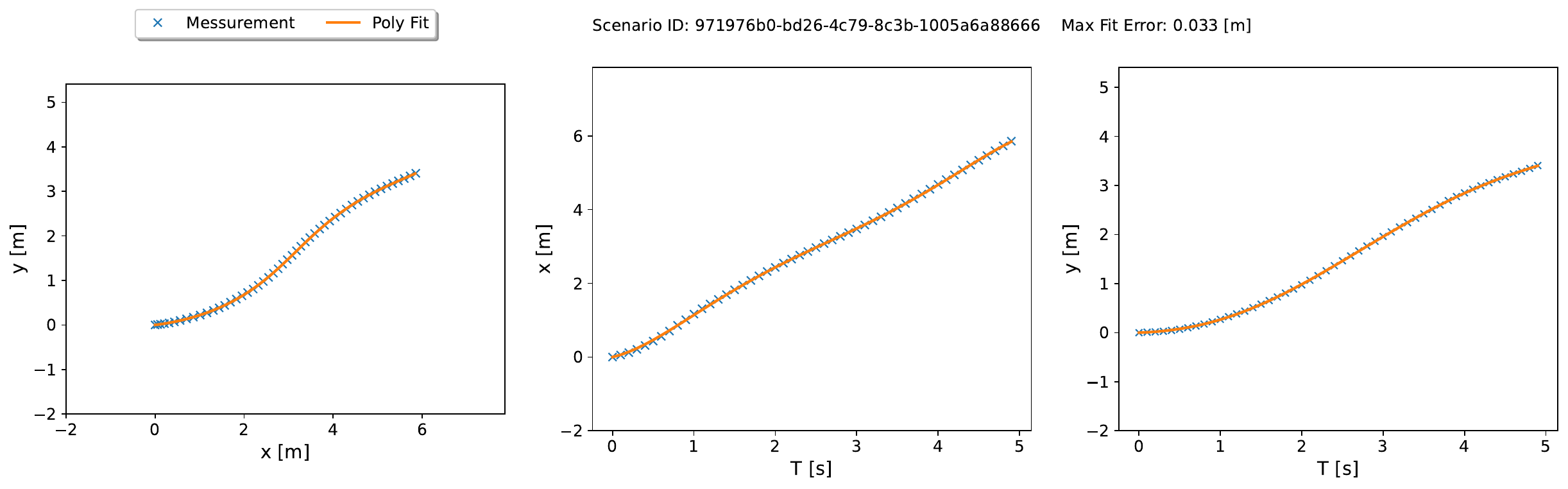} 

\includegraphics[width=\textwidth, height=1.7in]{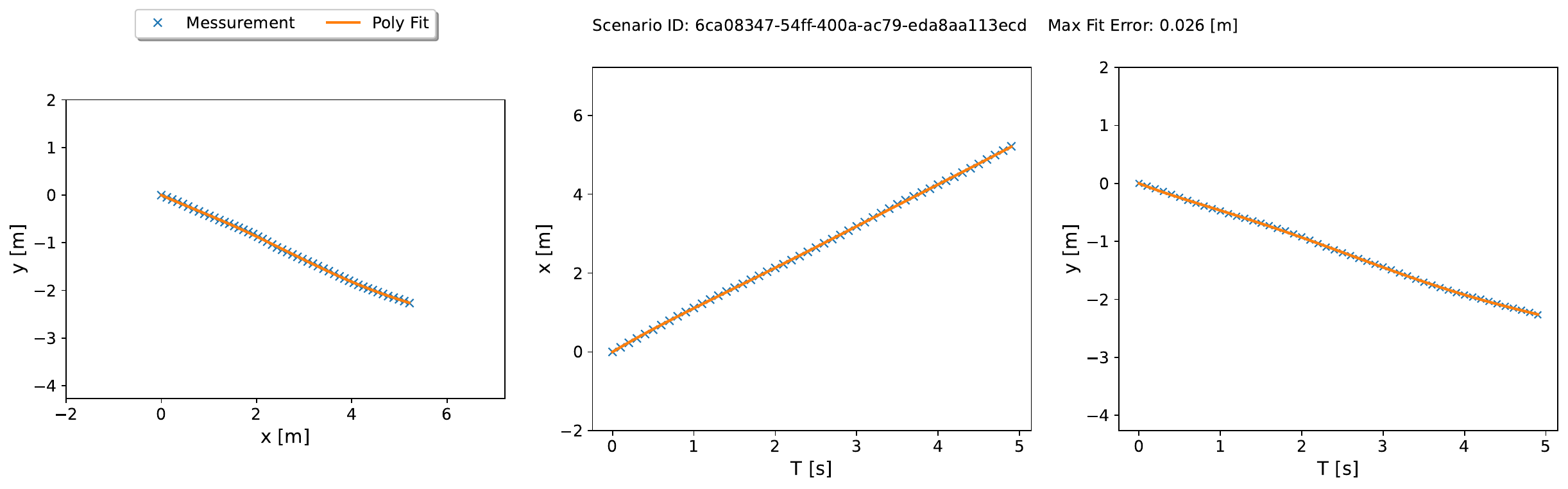} 






\subsubsection{Five 5-seconds vehicle trajectories with highest fit error in WO (Fitted with $\hat{N} = 5$)}
\centering
\includegraphics[width=\textwidth, height=1.8in]{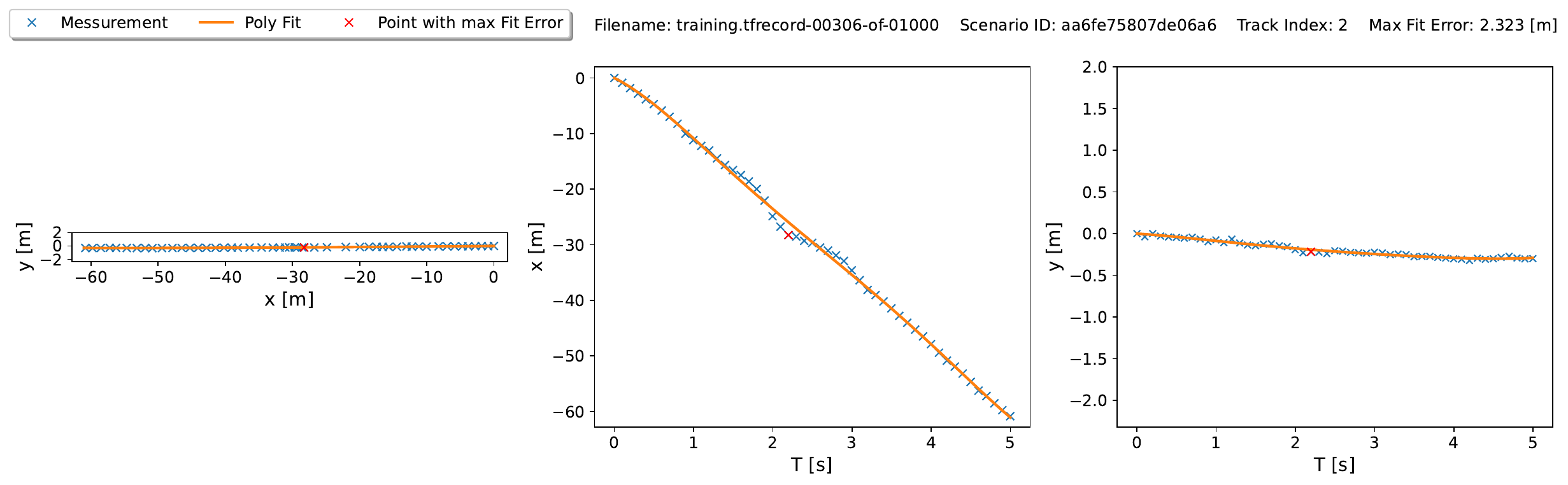} 

\includegraphics[width=\textwidth, height=1.8in]{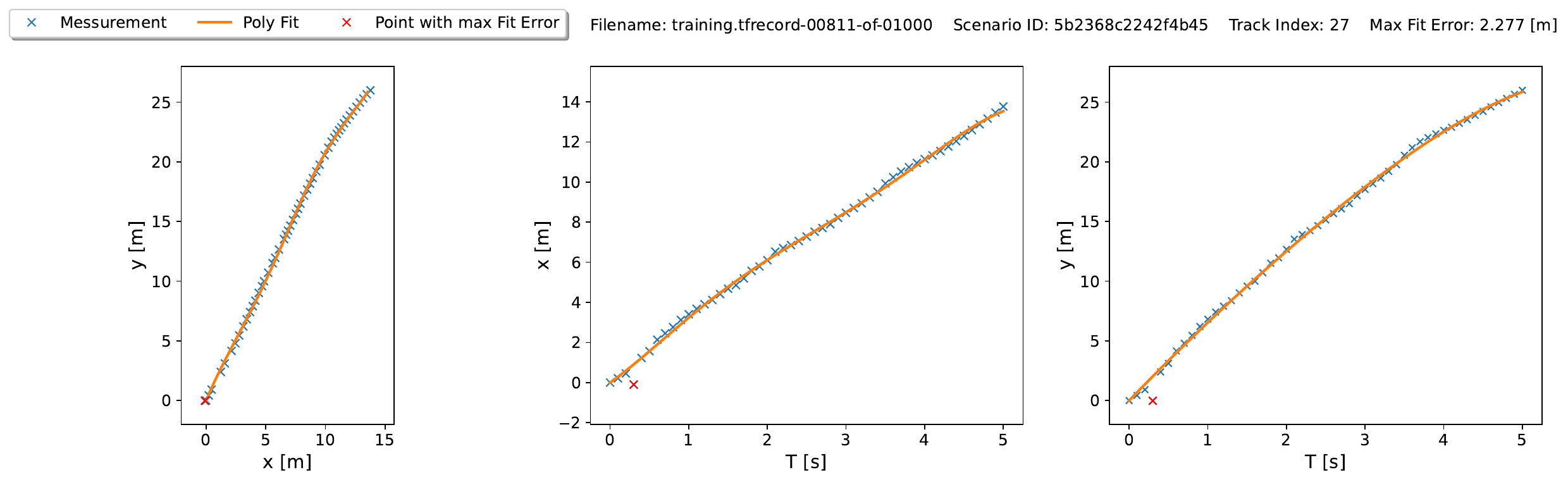} 

\includegraphics[width=\textwidth, height=1.8in]{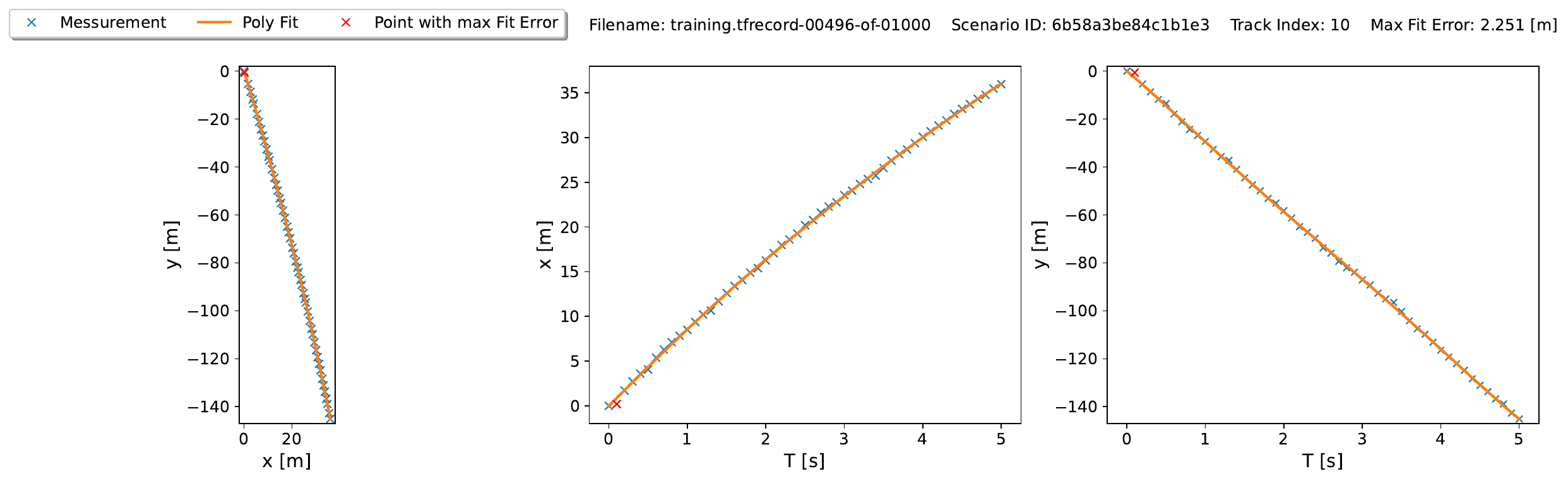} 

\includegraphics[width=\textwidth, height=1.8in]{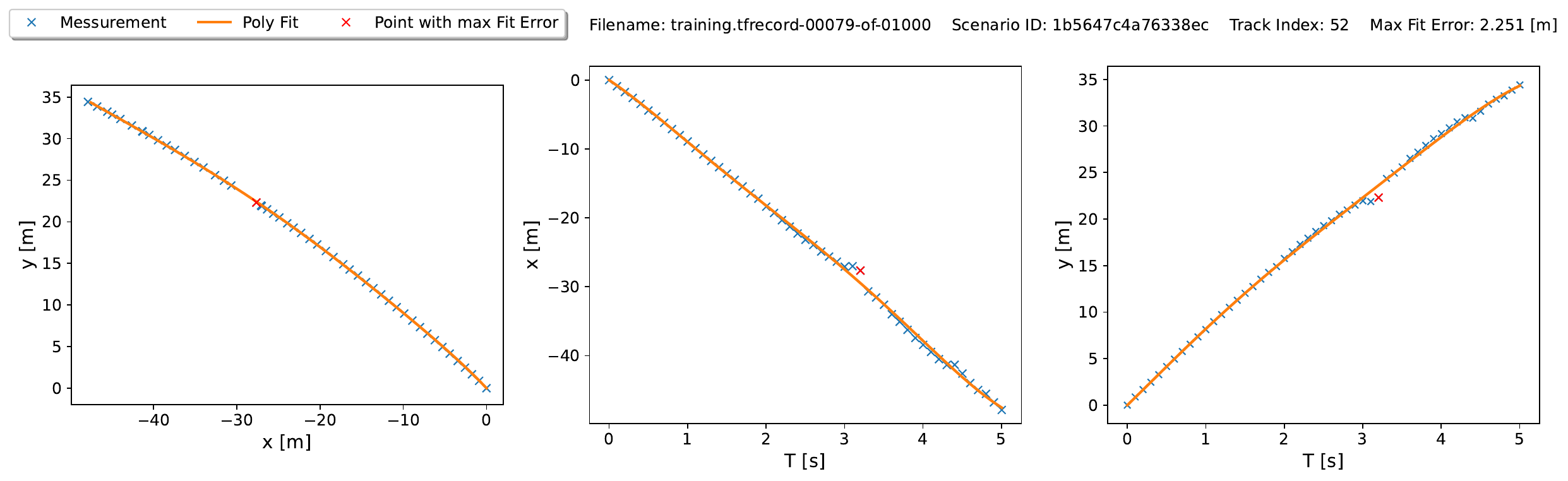} 

\includegraphics[width=\textwidth, height=1.7in]{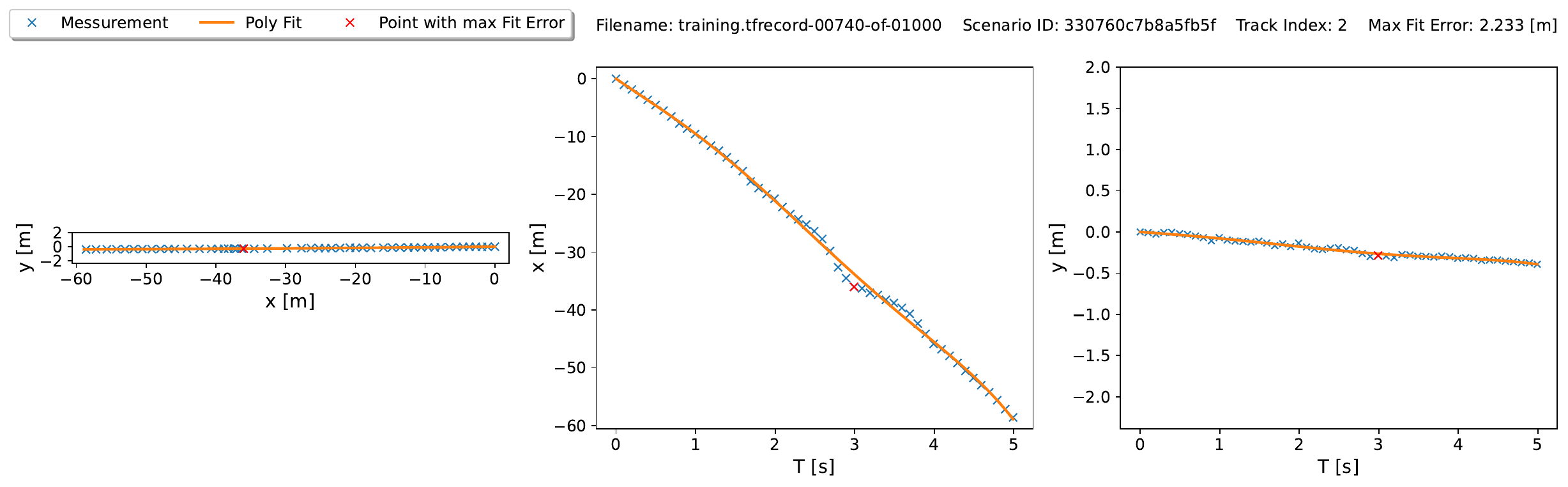}

\subsubsection{Five random 5-seconds vehicle trajectories in WO (Fitted with $\hat{N} = 5$)}
\centering
\includegraphics[width=\textwidth, height=1.8in]{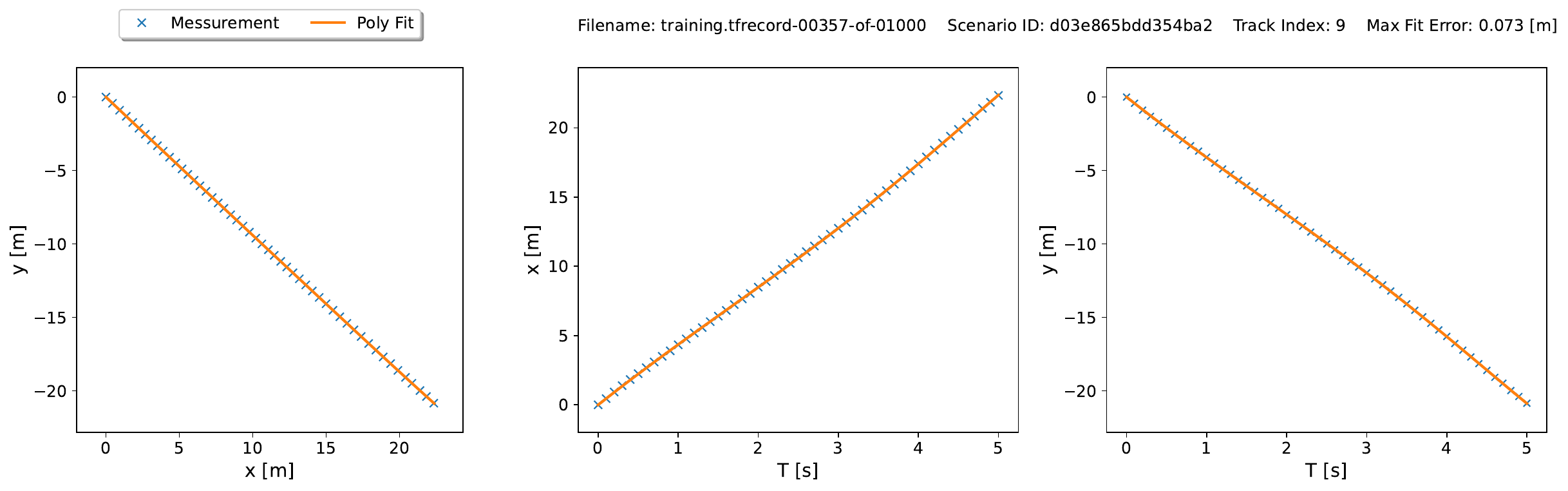} 

\includegraphics[width=\textwidth, height=1.8in]{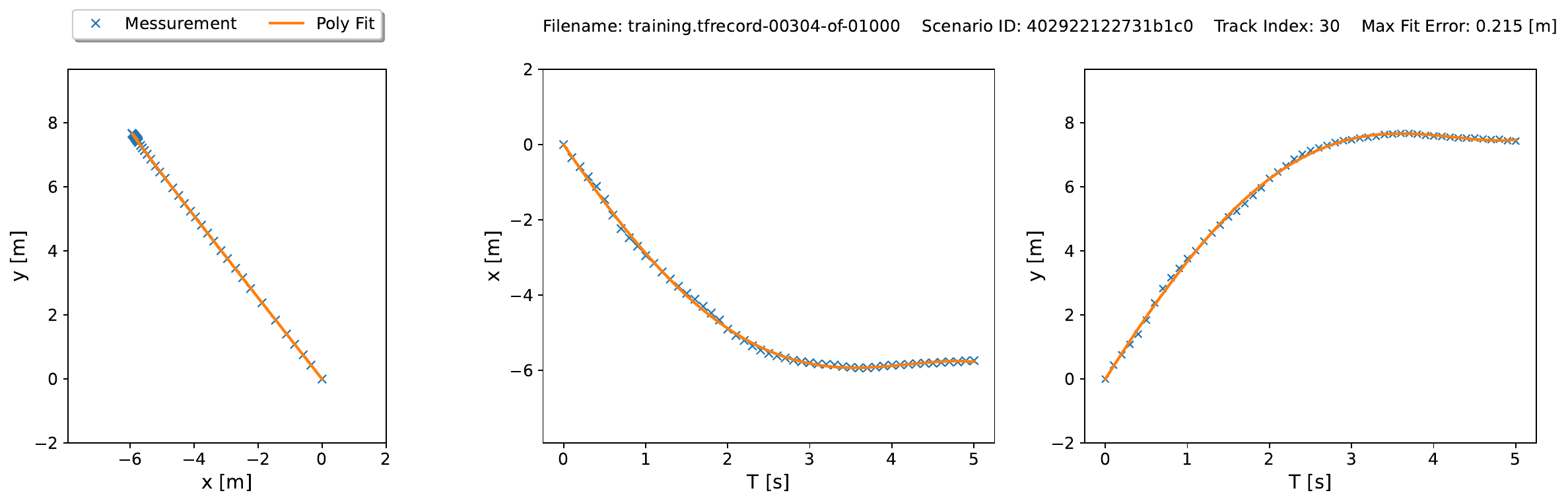} 

\includegraphics[width=\textwidth, height=1.8in]{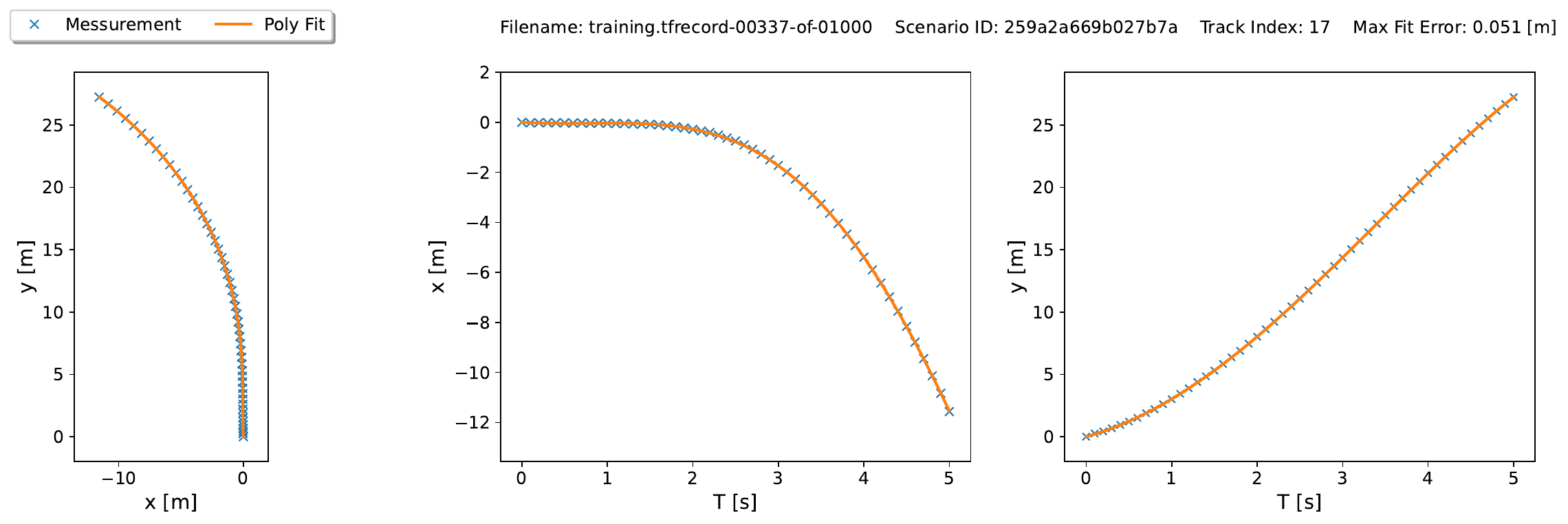} 

\includegraphics[width=\textwidth, height=1.8in]{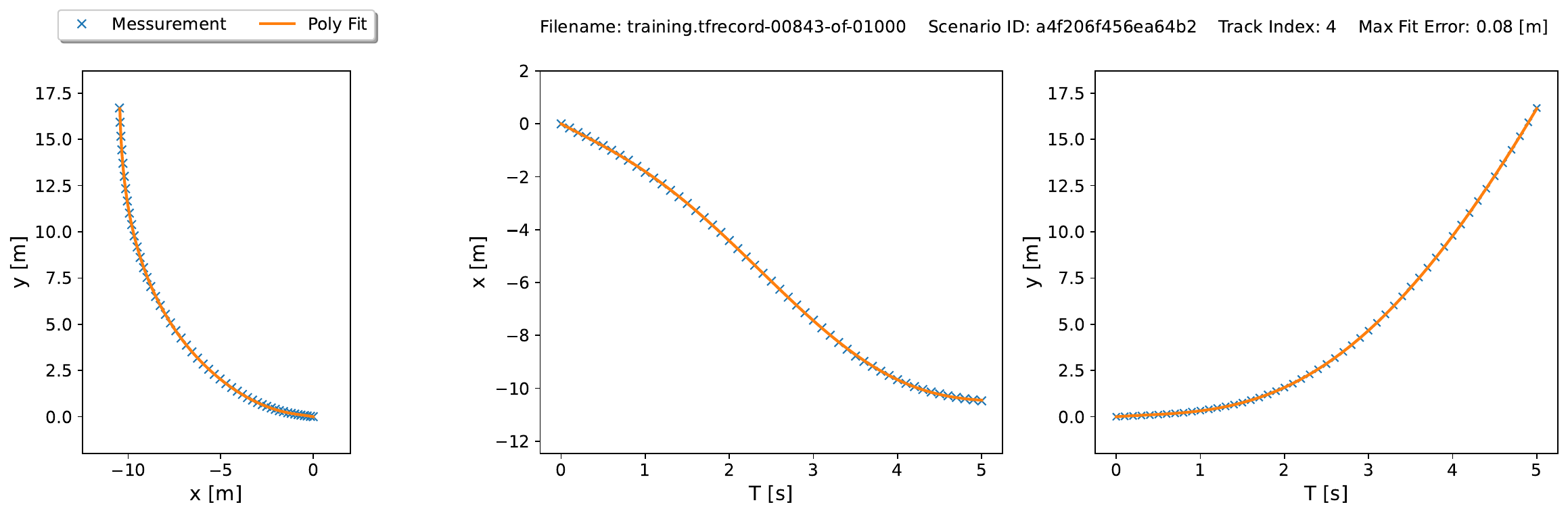} 

\includegraphics[width=\textwidth, height=1.7in]{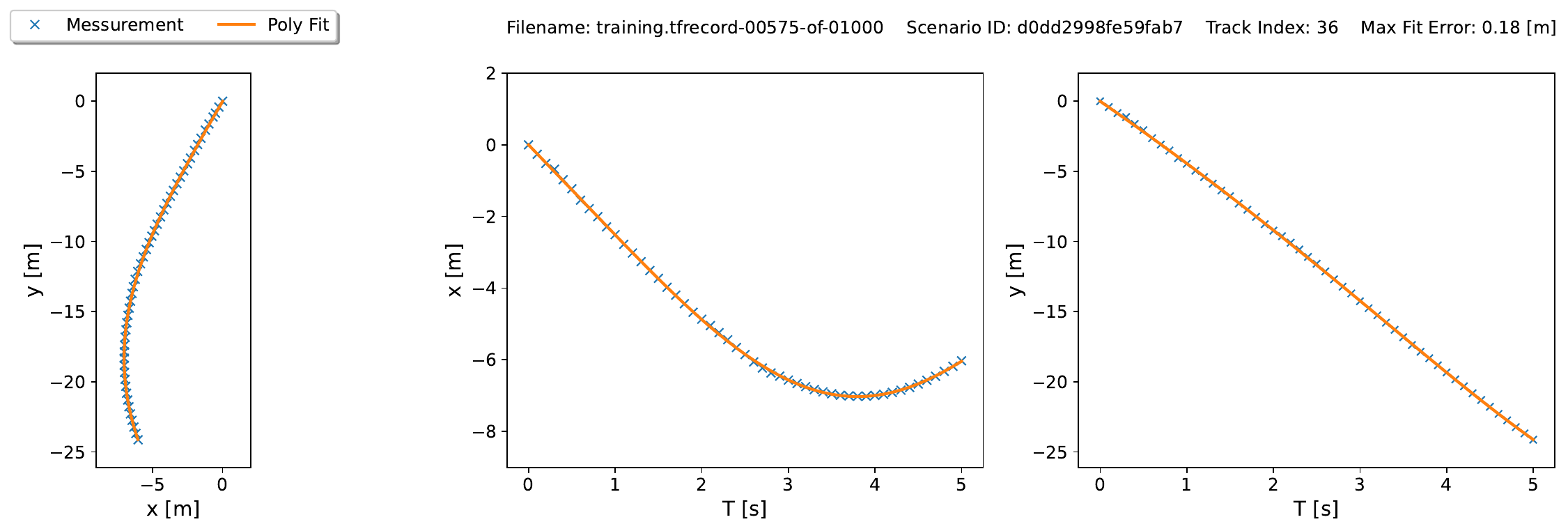}

\subsubsection{Five 5-seconds cyclist trajectories with highest fit error in WO (Fitted with $\hat{N} = 5$)}
\centering
\includegraphics[width=\textwidth, height=1.8in]{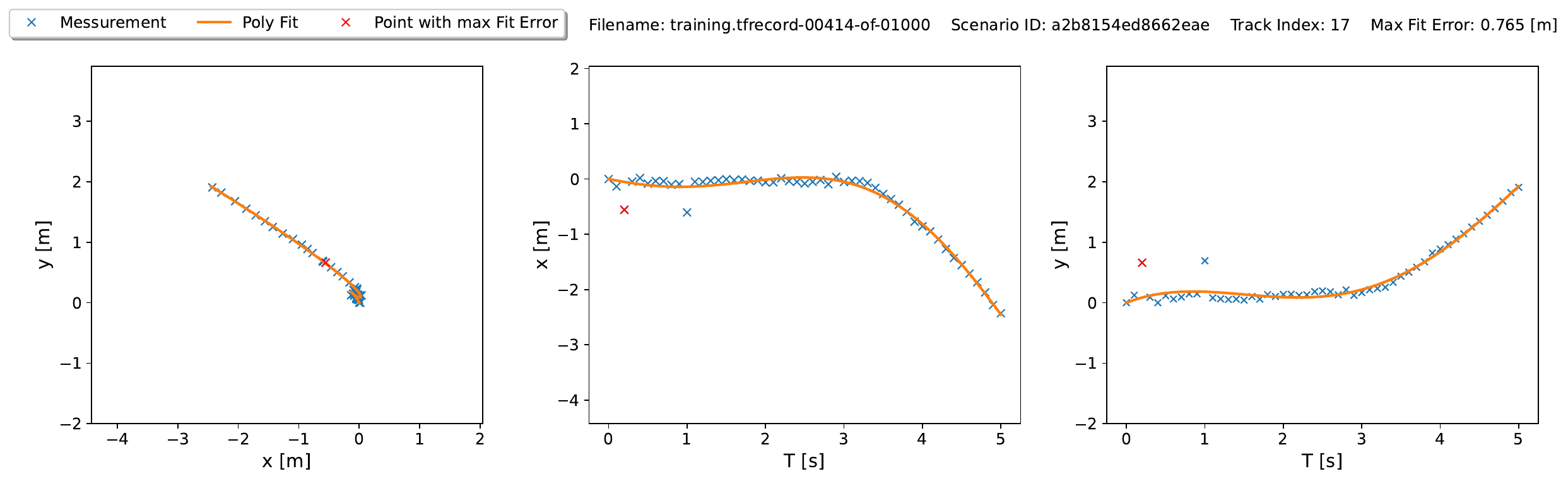} 

\includegraphics[width=\textwidth, height=1.8in]{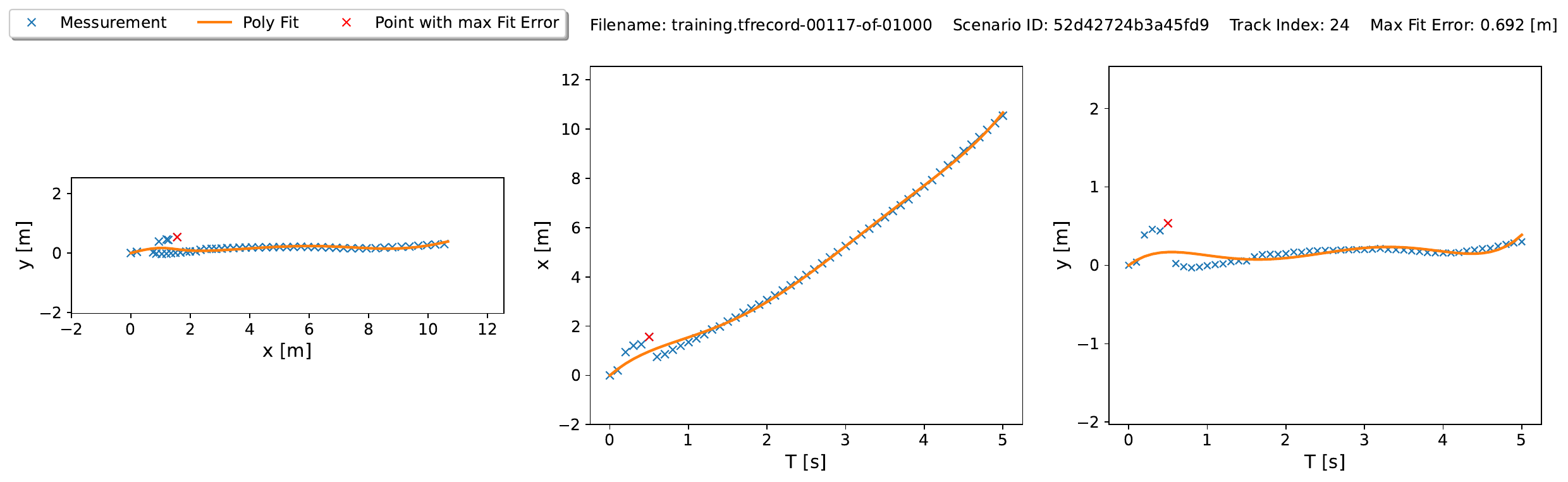} 

\includegraphics[width=\textwidth, height=1.8in]{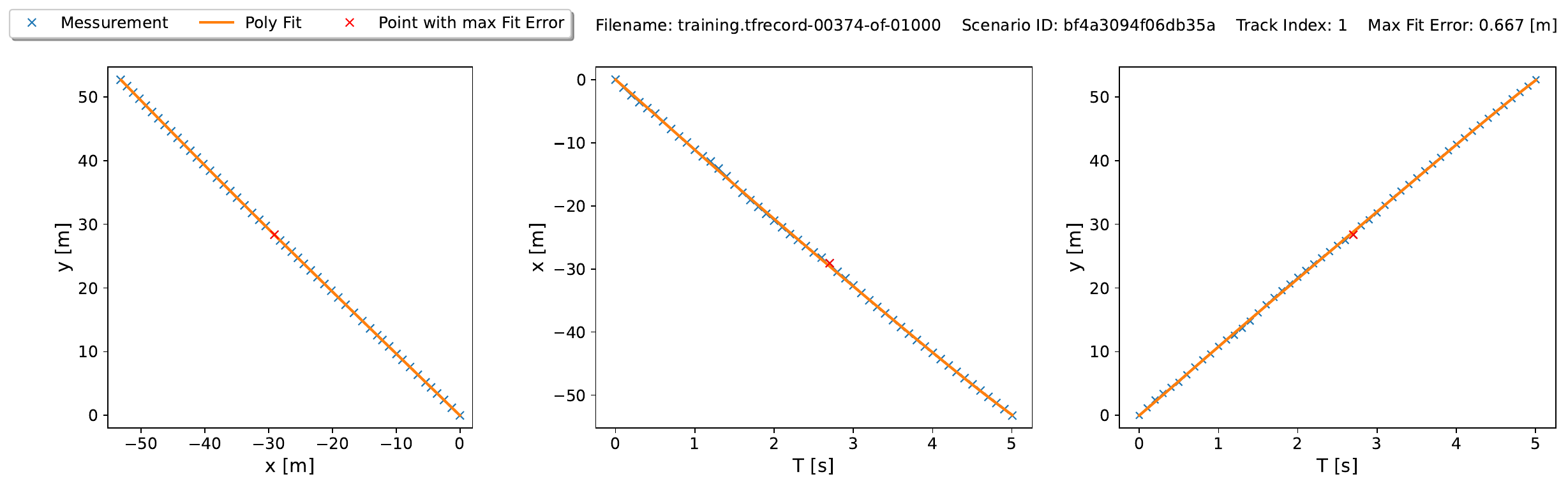} 

\includegraphics[width=\textwidth, height=1.8in]{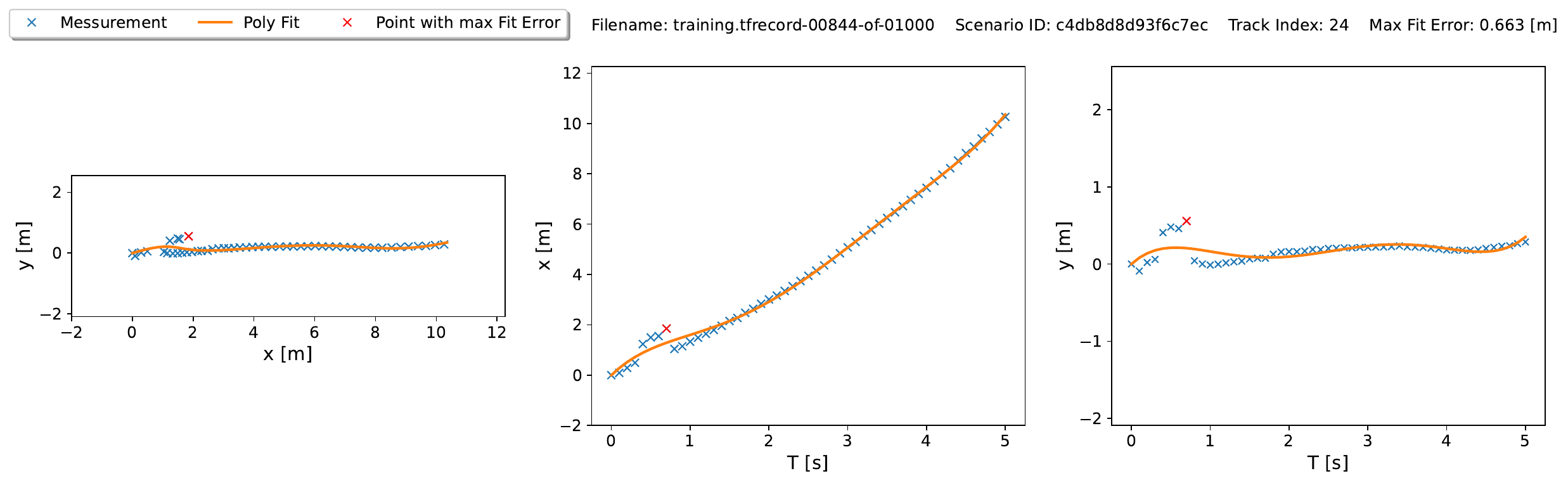} 

\includegraphics[width=\textwidth, height=1.7in]{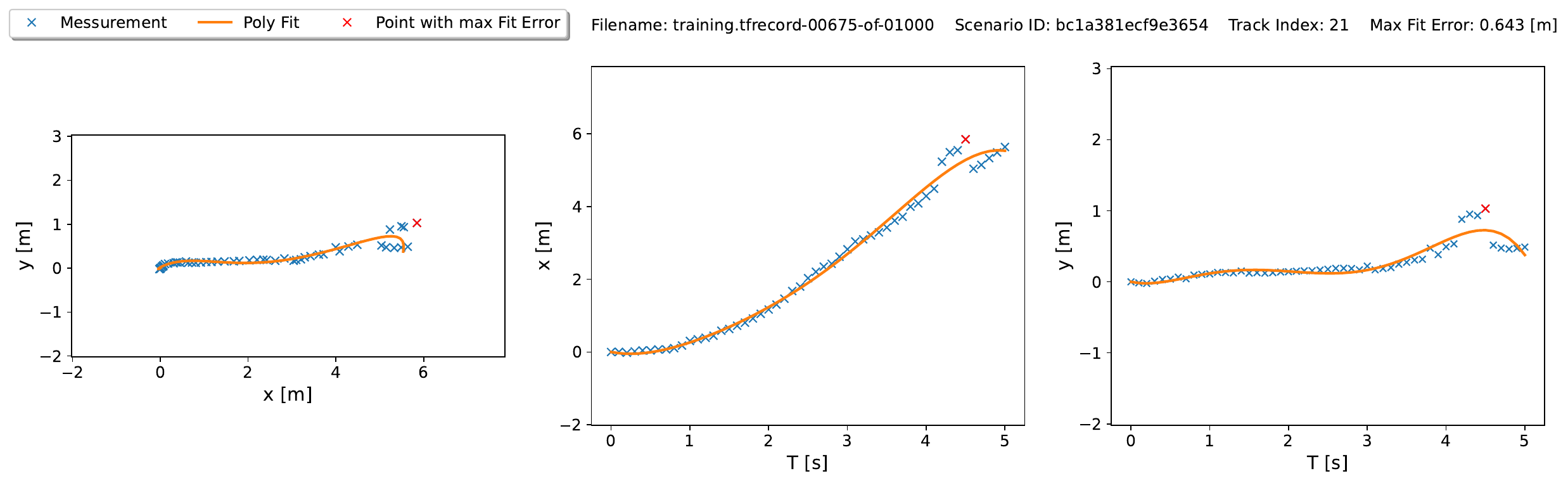}

\subsubsection{Five random 5-seconds cyclist trajectories in WO (Fitted with $\hat{N} = 5$)}
\centering
\includegraphics[width=\textwidth, height=1.8in]{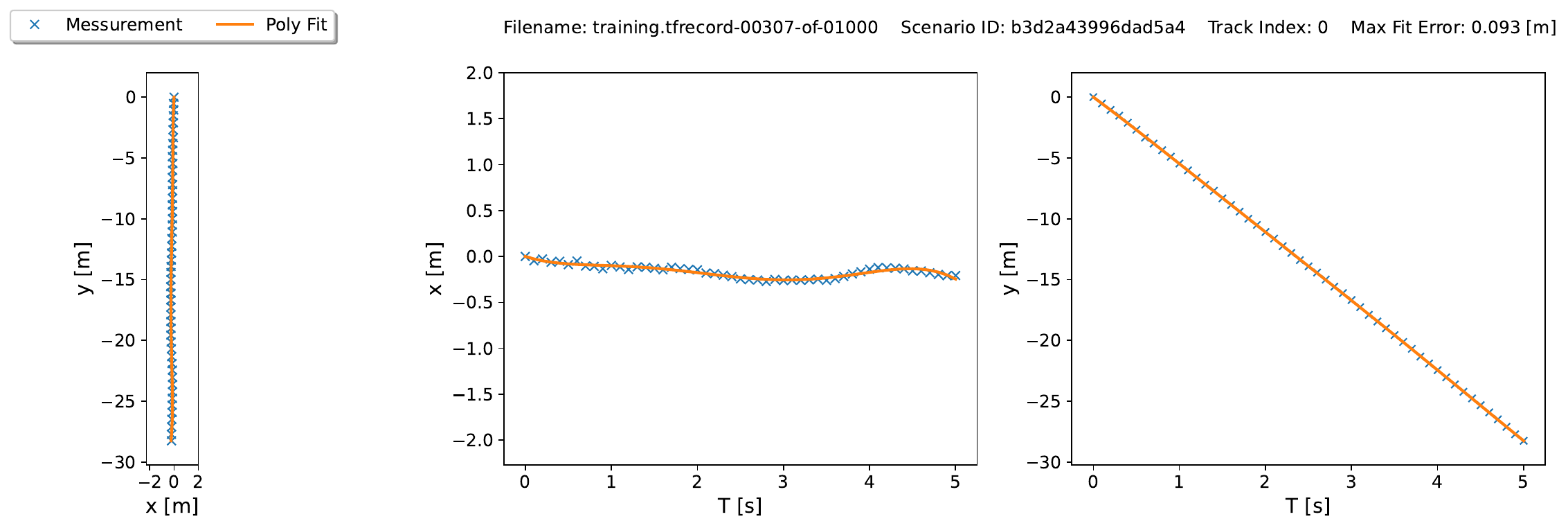} 

\includegraphics[width=\textwidth, height=1.8in]{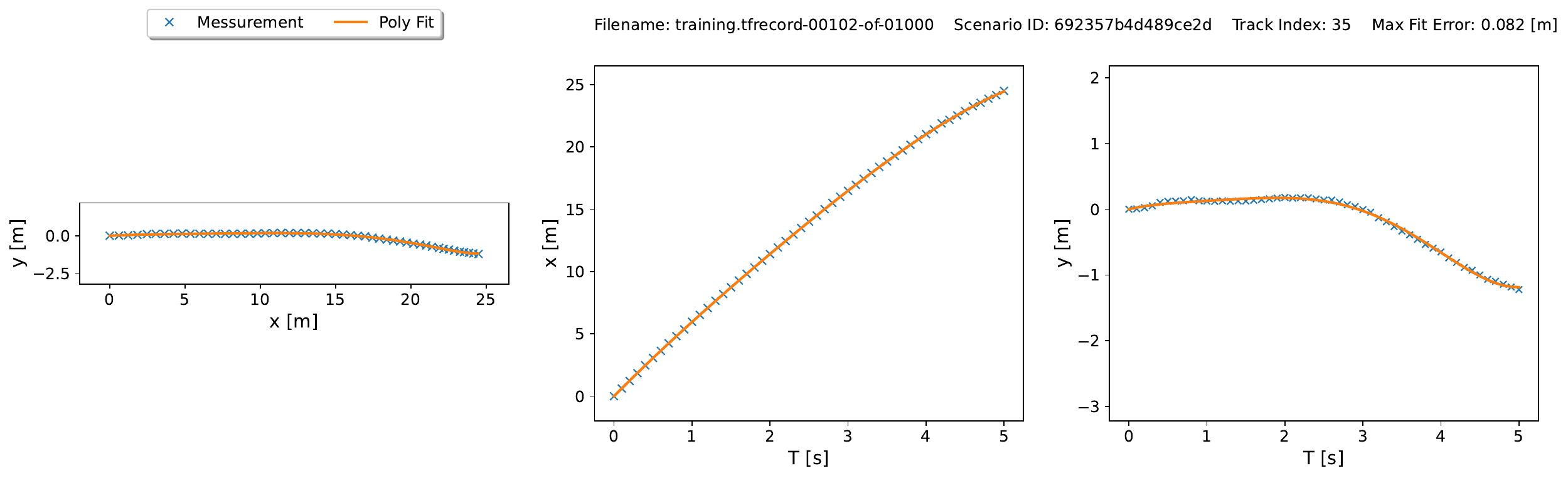} 

\includegraphics[width=\textwidth, height=1.8in]{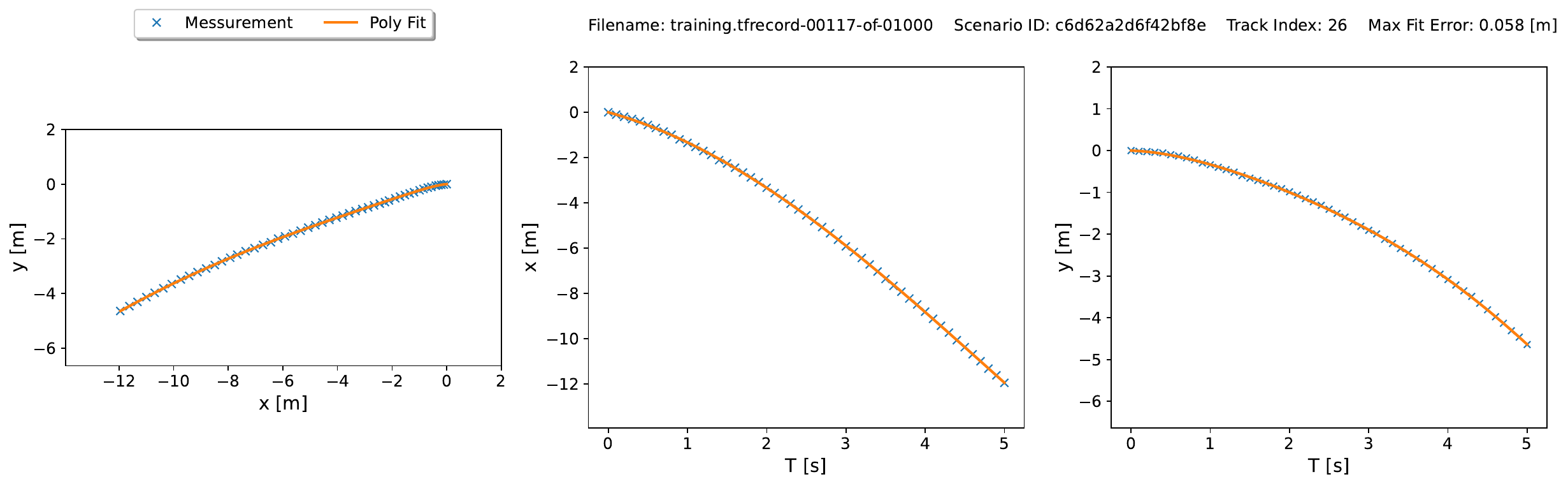} 

\includegraphics[width=\textwidth, height=1.8in]{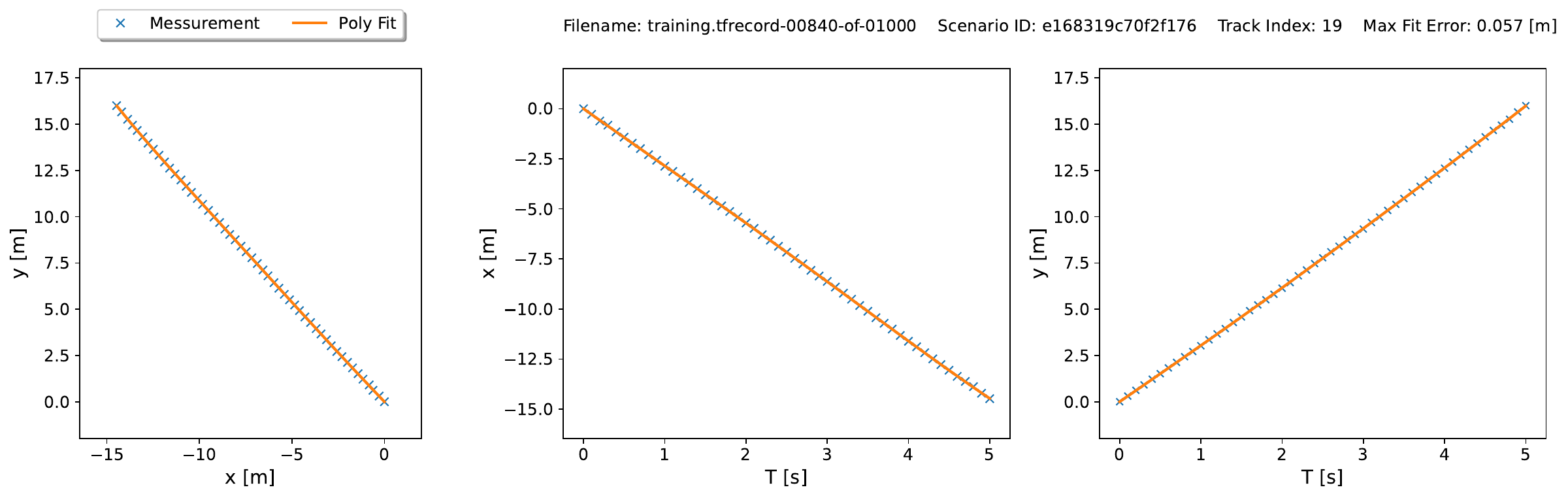} 

\includegraphics[width=\textwidth, height=1.7in]{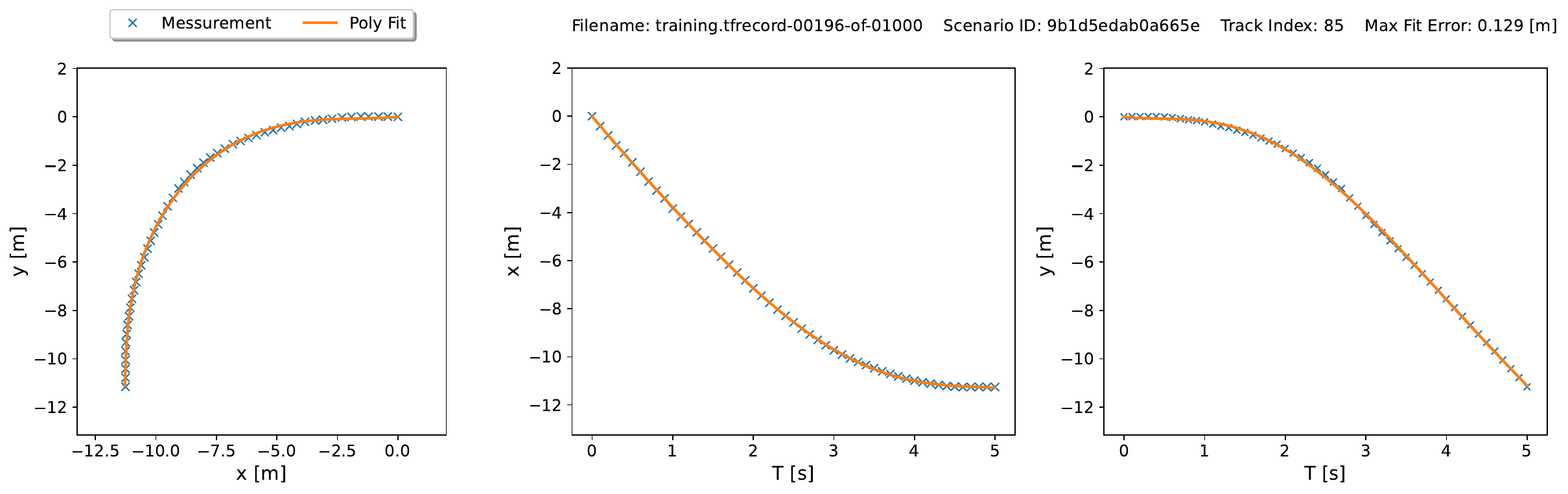}

\subsubsection{Five 5-seconds pedestrian trajectories with highest fit error in WO (Fitted with $\hat{N} = 5$)}
\centering
\includegraphics[width=\textwidth, height=1.8in]{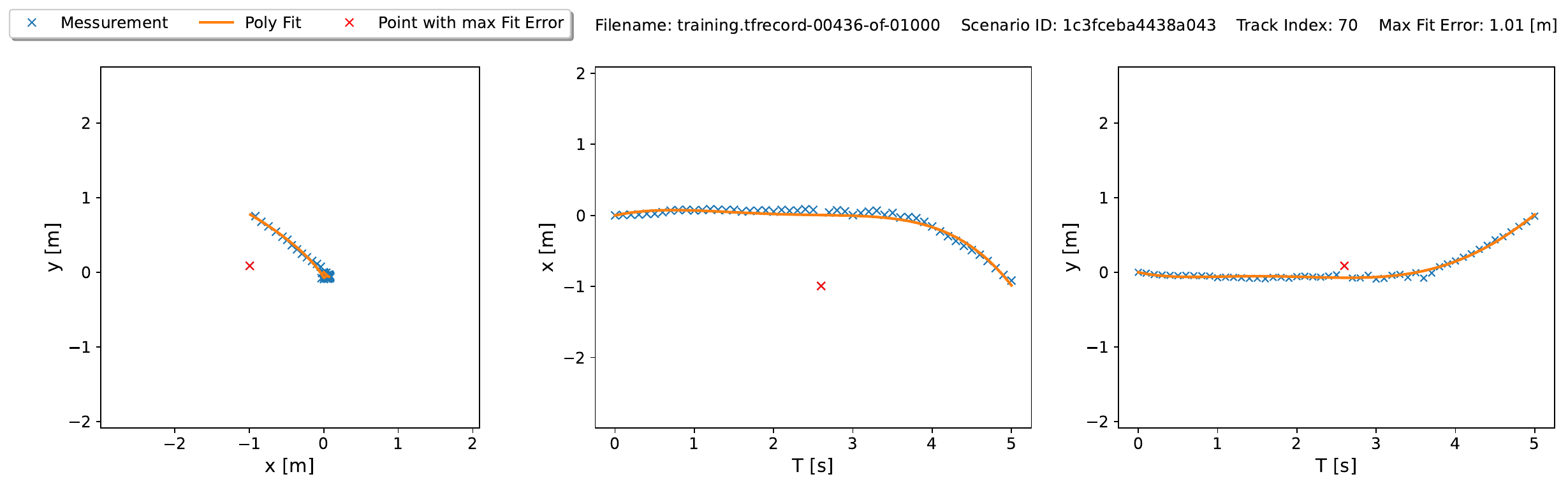} 

\includegraphics[width=\textwidth, height=1.8in]{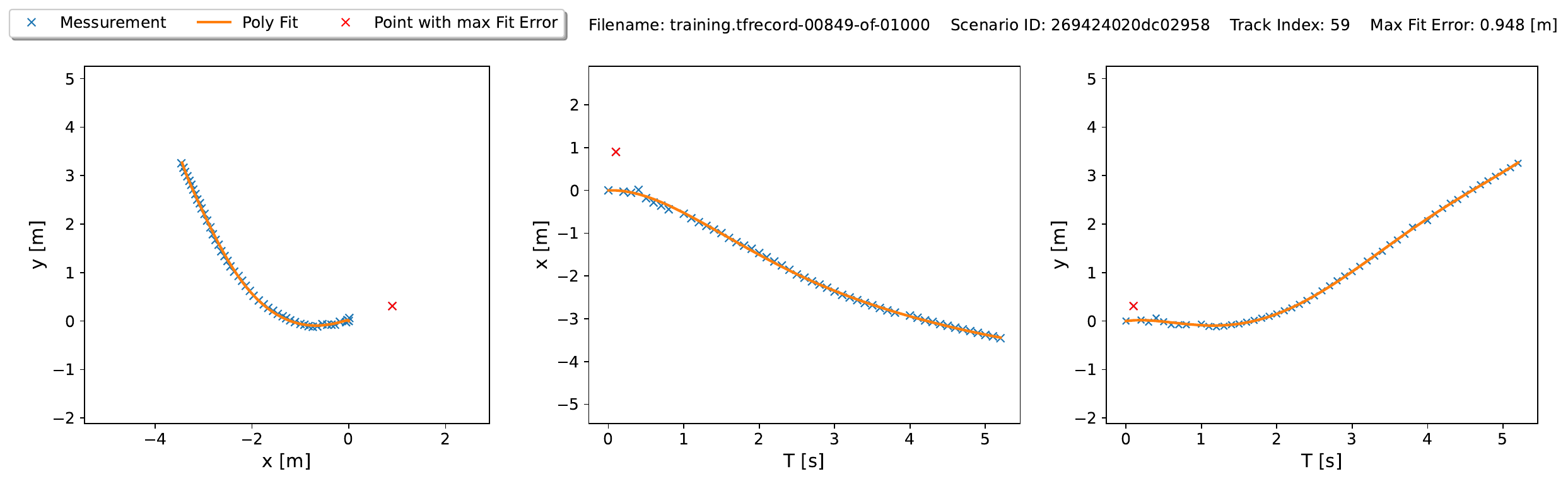} 

\includegraphics[width=\textwidth, height=1.8in]{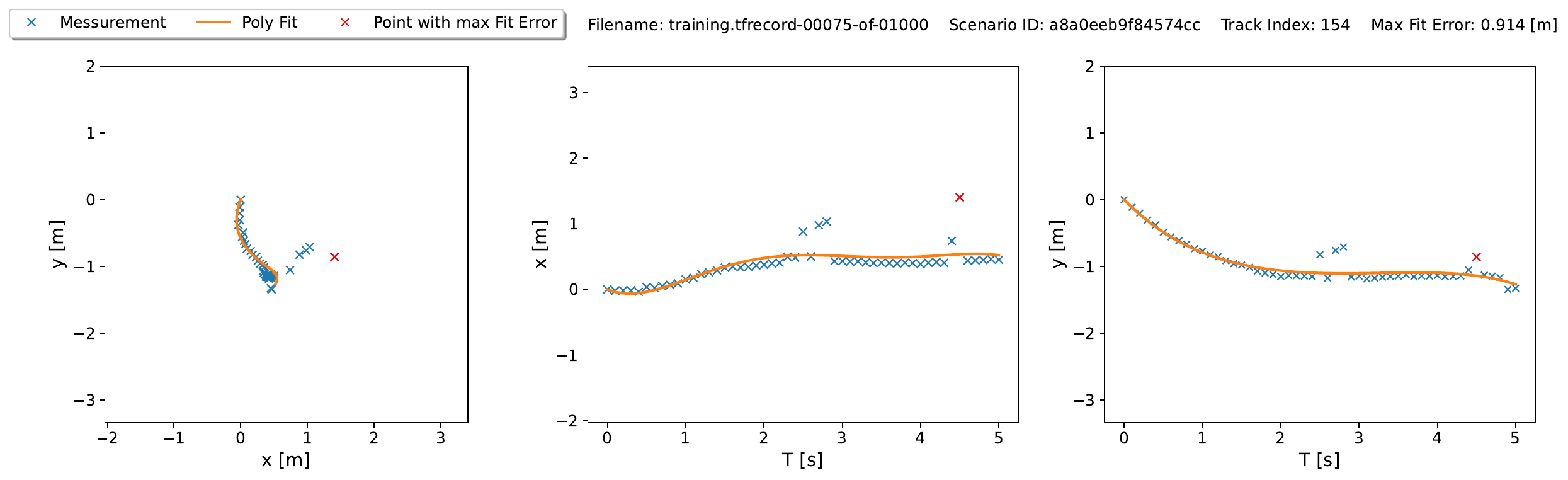} 

\includegraphics[width=\textwidth, height=1.8in]{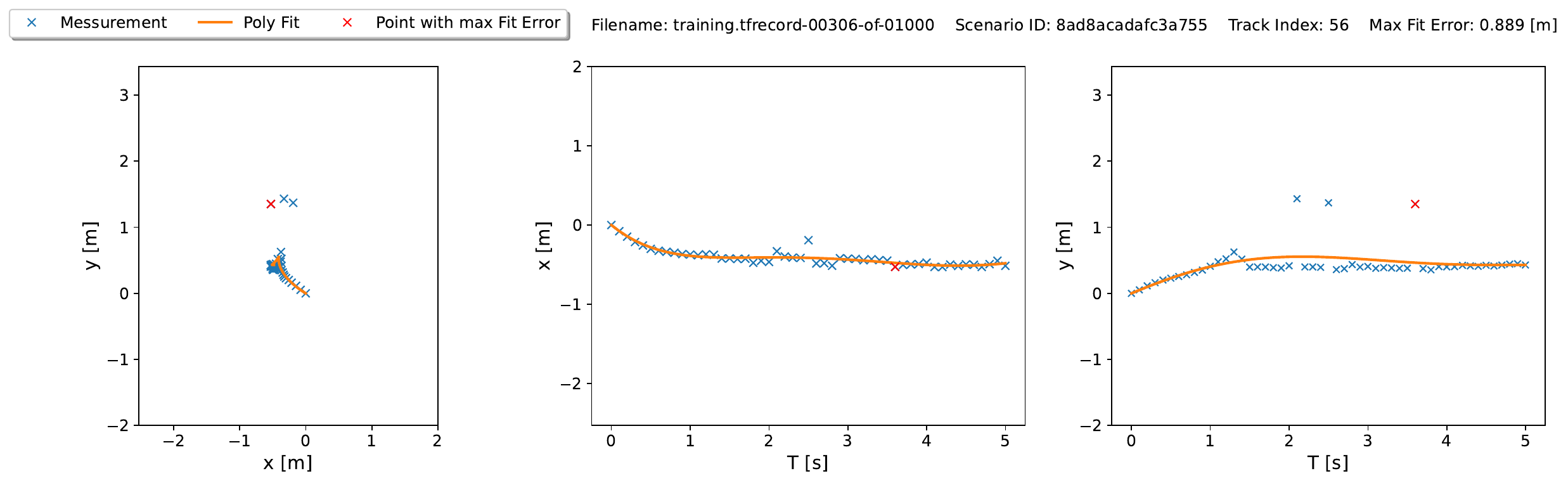} 

\includegraphics[width=\textwidth, height=1.7in]{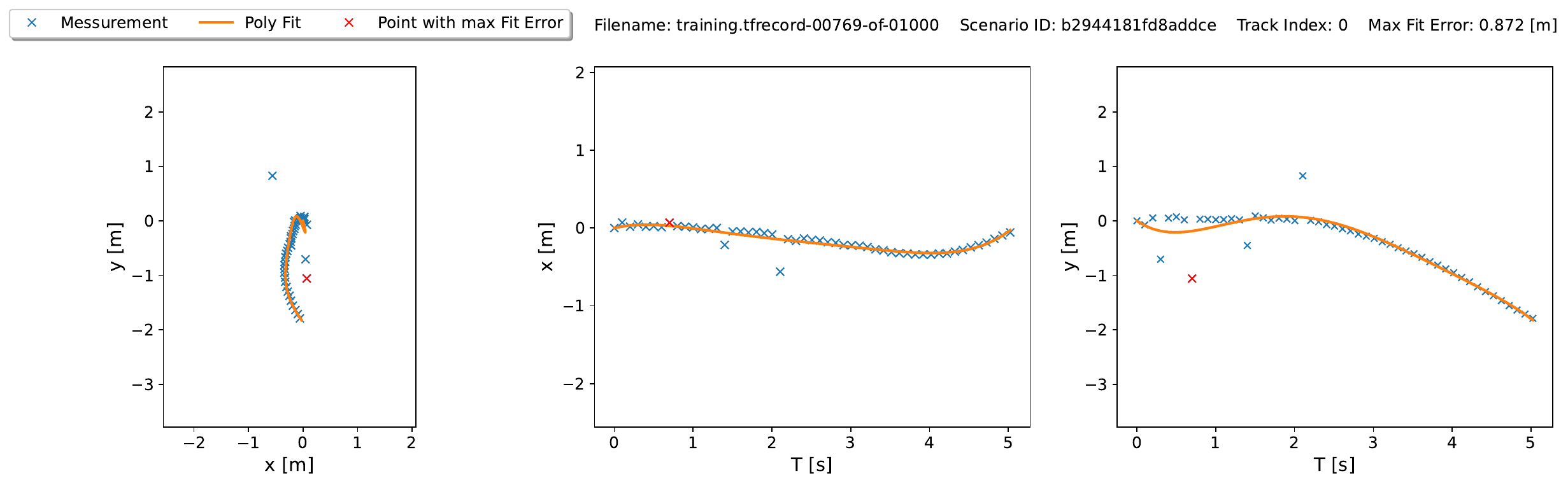}

\subsubsection{Five random 5-seconds pedestrian trajectories in WO (Fitted with $\hat{N} = 5$)}
\centering
\includegraphics[width=\textwidth, height=1.8in]{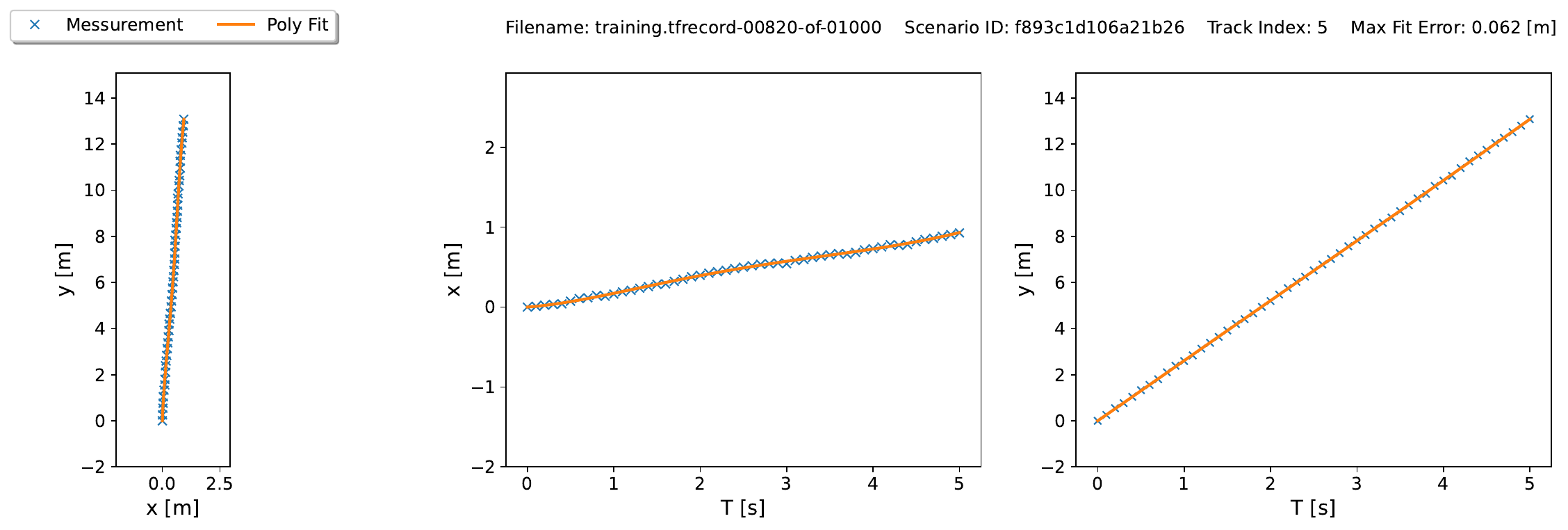} 

\includegraphics[width=\textwidth, height=1.8in]{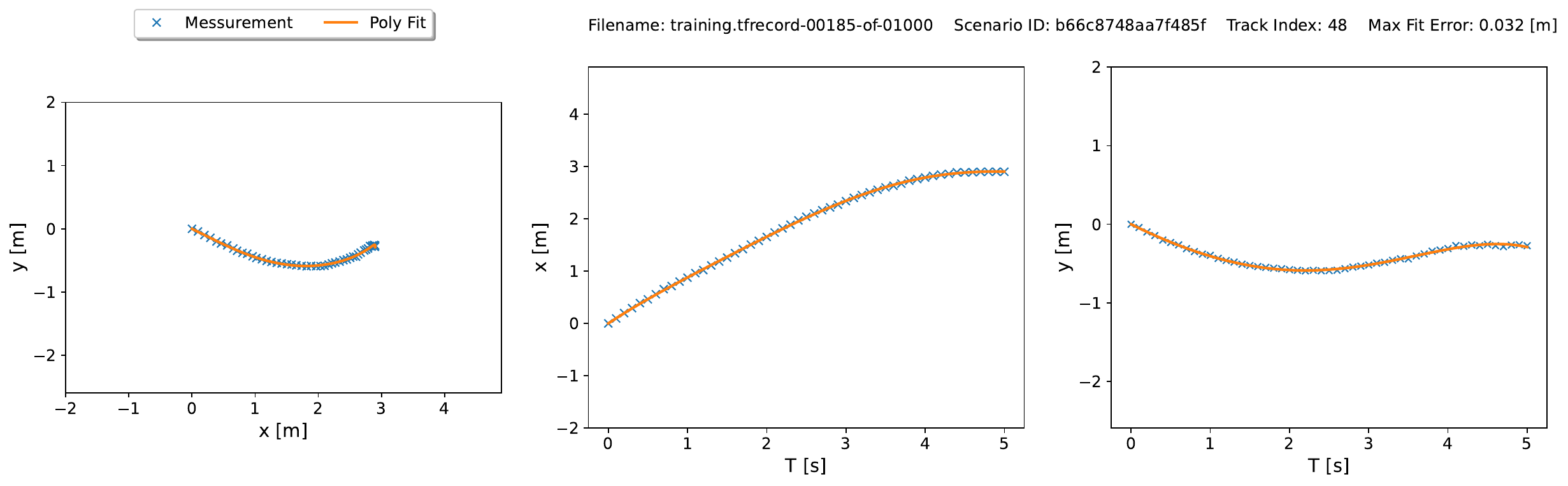} 

\includegraphics[width=\textwidth, height=1.8in]{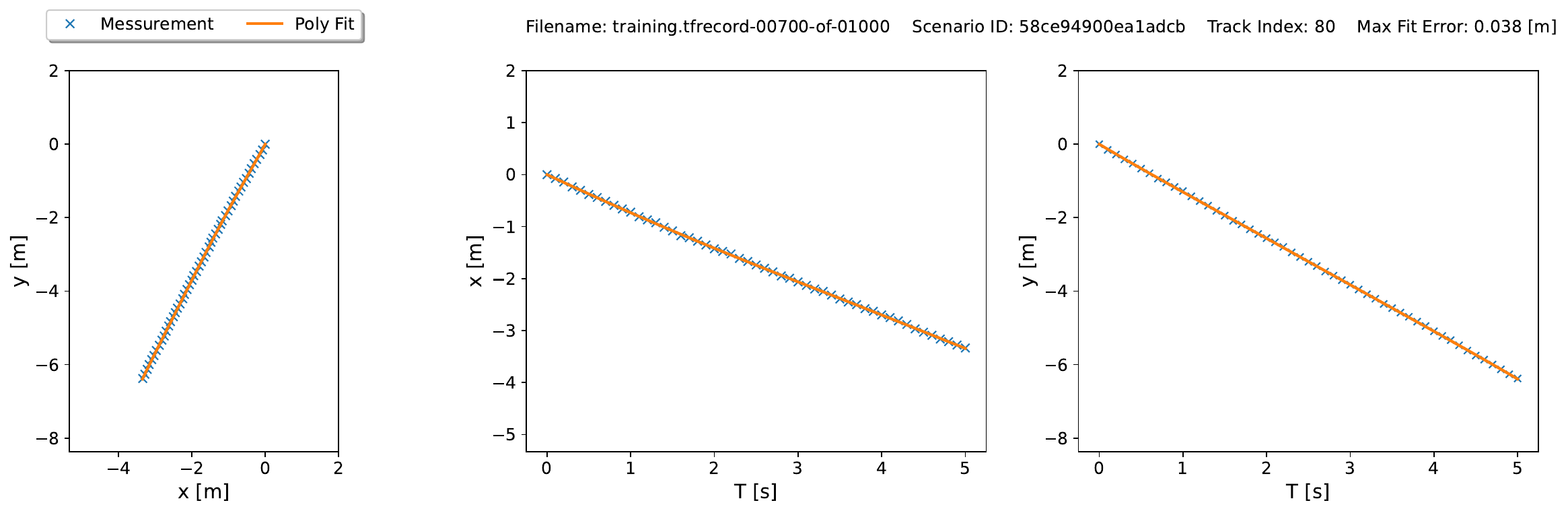} 

\includegraphics[width=\textwidth, height=1.8in]{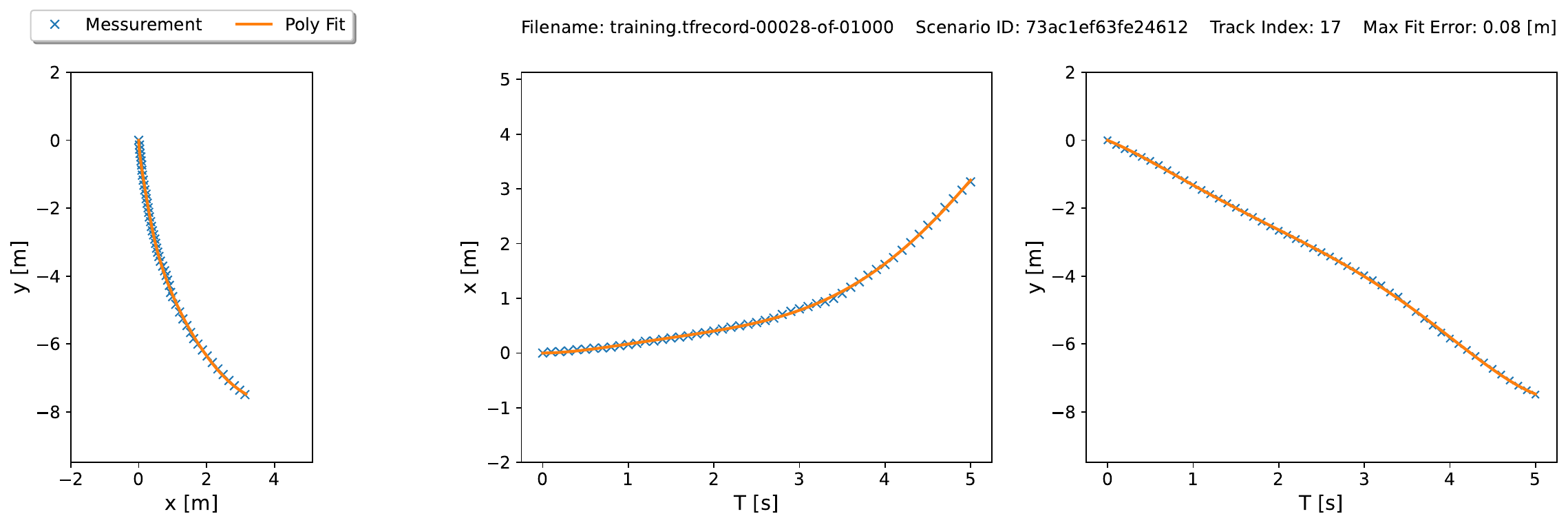} 

\includegraphics[width=\textwidth, height=1.7in]{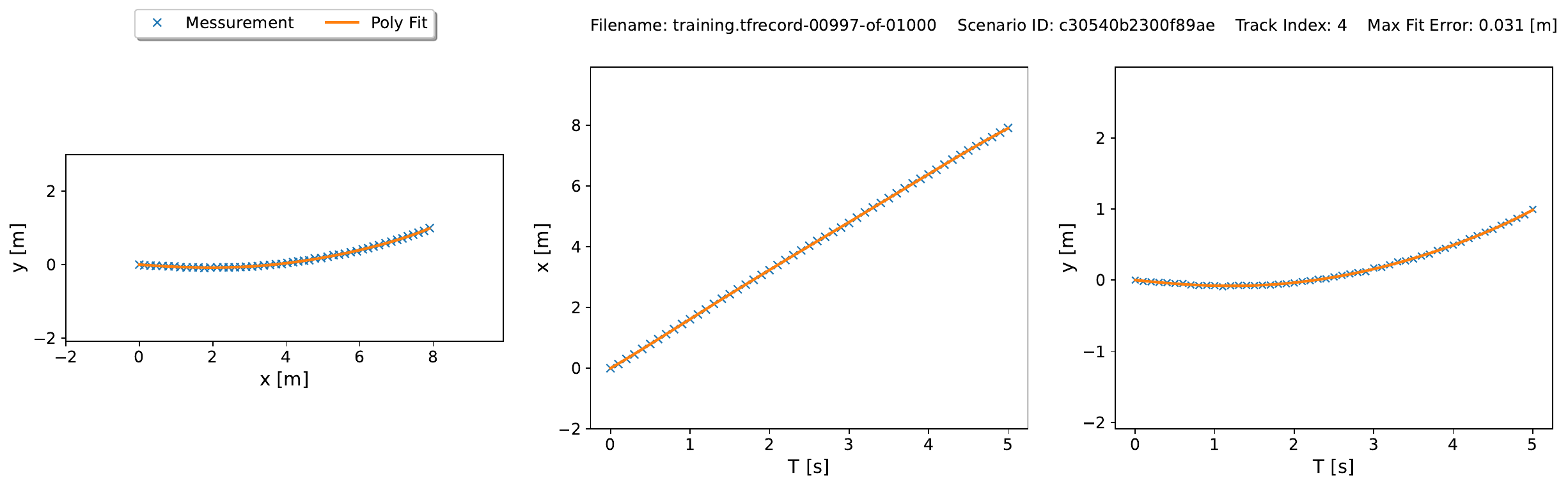}

\end{appendices}

\begingroup
\makeatletter
\PHD@frontmatterhead 
\makeatother
\chapter*{List of Publications}
\endgroup
\addcontentsline{toc}{chapter}{List of Publications}

\begin{itemize}
    \item \textbf{Y. Yao}, D. Goehring, and J. Reichardt, “An Empirical Bayes analysis of object trajectory representation models,” in \textit{Proceedings of the IEEE International Conference on Intelligent Transportation Systems (ITSC)}, pp.~902--909, 2023. DOI: \href{https://doi.org/10.1109/ITSC57777.2023.10422243}{10.1109/ITSC57777.2023.10422243}.

    \item \textbf{Y. Yao}, S. Yan, D. Goehring, W. Burgard, and J. Reichardt, “Improving out-of-distribution generalization of trajectory prediction for autonomous driving via polynomial representations,” in \textit{Proceedings of the IEEE/RSJ International Conference on Intelligent Robots and Systems (IROS)}, pp.~488--495, 2024. DOI: \href{https://doi.org/10.1109/IROS58592.2024.10801535}{10.1109/IROS58592.2024.10801535}.

    \item \textbf{Y. Yao}, D. Goehring, and J. Reichardt, “Beyond in-distribution performance: A cross-dataset study of trajectory prediction robustness,” \textit{16th UniDAS FAS-Workshop}, 2025. Available as arXiv preprint, DOI: \href{https://doi.org/10.48550/arXiv.2501.15842}{10.48550/arXiv.2501.15842}, licensed under CC BY 4.0 (\href{https://creativecommons.org/licenses/by/4.0/}{https://creativecommons.org/licenses/by/4.0/}).
    
    \item \textbf{Y. Yao}, M.-K. Bouzidi, D. Goehring, and J. Reichardt, “EP-Diffuser: An efficient diffusion model for traffic scene generation and prediction via polynomial representations,” \textit{IEEE Robotics and Automation Letters (RA-L)}, vol. 10, no. 9, pp. 9478–9485, 2025. DOI: \href{https://doi.org/10.1109/LRA.2025.3595042}{10.1109/LRA.2025.3595042}.
\end{itemize}

\begin{itemize}
    \item M.-K. Bouzidi, \textbf{Y. Yao}, D. Goehring, and J. Reichardt, “Learning-aided warmstart of model predictive control in uncertain fast-changing traffic,” in \textit{Proceedings of the  IEEE International Conference on Robotics and Automation (ICRA)}, pp.~14265--14271, 2024. DOI: \href{https://doi.org/10.1109/ICRA57147.2024.10610472}{10.1109/ICRA57147.2024.10610472}.

    \item M.-K. Bouzidi, C. Schlauch, N. Scheuerer, \textbf{Y. Yao}, N. Klein, D. Goehring, and J. Reichardt, “Closing the loop: Motion prediction models beyond open-loop benchmarks,” in \textit{Proceedings of the IEEE International Conference on Intelligent Transportation Systems (ITSC)}, 2025. DOI: \href{https://doi.org/10.1109/ITSC60802.2025.11423816}{10.1109/ITSC60802.2025.11423816}.

\end{itemize}

\printthesisindex 

\end{document}